\documentclass[11pt]{article}

\PassOptionsToPackage{table,dvipsnames}{xcolor} 
\usepackage[final]{acl}

\usepackage{times}
\usepackage{latexsym}
\usepackage[T1]{fontenc}
\usepackage[utf8]{inputenc}
\usepackage{microtype}
\usepackage{inconsolata}
\usepackage{graphicx}
\graphicspath{{figures/}}

\usepackage{titletoc}

\usepackage{amsmath,amssymb,mathtools}
\usepackage{bm}
\usepackage{booktabs}
\usepackage{multirow}
\usepackage{makecell}
\usepackage{array}
\usepackage{tabularx}
\usepackage{subcaption}
\usepackage{enumitem}
\usepackage{xspace}
\usepackage{pifont}
\usepackage{placeins}
\usepackage[most]{tcolorbox}
\usepackage[capitalize,nameinlink,noabbrev]{cleveref} 

\definecolor{cOurs}{HTML}{0072B2}     
\definecolor{cCmpA}{HTML}{E69F00}     
\definecolor{cCmpB}{HTML}{009E73}     
\definecolor{cInterv}{HTML}{D55E00}   
\definecolor{cJudge}{HTML}{CC79A7}    
\definecolor{cCmpC}{HTML}{56B4E9}     
\definecolor{cBase}{HTML}{8C8C8C}     
\definecolor{cInk}{HTML}{1A1A1A}      
\definecolor{cUp}{HTML}{0072B2}       
\definecolor{cDown}{HTML}{D55E00}     
\definecolor{cNeutral}{HTML}{8C8C8C}  
\definecolor{cRowHi}{HTML}{EAF2F8}    
\definecolor{cBoxBg}{HTML}{F2F2F2}    
\definecolor{cBoxFrame}{HTML}{A3A3A3} 
\definecolor{cMarkBg}{HTML}{FFF6E5}   

\newcommand{\method}{IST\xspace}              
\newcommand{\gci}{GCI\xspace}                 
\newcommand{\gcis}{GCIs\xspace}
\newcommand{\pci}{PCI\xspace}                 
\newcommand{\pcis}{PCIs\xspace}
\newcommand{\rlvr}{RLVR\xspace}               
\newcommand{\NT}{\textsc{nt}\xspace}          
\newcommand{\NAT}{\textsc{nat}\xspace}        
\newcommand{\rate}{R}                         

\newcommand{\form}[1]{\textit{#1}}
\newcommand{\gloss}[1]{`#1'}
\newcommand{\tok}[1]{\texttt{\small #1}}

\newcounter{hypothesis}
\newenvironment{hypothesis}[1][]{%
  \refstepcounter{hypothesis}\par\smallskip\noindent
  \textbf{H\thehypothesis}\if\relax\detokenize{#1}\relax\else\ (\textit{#1})\fi.\ \ignorespaces}%
  {\par\smallskip}
\crefformat{hypothesis}{#2H#1#3}
\Crefformat{hypothesis}{#2H#1#3}
\crefmultiformat{hypothesis}{#2H#1#3}{ and~#2H#1#3}{, #2H#1#3}{, and~#2H#1#3}
\Crefmultiformat{hypothesis}{#2H#1#3}{ and~#2H#1#3}{, #2H#1#3}{, and~#2H#1#3}
\crefrangeformat{hypothesis}{#3H#1#4--#5H#2#6}
\Crefrangeformat{hypothesis}{#3H#1#4--#5H#2#6}
\newcommand{\hyp}[1]{\textbf{H#1}\xspace}

\newtcolorbox{claimbox}{enhanced,colback=cBoxBg,colframe=cBoxFrame,boxrule=0.5pt,
  arc=1.5pt,left=5pt,right=5pt,top=3pt,bottom=3pt,fontupper=\itshape}

\newcommand{\std}[1]{{\scriptsize\,$\pm$\,#1}}
\newcommand{\pms}[2]{#1\std{#2}}

\newcommand{\up}[1]{{\scriptsize\textcolor{cUp}{$\uparrow$#1}}}
\newcommand{\down}[1]{{\scriptsize\textcolor{cDown}{$\downarrow$#1}}}
\newcommand{\upbad}[1]{{\scriptsize\textcolor{cDown}{$\uparrow$#1}}}
\newcommand{\downgood}[1]{{\scriptsize\textcolor{cUp}{$\downarrow$#1}}}

\newcommand{\sig}[1]{\textsuperscript{#1}}
\newcommand{\hup}{\,$\uparrow$}

\newcommand{\na}{--}

\newcommand{\resdef}[2]{\expandafter\def\csname res@#1\endcsname{#2}}
\usepackage{xstring}
\newcommand{\pilot}[1]{\ifdraft{\color{gray}#1}\else\errmessage{Pilot number left in final build}\fi}
\newcommand{\rv}[1]{\csname res@#1\endcsname}
\newcommand{\pending}{\ifdraft{\scriptsize\textcolor{gray}{\textit{pending}}}\else\errmessage{Pending result left in final build}\fi}
\newcommand{\resmissing}[1]{\typeout{MISSING RES KEY: #1}\pending}
\newcommand{\cellpm}[2]{%
  \ifcsname res@#1.#2.mean\endcsname
    \ifcsname res@#1.#2.std\endcsname\pms{\pv{#1.#2.mean}}{\rv{#1.#2.std}}\else\pv{#1.#2.mean}\fi
  \else\ifcsname res@pilot.#2.mean\endcsname
    \pilot{\ifcsname res@pilot.#2.std\endcsname\pms{\pv{pilot.#2.mean}}{\rv{pilot.#2.std}}\else\pv{pilot.#2.mean}\fi}%
  \else\resmissing{#1.#2.mean}\fi\fi}

\newcommand{\signeddelta}[2]{\IfBeginWith{#1}{-}%
  {\ifnum#2=1 \down{\StrGobbleLeft{#1}{1}}\else\downgood{\StrGobbleLeft{#1}{1}}\fi}%
  {\ifnum#2=1 \up{#1}\else\upbad{#1}\fi}}
\newcommand{\celldgen}[3]{%
  \ifcsname res@#2.#3.mean\endcsname\edef\pdtmp{\rv{#2.#3.mean}}\expandafter\signeddelta\expandafter{\pdtmp}{#1}%
  \else\ifcsname res@pilot.#3.mean\endcsname\edef\pdtmp{\rv{pilot.#3.mean}}\pilot{\expandafter\signeddelta\expandafter{\pdtmp}{#1}}%
  \else\resmissing{#2.#3.mean}\fi\fi}
\newcommand{\celld}[2]{\celldgen{1}{#1}{#2}}

\newcommand{\cistar}[2]{\ifcsname res@#1.#2.lo\endcsname\ifdim\rv{#1.#2.hi}pt<0pt\sig{*}\else\ifdim\rv{#1.#2.lo}pt>0pt\sig{*}\fi\fi\fi}
\newcommand{\celldpmgen}[3]{\celldgen{#1}{#2}{#3}\ifcsname res@#2.#3.std\endcsname\std{\rv{#2.#3.std}}\fi\cistar{#2}{#3}}
\newcommand{\celldpm}[2]{\celldpmgen{1}{#1}{#2}}

\newcommand{\pnum}[1]{\IfBeginWith{#1}{-}{\ensuremath{-}\StrGobbleLeft{#1}{1}}{#1}}
\newcommand{\pv}[1]{\edef\pvtmp{\rv{#1}}\expandafter\pnum\expandafter{\pvtmp}}
\newcommand{\res}[1]{\ifcsname res@#1\endcsname\pv{#1}\else\resmissing{#1}\fi}

\newif\ifdraft
\drafttrue

\newcommand{\figplaceholder}[2]{\ifdraft\fbox{\parbox[c][#2][c]{\dimexpr#1-2\fboxsep-2\fboxrule}{\centering\color{gray}\small\textit{Figure placeholder}}}\else\errmessage{Figure placeholder left in final build}\fi}

\newcommand{\vy}{\bm{y}}

\newcommand{\trace}{\vy}                      

\providecommand{\ourrow}{\rowcolor{gray!10}}

\resdef{pilot.qwen3-14b.a3-conn.inf.nat.dR.hi}{-6.7}  
\resdef{pilot.qwen3-14b.a3-conn.inf.nat.dR.lo}{-25.6}  
\resdef{pilot.qwen3-14b.a3-conn.inf.nat.dR.mean}{-15.6}  
\resdef{pilot.qwen3-14b.a3-conn.inf.nat.dR.std}{11.7}  
\resdef{pilot.qwen3-14b.a3-modal.inf.nat.dR.hi}{7.8}  
\resdef{pilot.qwen3-14b.a3-modal.inf.nat.dR.lo}{-4.4}  
\resdef{pilot.qwen3-14b.a3-modal.inf.nat.dR.mean}{1.1}  
\resdef{pilot.qwen3-14b.a3-modal.inf.nat.dR.std}{1.9}  
\resdef{pilot.qwen3-14b.a3-quant.inf.nat.dR.hi}{-13.3}  
\resdef{pilot.qwen3-14b.a3-quant.inf.nat.dR.lo}{-38.9}  
\resdef{pilot.qwen3-14b.a3-quant.inf.nat.dR.mean}{-25.6}  
\resdef{pilot.qwen3-14b.a3-quant.inf.nat.dR.std}{3.8}  
\resdef{pilot.qwen3-14b.a3.inf.nat.R.hi}{16.3}  
\resdef{pilot.qwen3-14b.a3.inf.nat.R.lo}{8.2}  
\resdef{pilot.qwen3-14b.a3.inf.nat.R.mean}{12.1}  
\resdef{pilot.qwen3-14b.a3.inf.nat.R.std}{1.8}  
\resdef{pilot.qwen3-14b.a3.inf.nat.dR.hi}{2.6}  
\resdef{pilot.qwen3-14b.a3.inf.nat.dR.lo}{-7.2}  
\resdef{pilot.qwen3-14b.a3.inf.nat.dR.mean}{-2.3}  
\resdef{pilot.qwen3-14b.a3.inf.nat.dR.std}{0.8}  
\resdef{pilot.qwen3-14b.a3.inf.off.R.hi}{18.9}  
\resdef{pilot.qwen3-14b.a3.inf.off.R.lo}{10.0}  
\resdef{pilot.qwen3-14b.a3.inf.off.R.mean}{14.4}  
\resdef{pilot.qwen3-14b.a3.inf.off.R.std}{2.5}  
\resdef{pilot.qwen3-14b.b2.yn.nat.acc.hi}{86.3}  
\resdef{pilot.qwen3-14b.b2.yn.nat.acc.lo}{75.9}  
\resdef{pilot.qwen3-14b.b2.yn.nat.acc.mean}{81.3}  
\resdef{pilot.qwen3-14b.b2.yn.nat.acc.std}{1.7}  
\resdef{pilot.qwen3-14b.b2.yn.nat.dacc.hi}{9.4}  
\resdef{pilot.qwen3-14b.b2.yn.nat.dacc.lo}{-0.6}  
\resdef{pilot.qwen3-14b.b2.yn.nat.dacc.mean}{4.4}  
\resdef{pilot.qwen3-14b.b2.yn.nat.dacc.std}{2.0}  
\resdef{pilot.qwen3-14b.b2.yn.off.acc.hi}{82.4}  
\resdef{pilot.qwen3-14b.b2.yn.off.acc.lo}{70.6}  
\resdef{pilot.qwen3-14b.b2.yn.off.acc.mean}{76.9}  
\resdef{pilot.qwen3-14b.b2.yn.off.acc.std}{0.3}  
\resdef{pilot.qwen3-4b.a3-conn.inf.nat.dR.hi}{-54.4}  
\resdef{pilot.qwen3-4b.a3-conn.inf.nat.dR.lo}{-76.7}  
\resdef{pilot.qwen3-4b.a3-conn.inf.nat.dR.mean}{-65.6}  
\resdef{pilot.qwen3-4b.a3-conn.inf.nat.dR.std}{3.8}  
\resdef{pilot.qwen3-4b.a3-modal.inf.nat.dR.hi}{-11.1}  
\resdef{pilot.qwen3-4b.a3-modal.inf.nat.dR.lo}{-47.8}  
\resdef{pilot.qwen3-4b.a3-modal.inf.nat.dR.mean}{-30.0}  
\resdef{pilot.qwen3-4b.a3-modal.inf.nat.dR.std}{17.6}  
\resdef{pilot.qwen3-4b.a3-quant.inf.nat.dR.hi}{-44.4}  
\resdef{pilot.qwen3-4b.a3-quant.inf.nat.dR.lo}{-72.2}  
\resdef{pilot.qwen3-4b.a3-quant.inf.nat.dR.mean}{-58.9}  
\resdef{pilot.qwen3-4b.a3-quant.inf.nat.dR.std}{8.4}  
\resdef{pilot.qwen3-4b.a3.inf.nat.R.hi}{27.0}  
\resdef{pilot.qwen3-4b.a3.inf.nat.R.lo}{16.8}  
\resdef{pilot.qwen3-4b.a3.inf.nat.R.mean}{21.9}  
\resdef{pilot.qwen3-4b.a3.inf.nat.R.std}{1.6}  
\resdef{pilot.qwen3-4b.a3.inf.nat.dR.hi}{-22.5}  
\resdef{pilot.qwen3-4b.a3.inf.nat.dR.lo}{-37.2}  
\resdef{pilot.qwen3-4b.a3.inf.nat.dR.mean}{-29.8}  
\resdef{pilot.qwen3-4b.a3.inf.nat.dR.std}{3.0}  
\resdef{pilot.qwen3-4b.a3.inf.off.R.hi}{56.3}  
\resdef{pilot.qwen3-4b.a3.inf.off.R.lo}{46.8}  
\resdef{pilot.qwen3-4b.a3.inf.off.R.mean}{51.8}  
\resdef{pilot.qwen3-4b.a3.inf.off.R.std}{1.7}  
\resdef{pilot.qwen3-4b.b2.yn.nat.acc.hi}{87.4}  
\resdef{pilot.qwen3-4b.b2.yn.nat.acc.lo}{77.0}  
\resdef{pilot.qwen3-4b.b2.yn.nat.acc.mean}{82.4}  
\resdef{pilot.qwen3-4b.b2.yn.nat.acc.std}{0.6}  
\resdef{pilot.qwen3-4b.b2.yn.nat.dacc.hi}{11.1}  
\resdef{pilot.qwen3-4b.b2.yn.nat.dacc.lo}{-1.3}  
\resdef{pilot.qwen3-4b.b2.yn.nat.dacc.mean}{5.0}  
\resdef{pilot.qwen3-4b.b2.yn.nat.dacc.std}{1.7}  
\resdef{pilot.qwen3-4b.b2.yn.off.acc.hi}{82.8}  
\resdef{pilot.qwen3-4b.b2.yn.off.acc.lo}{71.5}  
\resdef{pilot.qwen3-4b.b2.yn.off.acc.mean}{77.4}  
\resdef{pilot.qwen3-4b.b2.yn.off.acc.std}{1.2}  
\resdef{pilot.qwen3-8b.a3-conn.inf.nat.R.hi}{0.0}  
\resdef{pilot.qwen3-8b.a3-conn.inf.nat.R.lo}{0.0}  
\resdef{pilot.qwen3-8b.a3-conn.inf.nat.R.mean}{0.0}  
\resdef{pilot.qwen3-8b.a3-conn.inf.nat.R.std}{0.0}  
\resdef{pilot.qwen3-8b.a3-conn.inf.nat.dR.hi}{-25.6}  
\resdef{pilot.qwen3-8b.a3-conn.inf.nat.dR.lo}{-48.9}  
\resdef{pilot.qwen3-8b.a3-conn.inf.nat.dR.mean}{-36.7}  
\resdef{pilot.qwen3-8b.a3-conn.inf.nat.dR.std}{5.8}  
\resdef{pilot.qwen3-8b.a3-conn.inf.off.R.hi}{48.9}  
\resdef{pilot.qwen3-8b.a3-conn.inf.off.R.lo}{24.4}  
\resdef{pilot.qwen3-8b.a3-conn.inf.off.R.mean}{36.7}  
\resdef{pilot.qwen3-8b.a3-conn.inf.off.R.std}{5.8}  
\resdef{pilot.qwen3-8b.a3-modal.inf.nat.dR.hi}{-7.8}  
\resdef{pilot.qwen3-8b.a3-modal.inf.nat.dR.lo}{-31.1}  
\resdef{pilot.qwen3-8b.a3-modal.inf.nat.dR.mean}{-18.9}  
\resdef{pilot.qwen3-8b.a3-modal.inf.nat.dR.std}{6.9}  
\resdef{pilot.qwen3-8b.a3-quant.inf.nat.R.hi}{0.0}  
\resdef{pilot.qwen3-8b.a3-quant.inf.nat.R.lo}{0.0}  
\resdef{pilot.qwen3-8b.a3-quant.inf.nat.R.mean}{0.0}  
\resdef{pilot.qwen3-8b.a3-quant.inf.nat.R.std}{0.0}  
\resdef{pilot.qwen3-8b.a3-quant.inf.nat.dR.hi}{-20.0}  
\resdef{pilot.qwen3-8b.a3-quant.inf.nat.dR.lo}{-42.2}  
\resdef{pilot.qwen3-8b.a3-quant.inf.nat.dR.mean}{-31.1}  
\resdef{pilot.qwen3-8b.a3-quant.inf.nat.dR.std}{9.6}  
\resdef{pilot.qwen3-8b.a3-quant.inf.off.R.hi}{42.2}  
\resdef{pilot.qwen3-8b.a3-quant.inf.off.R.lo}{20.0}  
\resdef{pilot.qwen3-8b.a3-quant.inf.off.R.mean}{31.1}  
\resdef{pilot.qwen3-8b.a3-quant.inf.off.R.std}{9.6}  
\resdef{pilot.qwen3-8b.a3.inf.nat.R.hi}{14.9}  
\resdef{pilot.qwen3-8b.a3.inf.nat.R.lo}{7.5}  
\resdef{pilot.qwen3-8b.a3.inf.nat.R.mean}{11.1}  
\resdef{pilot.qwen3-8b.a3.inf.nat.R.std}{1.1}  
\resdef{pilot.qwen3-8b.a3.inf.nat.dR.hi}{-7.2}  
\resdef{pilot.qwen3-8b.a3.inf.nat.dR.lo}{-17.4}  
\resdef{pilot.qwen3-8b.a3.inf.nat.dR.mean}{-12.3}  
\resdef{pilot.qwen3-8b.a3.inf.nat.dR.std}{2.1}  
\resdef{pilot.qwen3-8b.a3.inf.off.R.hi}{28.2}  
\resdef{pilot.qwen3-8b.a3.inf.off.R.lo}{18.4}  
\resdef{pilot.qwen3-8b.a3.inf.off.R.mean}{23.3}  
\resdef{pilot.qwen3-8b.a3.inf.off.R.std}{1.3}  
\resdef{pilot.qwen3-8b.b2.yn.nat.acc.hi}{87.8}  
\resdef{pilot.qwen3-8b.b2.yn.nat.acc.lo}{77.8}  
\resdef{pilot.qwen3-8b.b2.yn.nat.acc.mean}{83.0}  
\resdef{pilot.qwen3-8b.b2.yn.nat.acc.std}{0.8}  
\resdef{pilot.qwen3-8b.b2.yn.nat.dacc.hi}{10.2}  
\resdef{pilot.qwen3-8b.b2.yn.nat.dacc.lo}{-0.6}  
\resdef{pilot.qwen3-8b.b2.yn.nat.dacc.mean}{4.6}  
\resdef{pilot.qwen3-8b.b2.yn.nat.dacc.std}{2.0}  
\resdef{pilot.qwen3-8b.b2.yn.off.acc.hi}{83.7}  
\resdef{pilot.qwen3-8b.b2.yn.off.acc.lo}{72.6}  
\resdef{pilot.qwen3-8b.b2.yn.off.acc.mean}{78.3}  
\resdef{pilot.qwen3-8b.b2.yn.off.acc.std}{1.1}  
\resdef{e1.qwen3-14b.a1.ter.nat.R.hi}{0.0}  
\resdef{e1.qwen3-14b.a1.ter.nat.R.lo}{0.0}  
\resdef{e1.qwen3-14b.a1.ter.nat.R.mean}{0.0}  
\resdef{e1.qwen3-14b.a1.ter.nat.R.std}{0.0}  
\resdef{e1.qwen3-14b.a1.ter.nat.dR.hi}{0.0}  
\resdef{e1.qwen3-14b.a1.ter.nat.dR.lo}{-7.5}  
\resdef{e1.qwen3-14b.a1.ter.nat.dR.mean}{-2.5}  
\resdef{e1.qwen3-14b.a1.ter.nat.dR.std}{0.0}  
\resdef{e1.qwen3-14b.a1.ter.off.R.hi}{7.5}  
\resdef{e1.qwen3-14b.a1.ter.off.R.lo}{0.0}  
\resdef{e1.qwen3-14b.a1.ter.off.R.mean}{2.5}  
\resdef{e1.qwen3-14b.a1.ter.off.R.std}{0.0}  
\resdef{e1.qwen3-14b.a2.inf.nat.R.hi}{13.2}  
\resdef{e1.qwen3-14b.a2.inf.nat.R.lo}{4.9}  
\resdef{e1.qwen3-14b.a2.inf.nat.R.mean}{8.8}  
\resdef{e1.qwen3-14b.a2.inf.nat.R.std}{1.4}  
\resdef{e1.qwen3-14b.a2.inf.nat.dR.hi}{10.8}  
\resdef{e1.qwen3-14b.a2.inf.nat.dR.lo}{3.3}  
\resdef{e1.qwen3-14b.a2.inf.nat.dR.mean}{6.8}  
\resdef{e1.qwen3-14b.a2.inf.nat.dR.std}{1.7}  
\resdef{e1.qwen3-14b.a2.inf.off.R.hi}{4.2}  
\resdef{e1.qwen3-14b.a2.inf.off.R.lo}{0.4}  
\resdef{e1.qwen3-14b.a2.inf.off.R.mean}{2.0}  
\resdef{e1.qwen3-14b.a2.inf.off.R.std}{0.7}  
\resdef{e1.qwen3-14b.a3-conn.inf.nat.dR.hi}{-6.1}  
\resdef{e1.qwen3-14b.a3-conn.inf.nat.dR.lo}{-21.2}  
\resdef{e1.qwen3-14b.a3-conn.inf.nat.dR.mean}{-13.1}  
\resdef{e1.qwen3-14b.a3-conn.inf.nat.dR.std}{6.3}  
\resdef{e1.qwen3-14b.a3-modal.inf.nat.dR.hi}{13.5}  
\resdef{e1.qwen3-14b.a3-modal.inf.nat.dR.lo}{-3.1}  
\resdef{e1.qwen3-14b.a3-modal.inf.nat.dR.mean}{4.2}  
\resdef{e1.qwen3-14b.a3-modal.inf.nat.dR.std}{6.5}  
\resdef{e1.qwen3-14b.a3-quant.inf.nat.dR.hi}{-9.0}  
\resdef{e1.qwen3-14b.a3-quant.inf.nat.dR.lo}{-37.2}  
\resdef{e1.qwen3-14b.a3-quant.inf.nat.dR.mean}{-21.8}  
\resdef{e1.qwen3-14b.a3-quant.inf.nat.dR.std}{8.0}  
\resdef{e1.qwen3-14b.a3.inf.nat.R.hi}{19.5}  
\resdef{e1.qwen3-14b.a3.inf.nat.R.lo}{10.9}  
\resdef{e1.qwen3-14b.a3.inf.nat.R.mean}{15.0}  
\resdef{e1.qwen3-14b.a3.inf.nat.R.std}{2.1}  
\resdef{e1.qwen3-14b.a3.inf.nat.dR.hi}{4.8}  
\resdef{e1.qwen3-14b.a3.inf.nat.dR.lo}{-3.8}  
\resdef{e1.qwen3-14b.a3.inf.nat.dR.mean}{0.5}  
\resdef{e1.qwen3-14b.a3.inf.nat.dR.std}{1.5}  
\resdef{e1.qwen3-14b.a3.inf.off.R.hi}{19.0}  
\resdef{e1.qwen3-14b.a3.inf.off.R.lo}{10.4}  
\resdef{e1.qwen3-14b.a3.inf.off.R.mean}{14.5}  
\resdef{e1.qwen3-14b.a3.inf.off.R.std}{1.5}  
\resdef{e1.qwen3-14b.a4.nli.nat.acc.hi}{83.5}  
\resdef{e1.qwen3-14b.a4.nli.nat.acc.lo}{69.7}  
\resdef{e1.qwen3-14b.a4.nli.nat.acc.mean}{76.9}  
\resdef{e1.qwen3-14b.a4.nli.nat.acc.std}{2.6}  
\resdef{e1.qwen3-14b.a4.nli.nat.dacc.hi}{31.5}  
\resdef{e1.qwen3-14b.a4.nli.nat.dacc.lo}{14.4}  
\resdef{e1.qwen3-14b.a4.nli.nat.dacc.mean}{22.8}  
\resdef{e1.qwen3-14b.a4.nli.nat.dacc.std}{2.6}  
\resdef{e1.qwen3-14b.a4.nli.off.acc.hi}{62.2}  
\resdef{e1.qwen3-14b.a4.nli.off.acc.lo}{45.3}  
\resdef{e1.qwen3-14b.a4.nli.off.acc.mean}{54.1}  
\resdef{e1.qwen3-14b.a4.nli.off.acc.std}{0.0}  
\resdef{e1.qwen3-14b.b2.yn.nat.acc.hi}{83.5}  
\resdef{e1.qwen3-14b.b2.yn.nat.acc.lo}{67.4}  
\resdef{e1.qwen3-14b.b2.yn.nat.acc.mean}{75.6}  
\resdef{e1.qwen3-14b.b2.yn.nat.acc.std}{0.6}  
\resdef{e1.qwen3-14b.b2.yn.nat.dacc.hi}{8.2}  
\resdef{e1.qwen3-14b.b2.yn.nat.dacc.lo}{-5.4}  
\resdef{e1.qwen3-14b.b2.yn.nat.dacc.mean}{1.4}  
\resdef{e1.qwen3-14b.b2.yn.nat.dacc.std}{0.6}  
\resdef{e1.qwen3-14b.b2.yn.off.acc.hi}{82.4}  
\resdef{e1.qwen3-14b.b2.yn.off.acc.lo}{65.6}  
\resdef{e1.qwen3-14b.b2.yn.off.acc.mean}{74.2}  
\resdef{e1.qwen3-14b.b2.yn.off.acc.std}{1.1}  
\resdef{e1.qwen3-4b.a1.ter.nat.R.hi}{7.5}  
\resdef{e1.qwen3-4b.a1.ter.nat.R.lo}{0.0}  
\resdef{e1.qwen3-4b.a1.ter.nat.R.mean}{2.5}  
\resdef{e1.qwen3-4b.a1.ter.nat.R.std}{0.0}  
\resdef{e1.qwen3-4b.a1.ter.nat.dR.hi}{0.0}  
\resdef{e1.qwen3-4b.a1.ter.nat.dR.lo}{-7.5}  
\resdef{e1.qwen3-4b.a1.ter.nat.dR.mean}{-2.5}  
\resdef{e1.qwen3-4b.a1.ter.nat.dR.std}{0.0}  
\resdef{e1.qwen3-4b.a1.ter.off.R.hi}{12.5}  
\resdef{e1.qwen3-4b.a1.ter.off.R.lo}{0.0}  
\resdef{e1.qwen3-4b.a1.ter.off.R.mean}{5.0}  
\resdef{e1.qwen3-4b.a1.ter.off.R.std}{0.0}  
\resdef{e1.qwen3-4b.a2.inf.nat.R.hi}{13.2}  
\resdef{e1.qwen3-4b.a2.inf.nat.R.lo}{4.9}  
\resdef{e1.qwen3-4b.a2.inf.nat.R.mean}{8.8}  
\resdef{e1.qwen3-4b.a2.inf.nat.R.std}{1.0}  
\resdef{e1.qwen3-4b.a2.inf.nat.dR.hi}{-4.0}  
\resdef{e1.qwen3-4b.a2.inf.nat.dR.lo}{-11.9}  
\resdef{e1.qwen3-4b.a2.inf.nat.dR.mean}{-7.7}  
\resdef{e1.qwen3-4b.a2.inf.nat.dR.std}{2.0}  
\resdef{e1.qwen3-4b.a2.inf.off.R.hi}{21.9}  
\resdef{e1.qwen3-4b.a2.inf.off.R.lo}{11.5}  
\resdef{e1.qwen3-4b.a2.inf.off.R.mean}{16.6}  
\resdef{e1.qwen3-4b.a2.inf.off.R.std}{1.1}  
\resdef{e1.qwen3-4b.a3-conn.inf.nat.dR.hi}{-47.5}  
\resdef{e1.qwen3-4b.a3-conn.inf.nat.dR.lo}{-74.7}  
\resdef{e1.qwen3-4b.a3-conn.inf.nat.dR.mean}{-61.6}  
\resdef{e1.qwen3-4b.a3-conn.inf.nat.dR.std}{4.6}  
\resdef{e1.qwen3-4b.a3-modal.inf.nat.dR.hi}{-15.6}  
\resdef{e1.qwen3-4b.a3-modal.inf.nat.dR.lo}{-47.9}  
\resdef{e1.qwen3-4b.a3-modal.inf.nat.dR.mean}{-32.3}  
\resdef{e1.qwen3-4b.a3-modal.inf.nat.dR.std}{14.4}  
\resdef{e1.qwen3-4b.a3-quant.inf.nat.dR.hi}{-17.9}  
\resdef{e1.qwen3-4b.a3-quant.inf.nat.dR.lo}{-55.1}  
\resdef{e1.qwen3-4b.a3-quant.inf.nat.dR.mean}{-37.2}  
\resdef{e1.qwen3-4b.a3-quant.inf.nat.dR.std}{4.4}  
\resdef{e1.qwen3-4b.a3.inf.nat.R.hi}{28.4}  
\resdef{e1.qwen3-4b.a3.inf.nat.R.lo}{18.1}  
\resdef{e1.qwen3-4b.a3.inf.nat.R.mean}{23.2}  
\resdef{e1.qwen3-4b.a3.inf.nat.R.std}{1.3}  
\resdef{e1.qwen3-4b.a3.inf.nat.dR.hi}{-19.1}  
\resdef{e1.qwen3-4b.a3.inf.nat.dR.lo}{-33.8}  
\resdef{e1.qwen3-4b.a3.inf.nat.dR.mean}{-26.5}  
\resdef{e1.qwen3-4b.a3.inf.nat.dR.std}{2.8}  
\resdef{e1.qwen3-4b.a3.inf.off.R.hi}{54.4}  
\resdef{e1.qwen3-4b.a3.inf.off.R.lo}{45.1}  
\resdef{e1.qwen3-4b.a3.inf.off.R.mean}{49.7}  
\resdef{e1.qwen3-4b.a3.inf.off.R.std}{1.8}  
\resdef{e1.qwen3-4b.a4.nli.nat.acc.hi}{78.4}  
\resdef{e1.qwen3-4b.a4.nli.nat.acc.lo}{63.7}  
\resdef{e1.qwen3-4b.a4.nli.nat.acc.mean}{71.2}  
\resdef{e1.qwen3-4b.a4.nli.nat.acc.std}{0.9}  
\resdef{e1.qwen3-4b.a4.nli.nat.dacc.hi}{20.7}  
\resdef{e1.qwen3-4b.a4.nli.nat.dacc.lo}{3.0}  
\resdef{e1.qwen3-4b.a4.nli.nat.dacc.mean}{12.0}  
\resdef{e1.qwen3-4b.a4.nli.nat.dacc.std}{3.2}  
\resdef{e1.qwen3-4b.a4.nli.off.acc.hi}{66.1}  
\resdef{e1.qwen3-4b.a4.nli.off.acc.lo}{52.0}  
\resdef{e1.qwen3-4b.a4.nli.off.acc.mean}{59.2}  
\resdef{e1.qwen3-4b.a4.nli.off.acc.std}{2.8}  
\resdef{e1.qwen3-4b.b2.yn.nat.acc.hi}{84.6}  
\resdef{e1.qwen3-4b.b2.yn.nat.acc.lo}{67.4}  
\resdef{e1.qwen3-4b.b2.yn.nat.acc.mean}{76.0}  
\resdef{e1.qwen3-4b.b2.yn.nat.acc.std}{0.6}  
\resdef{e1.qwen3-4b.b2.yn.nat.dacc.hi}{6.8}  
\resdef{e1.qwen3-4b.b2.yn.nat.dacc.lo}{-7.2}  
\resdef{e1.qwen3-4b.b2.yn.nat.dacc.mean}{-0.4}  
\resdef{e1.qwen3-4b.b2.yn.nat.dacc.std}{2.7}  
\resdef{e1.qwen3-4b.b2.yn.off.acc.hi}{84.6}  
\resdef{e1.qwen3-4b.b2.yn.off.acc.lo}{67.7}  
\resdef{e1.qwen3-4b.b2.yn.off.acc.mean}{76.3}  
\resdef{e1.qwen3-4b.b2.yn.off.acc.std}{2.2}  
\resdef{e1.qwen3-8b.a1.fel.off.lit.mean}{96.7}  
\resdef{e1.qwen3-8b.a1.ter.nat.R.hi}{0.0}  
\resdef{e1.qwen3-8b.a1.ter.nat.R.lo}{0.0}  
\resdef{e1.qwen3-8b.a1.ter.nat.R.mean}{0.0}  
\resdef{e1.qwen3-8b.a1.ter.nat.R.std}{0.0}  
\resdef{e1.qwen3-8b.a1.ter.nat.dR.hi}{0.0}  
\resdef{e1.qwen3-8b.a1.ter.nat.dR.lo}{-7.5}  
\resdef{e1.qwen3-8b.a1.ter.nat.dR.mean}{-2.5}  
\resdef{e1.qwen3-8b.a1.ter.nat.dR.std}{0.0}  
\resdef{e1.qwen3-8b.a1.ter.off.R.hi}{7.5}  
\resdef{e1.qwen3-8b.a1.ter.off.R.lo}{0.0}  
\resdef{e1.qwen3-8b.a1.ter.off.R.mean}{2.5}  
\resdef{e1.qwen3-8b.a1.ter.off.R.std}{0.0}  
\resdef{e1.qwen3-8b.a2.inf.nat.R.hi}{15.0}  
\resdef{e1.qwen3-8b.a2.inf.nat.R.lo}{6.2}  
\resdef{e1.qwen3-8b.a2.inf.nat.R.mean}{10.4}  
\resdef{e1.qwen3-8b.a2.inf.nat.R.std}{1.0}  
\resdef{e1.qwen3-8b.a2.inf.nat.dR.hi}{8.8}  
\resdef{e1.qwen3-8b.a2.inf.nat.dR.lo}{2.4}  
\resdef{e1.qwen3-8b.a2.inf.nat.dR.mean}{5.5}  
\resdef{e1.qwen3-8b.a2.inf.nat.dR.std}{1.4}  
\resdef{e1.qwen3-8b.a2.inf.off.R.hi}{8.4}  
\resdef{e1.qwen3-8b.a2.inf.off.R.lo}{2.0}  
\resdef{e1.qwen3-8b.a2.inf.off.R.mean}{4.9}  
\resdef{e1.qwen3-8b.a2.inf.off.R.std}{0.4}  
\resdef{e1.qwen3-8b.a3-conn.inf.c0128.dR.mean}{4.0}  
\resdef{e1.qwen3-8b.a3-conn.inf.c0512.dR.mean}{-15.2}  
\resdef{e1.qwen3-8b.a3-conn.inf.c1024.dR.mean}{-31.3}  
\resdef{e1.qwen3-8b.a3-conn.inf.nat.dR.hi}{-24.2}  
\resdef{e1.qwen3-8b.a3-conn.inf.nat.dR.lo}{-49.5}  
\resdef{e1.qwen3-8b.a3-conn.inf.nat.dR.mean}{-36.4}  
\resdef{e1.qwen3-8b.a3-conn.inf.nat.dR.std}{5.2}  
\resdef{e1.qwen3-8b.a3-conn.inf.nat.dRb0.mean}{-33.3}  
\resdef{e1.qwen3-8b.a3-conn.inf.p1024.dR.mean}{-34.3}  
\resdef{e1.qwen3-8b.a3-modal.inf.nat.dR.hi}{-2.1}  
\resdef{e1.qwen3-8b.a3-modal.inf.nat.dR.lo}{-22.9}  
\resdef{e1.qwen3-8b.a3-modal.inf.nat.dR.mean}{-12.5}  
\resdef{e1.qwen3-8b.a3-modal.inf.nat.dR.std}{6.2}  
\resdef{e1.qwen3-8b.a3-quant.inf.c0128.R.mean}{66.7}  
\resdef{e1.qwen3-8b.a3-quant.inf.c0128.dR.mean}{37.2}  
\resdef{e1.qwen3-8b.a3-quant.inf.c0512.dR.mean}{5.1}  
\resdef{e1.qwen3-8b.a3-quant.inf.c1024.dR.mean}{-21.8}  
\resdef{e1.qwen3-8b.a3-quant.inf.nat.dR.hi}{-16.7}  
\resdef{e1.qwen3-8b.a3-quant.inf.nat.dR.lo}{-41.0}  
\resdef{e1.qwen3-8b.a3-quant.inf.nat.dR.mean}{-28.2}  
\resdef{e1.qwen3-8b.a3-quant.inf.nat.dR.std}{5.9}  
\resdef{e1.qwen3-8b.a3-quant.inf.nat.dRb0.mean}{-39.7}  
\resdef{e1.qwen3-8b.a3-quant.inf.p1024.dR.mean}{-24.4}  
\resdef{e1.qwen3-8b.a3.inf.nat.R.hi}{16.9}  
\resdef{e1.qwen3-8b.a3.inf.nat.R.lo}{8.9}  
\resdef{e1.qwen3-8b.a3.inf.nat.R.mean}{12.7}  
\resdef{e1.qwen3-8b.a3.inf.nat.R.std}{1.2}  
\resdef{e1.qwen3-8b.a3.inf.nat.dR.hi}{-5.1}  
\resdef{e1.qwen3-8b.a3.inf.nat.dR.lo}{-14.7}  
\resdef{e1.qwen3-8b.a3.inf.nat.dR.mean}{-9.9}  
\resdef{e1.qwen3-8b.a3.inf.nat.dR.std}{2.4}  
\resdef{e1.qwen3-8b.a3.inf.off.R.hi}{27.4}  
\resdef{e1.qwen3-8b.a3.inf.off.R.lo}{18.1}  
\resdef{e1.qwen3-8b.a3.inf.off.R.mean}{22.6}  
\resdef{e1.qwen3-8b.a3.inf.off.R.std}{1.8}  
\resdef{e1.qwen3-8b.a4.nli.nat.acc.hi}{79.3}  
\resdef{e1.qwen3-8b.a4.nli.nat.acc.lo}{64.0}  
\resdef{e1.qwen3-8b.a4.nli.nat.acc.mean}{71.8}  
\resdef{e1.qwen3-8b.a4.nli.nat.acc.std}{0.5}  
\resdef{e1.qwen3-8b.a4.nli.nat.dacc.hi}{29.4}  
\resdef{e1.qwen3-8b.a4.nli.nat.dacc.lo}{15.0}  
\resdef{e1.qwen3-8b.a4.nli.nat.dacc.mean}{22.2}  
\resdef{e1.qwen3-8b.a4.nli.nat.dacc.std}{5.0}  
\resdef{e1.qwen3-8b.a4.nli.off.acc.hi}{56.8}  
\resdef{e1.qwen3-8b.a4.nli.off.acc.lo}{41.7}  
\resdef{e1.qwen3-8b.a4.nli.off.acc.mean}{49.5}  
\resdef{e1.qwen3-8b.a4.nli.off.acc.std}{4.5}  
\resdef{e1.qwen3-8b.b2.yn.nat.acc.hi}{83.5}  
\resdef{e1.qwen3-8b.b2.yn.nat.acc.lo}{66.7}  
\resdef{e1.qwen3-8b.b2.yn.nat.acc.mean}{75.3}  
\resdef{e1.qwen3-8b.b2.yn.nat.acc.std}{2.2}  
\resdef{e1.qwen3-8b.b2.yn.nat.dacc.hi}{14.7}  
\resdef{e1.qwen3-8b.b2.yn.nat.dacc.lo}{-1.1}  
\resdef{e1.qwen3-8b.b2.yn.nat.dacc.mean}{6.5}  
\resdef{e1.qwen3-8b.b2.yn.nat.dacc.std}{2.8}  
\resdef{e1.qwen3-8b.b2.yn.off.acc.hi}{77.4}  
\resdef{e1.qwen3-8b.b2.yn.off.acc.lo}{59.9}  
\resdef{e1.qwen3-8b.b2.yn.off.acc.mean}{68.8}  
\resdef{e1.qwen3-8b.b2.yn.off.acc.std}{1.1}  
\resdef{e1s.olmo3-32b-think.a2.inf.nat.dR.hi}{9.5}  
\resdef{e1s.olmo3-32b-think.a2.inf.nat.dR.lo}{3.0}  
\resdef{e1s.olmo3-32b-think.a2.inf.nat.dR.mean}{6.0}  
\resdef{e1s.olmo3-32b-think.a2.inf.nat.dR.std}{1.6}  
\resdef{e1s.olmo3-32b-think.a3-conn.inf.nat.dR.hi}{3.0}  
\resdef{e1s.olmo3-32b-think.a3-conn.inf.nat.dR.lo}{-6.1}  
\resdef{e1s.olmo3-32b-think.a3-conn.inf.nat.dR.mean}{-1.0}  
\resdef{e1s.olmo3-32b-think.a3-conn.inf.nat.dR.std}{3.1}  
\resdef{e1s.olmo3-32b-think.a3-modal.inf.nat.dR.hi}{41.1}  
\resdef{e1s.olmo3-32b-think.a3-modal.inf.nat.dR.lo}{19.3}  
\resdef{e1s.olmo3-32b-think.a3-modal.inf.nat.dR.mean}{29.7}  
\resdef{e1s.olmo3-32b-think.a3-modal.inf.nat.dR.std}{8.3}  
\resdef{e1s.olmo3-32b-think.a3-quant.inf.nat.dR.hi}{-3.8}  
\resdef{e1s.olmo3-32b-think.a3-quant.inf.nat.dR.lo}{-20.5}  
\resdef{e1s.olmo3-32b-think.a3-quant.inf.nat.dR.mean}{-11.5}  
\resdef{e1s.olmo3-32b-think.a3-quant.inf.nat.dR.std}{2.4}  
\resdef{e1s.olmo3-32b-think.a3.inf.nat.dR.hi}{24.4}  
\resdef{e1s.olmo3-32b-think.a3.inf.nat.dR.lo}{13.4}  
\resdef{e1s.olmo3-32b-think.a3.inf.nat.dR.mean}{19.0}  
\resdef{e1s.olmo3-32b-think.a3.inf.nat.dR.std}{3.9}  
\resdef{e1s.olmo3-32b-think.b2.yn.nat.dacc.hi}{13.9}  
\resdef{e1s.olmo3-32b-think.b2.yn.nat.dacc.lo}{-1.6}  
\resdef{e1s.olmo3-32b-think.b2.yn.nat.dacc.mean}{6.0}  
\resdef{e1s.olmo3-32b-think.b2.yn.nat.dacc.std}{2.3}  
\resdef{e1s.olmo3-7b-think.a2.inf.nat.dR.hi}{16.3}  
\resdef{e1s.olmo3-7b-think.a2.inf.nat.dR.lo}{6.8}  
\resdef{e1s.olmo3-7b-think.a2.inf.nat.dR.mean}{11.3}  
\resdef{e1s.olmo3-7b-think.a2.inf.nat.dR.std}{1.8}  
\resdef{e1s.olmo3-7b-think.a3-conn.inf.nat.dR.hi}{3.0}  
\resdef{e1s.olmo3-7b-think.a3-conn.inf.nat.dR.lo}{-14.1}  
\resdef{e1s.olmo3-7b-think.a3-conn.inf.nat.dR.mean}{-5.1}  
\resdef{e1s.olmo3-7b-think.a3-conn.inf.nat.dR.std}{4.6}  
\resdef{e1s.olmo3-7b-think.a3-modal.inf.nat.dR.hi}{6.2}  
\resdef{e1s.olmo3-7b-think.a3-modal.inf.nat.dR.lo}{-14.6}  
\resdef{e1s.olmo3-7b-think.a3-modal.inf.nat.dR.mean}{-4.2}  
\resdef{e1s.olmo3-7b-think.a3-modal.inf.nat.dR.std}{1.8}  
\resdef{e1s.olmo3-7b-think.a3-quant.inf.nat.dR.hi}{7.7}  
\resdef{e1s.olmo3-7b-think.a3-quant.inf.nat.dR.lo}{-2.6}  
\resdef{e1s.olmo3-7b-think.a3-quant.inf.nat.dR.mean}{2.6}  
\resdef{e1s.olmo3-7b-think.a3-quant.inf.nat.dR.std}{8.0}  
\resdef{e1s.olmo3-7b-think.a3.inf.nat.dR.hi}{6.0}  
\resdef{e1s.olmo3-7b-think.a3.inf.nat.dR.lo}{-1.4}  
\resdef{e1s.olmo3-7b-think.a3.inf.nat.dR.mean}{2.4}  
\resdef{e1s.olmo3-7b-think.a3.inf.nat.dR.std}{1.6}  
\resdef{e1s.olmo3-7b-think.b2.yn.nat.dacc.hi}{17.9}  
\resdef{e1s.olmo3-7b-think.b2.yn.nat.dacc.lo}{0.4}  
\resdef{e1s.olmo3-7b-think.b2.yn.nat.dacc.mean}{9.5}  
\resdef{e1s.olmo3-7b-think.b2.yn.nat.dacc.std}{3.6}  
\resdef{e1s.pragrest-8b.a3-conn.inf.nat.R.hi}{34.3}  
\resdef{e1s.pragrest-8b.a3-conn.inf.nat.R.lo}{10.1}  
\resdef{e1s.pragrest-8b.a3-conn.inf.nat.R.mean}{21.2}  
\resdef{e1s.pragrest-8b.a3-conn.inf.nat.R.std}{0.0}  
\resdef{e1s.pragrest-8b.a3-quant.inf.nat.R.hi}{32.1}  
\resdef{e1s.pragrest-8b.a3-quant.inf.nat.R.lo}{7.7}  
\resdef{e1s.pragrest-8b.a3-quant.inf.nat.R.mean}{19.2}  
\resdef{e1s.pragrest-8b.a3-quant.inf.nat.R.std}{3.8}  
\resdef{e1s.pragrest-8b.b2.yn.nat.acc.hi}{88.4}  
\resdef{e1s.pragrest-8b.b2.yn.nat.acc.lo}{75.1}  
\resdef{e1s.pragrest-8b.b2.yn.nat.acc.mean}{82.1}  
\resdef{e1s.pragrest-8b.b2.yn.nat.acc.std}{1.1}  
\resdef{e1s.qwen3-0p6b.a2.inf.nat.dR.hi}{-65.8}  
\resdef{e1s.qwen3-0p6b.a2.inf.nat.dR.lo}{-74.8}  
\resdef{e1s.qwen3-0p6b.a2.inf.nat.dR.mean}{-70.4}  
\resdef{e1s.qwen3-0p6b.a2.inf.nat.dR.std}{2.0}  
\resdef{e1s.qwen3-0p6b.a3-conn.inf.nat.dR.hi}{-74.7}  
\resdef{e1s.qwen3-0p6b.a3-conn.inf.nat.dR.lo}{-91.9}  
\resdef{e1s.qwen3-0p6b.a3-conn.inf.nat.dR.mean}{-83.8}  
\resdef{e1s.qwen3-0p6b.a3-conn.inf.nat.dR.std}{4.6}  
\resdef{e1s.qwen3-0p6b.a3-modal.inf.nat.dR.hi}{-49.0}  
\resdef{e1s.qwen3-0p6b.a3-modal.inf.nat.dR.lo}{-74.0}  
\resdef{e1s.qwen3-0p6b.a3-modal.inf.nat.dR.mean}{-62.5}  
\resdef{e1s.qwen3-0p6b.a3-modal.inf.nat.dR.std}{12.5}  
\resdef{e1s.qwen3-0p6b.a3-quant.inf.nat.dR.hi}{-32.1}  
\resdef{e1s.qwen3-0p6b.a3-quant.inf.nat.dR.lo}{-66.7}  
\resdef{e1s.qwen3-0p6b.a3-quant.inf.nat.dR.mean}{-50.0}  
\resdef{e1s.qwen3-0p6b.a3-quant.inf.nat.dR.std}{10.2}  
\resdef{e1s.qwen3-0p6b.a3.inf.nat.dR.hi}{-52.0}  
\resdef{e1s.qwen3-0p6b.a3.inf.nat.dR.lo}{-65.6}  
\resdef{e1s.qwen3-0p6b.a3.inf.nat.dR.mean}{-59.1}  
\resdef{e1s.qwen3-0p6b.a3.inf.nat.dR.std}{1.6}  
\resdef{e1s.qwen3-0p6b.b2.yn.nat.dacc.hi}{17.5}  
\resdef{e1s.qwen3-0p6b.b2.yn.nat.dacc.lo}{2.1}  
\resdef{e1s.qwen3-0p6b.b2.yn.nat.dacc.mean}{9.8}  
\resdef{e1s.qwen3-0p6b.b2.yn.nat.dacc.std}{3.2}  
\resdef{e1s.qwen3-1p7b.a2.inf.nat.dR.hi}{-19.4}  
\resdef{e1s.qwen3-1p7b.a2.inf.nat.dR.lo}{-31.3}  
\resdef{e1s.qwen3-1p7b.a2.inf.nat.dR.mean}{-25.4}  
\resdef{e1s.qwen3-1p7b.a2.inf.nat.dR.std}{7.3}  
\resdef{e1s.qwen3-1p7b.a3-conn.inf.nat.dR.hi}{-34.3}  
\resdef{e1s.qwen3-1p7b.a3-conn.inf.nat.dR.lo}{-60.6}  
\resdef{e1s.qwen3-1p7b.a3-conn.inf.nat.dR.mean}{-47.5}  
\resdef{e1s.qwen3-1p7b.a3-conn.inf.nat.dR.std}{4.6}  
\resdef{e1s.qwen3-1p7b.a3-modal.inf.nat.dR.hi}{-19.8}  
\resdef{e1s.qwen3-1p7b.a3-modal.inf.nat.dR.lo}{-46.9}  
\resdef{e1s.qwen3-1p7b.a3-modal.inf.nat.dR.mean}{-33.3}  
\resdef{e1s.qwen3-1p7b.a3-modal.inf.nat.dR.std}{9.5}  
\resdef{e1s.qwen3-1p7b.a3-quant.inf.nat.dR.hi}{5.1}  
\resdef{e1s.qwen3-1p7b.a3-quant.inf.nat.dR.lo}{-25.6}  
\resdef{e1s.qwen3-1p7b.a3-quant.inf.nat.dR.mean}{-10.3}  
\resdef{e1s.qwen3-1p7b.a3-quant.inf.nat.dR.std}{8.9}  
\resdef{e1s.qwen3-1p7b.a3.inf.nat.dR.hi}{-17.4}  
\resdef{e1s.qwen3-1p7b.a3.inf.nat.dR.lo}{-29.9}  
\resdef{e1s.qwen3-1p7b.a3.inf.nat.dR.mean}{-23.6}  
\resdef{e1s.qwen3-1p7b.a3.inf.nat.dR.std}{1.8}  
\resdef{e1s.qwen3-1p7b.b2.yn.nat.dacc.hi}{25.3}  
\resdef{e1s.qwen3-1p7b.b2.yn.nat.dacc.lo}{8.8}  
\resdef{e1s.qwen3-1p7b.b2.yn.nat.dacc.mean}{17.2}  
\resdef{e1s.qwen3-1p7b.b2.yn.nat.dacc.std}{0.6}  
\resdef{e1s.qwen3-30b-a3b.a2.inf.nat.dR.hi}{4.0}  
\resdef{e1s.qwen3-30b-a3b.a2.inf.nat.dR.lo}{-2.6}  
\resdef{e1s.qwen3-30b-a3b.a2.inf.nat.dR.mean}{0.7}  
\resdef{e1s.qwen3-30b-a3b.a2.inf.nat.dR.std}{2.9}  
\resdef{e1s.qwen3-30b-a3b.a3-conn.inf.nat.dR.hi}{-26.3}  
\resdef{e1s.qwen3-30b-a3b.a3-conn.inf.nat.dR.lo}{-51.5}  
\resdef{e1s.qwen3-30b-a3b.a3-conn.inf.nat.dR.mean}{-39.4}  
\resdef{e1s.qwen3-30b-a3b.a3-conn.inf.nat.dR.std}{6.1}  
\resdef{e1s.qwen3-30b-a3b.a3-modal.inf.nat.dR.hi}{-2.1}  
\resdef{e1s.qwen3-30b-a3b.a3-modal.inf.nat.dR.lo}{-36.5}  
\resdef{e1s.qwen3-30b-a3b.a3-modal.inf.nat.dR.mean}{-19.8}  
\resdef{e1s.qwen3-30b-a3b.a3-modal.inf.nat.dR.std}{6.5}  
\resdef{e1s.qwen3-30b-a3b.a3-quant.inf.nat.dR.hi}{-17.9}  
\resdef{e1s.qwen3-30b-a3b.a3-quant.inf.nat.dR.lo}{-39.7}  
\resdef{e1s.qwen3-30b-a3b.a3-quant.inf.nat.dR.mean}{-28.2}  
\resdef{e1s.qwen3-30b-a3b.a3-quant.inf.nat.dR.std}{2.2}  
\resdef{e1s.qwen3-30b-a3b.a3.inf.nat.dR.hi}{-9.4}  
\resdef{e1s.qwen3-30b-a3b.a3.inf.nat.dR.lo}{-22.2}  
\resdef{e1s.qwen3-30b-a3b.a3.inf.nat.dR.mean}{-15.9}  
\resdef{e1s.qwen3-30b-a3b.a3.inf.nat.dR.std}{3.4}  
\resdef{e1s.qwen3-30b-a3b.b2.yn.nat.dacc.hi}{16.8}  
\resdef{e1s.qwen3-30b-a3b.b2.yn.nat.dacc.lo}{2.5}  
\resdef{e1s.qwen3-30b-a3b.b2.yn.nat.dacc.mean}{9.5}  
\resdef{e1s.qwen3-30b-a3b.b2.yn.nat.dacc.std}{2.1}  
\resdef{e1s.qwen3-32b.a2.inf.nat.dR.hi}{7.3}  
\resdef{e1s.qwen3-32b.a2.inf.nat.dR.lo}{1.1}  
\resdef{e1s.qwen3-32b.a2.inf.nat.dR.mean}{4.0}  
\resdef{e1s.qwen3-32b.a2.inf.nat.dR.std}{2.4}  
\resdef{e1s.qwen3-32b.a3-conn.inf.nat.dR.hi}{-32.3}  
\resdef{e1s.qwen3-32b.a3-conn.inf.nat.dR.lo}{-60.6}  
\resdef{e1s.qwen3-32b.a3-conn.inf.nat.dR.mean}{-46.5}  
\resdef{e1s.qwen3-32b.a3-conn.inf.nat.dR.std}{4.6}  
\resdef{e1s.qwen3-32b.a3-modal.inf.nat.dR.hi}{-0.0}  
\resdef{e1s.qwen3-32b.a3-modal.inf.nat.dR.lo}{-22.9}  
\resdef{e1s.qwen3-32b.a3-modal.inf.nat.dR.mean}{-11.5}  
\resdef{e1s.qwen3-32b.a3-modal.inf.nat.dR.std}{7.9}  
\resdef{e1s.qwen3-32b.a3-quant.inf.nat.dR.hi}{-37.2}  
\resdef{e1s.qwen3-32b.a3-quant.inf.nat.dR.lo}{-65.4}  
\resdef{e1s.qwen3-32b.a3-quant.inf.nat.dR.mean}{-51.3}  
\resdef{e1s.qwen3-32b.a3-quant.inf.nat.dR.std}{8.0}  
\resdef{e1s.qwen3-32b.a3.inf.nat.dR.hi}{-12.3}  
\resdef{e1s.qwen3-32b.a3.inf.nat.dR.lo}{-24.8}  
\resdef{e1s.qwen3-32b.a3.inf.nat.dR.mean}{-18.5}  
\resdef{e1s.qwen3-32b.a3.inf.nat.dR.std}{1.4}  
\resdef{e1s.qwen3-32b.b2.yn.nat.dacc.hi}{8.8}  
\resdef{e1s.qwen3-32b.b2.yn.nat.dacc.lo}{-3.9}  
\resdef{e1s.qwen3-32b.b2.yn.nat.dacc.mean}{2.5}  
\resdef{e1s.qwen3-32b.b2.yn.nat.dacc.std}{3.2}  
\resdef{e1s.r1distill-14b.a2.inf.nat.dR.hi}{7.1}  
\resdef{e1s.r1distill-14b.a2.inf.nat.dR.lo}{0.9}  
\resdef{e1s.r1distill-14b.a2.inf.nat.dR.mean}{3.8}  
\resdef{e1s.r1distill-14b.a2.inf.nat.dR.std}{1.0}  
\resdef{e1s.r1distill-14b.a3-conn.inf.nat.dR.hi}{18.2}  
\resdef{e1s.r1distill-14b.a3-conn.inf.nat.dR.lo}{-4.0}  
\resdef{e1s.r1distill-14b.a3-conn.inf.nat.dR.mean}{7.1}  
\resdef{e1s.r1distill-14b.a3-conn.inf.nat.dR.std}{7.6}  
\resdef{e1s.r1distill-14b.a3-modal.inf.nat.dR.hi}{52.1}  
\resdef{e1s.r1distill-14b.a3-modal.inf.nat.dR.lo}{24.0}  
\resdef{e1s.r1distill-14b.a3-modal.inf.nat.dR.mean}{37.5}  
\resdef{e1s.r1distill-14b.a3-modal.inf.nat.dR.std}{5.4}  
\resdef{e1s.r1distill-14b.a3-quant.inf.nat.dR.hi}{2.6}  
\resdef{e1s.r1distill-14b.a3-quant.inf.nat.dR.lo}{-32.1}  
\resdef{e1s.r1distill-14b.a3-quant.inf.nat.dR.mean}{-14.1}  
\resdef{e1s.r1distill-14b.a3-quant.inf.nat.dR.std}{5.9}  
\resdef{e1s.r1distill-14b.a3.inf.nat.dR.hi}{23.6}  
\resdef{e1s.r1distill-14b.a3.inf.nat.dR.lo}{9.9}  
\resdef{e1s.r1distill-14b.a3.inf.nat.dR.mean}{16.8}  
\resdef{e1s.r1distill-14b.a3.inf.nat.dR.std}{3.1}  
\resdef{e1s.r1distill-14b.b2.yn.nat.dacc.hi}{15.8}  
\resdef{e1s.r1distill-14b.b2.yn.nat.dacc.lo}{1.8}  
\resdef{e1s.r1distill-14b.b2.yn.nat.dacc.mean}{8.8}  
\resdef{e1s.r1distill-14b.b2.yn.nat.dacc.std}{3.2}  
\resdef{e1s.r1distill-7b.a2.inf.nat.dR.hi}{-5.7}  
\resdef{e1s.r1distill-7b.a2.inf.nat.dR.lo}{-19.9}  
\resdef{e1s.r1distill-7b.a2.inf.nat.dR.mean}{-13.0}  
\resdef{e1s.r1distill-7b.a2.inf.nat.dR.std}{4.8}  
\resdef{e1s.r1distill-7b.a3-conn.inf.nat.dR.hi}{-18.2}  
\resdef{e1s.r1distill-7b.a3-conn.inf.nat.dR.lo}{-51.5}  
\resdef{e1s.r1distill-7b.a3-conn.inf.nat.dR.mean}{-35.4}  
\resdef{e1s.r1distill-7b.a3-conn.inf.nat.dR.std}{14.3}  
\resdef{e1s.r1distill-7b.a3-modal.inf.nat.dR.hi}{9.4}  
\resdef{e1s.r1distill-7b.a3-modal.inf.nat.dR.lo}{-24.0}  
\resdef{e1s.r1distill-7b.a3-modal.inf.nat.dR.mean}{-7.3}  
\resdef{e1s.r1distill-7b.a3-modal.inf.nat.dR.std}{10.0}  
\resdef{e1s.r1distill-7b.a3-quant.inf.nat.dR.hi}{-5.1}  
\resdef{e1s.r1distill-7b.a3-quant.inf.nat.dR.lo}{-38.5}  
\resdef{e1s.r1distill-7b.a3-quant.inf.nat.dR.mean}{-21.8}  
\resdef{e1s.r1distill-7b.a3-quant.inf.nat.dR.std}{4.4}  
\resdef{e1s.r1distill-7b.a3.inf.nat.dR.hi}{-4.6}  
\resdef{e1s.r1distill-7b.a3.inf.nat.dR.lo}{-18.8}  
\resdef{e1s.r1distill-7b.a3.inf.nat.dR.mean}{-11.8}  
\resdef{e1s.r1distill-7b.a3.inf.nat.dR.std}{3.6}  
\resdef{e1s.r1distill-7b.b2.yn.nat.dacc.hi}{16.1}  
\resdef{e1s.r1distill-7b.b2.yn.nat.dacc.lo}{-1.8}  
\resdef{e1s.r1distill-7b.b2.yn.nat.dacc.mean}{7.0}  
\resdef{e1s.r1distill-7b.b2.yn.nat.dacc.std}{2.2}  
\resdef{a1ctx.gpt-oss-120b.a1ctx-fel.fel.high.ksi.mean}{62.5}  
\resdef{a1ctx.gpt-oss-120b.a1ctx-fel.fel.low.ksi.mean}{84.0}  
\resdef{a1ctx.qwen3-32b.a1ctx-fel.fel.nat.ksi.mean}{4.2}  
\resdef{e3.gpt-oss-120b.a2.inf.high.dR.hi}{9.9}  
\resdef{e3.gpt-oss-120b.a2.inf.high.dR.lo}{3.5}  
\resdef{e3.gpt-oss-120b.a2.inf.high.dR.mean}{6.7}  
\resdef{e3.gpt-oss-120b.a2.inf.high.dR.std}{1.7}  
\resdef{e3.gpt-oss-120b.a3-conn.inf.high.dR.hi}{-2.5}  
\resdef{e3.gpt-oss-120b.a3-conn.inf.high.dR.lo}{-20.7}  
\resdef{e3.gpt-oss-120b.a3-conn.inf.high.dR.mean}{-11.1}  
\resdef{e3.gpt-oss-120b.a3-conn.inf.high.dR.std}{5.6}  
\resdef{e3.gpt-oss-120b.a3-modal.inf.high.dR.hi}{-15.6}  
\resdef{e3.gpt-oss-120b.a3-modal.inf.high.dR.lo}{-35.4}  
\resdef{e3.gpt-oss-120b.a3-modal.inf.high.dR.mean}{-25.0}  
\resdef{e3.gpt-oss-120b.a3-modal.inf.high.dR.std}{7.7}  
\resdef{e3.gpt-oss-120b.a3-quant.inf.high.dR.hi}{6.4}  
\resdef{e3.gpt-oss-120b.a3-quant.inf.high.dR.lo}{-18.0}  
\resdef{e3.gpt-oss-120b.a3-quant.inf.high.dR.mean}{-6.4}  
\resdef{e3.gpt-oss-120b.a3-quant.inf.high.dR.std}{7.2}  
\resdef{e3.gpt-oss-120b.a3.inf.high.dR.hi}{-7.7}  
\resdef{e3.gpt-oss-120b.a3.inf.high.dR.lo}{-16.4}  
\resdef{e3.gpt-oss-120b.a3.inf.high.dR.mean}{-12.0}  
\resdef{e3.gpt-oss-120b.a3.inf.high.dR.std}{1.9}  
\resdef{e3.gpt-oss-120b.b2.yn.high.dacc.hi}{7.0}  
\resdef{e3.gpt-oss-120b.b2.yn.high.dacc.lo}{-0.5}  
\resdef{e3.gpt-oss-120b.b2.yn.high.dacc.mean}{3.0}  
\resdef{e3.gpt-oss-120b.b2.yn.high.dacc.std}{1.4}  
\resdef{e3.gpt-oss-20b.a2.inf.high.dR.hi}{1.9}  
\resdef{e3.gpt-oss-20b.a2.inf.high.dR.lo}{-2.5}  
\resdef{e3.gpt-oss-20b.a2.inf.high.dR.mean}{-0.3}  
\resdef{e3.gpt-oss-20b.a2.inf.high.dR.std}{2.7}  
\resdef{e3.gpt-oss-20b.a3-conn.inf.high.dR.hi}{-16.2}  
\resdef{e3.gpt-oss-20b.a3-conn.inf.high.dR.lo}{-33.3}  
\resdef{e3.gpt-oss-20b.a3-conn.inf.high.dR.mean}{-24.7}  
\resdef{e3.gpt-oss-20b.a3-conn.inf.high.dR.std}{10.0}  
\resdef{e3.gpt-oss-20b.a3-modal.inf.high.dR.hi}{-12.0}  
\resdef{e3.gpt-oss-20b.a3-modal.inf.high.dR.lo}{-29.2}  
\resdef{e3.gpt-oss-20b.a3-modal.inf.high.dR.mean}{-20.3}  
\resdef{e3.gpt-oss-20b.a3-modal.inf.high.dR.std}{8.8}  
\resdef{e3.gpt-oss-20b.a3-quant.inf.high.dR.hi}{-19.2}  
\resdef{e3.gpt-oss-20b.a3-quant.inf.high.dR.lo}{-42.9}  
\resdef{e3.gpt-oss-20b.a3-quant.inf.high.dR.mean}{-30.8}  
\resdef{e3.gpt-oss-20b.a3-quant.inf.high.dR.std}{6.0}  
\resdef{e3.gpt-oss-20b.a3.inf.high.dR.hi}{-12.7}  
\resdef{e3.gpt-oss-20b.a3.inf.high.dR.lo}{-20.7}  
\resdef{e3.gpt-oss-20b.a3.inf.high.dR.mean}{-16.7}  
\resdef{e3.gpt-oss-20b.a3.inf.high.dR.std}{3.2}  
\resdef{e3.gpt-oss-20b.b2.yn.high.dacc.hi}{7.4}  
\resdef{e3.gpt-oss-20b.b2.yn.high.dacc.lo}{-0.7}  
\resdef{e3.gpt-oss-20b.b2.yn.high.dacc.mean}{3.2}  
\resdef{e3.gpt-oss-20b.b2.yn.high.dacc.std}{2.1}  
\resdef{e2.olmo2-7b-base.a2.inf.lp4.R.hi}{25.4}  
\resdef{e2.olmo2-7b-base.a2.inf.lp4.R.lo}{14.1}  
\resdef{e2.olmo2-7b-base.a2.inf.lp4.R.mean}{19.6}  
\resdef{e2.olmo2-7b-base.a2.inf.lp4.R.std}{1.5}  
\resdef{e2.olmo2-7b-base.a3-conn.inf.lp4.R.hi}{42.4}  
\resdef{e2.olmo2-7b-base.a3-conn.inf.lp4.R.lo}{21.2}  
\resdef{e2.olmo2-7b-base.a3-conn.inf.lp4.R.mean}{31.3}  
\resdef{e2.olmo2-7b-base.a3-conn.inf.lp4.R.std}{3.5}  
\resdef{e2.olmo2-7b-base.a3-quant.inf.lp4.R.hi}{59.0}  
\resdef{e2.olmo2-7b-base.a3-quant.inf.lp4.R.lo}{28.2}  
\resdef{e2.olmo2-7b-base.a3-quant.inf.lp4.R.mean}{43.6}  
\resdef{e2.olmo2-7b-base.a3-quant.inf.lp4.R.std}{2.2}  
\resdef{e2.olmo2-7b-base.b2.yn.lp4.acc.hi}{78.6}  
\resdef{e2.olmo2-7b-base.b2.yn.lp4.acc.lo}{61.1}  
\resdef{e2.olmo2-7b-base.b2.yn.lp4.acc.mean}{69.8}  
\resdef{e2.olmo2-7b-base.b2.yn.lp4.acc.std}{1.6}  
\resdef{e2.olmo2-7b-dpo.a2.inf.lp4.R.hi}{38.6}  
\resdef{e2.olmo2-7b-dpo.a2.inf.lp4.R.lo}{27.6}  
\resdef{e2.olmo2-7b-dpo.a2.inf.lp4.R.mean}{32.9}  
\resdef{e2.olmo2-7b-dpo.a2.inf.lp4.R.std}{2.8}  
\resdef{e2.olmo2-7b-dpo.a3-conn.inf.lp4.R.hi}{27.3}  
\resdef{e2.olmo2-7b-dpo.a3-conn.inf.lp4.R.lo}{11.1}  
\resdef{e2.olmo2-7b-dpo.a3-conn.inf.lp4.R.mean}{19.2}  
\resdef{e2.olmo2-7b-dpo.a3-conn.inf.lp4.R.std}{1.8}  
\resdef{e2.olmo2-7b-dpo.a3-quant.inf.lp4.R.hi}{53.8}  
\resdef{e2.olmo2-7b-dpo.a3-quant.inf.lp4.R.lo}{21.8}  
\resdef{e2.olmo2-7b-dpo.a3-quant.inf.lp4.R.mean}{37.2}  
\resdef{e2.olmo2-7b-dpo.a3-quant.inf.lp4.R.std}{4.4}  
\resdef{e2.olmo2-7b-dpo.b2.yn.lp4.acc.hi}{78.9}  
\resdef{e2.olmo2-7b-dpo.b2.yn.lp4.acc.lo}{62.1}  
\resdef{e2.olmo2-7b-dpo.b2.yn.lp4.acc.mean}{70.5}  
\resdef{e2.olmo2-7b-dpo.b2.yn.lp4.acc.std}{1.1}  
\resdef{e2.olmo2-7b-instruct.a2.inf.lp4.R.hi}{40.0}  
\resdef{e2.olmo2-7b-instruct.a2.inf.lp4.R.lo}{29.1}  
\resdef{e2.olmo2-7b-instruct.a2.inf.lp4.R.mean}{34.4}  
\resdef{e2.olmo2-7b-instruct.a2.inf.lp4.R.std}{2.6}  
\resdef{e2.olmo2-7b-instruct.a3-conn.inf.lp4.R.hi}{25.3}  
\resdef{e2.olmo2-7b-instruct.a3-conn.inf.lp4.R.lo}{8.1}  
\resdef{e2.olmo2-7b-instruct.a3-conn.inf.lp4.R.mean}{16.2}  
\resdef{e2.olmo2-7b-instruct.a3-conn.inf.lp4.R.std}{3.5}  
\resdef{e2.olmo2-7b-instruct.a3-quant.inf.lp4.R.hi}{38.5}  
\resdef{e2.olmo2-7b-instruct.a3-quant.inf.lp4.R.lo}{10.3}  
\resdef{e2.olmo2-7b-instruct.a3-quant.inf.lp4.R.mean}{24.4}  
\resdef{e2.olmo2-7b-instruct.a3-quant.inf.lp4.R.std}{2.2}  
\resdef{e2.olmo2-7b-instruct.b2.yn.lp4.acc.hi}{78.6}  
\resdef{e2.olmo2-7b-instruct.b2.yn.lp4.acc.lo}{62.1}  
\resdef{e2.olmo2-7b-instruct.b2.yn.lp4.acc.mean}{70.5}  
\resdef{e2.olmo2-7b-instruct.b2.yn.lp4.acc.std}{2.1}  
\resdef{e2.olmo2-7b-sft.a2.inf.lp4.R.hi}{39.3}  
\resdef{e2.olmo2-7b-sft.a2.inf.lp4.R.lo}{26.5}  
\resdef{e2.olmo2-7b-sft.a2.inf.lp4.R.mean}{32.9}  
\resdef{e2.olmo2-7b-sft.a2.inf.lp4.R.std}{2.0}  
\resdef{e2.olmo2-7b-sft.a3-conn.inf.lp4.R.hi}{42.4}  
\resdef{e2.olmo2-7b-sft.a3-conn.inf.lp4.R.lo}{21.2}  
\resdef{e2.olmo2-7b-sft.a3-conn.inf.lp4.R.mean}{31.3}  
\resdef{e2.olmo2-7b-sft.a3-conn.inf.lp4.R.std}{4.6}  
\resdef{e2.olmo2-7b-sft.a3-quant.inf.lp4.R.hi}{56.4}  
\resdef{e2.olmo2-7b-sft.a3-quant.inf.lp4.R.lo}{28.2}  
\resdef{e2.olmo2-7b-sft.a3-quant.inf.lp4.R.mean}{42.3}  
\resdef{e2.olmo2-7b-sft.a3-quant.inf.lp4.R.std}{0.0}  
\resdef{e2.olmo2-7b-sft.b2.yn.lp4.acc.hi}{81.4}  
\resdef{e2.olmo2-7b-sft.b2.yn.lp4.acc.lo}{64.6}  
\resdef{e2.olmo2-7b-sft.b2.yn.lp4.acc.mean}{73.3}  
\resdef{e2.olmo2-7b-sft.b2.yn.lp4.acc.std}{0.6}  
\resdef{e2.olmo3-7b-base.a2.inf.lp4.R.hi}{46.6}  
\resdef{e2.olmo3-7b-base.a2.inf.lp4.R.lo}{32.4}  
\resdef{e2.olmo3-7b-base.a2.inf.lp4.R.mean}{39.5}  
\resdef{e2.olmo3-7b-base.a2.inf.lp4.R.std}{0.4}  
\resdef{e2.olmo3-7b-base.a3-conn.inf.lp4.R.hi}{86.9}  
\resdef{e2.olmo3-7b-base.a3-conn.inf.lp4.R.lo}{61.6}  
\resdef{e2.olmo3-7b-base.a3-conn.inf.lp4.R.mean}{74.7}  
\resdef{e2.olmo3-7b-base.a3-conn.inf.lp4.R.std}{1.8}  
\resdef{e2.olmo3-7b-base.a3-quant.inf.lp4.R.hi}{89.7}  
\resdef{e2.olmo3-7b-base.a3-quant.inf.lp4.R.lo}{59.0}  
\resdef{e2.olmo3-7b-base.a3-quant.inf.lp4.R.mean}{75.6}  
\resdef{e2.olmo3-7b-base.a3-quant.inf.lp4.R.std}{4.4}  
\resdef{e2.olmo3-7b-base.b2.yn.lp4.acc.hi}{75.1}  
\resdef{e2.olmo3-7b-base.b2.yn.lp4.acc.lo}{58.9}  
\resdef{e2.olmo3-7b-base.b2.yn.lp4.acc.mean}{67.0}  
\resdef{e2.olmo3-7b-base.b2.yn.lp4.acc.std}{4.0}  
\resdef{e2.olmo3-rlzero-general-final.a2.inf.lp4.R.hi}{53.0}  
\resdef{e2.olmo3-rlzero-general-final.a2.inf.lp4.R.lo}{37.5}  
\resdef{e2.olmo3-rlzero-general-final.a2.inf.lp4.R.mean}{45.3}  
\resdef{e2.olmo3-rlzero-general-final.a2.inf.lp4.R.std}{0.8}  
\resdef{e2.olmo3-rlzero-general-final.a3-conn.inf.lp4.R.hi}{85.9}  
\resdef{e2.olmo3-rlzero-general-final.a3-conn.inf.lp4.R.lo}{59.6}  
\resdef{e2.olmo3-rlzero-general-final.a3-conn.inf.lp4.R.mean}{73.7}  
\resdef{e2.olmo3-rlzero-general-final.a3-conn.inf.lp4.R.std}{1.8}  
\resdef{e2.olmo3-rlzero-general-final.a3-quant.inf.lp4.R.hi}{92.3}  
\resdef{e2.olmo3-rlzero-general-final.a3-quant.inf.lp4.R.lo}{62.8}  
\resdef{e2.olmo3-rlzero-general-final.a3-quant.inf.lp4.R.mean}{78.2}  
\resdef{e2.olmo3-rlzero-general-final.a3-quant.inf.lp4.R.std}{2.2}  
\resdef{e2.olmo3-rlzero-general-final.b2.yn.lp4.acc.hi}{74.0}  
\resdef{e2.olmo3-rlzero-general-final.b2.yn.lp4.acc.lo}{57.9}  
\resdef{e2.olmo3-rlzero-general-final.b2.yn.lp4.acc.mean}{66.0}  
\resdef{e2.olmo3-rlzero-general-final.b2.yn.lp4.acc.std}{3.0}  
\resdef{e2.olmo3-rlzero-if-final.a2.inf.lp4.R.hi}{43.0}  
\resdef{e2.olmo3-rlzero-if-final.a2.inf.lp4.R.lo}{28.0}  
\resdef{e2.olmo3-rlzero-if-final.a2.inf.lp4.R.mean}{35.3}  
\resdef{e2.olmo3-rlzero-if-final.a2.inf.lp4.R.std}{1.4}  
\resdef{e2.olmo3-rlzero-if-final.a3-conn.inf.lp4.R.hi}{93.9}  
\resdef{e2.olmo3-rlzero-if-final.a3-conn.inf.lp4.R.lo}{71.7}  
\resdef{e2.olmo3-rlzero-if-final.a3-conn.inf.lp4.R.mean}{83.8}  
\resdef{e2.olmo3-rlzero-if-final.a3-conn.inf.lp4.R.std}{1.8}  
\resdef{e2.olmo3-rlzero-if-final.a3-quant.inf.lp4.R.hi}{100.0}  
\resdef{e2.olmo3-rlzero-if-final.a3-quant.inf.lp4.R.lo}{76.9}  
\resdef{e2.olmo3-rlzero-if-final.a3-quant.inf.lp4.R.mean}{89.7}  
\resdef{e2.olmo3-rlzero-if-final.a3-quant.inf.lp4.R.std}{2.2}  
\resdef{e2.olmo3-rlzero-if-final.b2.yn.lp4.acc.hi}{78.6}  
\resdef{e2.olmo3-rlzero-if-final.b2.yn.lp4.acc.lo}{61.8}  
\resdef{e2.olmo3-rlzero-if-final.b2.yn.lp4.acc.mean}{70.5}  
\resdef{e2.olmo3-rlzero-if-final.b2.yn.lp4.acc.std}{1.1}  
\resdef{e2.olmo3-rlzero-math-final.a2.inf.lp4.R.hi}{45.9}  
\resdef{e2.olmo3-rlzero-math-final.a2.inf.lp4.R.lo}{31.8}  
\resdef{e2.olmo3-rlzero-math-final.a2.inf.lp4.R.mean}{38.9}  
\resdef{e2.olmo3-rlzero-math-final.a2.inf.lp4.R.std}{1.7}  
\resdef{e2.olmo3-rlzero-math-final.a3-conn.inf.lp4.R.hi}{88.9}  
\resdef{e2.olmo3-rlzero-math-final.a3-conn.inf.lp4.R.lo}{65.7}  
\resdef{e2.olmo3-rlzero-math-final.a3-conn.inf.lp4.R.mean}{77.8}  
\resdef{e2.olmo3-rlzero-math-final.a3-conn.inf.lp4.R.std}{1.8}  
\resdef{e2.olmo3-rlzero-math-final.a3-quant.inf.lp4.R.hi}{92.3}  
\resdef{e2.olmo3-rlzero-math-final.a3-quant.inf.lp4.R.lo}{62.8}  
\resdef{e2.olmo3-rlzero-math-final.a3-quant.inf.lp4.R.mean}{78.2}  
\resdef{e2.olmo3-rlzero-math-final.a3-quant.inf.lp4.R.std}{2.2}  
\resdef{e2.olmo3-rlzero-math-final.b2.yn.lp4.acc.hi}{75.8}  
\resdef{e2.olmo3-rlzero-math-final.b2.yn.lp4.acc.lo}{59.3}  
\resdef{e2.olmo3-rlzero-math-final.b2.yn.lp4.acc.mean}{67.7}  
\resdef{e2.olmo3-rlzero-math-final.b2.yn.lp4.acc.std}{3.4}  
\resdef{e2.olmo3-rlzero-math-s0100.a2.inf.lp4.R.hi}{47.9}  
\resdef{e2.olmo3-rlzero-math-s0100.a2.inf.lp4.R.lo}{33.6}  
\resdef{e2.olmo3-rlzero-math-s0100.a2.inf.lp4.R.mean}{40.8}  
\resdef{e2.olmo3-rlzero-math-s0100.a2.inf.lp4.R.std}{1.0}  
\resdef{e2.olmo3-rlzero-math-s0100.a3-conn.inf.lp4.R.hi}{86.9}  
\resdef{e2.olmo3-rlzero-math-s0100.a3-conn.inf.lp4.R.lo}{61.6}  
\resdef{e2.olmo3-rlzero-math-s0100.a3-conn.inf.lp4.R.mean}{74.7}  
\resdef{e2.olmo3-rlzero-math-s0100.a3-conn.inf.lp4.R.std}{1.8}  
\resdef{e2.olmo3-rlzero-math-s0100.a3-quant.inf.lp4.R.hi}{89.7}  
\resdef{e2.olmo3-rlzero-math-s0100.a3-quant.inf.lp4.R.lo}{59.0}  
\resdef{e2.olmo3-rlzero-math-s0100.a3-quant.inf.lp4.R.mean}{75.6}  
\resdef{e2.olmo3-rlzero-math-s0100.a3-quant.inf.lp4.R.std}{4.4}  
\resdef{e2.olmo3-rlzero-math-s0100.b2.yn.lp4.acc.hi}{75.1}  
\resdef{e2.olmo3-rlzero-math-s0100.b2.yn.lp4.acc.lo}{58.9}  
\resdef{e2.olmo3-rlzero-math-s0100.b2.yn.lp4.acc.mean}{67.0}  
\resdef{e2.olmo3-rlzero-math-s0100.b2.yn.lp4.acc.std}{4.0}  
\resdef{e2.olmo3-rlzero-mix-final.a2.inf.lp4.R.hi}{58.9}  
\resdef{e2.olmo3-rlzero-mix-final.a2.inf.lp4.R.lo}{43.7}  
\resdef{e2.olmo3-rlzero-mix-final.a2.inf.lp4.R.mean}{51.4}  
\resdef{e2.olmo3-rlzero-mix-final.a2.inf.lp4.R.std}{1.5}  
\resdef{e2.olmo3-rlzero-mix-final.a3-conn.inf.lp4.R.hi}{93.9}  
\resdef{e2.olmo3-rlzero-mix-final.a3-conn.inf.lp4.R.lo}{69.7}  
\resdef{e2.olmo3-rlzero-mix-final.a3-conn.inf.lp4.R.mean}{82.8}  
\resdef{e2.olmo3-rlzero-mix-final.a3-conn.inf.lp4.R.std}{1.8}  
\resdef{e2.olmo3-rlzero-mix-final.a3-quant.inf.lp4.R.hi}{100.0}  
\resdef{e2.olmo3-rlzero-mix-final.a3-quant.inf.lp4.R.lo}{88.5}  
\resdef{e2.olmo3-rlzero-mix-final.a3-quant.inf.lp4.R.mean}{94.9}  
\resdef{e2.olmo3-rlzero-mix-final.a3-quant.inf.lp4.R.std}{4.4}  
\resdef{e2.olmo3-rlzero-mix-final.b2.yn.lp4.acc.hi}{76.8}  
\resdef{e2.olmo3-rlzero-mix-final.b2.yn.lp4.acc.lo}{60.7}  
\resdef{e2.olmo3-rlzero-mix-final.b2.yn.lp4.acc.mean}{68.8}  
\resdef{e2.olmo3-rlzero-mix-final.b2.yn.lp4.acc.std}{3.4}  
\resdef{e2.olmo3-think-dpo.a2.inf.lp4.R.hi}{7.7}  
\resdef{e2.olmo3-think-dpo.a2.inf.lp4.R.lo}{1.8}  
\resdef{e2.olmo3-think-dpo.a2.inf.lp4.R.mean}{4.6}  
\resdef{e2.olmo3-think-dpo.a2.inf.lp4.R.std}{0.7}  
\resdef{e2.olmo3-think-dpo.a3-conn.inf.lp4.R.hi}{64.6}  
\resdef{e2.olmo3-think-dpo.a3-conn.inf.lp4.R.lo}{33.3}  
\resdef{e2.olmo3-think-dpo.a3-conn.inf.lp4.R.mean}{48.5}  
\resdef{e2.olmo3-think-dpo.a3-conn.inf.lp4.R.std}{5.2}  
\resdef{e2.olmo3-think-dpo.a3-quant.inf.lp4.R.hi}{70.5}  
\resdef{e2.olmo3-think-dpo.a3-quant.inf.lp4.R.lo}{37.2}  
\resdef{e2.olmo3-think-dpo.a3-quant.inf.lp4.R.mean}{53.8}  
\resdef{e2.olmo3-think-dpo.a3-quant.inf.lp4.R.std}{3.8}  
\resdef{e2.olmo3-think-dpo.b2.yn.lp4.acc.hi}{82.8}  
\resdef{e2.olmo3-think-dpo.b2.yn.lp4.acc.lo}{66.3}  
\resdef{e2.olmo3-think-dpo.b2.yn.lp4.acc.mean}{75.1}  
\resdef{e2.olmo3-think-dpo.b2.yn.lp4.acc.std}{2.4}  
\resdef{e2.olmo3-think-s0100.a2.inf.lp4.R.hi}{8.8}  
\resdef{e2.olmo3-think-s0100.a2.inf.lp4.R.lo}{2.2}  
\resdef{e2.olmo3-think-s0100.a2.inf.lp4.R.mean}{5.3}  
\resdef{e2.olmo3-think-s0100.a2.inf.lp4.R.std}{0.7}  
\resdef{e2.olmo3-think-s0100.a3-conn.inf.lp4.R.hi}{67.7}  
\resdef{e2.olmo3-think-s0100.a3-conn.inf.lp4.R.lo}{35.4}  
\resdef{e2.olmo3-think-s0100.a3-conn.inf.lp4.R.mean}{51.5}  
\resdef{e2.olmo3-think-s0100.a3-conn.inf.lp4.R.std}{0.0}  
\resdef{e2.olmo3-think-s0100.a3-quant.inf.lp4.R.hi}{65.4}  
\resdef{e2.olmo3-think-s0100.a3-quant.inf.lp4.R.lo}{29.5}  
\resdef{e2.olmo3-think-s0100.a3-quant.inf.lp4.R.mean}{47.4}  
\resdef{e2.olmo3-think-s0100.a3-quant.inf.lp4.R.std}{2.2}  
\resdef{e2.olmo3-think-s0100.b2.yn.lp4.acc.hi}{82.8}  
\resdef{e2.olmo3-think-s0100.b2.yn.lp4.acc.lo}{66.7}  
\resdef{e2.olmo3-think-s0100.b2.yn.lp4.acc.mean}{75.1}  
\resdef{e2.olmo3-think-s0100.b2.yn.lp4.acc.std}{3.4}  
\resdef{e2.olmo3-think-s0500.a2.inf.lp4.R.hi}{8.2}  
\resdef{e2.olmo3-think-s0500.a2.inf.lp4.R.lo}{2.0}  
\resdef{e2.olmo3-think-s0500.a2.inf.lp4.R.mean}{4.9}  
\resdef{e2.olmo3-think-s0500.a2.inf.lp4.R.std}{0.4}  
\resdef{e2.olmo3-think-s0500.a3-conn.inf.lp4.R.hi}{66.7}  
\resdef{e2.olmo3-think-s0500.a3-conn.inf.lp4.R.lo}{35.4}  
\resdef{e2.olmo3-think-s0500.a3-conn.inf.lp4.R.mean}{50.5}  
\resdef{e2.olmo3-think-s0500.a3-conn.inf.lp4.R.std}{3.5}  
\resdef{e2.olmo3-think-s0500.a3-quant.inf.lp4.R.hi}{59.0}  
\resdef{e2.olmo3-think-s0500.a3-quant.inf.lp4.R.lo}{24.4}  
\resdef{e2.olmo3-think-s0500.a3-quant.inf.lp4.R.mean}{41.0}  
\resdef{e2.olmo3-think-s0500.a3-quant.inf.lp4.R.std}{4.4}  
\resdef{e2.olmo3-think-s0500.b2.yn.lp4.acc.hi}{83.5}  
\resdef{e2.olmo3-think-s0500.b2.yn.lp4.acc.lo}{67.4}  
\resdef{e2.olmo3-think-s0500.b2.yn.lp4.acc.mean}{75.8}  
\resdef{e2.olmo3-think-s0500.b2.yn.lp4.acc.std}{3.6}  
\resdef{e2.olmo3-think-s1375.a2.inf.lp4.R.hi}{7.7}  
\resdef{e2.olmo3-think-s1375.a2.inf.lp4.R.lo}{1.3}  
\resdef{e2.olmo3-think-s1375.a2.inf.lp4.R.mean}{4.4}  
\resdef{e2.olmo3-think-s1375.a2.inf.lp4.R.std}{0.4}  
\resdef{e2.olmo3-think-s1375.a3-conn.inf.lp4.R.hi}{57.6}  
\resdef{e2.olmo3-think-s1375.a3-conn.inf.lp4.R.lo}{28.3}  
\resdef{e2.olmo3-think-s1375.a3-conn.inf.lp4.R.mean}{42.4}  
\resdef{e2.olmo3-think-s1375.a3-conn.inf.lp4.R.std}{3.0}  
\resdef{e2.olmo3-think-s1375.a3-quant.inf.lp4.R.hi}{48.7}  
\resdef{e2.olmo3-think-s1375.a3-quant.inf.lp4.R.lo}{16.7}  
\resdef{e2.olmo3-think-s1375.a3-quant.inf.lp4.R.mean}{32.1}  
\resdef{e2.olmo3-think-s1375.a3-quant.inf.lp4.R.std}{4.4}  
\resdef{e2.olmo3-think-s1375.b2.yn.lp4.acc.hi}{82.5}  
\resdef{e2.olmo3-think-s1375.b2.yn.lp4.acc.lo}{66.3}  
\resdef{e2.olmo3-think-s1375.b2.yn.lp4.acc.mean}{74.7}  
\resdef{e2.olmo3-think-s1375.b2.yn.lp4.acc.std}{2.8}  
\resdef{e2.olmo3-think-sft.a2.inf.lp4.R.hi}{8.8}  
\resdef{e2.olmo3-think-sft.a2.inf.lp4.R.lo}{2.0}  
\resdef{e2.olmo3-think-sft.a2.inf.lp4.R.mean}{5.3}  
\resdef{e2.olmo3-think-sft.a2.inf.lp4.R.std}{0.7}  
\resdef{e2.olmo3-think-sft.a3-conn.inf.lp4.R.hi}{65.7}  
\resdef{e2.olmo3-think-sft.a3-conn.inf.lp4.R.lo}{35.4}  
\resdef{e2.olmo3-think-sft.a3-conn.inf.lp4.R.mean}{50.5}  
\resdef{e2.olmo3-think-sft.a3-conn.inf.lp4.R.std}{3.5}  
\resdef{e2.olmo3-think-sft.a3-quant.inf.lp4.R.hi}{56.4}  
\resdef{e2.olmo3-think-sft.a3-quant.inf.lp4.R.lo}{24.4}  
\resdef{e2.olmo3-think-sft.a3-quant.inf.lp4.R.mean}{39.7}  
\resdef{e2.olmo3-think-sft.a3-quant.inf.lp4.R.std}{5.9}  
\resdef{e2.olmo3-think-sft.b2.yn.lp4.acc.hi}{84.6}  
\resdef{e2.olmo3-think-sft.b2.yn.lp4.acc.lo}{68.4}  
\resdef{e2.olmo3-think-sft.b2.yn.lp4.acc.mean}{76.8}  
\resdef{e2.olmo3-think-sft.b2.yn.lp4.acc.std}{2.8}  
\resdef{e2.orz-7b.a2.inf.lp4.R.hi}{11.0}  
\resdef{e2.orz-7b.a2.inf.lp4.R.lo}{3.5}  
\resdef{e2.orz-7b.a2.inf.lp4.R.mean}{7.1}  
\resdef{e2.orz-7b.a2.inf.lp4.R.std}{1.0}  
\resdef{e2.orz-7b.a3-conn.inf.lp4.R.hi}{17.2}  
\resdef{e2.orz-7b.a3-conn.inf.lp4.R.lo}{1.0}  
\resdef{e2.orz-7b.a3-conn.inf.lp4.R.mean}{8.1}  
\resdef{e2.orz-7b.a3-conn.inf.lp4.R.std}{3.5}  
\resdef{e2.orz-7b.a3-quant.inf.lp4.R.hi}{59.0}  
\resdef{e2.orz-7b.a3-quant.inf.lp4.R.lo}{23.1}  
\resdef{e2.orz-7b.a3-quant.inf.lp4.R.mean}{41.0}  
\resdef{e2.orz-7b.a3-quant.inf.lp4.R.std}{2.2}  
\resdef{e2.orz-7b.b2.yn.lp4.acc.hi}{74.4}  
\resdef{e2.orz-7b.b2.yn.lp4.acc.lo}{57.2}  
\resdef{e2.orz-7b.b2.yn.lp4.acc.mean}{66.0}  
\resdef{e2.orz-7b.b2.yn.lp4.acc.std}{3.0}  
\resdef{e2.qwen25-7b.a2.inf.lp4.R.hi}{40.0}  
\resdef{e2.qwen25-7b.a2.inf.lp4.R.lo}{26.3}  
\resdef{e2.qwen25-7b.a2.inf.lp4.R.mean}{32.9}  
\resdef{e2.qwen25-7b.a2.inf.lp4.R.std}{1.7}  
\resdef{e2.qwen25-7b.a3-conn.inf.lp4.R.hi}{16.2}  
\resdef{e2.qwen25-7b.a3-conn.inf.lp4.R.lo}{0.0}  
\resdef{e2.qwen25-7b.a3-conn.inf.lp4.R.mean}{7.1}  
\resdef{e2.qwen25-7b.a3-conn.inf.lp4.R.std}{1.8}  
\resdef{e2.qwen25-7b.a3-quant.inf.lp4.R.hi}{60.3}  
\resdef{e2.qwen25-7b.a3-quant.inf.lp4.R.lo}{23.1}  
\resdef{e2.qwen25-7b.a3-quant.inf.lp4.R.mean}{41.0}  
\resdef{e2.qwen25-7b.a3-quant.inf.lp4.R.std}{2.2}  
\resdef{e2.qwen25-7b.b2.yn.lp4.acc.hi}{75.1}  
\resdef{e2.qwen25-7b.b2.yn.lp4.acc.lo}{58.2}  
\resdef{e2.qwen25-7b.b2.yn.lp4.acc.mean}{66.7}  
\resdef{e2.qwen25-7b.b2.yn.lp4.acc.std}{3.4}  
\resdef{e2.r1distill-qwen-7b.a2.inf.lp4.R.hi}{52.3}  
\resdef{e2.r1distill-qwen-7b.a2.inf.lp4.R.lo}{41.3}  
\resdef{e2.r1distill-qwen-7b.a2.inf.lp4.R.mean}{46.8}  
\resdef{e2.r1distill-qwen-7b.a2.inf.lp4.R.std}{2.7}  
\resdef{e2.r1distill-qwen-7b.a3-conn.inf.lp4.R.hi}{78.8}  
\resdef{e2.r1distill-qwen-7b.a3-conn.inf.lp4.R.lo}{52.5}  
\resdef{e2.r1distill-qwen-7b.a3-conn.inf.lp4.R.mean}{65.7}  
\resdef{e2.r1distill-qwen-7b.a3-conn.inf.lp4.R.std}{7.6}  
\resdef{e2.r1distill-qwen-7b.a3-quant.inf.lp4.R.hi}{94.9}  
\resdef{e2.r1distill-qwen-7b.a3-quant.inf.lp4.R.lo}{74.4}  
\resdef{e2.r1distill-qwen-7b.a3-quant.inf.lp4.R.mean}{85.9}  
\resdef{e2.r1distill-qwen-7b.a3-quant.inf.lp4.R.std}{2.2}  
\resdef{e2.r1distill-qwen-7b.b2.yn.lp4.acc.hi}{81.8}  
\resdef{e2.r1distill-qwen-7b.b2.yn.lp4.acc.lo}{65.3}  
\resdef{e2.r1distill-qwen-7b.b2.yn.lp4.acc.mean}{73.7}  
\resdef{e2.r1distill-qwen-7b.b2.yn.lp4.acc.std}{3.8}  
\resdef{e2.simplerl-zoo-7b.a2.inf.lp4.R.hi}{38.6}  
\resdef{e2.simplerl-zoo-7b.a2.inf.lp4.R.lo}{25.4}  
\resdef{e2.simplerl-zoo-7b.a2.inf.lp4.R.mean}{31.8}  
\resdef{e2.simplerl-zoo-7b.a2.inf.lp4.R.std}{2.4}  
\resdef{e2.simplerl-zoo-7b.a3-conn.inf.lp4.R.hi}{16.2}  
\resdef{e2.simplerl-zoo-7b.a3-conn.inf.lp4.R.lo}{0.0}  
\resdef{e2.simplerl-zoo-7b.a3-conn.inf.lp4.R.mean}{7.1}  
\resdef{e2.simplerl-zoo-7b.a3-conn.inf.lp4.R.std}{1.8}  
\resdef{e2.simplerl-zoo-7b.a3-quant.inf.lp4.R.hi}{61.5}  
\resdef{e2.simplerl-zoo-7b.a3-quant.inf.lp4.R.lo}{25.6}  
\resdef{e2.simplerl-zoo-7b.a3-quant.inf.lp4.R.mean}{43.6}  
\resdef{e2.simplerl-zoo-7b.a3-quant.inf.lp4.R.std}{2.2}  
\resdef{e2.simplerl-zoo-7b.b2.yn.lp4.acc.hi}{74.0}  
\resdef{e2.simplerl-zoo-7b.b2.yn.lp4.acc.lo}{57.2}  
\resdef{e2.simplerl-zoo-7b.b2.yn.lp4.acc.mean}{65.6}  
\resdef{e2.simplerl-zoo-7b.b2.yn.lp4.acc.std}{3.4}  
\resdef{e2.tulu3-8b-dpo.a2.inf.lp4.R.hi}{7.1}  
\resdef{e2.tulu3-8b-dpo.a2.inf.lp4.R.lo}{1.3}  
\resdef{e2.tulu3-8b-dpo.a2.inf.lp4.R.mean}{4.0}  
\resdef{e2.tulu3-8b-dpo.a2.inf.lp4.R.std}{0.0}  
\resdef{e2.tulu3-8b-dpo.a3-conn.inf.lp4.R.hi}{5.1}  
\resdef{e2.tulu3-8b-dpo.a3-conn.inf.lp4.R.lo}{0.0}  
\resdef{e2.tulu3-8b-dpo.a3-conn.inf.lp4.R.mean}{2.0}  
\resdef{e2.tulu3-8b-dpo.a3-conn.inf.lp4.R.std}{1.8}  
\resdef{e2.tulu3-8b-dpo.a3-quant.inf.lp4.R.hi}{28.2}  
\resdef{e2.tulu3-8b-dpo.a3-quant.inf.lp4.R.lo}{2.6}  
\resdef{e2.tulu3-8b-dpo.a3-quant.inf.lp4.R.mean}{14.1}  
\resdef{e2.tulu3-8b-dpo.a3-quant.inf.lp4.R.std}{2.2}  
\resdef{e2.tulu3-8b-dpo.b2.yn.lp4.acc.hi}{80.7}  
\resdef{e2.tulu3-8b-dpo.b2.yn.lp4.acc.lo}{66.0}  
\resdef{e2.tulu3-8b-dpo.b2.yn.lp4.acc.mean}{73.7}  
\resdef{e2.tulu3-8b-dpo.b2.yn.lp4.acc.std}{2.8}  
\resdef{e2.tulu3-8b-rlvr.a2.inf.lp4.R.hi}{7.5}  
\resdef{e2.tulu3-8b-rlvr.a2.inf.lp4.R.lo}{1.8}  
\resdef{e2.tulu3-8b-rlvr.a2.inf.lp4.R.mean}{4.4}  
\resdef{e2.tulu3-8b-rlvr.a2.inf.lp4.R.std}{0.4}  
\resdef{e2.tulu3-8b-rlvr.a3-conn.inf.lp4.R.hi}{5.1}  
\resdef{e2.tulu3-8b-rlvr.a3-conn.inf.lp4.R.lo}{0.0}  
\resdef{e2.tulu3-8b-rlvr.a3-conn.inf.lp4.R.mean}{2.0}  
\resdef{e2.tulu3-8b-rlvr.a3-conn.inf.lp4.R.std}{1.8}  
\resdef{e2.tulu3-8b-rlvr.a3-quant.inf.lp4.R.hi}{28.2}  
\resdef{e2.tulu3-8b-rlvr.a3-quant.inf.lp4.R.lo}{2.6}  
\resdef{e2.tulu3-8b-rlvr.a3-quant.inf.lp4.R.mean}{14.1}  
\resdef{e2.tulu3-8b-rlvr.a3-quant.inf.lp4.R.std}{2.2}  
\resdef{e2.tulu3-8b-rlvr.b2.yn.lp4.acc.hi}{83.5}  
\resdef{e2.tulu3-8b-rlvr.b2.yn.lp4.acc.lo}{67.4}  
\resdef{e2.tulu3-8b-rlvr.b2.yn.lp4.acc.mean}{75.8}  
\resdef{e2.tulu3-8b-rlvr.b2.yn.lp4.acc.std}{2.8}  
\resdef{e2.tulu3-8b-sft.a2.inf.lp4.R.hi}{14.1}  
\resdef{e2.tulu3-8b-sft.a2.inf.lp4.R.lo}{6.4}  
\resdef{e2.tulu3-8b-sft.a2.inf.lp4.R.mean}{10.2}  
\resdef{e2.tulu3-8b-sft.a2.inf.lp4.R.std}{0.4}  
\resdef{e2.tulu3-8b-sft.a3-conn.inf.lp4.R.hi}{11.1}  
\resdef{e2.tulu3-8b-sft.a3-conn.inf.lp4.R.lo}{1.0}  
\resdef{e2.tulu3-8b-sft.a3-conn.inf.lp4.R.mean}{5.1}  
\resdef{e2.tulu3-8b-sft.a3-conn.inf.lp4.R.std}{1.8}  
\resdef{e2.tulu3-8b-sft.a3-quant.inf.lp4.R.hi}{64.1}  
\resdef{e2.tulu3-8b-sft.a3-quant.inf.lp4.R.lo}{29.5}  
\resdef{e2.tulu3-8b-sft.a3-quant.inf.lp4.R.mean}{46.2}  
\resdef{e2.tulu3-8b-sft.a3-quant.inf.lp4.R.std}{0.0}  
\resdef{e2.tulu3-8b-sft.b2.yn.lp4.acc.hi}{83.9}  
\resdef{e2.tulu3-8b-sft.b2.yn.lp4.acc.lo}{68.8}  
\resdef{e2.tulu3-8b-sft.b2.yn.lp4.acc.mean}{76.5}  
\resdef{e2.tulu3-8b-sft.b2.yn.lp4.acc.std}{1.6}  
\resdef{e2.tulu31-8b.a2.inf.lp4.R.hi}{15.5}  
\resdef{e2.tulu31-8b.a2.inf.lp4.R.lo}{6.0}  
\resdef{e2.tulu31-8b.a2.inf.lp4.R.mean}{10.4}  
\resdef{e2.tulu31-8b.a2.inf.lp4.R.std}{0.4}  
\resdef{e2.tulu31-8b.a3-conn.inf.lp4.R.hi}{7.1}  
\resdef{e2.tulu31-8b.a3-conn.inf.lp4.R.lo}{0.0}  
\resdef{e2.tulu31-8b.a3-conn.inf.lp4.R.mean}{3.0}  
\resdef{e2.tulu31-8b.a3-conn.inf.lp4.R.std}{0.0}  
\resdef{e2.tulu31-8b.a3-quant.inf.lp4.R.hi}{26.9}  
\resdef{e2.tulu31-8b.a3-quant.inf.lp4.R.lo}{3.8}  
\resdef{e2.tulu31-8b.a3-quant.inf.lp4.R.mean}{14.1}  
\resdef{e2.tulu31-8b.a3-quant.inf.lp4.R.std}{2.2}  
\resdef{e2.tulu31-8b.b2.yn.lp4.acc.hi}{84.2}  
\resdef{e2.tulu31-8b.b2.yn.lp4.acc.lo}{68.1}  
\resdef{e2.tulu31-8b.b2.yn.lp4.acc.mean}{76.1}  
\resdef{e2.tulu31-8b.b2.yn.lp4.acc.std}{2.2}  
\resdef{e2t.olmo3-think-dpo.a3-quant.inf.nat.dR.mean}{-32.1}  
\resdef{e2t.olmo3-think-s1375.a3-quant.inf.nat.dR.mean}{-2.6}  
\resdef{e8.qwen3-32b.a3-conn.inf.cot.R.hi}{57.6}  
\resdef{e8.qwen3-32b.a3-conn.inf.cot.R.lo}{34.3}  
\resdef{e8.qwen3-32b.a3-conn.inf.cot.R.mean}{45.5}  
\resdef{e8.qwen3-32b.a3-conn.inf.cot.R.std}{13.9}  
\resdef{e8.qwen3-32b.a3-quant.inf.cot.R.hi}{59.0}  
\resdef{e8.qwen3-32b.a3-quant.inf.cot.R.lo}{34.6}  
\resdef{e8.qwen3-32b.a3-quant.inf.cot.R.mean}{46.2}  
\resdef{e8.qwen3-32b.a3-quant.inf.cot.R.std}{10.2}  
\resdef{e8.qwen3-32b.b2.yn.cot.acc.hi}{87.4}  
\resdef{e8.qwen3-32b.b2.yn.cot.acc.lo}{73.3}  
\resdef{e8.qwen3-32b.b2.yn.cot.acc.mean}{80.7}  
\resdef{e8.qwen3-32b.b2.yn.cot.acc.std}{2.2}  
\resdef{e8.qwen3-8b.a3-conn.inf.cot.R.hi}{11.1}  
\resdef{e8.qwen3-8b.a3-conn.inf.cot.R.lo}{1.0}  
\resdef{e8.qwen3-8b.a3-conn.inf.cot.R.mean}{6.1}  
\resdef{e8.qwen3-8b.a3-conn.inf.cot.R.std}{3.0}  
\resdef{e8.qwen3-8b.a3-quant.inf.cot.R.hi}{19.2}  
\resdef{e8.qwen3-8b.a3-quant.inf.cot.R.lo}{2.6}  
\resdef{e8.qwen3-8b.a3-quant.inf.cot.R.mean}{10.3}  
\resdef{e8.qwen3-8b.a3-quant.inf.cot.R.std}{5.9}  
\resdef{e8.qwen3-8b.b2.yn.cot.acc.hi}{83.2}  
\resdef{e8.qwen3-8b.b2.yn.cot.acc.lo}{67.4}  
\resdef{e8.qwen3-8b.b2.yn.cot.acc.mean}{75.4}  
\resdef{e8.qwen3-8b.b2.yn.cot.acc.std}{2.2}  
\resdef{e1f.claude-opus-4p8.a2.inf.eff-high.dR.hi}{1.7}  
\resdef{e1f.claude-opus-4p8.a2.inf.eff-high.dR.lo}{-16.7}  
\resdef{e1f.claude-opus-4p8.a2.inf.eff-high.dR.mean}{-7.5}  
\resdef{e1f.claude-opus-4p8.a2.inf.eff-high.dR.std}{3.5}  
\resdef{e1f.claude-opus-4p8.a3-conn.inf.eff-high.dR.hi}{16.7}  
\resdef{e1f.claude-opus-4p8.a3-conn.inf.eff-high.dR.lo}{-2.9}  
\resdef{e1f.claude-opus-4p8.a3-conn.inf.eff-high.dR.mean}{7.2}  
\resdef{e1f.claude-opus-4p8.a3-conn.inf.eff-high.dR.std}{1.1}  
\resdef{e1f.claude-opus-4p8.a3-gradadj.inf.eff-high.dR.hi}{17.5}  
\resdef{e1f.claude-opus-4p8.a3-gradadj.inf.eff-high.dR.lo}{-5.0}  
\resdef{e1f.claude-opus-4p8.a3-gradadj.inf.eff-high.dR.mean}{6.2}  
\resdef{e1f.claude-opus-4p8.a3-gradadj.inf.eff-high.dR.std}{1.8}  
\resdef{e1f.claude-opus-4p8.a3-modal.inf.eff-high.dR.hi}{21.2}  
\resdef{e1f.claude-opus-4p8.a3-modal.inf.eff-high.dR.lo}{0.0}  
\resdef{e1f.claude-opus-4p8.a3-modal.inf.eff-high.dR.mean}{10.0}  
\resdef{e1f.claude-opus-4p8.a3-modal.inf.eff-high.dR.std}{3.4}  
\resdef{e1f.claude-opus-4p8.a3-quant.inf.eff-high.dR.hi}{25.0}  
\resdef{e1f.claude-opus-4p8.a3-quant.inf.eff-high.dR.lo}{9.0}  
\resdef{e1f.claude-opus-4p8.a3-quant.inf.eff-high.dR.mean}{16.7}  
\resdef{e1f.claude-opus-4p8.a3-quant.inf.eff-high.dR.std}{2.2}  
\resdef{e1f.claude-opus-4p8.a3.inf.eff-high.dRx.hi}{15.8}  
\resdef{e1f.claude-opus-4p8.a3.inf.eff-high.dRx.lo}{5.9}  
\resdef{e1f.claude-opus-4p8.a3.inf.eff-high.dRx.mean}{10.8}  
\resdef{e1f.claude-opus-4p8.a3.inf.off512.fail.any.mean}{8.1}  
\resdef{e1f.claude-opus-4p8.a3.infflip.eff-high.dR.hi}{8.1}  
\resdef{e1f.claude-opus-4p8.a3.infflip.eff-high.dR.lo}{2.2}  
\resdef{e1f.claude-opus-4p8.a3.infflip.eff-high.dR.mean}{5.0}  
\resdef{e1f.claude-opus-4p8.a3.infprior.eff-high.dYes.hi}{7.9}  
\resdef{e1f.claude-opus-4p8.a3.infprior.eff-high.dYes.lo}{0.0}  
\resdef{e1f.claude-opus-4p8.a3.infprior.eff-high.dYes.mean}{3.9}  
\resdef{e1f.claude-opus-4p8.a3.nli3.eff-high.dRx.hi}{0.0}  
\resdef{e1f.claude-opus-4p8.a3.nli3.eff-high.dRx.lo}{-3.2}  
\resdef{e1f.claude-opus-4p8.a3.nli3.eff-high.dRx.mean}{-1.3}  
\resdef{e1f.claude-opus-4p8.b2.yn.eff-high.dacc.hi}{10.0}  
\resdef{e1f.claude-opus-4p8.b2.yn.eff-high.dacc.lo}{-5.0}  
\resdef{e1f.claude-opus-4p8.b2.yn.eff-high.dacc.mean}{2.5}  
\resdef{e1f.claude-opus-4p8.b2.yn.eff-high.dacc.std}{0.0}  
\resdef{e1f.claude-sonnet-4p6.a2.inf.bud16k.dR.hi}{30.8}  
\resdef{e1f.claude-sonnet-4p6.a2.inf.bud16k.dR.lo}{11.7}  
\resdef{e1f.claude-sonnet-4p6.a2.inf.bud16k.dR.mean}{20.8}  
\resdef{e1f.claude-sonnet-4p6.a2.inf.bud16k.dR.std}{5.9}  
\resdef{e1f.claude-sonnet-4p6.a3-conn.inf.bud16k.dR.hi}{24.4}  
\resdef{e1f.claude-sonnet-4p6.a3-conn.inf.bud16k.dR.lo}{-3.1}  
\resdef{e1f.claude-sonnet-4p6.a3-conn.inf.bud16k.dR.mean}{10.6}  
\resdef{e1f.claude-sonnet-4p6.a3-conn.inf.bud16k.dR.std}{4.6}  
\resdef{e1f.claude-sonnet-4p6.a3-conn.nli3.bud16k.dRx.mean}{-15.6}  
\resdef{e1f.claude-sonnet-4p6.a3-gradadj.inf.bud16k.dR.hi}{40.0}  
\resdef{e1f.claude-sonnet-4p6.a3-gradadj.inf.bud16k.dR.lo}{10.0}  
\resdef{e1f.claude-sonnet-4p6.a3-gradadj.inf.bud16k.dR.mean}{25.0}  
\resdef{e1f.claude-sonnet-4p6.a3-gradadj.inf.bud16k.dR.std}{3.5}  
\resdef{e1f.claude-sonnet-4p6.a3-modal.inf.bud16k.dR.hi}{38.8}  
\resdef{e1f.claude-sonnet-4p6.a3-modal.inf.bud16k.dR.lo}{13.8}  
\resdef{e1f.claude-sonnet-4p6.a3-modal.inf.bud16k.dR.mean}{26.2}  
\resdef{e1f.claude-sonnet-4p6.a3-modal.inf.bud16k.dR.std}{1.8}  
\resdef{e1f.claude-sonnet-4p6.a3-quant.inf.bud16k.dR.hi}{30.6}  
\resdef{e1f.claude-sonnet-4p6.a3-quant.inf.bud16k.dR.lo}{12.5}  
\resdef{e1f.claude-sonnet-4p6.a3-quant.inf.bud16k.dR.mean}{21.2}  
\resdef{e1f.claude-sonnet-4p6.a3-quant.inf.bud16k.dR.std}{1.6}  
\resdef{e1f.claude-sonnet-4p6.a3-quant.nli3.bud16k.dRx.mean}{-5.1}  
\resdef{e1f.claude-sonnet-4p6.a3.inf.bud16k.dRx.hi}{25.4}  
\resdef{e1f.claude-sonnet-4p6.a3.inf.bud16k.dRx.lo}{13.0}  
\resdef{e1f.claude-sonnet-4p6.a3.inf.bud16k.dRx.mean}{19.2}  
\resdef{e1f.claude-sonnet-4p6.a3.infflip.bud16k.dR.hi}{-4.5}  
\resdef{e1f.claude-sonnet-4p6.a3.infflip.bud16k.dR.lo}{-21.4}  
\resdef{e1f.claude-sonnet-4p6.a3.infflip.bud16k.dR.mean}{-13.0}  
\resdef{e1f.claude-sonnet-4p6.a3.infprior.bud16k.dYes.hi}{12.0}  
\resdef{e1f.claude-sonnet-4p6.a3.infprior.bud16k.dYes.lo}{4.0}  
\resdef{e1f.claude-sonnet-4p6.a3.infprior.bud16k.dYes.mean}{7.9}  
\resdef{e1f.claude-sonnet-4p6.a3.nli3.bud16k.dRx.hi}{-5.4}  
\resdef{e1f.claude-sonnet-4p6.a3.nli3.bud16k.dRx.lo}{-14.5}  
\resdef{e1f.claude-sonnet-4p6.a3.nli3.bud16k.dRx.mean}{-9.7}  
\resdef{e1f.claude-sonnet-4p6.b2.yn.bud16k.dacc.hi}{10.0}  
\resdef{e1f.claude-sonnet-4p6.b2.yn.bud16k.dacc.lo}{-5.0}  
\resdef{e1f.claude-sonnet-4p6.b2.yn.bud16k.dacc.mean}{2.5}  
\resdef{e1f.claude-sonnet-4p6.b2.yn.bud16k.dacc.std}{0.0}  
\resdef{e1f.deepseek-v4-flash.a2.inf.think.dR.hi}{18.3}  
\resdef{e1f.deepseek-v4-flash.a2.inf.think.dR.lo}{4.2}  
\resdef{e1f.deepseek-v4-flash.a2.inf.think.dR.mean}{10.8}  
\resdef{e1f.deepseek-v4-flash.a2.inf.think.dR.std}{1.2}  
\resdef{e1f.deepseek-v4-flash.a3-conn.inf.think.dR.hi}{10.6}  
\resdef{e1f.deepseek-v4-flash.a3-conn.inf.think.dR.lo}{-1.9}  
\resdef{e1f.deepseek-v4-flash.a3-conn.inf.think.dR.mean}{4.4}  
\resdef{e1f.deepseek-v4-flash.a3-conn.inf.think.dR.std}{2.7}  
\resdef{e1f.deepseek-v4-flash.a3-gradadj.inf.think.dR.hi}{11.2}  
\resdef{e1f.deepseek-v4-flash.a3-gradadj.inf.think.dR.lo}{0.0}  
\resdef{e1f.deepseek-v4-flash.a3-gradadj.inf.think.dR.mean}{5.0}  
\resdef{e1f.deepseek-v4-flash.a3-gradadj.inf.think.dR.std}{3.5}  
\resdef{e1f.deepseek-v4-flash.a3-modal.inf.think.dR.hi}{0.0}  
\resdef{e1f.deepseek-v4-flash.a3-modal.inf.think.dR.lo}{-3.8}  
\resdef{e1f.deepseek-v4-flash.a3-modal.inf.think.dR.mean}{-1.2}  
\resdef{e1f.deepseek-v4-flash.a3-modal.inf.think.dR.std}{1.8}  
\resdef{e1f.deepseek-v4-flash.a3-quant.inf.think.dR.hi}{25.6}  
\resdef{e1f.deepseek-v4-flash.a3-quant.inf.think.dR.lo}{8.1}  
\resdef{e1f.deepseek-v4-flash.a3-quant.inf.think.dR.mean}{16.9}  
\resdef{e1f.deepseek-v4-flash.a3-quant.inf.think.dR.std}{2.7}  
\resdef{e1f.deepseek-v4-flash.b2.yn.think.dacc.hi}{13.8}  
\resdef{e1f.deepseek-v4-flash.b2.yn.think.dacc.lo}{-3.8}  
\resdef{e1f.deepseek-v4-flash.b2.yn.think.dacc.mean}{5.0}  
\resdef{e1f.deepseek-v4-flash.b2.yn.think.dacc.std}{0.0}  
\resdef{e1f.deepseek-v4-pro.a2.inf.think.dR.hi}{53.3}  
\resdef{e1f.deepseek-v4-pro.a2.inf.think.dR.lo}{30.8}  
\resdef{e1f.deepseek-v4-pro.a2.inf.think.dR.mean}{41.7}  
\resdef{e1f.deepseek-v4-pro.a2.inf.think.dR.std}{2.2}  
\resdef{e1f.deepseek-v4-pro.a3-conn.inf.think.dR.hi}{7.5}  
\resdef{e1f.deepseek-v4-pro.a3-conn.inf.think.dR.lo}{-3.1}  
\resdef{e1f.deepseek-v4-pro.a3-conn.inf.think.dR.mean}{1.9}  
\resdef{e1f.deepseek-v4-pro.a3-conn.inf.think.dR.std}{6.3}  
\resdef{e1f.deepseek-v4-pro.a3-gradadj.inf.think.dR.hi}{17.5}  
\resdef{e1f.deepseek-v4-pro.a3-gradadj.inf.think.dR.lo}{1.2}  
\resdef{e1f.deepseek-v4-pro.a3-gradadj.inf.think.dR.mean}{8.8}  
\resdef{e1f.deepseek-v4-pro.a3-gradadj.inf.think.dR.std}{4.8}  
\resdef{e1f.deepseek-v4-pro.a3-modal.inf.think.dR.hi}{7.5}  
\resdef{e1f.deepseek-v4-pro.a3-modal.inf.think.dR.lo}{-3.8}  
\resdef{e1f.deepseek-v4-pro.a3-modal.inf.think.dR.mean}{1.2}  
\resdef{e1f.deepseek-v4-pro.a3-modal.inf.think.dR.std}{1.8}  
\resdef{e1f.deepseek-v4-pro.a3-quant.inf.think.dR.hi}{16.5}  
\resdef{e1f.deepseek-v4-pro.a3-quant.inf.think.dR.lo}{1.3}  
\resdef{e1f.deepseek-v4-pro.a3-quant.inf.think.dR.mean}{8.9}  
\resdef{e1f.deepseek-v4-pro.a3-quant.inf.think.dR.std}{3.6}  
\resdef{e1f.deepseek-v4-pro.b2.yn.think.dacc.hi}{18.8}  
\resdef{e1f.deepseek-v4-pro.b2.yn.think.dacc.lo}{2.5}  
\resdef{e1f.deepseek-v4-pro.b2.yn.think.dacc.mean}{10.0}  
\resdef{e1f.deepseek-v4-pro.b2.yn.think.dacc.std}{3.5}  
\resdef{e1f.gemini-2p5-flash.a2.inf.bud8k.dR.hi}{58.3}  
\resdef{e1f.gemini-2p5-flash.a2.inf.bud8k.dR.lo}{31.7}  
\resdef{e1f.gemini-2p5-flash.a2.inf.bud8k.dR.mean}{45.0}  
\resdef{e1f.gemini-2p5-flash.a3-conn.inf.bud8k.dR.hi}{-4.8}  
\resdef{e1f.gemini-2p5-flash.a3-conn.inf.bud8k.dR.lo}{-42.9}  
\resdef{e1f.gemini-2p5-flash.a3-conn.inf.bud8k.dR.mean}{-23.8}  
\resdef{e1f.gemini-2p5-flash.a3-gradadj.inf.bud8k.dR.hi}{20.0}  
\resdef{e1f.gemini-2p5-flash.a3-gradadj.inf.bud8k.dR.lo}{-5.0}  
\resdef{e1f.gemini-2p5-flash.a3-gradadj.inf.bud8k.dR.mean}{7.5}  
\resdef{e1f.gemini-2p5-flash.a3-modal.inf.bud8k.dR.hi}{20.6}  
\resdef{e1f.gemini-2p5-flash.a3-modal.inf.bud8k.dR.lo}{0.0}  
\resdef{e1f.gemini-2p5-flash.a3-modal.inf.bud8k.dR.mean}{8.8}  
\resdef{e1f.gemini-2p5-flash.a3-quant.inf.bud8k.dR.hi}{10.6}  
\resdef{e1f.gemini-2p5-flash.a3-quant.inf.bud8k.dR.lo}{-16.7}  
\resdef{e1f.gemini-2p5-flash.a3-quant.inf.bud8k.dR.mean}{-3.0}  
\resdef{e1f.gemini-2p5-flash.a3.inf.bud8k.fail.any.mean}{26.2}  
\resdef{e1f.gemini-2p5-flash.b2.yn.bud8k.dacc.hi}{23.1}  
\resdef{e1f.gemini-2p5-flash.b2.yn.bud8k.dacc.lo}{0.0}  
\resdef{e1f.gemini-2p5-flash.b2.yn.bud8k.dacc.mean}{10.3}  
\resdef{e1f.gemini-2p5-pro.a2.inf.bud8k.dR.hi}{70.0}  
\resdef{e1f.gemini-2p5-pro.a2.inf.bud8k.dR.lo}{45.0}  
\resdef{e1f.gemini-2p5-pro.a2.inf.bud8k.dR.mean}{58.3}  
\resdef{e1f.gemini-2p5-pro.a3-conn.inf.bud8k.dR.hi}{3.9}  
\resdef{e1f.gemini-2p5-pro.a3-conn.inf.bud8k.dR.lo}{-25.0}  
\resdef{e1f.gemini-2p5-pro.a3-conn.inf.bud8k.dR.mean}{-10.5}  
\resdef{e1f.gemini-2p5-pro.a3-gradadj.inf.bud8k.dR.hi}{40.0}  
\resdef{e1f.gemini-2p5-pro.a3-gradadj.inf.bud8k.dR.lo}{10.0}  
\resdef{e1f.gemini-2p5-pro.a3-gradadj.inf.bud8k.dR.mean}{25.0}  
\resdef{e1f.gemini-2p5-pro.a3-modal.inf.bud8k.dR.hi}{48.7}  
\resdef{e1f.gemini-2p5-pro.a3-modal.inf.bud8k.dR.lo}{12.8}  
\resdef{e1f.gemini-2p5-pro.a3-modal.inf.bud8k.dR.mean}{30.8}  
\resdef{e1f.gemini-2p5-pro.a3-quant.inf.bud8k.dR.hi}{5.1}  
\resdef{e1f.gemini-2p5-pro.a3-quant.inf.bud8k.dR.lo}{-7.6}  
\resdef{e1f.gemini-2p5-pro.a3-quant.inf.bud8k.dR.mean}{-1.3}  
\resdef{e1f.gemini-2p5-pro.b2.yn.bud8k.dacc.hi}{7.5}  
\resdef{e1f.gemini-2p5-pro.b2.yn.bud8k.dacc.lo}{-7.5}  
\resdef{e1f.gemini-2p5-pro.b2.yn.bud8k.dacc.mean}{0.0}  
\resdef{e1f.gpt-5.a2.inf.eff-high.dR.hi}{13.3}  
\resdef{e1f.gpt-5.a2.inf.eff-high.dR.lo}{1.7}  
\resdef{e1f.gpt-5.a2.inf.eff-high.dR.mean}{7.5}  
\resdef{e1f.gpt-5.a2.inf.eff-high.dR.std}{1.2}  
\resdef{e1f.gpt-5.a3-conn.inf.eff-high.dR.hi}{-10.0}  
\resdef{e1f.gpt-5.a3-conn.inf.eff-high.dR.lo}{-27.5}  
\resdef{e1f.gpt-5.a3-conn.inf.eff-high.dR.mean}{-18.8}  
\resdef{e1f.gpt-5.a3-conn.inf.eff-high.dR.std}{10.6}  
\resdef{e1f.gpt-5.a3-conn.nli3.eff-high.dRx.mean}{-11.2}  
\resdef{e1f.gpt-5.a3-gradadj.inf.eff-high.dR.hi}{11.2}  
\resdef{e1f.gpt-5.a3-gradadj.inf.eff-high.dR.lo}{-10.0}  
\resdef{e1f.gpt-5.a3-gradadj.inf.eff-high.dR.mean}{0.0}  
\resdef{e1f.gpt-5.a3-gradadj.inf.eff-high.dR.std}{3.5}  
\resdef{e1f.gpt-5.a3-modal.inf.eff-high.dR.hi}{-5.0}  
\resdef{e1f.gpt-5.a3-modal.inf.eff-high.dR.lo}{-20.0}  
\resdef{e1f.gpt-5.a3-modal.inf.eff-high.dR.mean}{-12.5}  
\resdef{e1f.gpt-5.a3-modal.inf.eff-high.dR.std}{3.5}  
\resdef{e1f.gpt-5.a3-quant.inf.eff-high.dR.hi}{-11.2}  
\resdef{e1f.gpt-5.a3-quant.inf.eff-high.dR.lo}{-31.2}  
\resdef{e1f.gpt-5.a3-quant.inf.eff-high.dR.mean}{-20.6}  
\resdef{e1f.gpt-5.a3-quant.inf.eff-high.dR.std}{4.4}  
\resdef{e1f.gpt-5.a3-quant.nli3.eff-high.dRx.mean}{-6.2}  
\resdef{e1f.gpt-5.a3.inf.eff-high.dRx.hi}{-10.3}  
\resdef{e1f.gpt-5.a3.inf.eff-high.dRx.lo}{-20.2}  
\resdef{e1f.gpt-5.a3.inf.eff-high.dRx.mean}{-15.2}  
\resdef{e1f.gpt-5.a3.infflip.eff-high.dR.hi}{4.4}  
\resdef{e1f.gpt-5.a3.infflip.eff-high.dR.lo}{-11.2}  
\resdef{e1f.gpt-5.a3.infflip.eff-high.dR.mean}{-3.4}  
\resdef{e1f.gpt-5.a3.infprior.eff-high.dYes.hi}{34.4}  
\resdef{e1f.gpt-5.a3.infprior.eff-high.dYes.lo}{19.4}  
\resdef{e1f.gpt-5.a3.infprior.eff-high.dYes.mean}{26.9}  
\resdef{e1f.gpt-5.a3.nli3.eff-high.dRx.hi}{-5.1}  
\resdef{e1f.gpt-5.a3.nli3.eff-high.dRx.lo}{-12.9}  
\resdef{e1f.gpt-5.a3.nli3.eff-high.dRx.mean}{-8.8}  
\resdef{e1f.gpt-5.b2.yn.eff-high.dacc.hi}{17.5}  
\resdef{e1f.gpt-5.b2.yn.eff-high.dacc.lo}{-5.0}  
\resdef{e1f.gpt-5.b2.yn.eff-high.dacc.mean}{6.2}  
\resdef{e1f.gpt-5.b2.yn.eff-high.dacc.std}{1.8}  
\resdef{e1f.gpt-5p4.a2.inf.eff-high.dR.hi}{25.0}  
\resdef{e1f.gpt-5p4.a2.inf.eff-high.dR.lo}{6.7}  
\resdef{e1f.gpt-5p4.a2.inf.eff-high.dR.mean}{15.0}  
\resdef{e1f.gpt-5p4.a3-conn.inf.eff-high.dR.hi}{-7.5}  
\resdef{e1f.gpt-5p4.a3-conn.inf.eff-high.dR.lo}{-32.5}  
\resdef{e1f.gpt-5p4.a3-conn.inf.eff-high.dR.mean}{-20.0}  
\resdef{e1f.gpt-5p4.a3-gradadj.inf.eff-high.dR.hi}{2.5}  
\resdef{e1f.gpt-5p4.a3-gradadj.inf.eff-high.dR.lo}{-20.0}  
\resdef{e1f.gpt-5p4.a3-gradadj.inf.eff-high.dR.mean}{-7.5}  
\resdef{e1f.gpt-5p4.a3-modal.inf.eff-high.dR.hi}{-5.0}  
\resdef{e1f.gpt-5p4.a3-modal.inf.eff-high.dR.lo}{-27.5}  
\resdef{e1f.gpt-5p4.a3-modal.inf.eff-high.dR.mean}{-15.0}  
\resdef{e1f.gpt-5p4.a3-quant.inf.eff-high.dR.hi}{17.5}  
\resdef{e1f.gpt-5p4.a3-quant.inf.eff-high.dR.lo}{-12.5}  
\resdef{e1f.gpt-5p4.a3-quant.inf.eff-high.dR.mean}{2.5}  
\resdef{e1f.gpt-5p4.b2.yn.eff-high.dacc.hi}{12.5}  
\resdef{e1f.gpt-5p4.b2.yn.eff-high.dacc.lo}{-7.5}  
\resdef{e1f.gpt-5p4.b2.yn.eff-high.dacc.mean}{2.5}  
\resdef{e1f.o4-mini.a2.inf.eff-high.dR.hi}{13.3}  
\resdef{e1f.o4-mini.a2.inf.eff-high.dR.lo}{-3.3}  
\resdef{e1f.o4-mini.a2.inf.eff-high.dR.mean}{5.0}  
\resdef{e1f.o4-mini.a3-conn.inf.eff-high.dR.hi}{2.5}  
\resdef{e1f.o4-mini.a3-conn.inf.eff-high.dR.lo}{-15.0}  
\resdef{e1f.o4-mini.a3-conn.inf.eff-high.dR.mean}{-6.2}  
\resdef{e1f.o4-mini.a3-gradadj.inf.eff-high.dR.hi}{15.0}  
\resdef{e1f.o4-mini.a3-gradadj.inf.eff-high.dR.lo}{-5.0}  
\resdef{e1f.o4-mini.a3-gradadj.inf.eff-high.dR.mean}{5.0}  
\resdef{e1f.o4-mini.a3-modal.inf.eff-high.dR.hi}{5.0}  
\resdef{e1f.o4-mini.a3-modal.inf.eff-high.dR.lo}{-30.0}  
\resdef{e1f.o4-mini.a3-modal.inf.eff-high.dR.mean}{-12.5}  
\resdef{e1f.o4-mini.a3-quant.inf.eff-high.dR.hi}{6.2}  
\resdef{e1f.o4-mini.a3-quant.inf.eff-high.dR.lo}{-8.8}  
\resdef{e1f.o4-mini.a3-quant.inf.eff-high.dR.mean}{-1.2}  
\resdef{e1f.o4-mini.b2.yn.eff-high.dacc.hi}{5.0}  
\resdef{e1f.o4-mini.b2.yn.eff-high.dacc.lo}{-10.0}  
\resdef{e1f.o4-mini.b2.yn.eff-high.dacc.mean}{-2.5}  
\resdef{e7.exaone-deep-7p8b.a3-quant.inf.nat.R.mean}{17.9}  
\resdef{e7.exaone-deep-7p8b.a3-quant.inf.p0128.R.mean}{61.5}  
\resdef{e7.llama-nemotron-nano-8b.a3-quant.inf.ntforced.R.mean}{3.8}  
\resdef{e7.nemotron-nano-9b-v2.a3-quant.inf.nat.R.mean}{11.5}  
\resdef{e7.nemotron-nano-9b-v2.a3-quant.inf.p0128.R.mean}{64.1}  
\resdef{e7.phi4-reasoning.a2.inf.ntforced.L.mean}{-1.08}  
\resdef{e7.phi4-reasoning.a2.inf.p0032.L.mean}{-1.00}  
\resdef{e7.phi4-reasoning.a2.inf.p0032.R.mean}{40.6}  
\resdef{e7.phi4-reasoning.a3-quant.inf.nat.R.mean}{9.0}  
\resdef{e7.phi4-reasoning.a3-quant.inf.ntforced.R.mean}{61.5}  
\resdef{e7.phi4-reasoning.a3-quant.inf.traj.a.share}{75.3}  
\resdef{e7.qwen3-8b.a2.inf.nat.dR.mean}{6.4}  
\resdef{e7.qwen3-8b.a3-quant.inf.p0128.dR.mean}{23.1}  
\resdef{e7.qwen3p5-4b.a2.inf.p0032.R.mean}{41.7}  
\resdef{e7.qwen3p5-4b.a2.inf.p0064.R.mean}{43.6}  
\resdef{e7.qwen3p5-9b.a2.inf.ntforced.L.mean}{-2.13}  
\resdef{e7.qwen3p5-9b.a2.inf.p0032.L.mean}{-1.17}  
\resdef{e7.qwen3p5-9b.a2.inf.p0032.R.mean}{44.1}  
\resdef{e7.qwen3p5-9b.a3-quant.inf.nat.R.mean}{0.0}  
\resdef{e7.qwen3p5-9b.a3-quant.inf.ntforced.R.mean}{35.9}  
\resdef{e7.qwen3p5-9b.a3-quant.inf.p0032.R.mean}{57.1}  
\resdef{e7.qwen3p8-27b.a3-conn.inf.nat.R.mean}{0.0}  
\resdef{e7.qwen3p8-27b.a3-conn.inf.ntforced.R.mean}{38.4}  
\resdef{e7.qwen3p8-27b.a3-conn.inf.traj.a.share}{55.0}  
\resdef{e7.qwen3p8-27b.a3-quant.inf.nat.R.mean}{16.7}  
\resdef{e7.qwen3p8-27b.a3-quant.inf.ntforced.R.mean}{28.2}  
\resdef{e7.skywork-or1-7b.a3-conn.inf.ntforced.R.mean}{75.8}  
\resdef{e7.skywork-or1-7b.a3-conn.inf.p0032.R.mean}{45.5}  
\resdef{e7f.qwen3-14b.a3f-conn.inf.nat.R.mean}{0.3}  
\resdef{e7f.qwen3-14b.a3f-conn.inf.p0128.dR.mean}{18.8}  
\resdef{e7f.qwen3-14b.a3f-conn.inf.traj.a.share}{41.4}  
\resdef{e7f.qwen3-32b.a3f-quant.inf.nat.dR.mean}{-56.7}  
\resdef{e7f.qwen3-32b.a3f-quant.inf.p0128.dR.mean}{-17.2}  
\resdef{e7f.qwen3-8b.a3f-conn.inf.p0128.R.n}{300}  
\resdef{e7f.qwen3-8b.a3f-conn.inf.p0128.dR.mean}{16.9}  
\resdef{e7f.qwen3-8b.a3f-conn.inf.traj.a.share}{68.1}  
\resdef{e7f.qwen3-8b.a3f-quant.inf.marker.ctrl_marker.mean}{39.1}  
\resdef{e7f.qwen3-8b.a3f-quant.inf.marker.pre.mean}{20.6}  
\resdef{e7f.qwen3-8b.a3f-quant.inf.nat.R.mean}{1.1}  
\resdef{e7f.qwen3-8b.a3f-quant.inf.ntforced.R.mean}{35.8}  
\resdef{e7f.qwen3-8b.a3f-quant.inf.p0128.R.mean}{52.0}  
\resdef{e7f.qwen3-8b.a3f-quant.inf.p0128.R.n}{300}  
\resdef{e7f.qwen3-8b.a3f-quant.inf.p0128.dR.hi}{19.8}  
\resdef{e7f.qwen3-8b.a3f-quant.inf.p0128.dR.lo}{12.8}  
\resdef{e7f.qwen3-8b.a3f-quant.inf.p0128.dR.mean}{16.2}  
\resdef{e7f.qwen3-8b.a3f-quant.inf.traj.a.share}{67.4}  
\resdef{e7f.qwen3-8b.a3f-quant.inf.traj.b.share}{31.4}  
\resdef{e7f.qwen3p5-9b.a3f-quant.inf.traj_n.dropped_invalid}{1{,}163}  
\resdef{e7f.qwen3p5-9b.a3f-quant.inf.traj_n.traces}{637}  
\resdef{e4.qwen3-14b.a3.L0.off.R.mean}{14.5}  
\resdef{e4.qwen3-14b.a3.Lx2048.off.R.mean}{20.9}  
\resdef{e4.qwen3-8b.a3.L0.off.R.mean}{22.6}  
\resdef{e4.qwen3-8b.a3.Lx2048.off.R.mean}{14.2}  
\resdef{e9.qwen3-8b.b2.k0.c0512.acc.mean}{84.0}  
\resdef{e9.qwen3-8b.b2.k0.nat.acc.mean}{82.0}  
\resdef{e9.qwen3-8b.b2.k0.off.acc.mean}{79.7}  
\resdef{e9.qwen3-8b.b2.k30.c0512.acc.mean}{78.1}  
\resdef{e9.qwen3-8b.b2.k30.nat.acc.mean}{78.7}  
\resdef{e9.qwen3-8b.b2.k30.off.acc.mean}{67.6}  
\resdef{e9.qwen3-8b.b2.k5.c0512.acc.mean}{79.6}  
\resdef{e9.qwen3-8b.b2.k5.nat.acc.mean}{78.6}  
\resdef{e9.qwen3-8b.b2.k5.off.acc.mean}{71.9}  
\resdef{e1p.magistral-small.a2.inf.nat.dR.hi}{5.7}  
\resdef{e1p.magistral-small.a2.inf.nat.dR.lo}{1.3}  
\resdef{e1p.magistral-small.a2.inf.nat.dR.mean}{3.3}  
\resdef{e1p.magistral-small.a2.inf.nat.dR.std}{2.6}  
\resdef{e1p.magistral-small.a3-conn.inf.nat.dR.hi}{29.3}  
\resdef{e1p.magistral-small.a3-conn.inf.nat.dR.lo}{11.1}  
\resdef{e1p.magistral-small.a3-conn.inf.nat.dR.mean}{20.2}  
\resdef{e1p.magistral-small.a3-conn.inf.nat.dR.std}{4.6}  
\resdef{e1p.magistral-small.a3-modal.inf.nat.dR.hi}{18.8}  
\resdef{e1p.magistral-small.a3-modal.inf.nat.dR.lo}{3.1}  
\resdef{e1p.magistral-small.a3-modal.inf.nat.dR.mean}{10.4}  
\resdef{e1p.magistral-small.a3-modal.inf.nat.dR.std}{3.6}  
\resdef{e1p.magistral-small.a3-quant.inf.nat.dR.hi}{16.7}  
\resdef{e1p.magistral-small.a3-quant.inf.nat.dR.lo}{2.6}  
\resdef{e1p.magistral-small.a3-quant.inf.nat.dR.mean}{9.0}  
\resdef{e1p.magistral-small.a3-quant.inf.nat.dR.std}{4.4}  
\resdef{e1p.magistral-small.a3.inf.nat.dR.hi}{14.5}  
\resdef{e1p.magistral-small.a3.inf.nat.dR.lo}{7.7}  
\resdef{e1p.magistral-small.a3.inf.nat.dR.mean}{10.9}  
\resdef{e1p.magistral-small.a3.inf.nat.dR.std}{0.6}  
\resdef{e1p.magistral-small.b2.yn.nat.dacc.hi}{23.9}  
\resdef{e1p.magistral-small.b2.yn.nat.dacc.lo}{7.4}  
\resdef{e1p.magistral-small.b2.yn.nat.dacc.mean}{15.8}  
\resdef{e1p.magistral-small.b2.yn.nat.dacc.std}{4.6}  
\resdef{e1p.phi4-reasoning.a2.inf.nat.dR.hi}{-1.3}  
\resdef{e1p.phi4-reasoning.a2.inf.nat.dR.lo}{-9.1}  
\resdef{e1p.phi4-reasoning.a2.inf.nat.dR.mean}{-5.1}  
\resdef{e1p.phi4-reasoning.a2.inf.nat.dR.std}{2.1}  
\resdef{e1p.phi4-reasoning.a3-conn.inf.nat.dR.hi}{-27.3}  
\resdef{e1p.phi4-reasoning.a3-conn.inf.nat.dR.lo}{-50.5}  
\resdef{e1p.phi4-reasoning.a3-conn.inf.nat.dR.mean}{-38.4}  
\resdef{e1p.phi4-reasoning.a3-conn.inf.nat.dR.std}{7.6}  
\resdef{e1p.phi4-reasoning.a3-modal.inf.nat.dR.hi}{0.0}  
\resdef{e1p.phi4-reasoning.a3-modal.inf.nat.dR.lo}{-32.3}  
\resdef{e1p.phi4-reasoning.a3-modal.inf.nat.dR.mean}{-16.7}  
\resdef{e1p.phi4-reasoning.a3-modal.inf.nat.dR.std}{15.8}  
\resdef{e1p.phi4-reasoning.a3-quant.inf.nat.dR.hi}{-38.5}  
\resdef{e1p.phi4-reasoning.a3-quant.inf.nat.dR.lo}{-67.9}  
\resdef{e1p.phi4-reasoning.a3-quant.inf.nat.dR.mean}{-53.8}  
\resdef{e1p.phi4-reasoning.a3-quant.inf.nat.dR.std}{7.7}  
\resdef{e1p.phi4-reasoning.a3.inf.nat.dR.hi}{-10.6}  
\resdef{e1p.phi4-reasoning.a3.inf.nat.dR.lo}{-23.6}  
\resdef{e1p.phi4-reasoning.a3.inf.nat.dR.mean}{-17.1}  
\resdef{e1p.phi4-reasoning.a3.inf.nat.dR.std}{2.3}  
\resdef{e1p.phi4-reasoning.b2.yn.nat.dacc.hi}{26.3}  
\resdef{e1p.phi4-reasoning.b2.yn.nat.dacc.lo}{8.8}  
\resdef{e1p.phi4-reasoning.b2.yn.nat.dacc.mean}{17.5}  
\resdef{e1p.phi4-reasoning.b2.yn.nat.dacc.std}{3.2}  
\resdef{e1p.r1distill-llama-70b.a2.inf.nat.dR.hi}{2.6}  
\resdef{e1p.r1distill-llama-70b.a2.inf.nat.dR.lo}{-5.7}  
\resdef{e1p.r1distill-llama-70b.a2.inf.nat.dR.mean}{-1.5}  
\resdef{e1p.r1distill-llama-70b.a2.inf.nat.dR.std}{1.4}  
\resdef{e1p.r1distill-llama-70b.a3-conn.inf.nat.dR.hi}{-27.3}  
\resdef{e1p.r1distill-llama-70b.a3-conn.inf.nat.dR.lo}{-51.5}  
\resdef{e1p.r1distill-llama-70b.a3-conn.inf.nat.dR.mean}{-39.4}  
\resdef{e1p.r1distill-llama-70b.a3-conn.inf.nat.dR.std}{3.0}  
\resdef{e1p.r1distill-llama-70b.a3-modal.inf.nat.dR.hi}{-15.6}  
\resdef{e1p.r1distill-llama-70b.a3-modal.inf.nat.dR.lo}{-41.7}  
\resdef{e1p.r1distill-llama-70b.a3-modal.inf.nat.dR.mean}{-28.1}  
\resdef{e1p.r1distill-llama-70b.a3-modal.inf.nat.dR.std}{10.8}  
\resdef{e1p.r1distill-llama-70b.a3-quant.inf.nat.dR.hi}{-33.3}  
\resdef{e1p.r1distill-llama-70b.a3-quant.inf.nat.dR.lo}{-66.7}  
\resdef{e1p.r1distill-llama-70b.a3-quant.inf.nat.dR.mean}{-50.0}  
\resdef{e1p.r1distill-llama-70b.a3-quant.inf.nat.dR.std}{7.7}  
\resdef{e1p.r1distill-llama-70b.a3.inf.nat.dR.hi}{-29.1}  
\resdef{e1p.r1distill-llama-70b.a3.inf.nat.dR.lo}{-40.3}  
\resdef{e1p.r1distill-llama-70b.a3.inf.nat.dR.mean}{-34.7}  
\resdef{e1p.r1distill-llama-70b.a3.inf.nat.dR.std}{0.6}  
\resdef{e1p.r1distill-llama-70b.b2.yn.nat.dacc.hi}{16.5}  
\resdef{e1p.r1distill-llama-70b.b2.yn.nat.dacc.lo}{3.5}  
\resdef{e1p.r1distill-llama-70b.b2.yn.nat.dacc.mean}{9.8}  
\resdef{e1p.r1distill-llama-70b.b2.yn.nat.dacc.std}{4.4}  
\resdef{e1p.r1distill-llama-8b.a2.inf.nat.dR.hi}{-26.3}  
\resdef{e1p.r1distill-llama-8b.a2.inf.nat.dR.lo}{-39.3}  
\resdef{e1p.r1distill-llama-8b.a2.inf.nat.dR.mean}{-32.9}  
\resdef{e1p.r1distill-llama-8b.a2.inf.nat.dR.std}{6.3}  
\resdef{e1p.r1distill-llama-8b.a3-conn.inf.nat.dR.hi}{-46.5}  
\resdef{e1p.r1distill-llama-8b.a3-conn.inf.nat.dR.lo}{-68.7}  
\resdef{e1p.r1distill-llama-8b.a3-conn.inf.nat.dR.mean}{-57.6}  
\resdef{e1p.r1distill-llama-8b.a3-conn.inf.nat.dR.std}{3.0}  
\resdef{e1p.r1distill-llama-8b.a3-modal.inf.nat.dR.hi}{-21.9}  
\resdef{e1p.r1distill-llama-8b.a3-modal.inf.nat.dR.lo}{-49.0}  
\resdef{e1p.r1distill-llama-8b.a3-modal.inf.nat.dR.mean}{-35.4}  
\resdef{e1p.r1distill-llama-8b.a3-modal.inf.nat.dR.std}{10.0}  
\resdef{e1p.r1distill-llama-8b.a3-quant.inf.nat.dR.hi}{-34.6}  
\resdef{e1p.r1distill-llama-8b.a3-quant.inf.nat.dR.lo}{-65.4}  
\resdef{e1p.r1distill-llama-8b.a3-quant.inf.nat.dR.mean}{-50.0}  
\resdef{e1p.r1distill-llama-8b.a3-quant.inf.nat.dR.std}{7.7}  
\resdef{e1p.r1distill-llama-8b.a3.inf.nat.dR.hi}{-31.8}  
\resdef{e1p.r1distill-llama-8b.a3.inf.nat.dR.lo}{-43.6}  
\resdef{e1p.r1distill-llama-8b.a3.inf.nat.dR.mean}{-37.8}  
\resdef{e1p.r1distill-llama-8b.a3.inf.nat.dR.std}{4.2}  
\resdef{e1p.r1distill-llama-8b.b2.yn.nat.dacc.hi}{35.4}  
\resdef{e1p.r1distill-llama-8b.b2.yn.nat.dacc.lo}{16.8}  
\resdef{e1p.r1distill-llama-8b.b2.yn.nat.dacc.mean}{26.0}  
\resdef{e1p.r1distill-llama-8b.b2.yn.nat.dacc.std}{3.7}  
\resdef{e1b.acereason-7b.a2.inf.nat.dR.hi}{-6.8}  
\resdef{e1b.acereason-7b.a2.inf.nat.dR.lo}{-22.3}  
\resdef{e1b.acereason-7b.a2.inf.nat.dR.mean}{-14.6}  
\resdef{e1b.acereason-7b.a2.inf.nat.dR.std}{6.5}  
\resdef{e1b.acereason-7b.a3-conn.inf.nat.dR.hi}{-22.2}  
\resdef{e1b.acereason-7b.a3-conn.inf.nat.dR.lo}{-53.5}  
\resdef{e1b.acereason-7b.a3-conn.inf.nat.dR.mean}{-38.4}  
\resdef{e1b.acereason-7b.a3-conn.inf.nat.dR.std}{12.6}  
\resdef{e1b.acereason-7b.a3-modal.inf.nat.dR.hi}{11.5}  
\resdef{e1b.acereason-7b.a3-modal.inf.nat.dR.lo}{-16.7}  
\resdef{e1b.acereason-7b.a3-modal.inf.nat.dR.mean}{-2.1}  
\resdef{e1b.acereason-7b.a3-modal.inf.nat.dR.std}{7.9}  
\resdef{e1b.acereason-7b.a3-quant.inf.nat.dR.hi}{-6.4}  
\resdef{e1b.acereason-7b.a3-quant.inf.nat.dR.lo}{-44.9}  
\resdef{e1b.acereason-7b.a3-quant.inf.nat.dR.mean}{-25.6}  
\resdef{e1b.acereason-7b.a3-quant.inf.nat.dR.std}{11.8}  
\resdef{e1b.acereason-7b.a3.inf.nat.dR.hi}{-5.0}  
\resdef{e1b.acereason-7b.a3.inf.nat.dR.lo}{-18.8}  
\resdef{e1b.acereason-7b.a3.inf.nat.dR.mean}{-12.0}  
\resdef{e1b.acereason-7b.a3.inf.nat.dR.std}{4.3}  
\resdef{e1b.acereason-7b.b2.yn.nat.dacc.hi}{25.3}  
\resdef{e1b.acereason-7b.b2.yn.nat.dacc.lo}{8.1}  
\resdef{e1b.acereason-7b.b2.yn.nat.dacc.mean}{16.8}  
\resdef{e1b.acereason-7b.b2.yn.nat.dacc.std}{5.6}  
\resdef{e1b.exaone-deep-7p8b.a2.inf.nat.dR.hi}{3.4}  
\resdef{e1b.exaone-deep-7p8b.a2.inf.nat.dR.lo}{-4.3}  
\resdef{e1b.exaone-deep-7p8b.a2.inf.nat.dR.mean}{-0.4}  
\resdef{e1b.exaone-deep-7p8b.a2.inf.nat.dR.std}{1.4}  
\resdef{e1b.exaone-deep-7p8b.a3-conn.inf.nat.dR.hi}{-12.1}  
\resdef{e1b.exaone-deep-7p8b.a3-conn.inf.nat.dR.lo}{-36.9}  
\resdef{e1b.exaone-deep-7p8b.a3-conn.inf.nat.dR.mean}{-24.7}  
\resdef{e1b.exaone-deep-7p8b.a3-conn.inf.nat.dR.std}{8.7}  
\resdef{e1b.exaone-deep-7p8b.a3-modal.inf.nat.dR.hi}{33.9}  
\resdef{e1b.exaone-deep-7p8b.a3-modal.inf.nat.dR.lo}{6.8}  
\resdef{e1b.exaone-deep-7p8b.a3-modal.inf.nat.dR.mean}{20.8}  
\resdef{e1b.exaone-deep-7p8b.a3-modal.inf.nat.dR.std}{13.4}  
\resdef{e1b.exaone-deep-7p8b.a3-quant.inf.nat.dR.hi}{-26.9}  
\resdef{e1b.exaone-deep-7p8b.a3-quant.inf.nat.dR.lo}{-53.2}  
\resdef{e1b.exaone-deep-7p8b.a3-quant.inf.nat.dR.mean}{-40.4}  
\resdef{e1b.exaone-deep-7p8b.a3-quant.inf.nat.dR.std}{6.8}  
\resdef{e1b.exaone-deep-7p8b.a3.inf.nat.dR.hi}{0.6}  
\resdef{e1b.exaone-deep-7p8b.a3.inf.nat.dR.lo}{-12.6}  
\resdef{e1b.exaone-deep-7p8b.a3.inf.nat.dR.mean}{-6.0}  
\resdef{e1b.exaone-deep-7p8b.a3.inf.nat.dR.std}{3.4}  
\resdef{e1b.exaone-deep-7p8b.b2.yn.nat.dacc.hi}{9.8}  
\resdef{e1b.exaone-deep-7p8b.b2.yn.nat.dacc.lo}{-4.2}  
\resdef{e1b.exaone-deep-7p8b.b2.yn.nat.dacc.mean}{2.8}  
\resdef{e1b.exaone-deep-7p8b.b2.yn.nat.dacc.std}{3.9}  
\resdef{e1b.granite-3p3-8b.a2.inf.nat.dR.hi}{1.5}  
\resdef{e1b.granite-3p3-8b.a2.inf.nat.dR.lo}{-2.6}  
\resdef{e1b.granite-3p3-8b.a2.inf.nat.dR.mean}{-0.4}  
\resdef{e1b.granite-3p3-8b.a2.inf.nat.dR.std}{2.5}  
\resdef{e1b.granite-3p3-8b.a3-conn.inf.nat.dR.hi}{32.3}  
\resdef{e1b.granite-3p3-8b.a3-conn.inf.nat.dR.lo}{14.1}  
\resdef{e1b.granite-3p3-8b.a3-conn.inf.nat.dR.mean}{23.2}  
\resdef{e1b.granite-3p3-8b.a3-conn.inf.nat.dR.std}{7.6}  
\resdef{e1b.granite-3p3-8b.a3-modal.inf.nat.dR.hi}{14.6}  
\resdef{e1b.granite-3p3-8b.a3-modal.inf.nat.dR.lo}{1.0}  
\resdef{e1b.granite-3p3-8b.a3-modal.inf.nat.dR.mean}{7.3}  
\resdef{e1b.granite-3p3-8b.a3-modal.inf.nat.dR.std}{3.6}  
\resdef{e1b.granite-3p3-8b.a3-quant.inf.nat.dR.hi}{21.8}  
\resdef{e1b.granite-3p3-8b.a3-quant.inf.nat.dR.lo}{-0.0}  
\resdef{e1b.granite-3p3-8b.a3-quant.inf.nat.dR.mean}{10.3}  
\resdef{e1b.granite-3p3-8b.a3-quant.inf.nat.dR.std}{5.9}  
\resdef{e1b.granite-3p3-8b.a3.inf.nat.dR.hi}{16.4}  
\resdef{e1b.granite-3p3-8b.a3.inf.nat.dR.lo}{9.1}  
\resdef{e1b.granite-3p3-8b.a3.inf.nat.dR.mean}{12.7}  
\resdef{e1b.granite-3p3-8b.a3.inf.nat.dR.std}{1.6}  
\resdef{e1b.granite-3p3-8b.b2.yn.nat.dacc.hi}{12.3}  
\resdef{e1b.granite-3p3-8b.b2.yn.nat.dacc.lo}{-1.4}  
\resdef{e1b.granite-3p3-8b.b2.yn.nat.dacc.mean}{5.3}  
\resdef{e1b.granite-3p3-8b.b2.yn.nat.dacc.std}{0.0}  
\resdef{e1b.llama-nemotron-nano-8b.a2.inf.nat.dR.hi}{15.5}  
\resdef{e1b.llama-nemotron-nano-8b.a2.inf.nat.dR.lo}{8.3}  
\resdef{e1b.llama-nemotron-nano-8b.a2.inf.nat.dR.mean}{11.7}  
\resdef{e1b.llama-nemotron-nano-8b.a2.inf.nat.dR.std}{3.5}  
\resdef{e1b.llama-nemotron-nano-8b.a3-conn.inf.nat.dR.hi}{7.1}  
\resdef{e1b.llama-nemotron-nano-8b.a3-conn.inf.nat.dR.lo}{1.0}  
\resdef{e1b.llama-nemotron-nano-8b.a3-conn.inf.nat.dR.mean}{3.5}  
\resdef{e1b.llama-nemotron-nano-8b.a3-conn.inf.nat.dR.std}{3.5}  
\resdef{e1b.llama-nemotron-nano-8b.a3-modal.inf.nat.dR.hi}{37.0}  
\resdef{e1b.llama-nemotron-nano-8b.a3-modal.inf.nat.dR.lo}{16.1}  
\resdef{e1b.llama-nemotron-nano-8b.a3-modal.inf.nat.dR.mean}{26.6}  
\resdef{e1b.llama-nemotron-nano-8b.a3-modal.inf.nat.dR.std}{5.8}  
\resdef{e1b.llama-nemotron-nano-8b.a3-quant.inf.nat.dR.hi}{19.9}  
\resdef{e1b.llama-nemotron-nano-8b.a3-quant.inf.nat.dR.lo}{7.7}  
\resdef{e1b.llama-nemotron-nano-8b.a3-quant.inf.nat.dR.mean}{13.5}  
\resdef{e1b.llama-nemotron-nano-8b.a3-quant.inf.nat.dR.std}{8.3}  
\resdef{e1b.llama-nemotron-nano-8b.a3.inf.nat.dR.hi}{17.9}  
\resdef{e1b.llama-nemotron-nano-8b.a3.inf.nat.dR.lo}{11.5}  
\resdef{e1b.llama-nemotron-nano-8b.a3.inf.nat.dR.mean}{14.6}  
\resdef{e1b.llama-nemotron-nano-8b.a3.inf.nat.dR.std}{1.8}  
\resdef{e1b.llama-nemotron-nano-8b.b2.yn.nat.dacc.hi}{23.5}  
\resdef{e1b.llama-nemotron-nano-8b.b2.yn.nat.dacc.lo}{7.0}  
\resdef{e1b.llama-nemotron-nano-8b.b2.yn.nat.dacc.mean}{15.1}  
\resdef{e1b.llama-nemotron-nano-8b.b2.yn.nat.dacc.std}{3.0}  
\resdef{e1b.llama-nemotron-super-49b.a2.inf.nat.dR.hi}{10.6}  
\resdef{e1b.llama-nemotron-super-49b.a2.inf.nat.dR.lo}{2.6}  
\resdef{e1b.llama-nemotron-super-49b.a2.inf.nat.dR.mean}{6.6}  
\resdef{e1b.llama-nemotron-super-49b.a2.inf.nat.dR.std}{2.3}  
\resdef{e1b.llama-nemotron-super-49b.a3-conn.inf.nat.dR.hi}{6.1}  
\resdef{e1b.llama-nemotron-super-49b.a3-conn.inf.nat.dR.lo}{-27.3}  
\resdef{e1b.llama-nemotron-super-49b.a3-conn.inf.nat.dR.mean}{-11.1}  
\resdef{e1b.llama-nemotron-super-49b.a3-conn.inf.nat.dR.std}{12.6}  
\resdef{e1b.llama-nemotron-super-49b.a3-modal.inf.nat.dR.hi}{13.6}  
\resdef{e1b.llama-nemotron-super-49b.a3-modal.inf.nat.dR.lo}{-9.4}  
\resdef{e1b.llama-nemotron-super-49b.a3-modal.inf.nat.dR.mean}{2.1}  
\resdef{e1b.llama-nemotron-super-49b.a3-modal.inf.nat.dR.std}{10.0}  
\resdef{e1b.llama-nemotron-super-49b.a3-quant.inf.nat.dR.hi}{-10.3}  
\resdef{e1b.llama-nemotron-super-49b.a3-quant.inf.nat.dR.lo}{-46.2}  
\resdef{e1b.llama-nemotron-super-49b.a3-quant.inf.nat.dR.mean}{-28.2}  
\resdef{e1b.llama-nemotron-super-49b.a3-quant.inf.nat.dR.std}{16.0}  
\resdef{e1b.llama-nemotron-super-49b.a3.inf.nat.dR.hi}{2.1}  
\resdef{e1b.llama-nemotron-super-49b.a3.inf.nat.dR.lo}{-9.6}  
\resdef{e1b.llama-nemotron-super-49b.a3.inf.nat.dR.mean}{-3.6}  
\resdef{e1b.llama-nemotron-super-49b.a3.inf.nat.dR.std}{3.9}  
\resdef{e1b.llama-nemotron-super-49b.b2.yn.nat.dacc.hi}{6.7}  
\resdef{e1b.llama-nemotron-super-49b.b2.yn.nat.dacc.lo}{-3.9}  
\resdef{e1b.llama-nemotron-super-49b.b2.yn.nat.dacc.mean}{1.4}  
\resdef{e1b.llama-nemotron-super-49b.b2.yn.nat.dacc.std}{1.6}  
\resdef{e1b.mimo-7b-rl.a2.inf.nat.dR.hi}{-0.9}  
\resdef{e1b.mimo-7b-rl.a2.inf.nat.dR.lo}{-14.3}  
\resdef{e1b.mimo-7b-rl.a2.inf.nat.dR.mean}{-7.5}  
\resdef{e1b.mimo-7b-rl.a2.inf.nat.dR.std}{1.7}  
\resdef{e1b.mimo-7b-rl.a3-conn.inf.nat.dR.hi}{-22.2}  
\resdef{e1b.mimo-7b-rl.a3-conn.inf.nat.dR.lo}{-54.5}  
\resdef{e1b.mimo-7b-rl.a3-conn.inf.nat.dR.mean}{-39.4}  
\resdef{e1b.mimo-7b-rl.a3-conn.inf.nat.dR.std}{5.2}  
\resdef{e1b.mimo-7b-rl.a3-modal.inf.nat.dR.hi}{12.5}  
\resdef{e1b.mimo-7b-rl.a3-modal.inf.nat.dR.lo}{-15.6}  
\resdef{e1b.mimo-7b-rl.a3-modal.inf.nat.dR.mean}{-1.0}  
\resdef{e1b.mimo-7b-rl.a3-modal.inf.nat.dR.std}{12.6}  
\resdef{e1b.mimo-7b-rl.a3-quant.inf.nat.dR.hi}{-32.0}  
\resdef{e1b.mimo-7b-rl.a3-quant.inf.nat.dR.lo}{-59.0}  
\resdef{e1b.mimo-7b-rl.a3-quant.inf.nat.dR.mean}{-44.9}  
\resdef{e1b.mimo-7b-rl.a3-quant.inf.nat.dR.std}{4.4}  
\resdef{e1b.mimo-7b-rl.a3.inf.nat.dR.hi}{-3.4}  
\resdef{e1b.mimo-7b-rl.a3.inf.nat.dR.lo}{-17.1}  
\resdef{e1b.mimo-7b-rl.a3.inf.nat.dR.mean}{-10.4}  
\resdef{e1b.mimo-7b-rl.a3.inf.nat.dR.std}{2.6}  
\resdef{e1b.mimo-7b-rl.b2.yn.nat.dacc.hi}{35.1}  
\resdef{e1b.mimo-7b-rl.b2.yn.nat.dacc.lo}{16.8}  
\resdef{e1b.mimo-7b-rl.b2.yn.nat.dacc.mean}{26.3}  
\resdef{e1b.mimo-7b-rl.b2.yn.nat.dacc.std}{1.1}  
\resdef{e1b.mimo-7b-sft.a2.inf.nat.dR.hi}{-15.9}  
\resdef{e1b.mimo-7b-sft.a2.inf.nat.dR.lo}{-30.0}  
\resdef{e1b.mimo-7b-sft.a2.inf.nat.dR.mean}{-23.0}  
\resdef{e1b.mimo-7b-sft.a2.inf.nat.dR.std}{3.0}  
\resdef{e1b.mimo-7b-sft.a3-conn.inf.nat.dR.hi}{-29.3}  
\resdef{e1b.mimo-7b-sft.a3-conn.inf.nat.dR.lo}{-50.5}  
\resdef{e1b.mimo-7b-sft.a3-conn.inf.nat.dR.mean}{-40.4}  
\resdef{e1b.mimo-7b-sft.a3-conn.inf.nat.dR.std}{7.6}  
\resdef{e1b.mimo-7b-sft.a3-modal.inf.nat.dR.hi}{-12.5}  
\resdef{e1b.mimo-7b-sft.a3-modal.inf.nat.dR.lo}{-37.5}  
\resdef{e1b.mimo-7b-sft.a3-modal.inf.nat.dR.mean}{-25.0}  
\resdef{e1b.mimo-7b-sft.a3-modal.inf.nat.dR.std}{6.2}  
\resdef{e1b.mimo-7b-sft.a3-quant.inf.nat.dR.hi}{-44.9}  
\resdef{e1b.mimo-7b-sft.a3-quant.inf.nat.dR.lo}{-74.4}  
\resdef{e1b.mimo-7b-sft.a3-quant.inf.nat.dR.mean}{-60.3}  
\resdef{e1b.mimo-7b-sft.a3-quant.inf.nat.dR.std}{2.2}  
\resdef{e1b.mimo-7b-sft.a3.inf.nat.dR.hi}{-19.3}  
\resdef{e1b.mimo-7b-sft.a3.inf.nat.dR.lo}{-32.1}  
\resdef{e1b.mimo-7b-sft.a3.inf.nat.dR.mean}{-25.8}  
\resdef{e1b.mimo-7b-sft.a3.inf.nat.dR.std}{3.9}  
\resdef{e1b.mimo-7b-sft.b2.yn.nat.dacc.hi}{30.2}  
\resdef{e1b.mimo-7b-sft.b2.yn.nat.dacc.lo}{10.9}  
\resdef{e1b.mimo-7b-sft.b2.yn.nat.dacc.mean}{20.4}  
\resdef{e1b.mimo-7b-sft.b2.yn.nat.dacc.std}{1.2}  
\resdef{e1b.nemotron-nano-9b-v2.a2.inf.nat.dR.hi}{7.4}  
\resdef{e1b.nemotron-nano-9b-v2.a2.inf.nat.dR.lo}{2.6}  
\resdef{e1b.nemotron-nano-9b-v2.a2.inf.nat.dR.mean}{4.9}  
\resdef{e1b.nemotron-nano-9b-v2.a2.inf.nat.dR.std}{1.3}  
\resdef{e1b.nemotron-nano-9b-v2.a3-conn.inf.nat.dR.hi}{-27.3}  
\resdef{e1b.nemotron-nano-9b-v2.a3-conn.inf.nat.dR.lo}{-60.1}  
\resdef{e1b.nemotron-nano-9b-v2.a3-conn.inf.nat.dR.mean}{-43.4}  
\resdef{e1b.nemotron-nano-9b-v2.a3-conn.inf.nat.dR.std}{6.5}  
\resdef{e1b.nemotron-nano-9b-v2.a3-modal.inf.nat.dR.hi}{0.5}  
\resdef{e1b.nemotron-nano-9b-v2.a3-modal.inf.nat.dR.lo}{-18.8}  
\resdef{e1b.nemotron-nano-9b-v2.a3-modal.inf.nat.dR.mean}{-8.3}  
\resdef{e1b.nemotron-nano-9b-v2.a3-modal.inf.nat.dR.std}{10.0}  
\resdef{e1b.nemotron-nano-9b-v2.a3-quant.inf.nat.dR.hi}{-41.7}  
\resdef{e1b.nemotron-nano-9b-v2.a3-quant.inf.nat.dR.lo}{-75.6}  
\resdef{e1b.nemotron-nano-9b-v2.a3-quant.inf.nat.dR.mean}{-59.0}  
\resdef{e1b.nemotron-nano-9b-v2.a3-quant.inf.nat.dR.std}{4.0}  
\resdef{e1b.nemotron-nano-9b-v2.a3.inf.nat.dR.hi}{-10.6}  
\resdef{e1b.nemotron-nano-9b-v2.a3.inf.nat.dR.lo}{-24.1}  
\resdef{e1b.nemotron-nano-9b-v2.a3.inf.nat.dR.mean}{-17.4}  
\resdef{e1b.nemotron-nano-9b-v2.a3.inf.nat.dR.std}{3.6}  
\resdef{e1b.nemotron-nano-9b-v2.b2.yn.nat.dacc.hi}{14.2}  
\resdef{e1b.nemotron-nano-9b-v2.b2.yn.nat.dacc.lo}{1.4}  
\resdef{e1b.nemotron-nano-9b-v2.b2.yn.nat.dacc.mean}{7.7}  
\resdef{e1b.nemotron-nano-9b-v2.b2.yn.nat.dacc.std}{2.5}  
\resdef{e1b.phi4-mini-reasoning.a2.inf.nat.dR.hi}{-12.9}  
\resdef{e1b.phi4-mini-reasoning.a2.inf.nat.dR.lo}{-22.2}  
\resdef{e1b.phi4-mini-reasoning.a2.inf.nat.dR.mean}{-17.6}  
\resdef{e1b.phi4-mini-reasoning.a2.inf.nat.dR.std}{1.0}  
\resdef{e1b.phi4-mini-reasoning.a3-conn.inf.nat.dR.hi}{-2.0}  
\resdef{e1b.phi4-mini-reasoning.a3-conn.inf.nat.dR.lo}{-25.3}  
\resdef{e1b.phi4-mini-reasoning.a3-conn.inf.nat.dR.mean}{-13.6}  
\resdef{e1b.phi4-mini-reasoning.a3-conn.inf.nat.dR.std}{3.2}  
\resdef{e1b.phi4-mini-reasoning.a3-modal.inf.nat.dR.hi}{0.5}  
\resdef{e1b.phi4-mini-reasoning.a3-modal.inf.nat.dR.lo}{-21.4}  
\resdef{e1b.phi4-mini-reasoning.a3-modal.inf.nat.dR.mean}{-10.4}  
\resdef{e1b.phi4-mini-reasoning.a3-modal.inf.nat.dR.std}{10.3}  
\resdef{e1b.phi4-mini-reasoning.a3-quant.inf.nat.dR.hi}{2.6}  
\resdef{e1b.phi4-mini-reasoning.a3-quant.inf.nat.dR.lo}{-17.9}  
\resdef{e1b.phi4-mini-reasoning.a3-quant.inf.nat.dR.mean}{-7.7}  
\resdef{e1b.phi4-mini-reasoning.a3-quant.inf.nat.dR.std}{8.3}  
\resdef{e1b.phi4-mini-reasoning.a3.inf.nat.dR.hi}{-8.4}  
\resdef{e1b.phi4-mini-reasoning.a3.inf.nat.dR.lo}{-17.9}  
\resdef{e1b.phi4-mini-reasoning.a3.inf.nat.dR.mean}{-13.1}  
\resdef{e1b.phi4-mini-reasoning.a3.inf.nat.dR.std}{2.9}  
\resdef{e1b.phi4-mini-reasoning.b2.yn.nat.dacc.hi}{26.7}  
\resdef{e1b.phi4-mini-reasoning.b2.yn.nat.dacc.lo}{10.0}  
\resdef{e1b.phi4-mini-reasoning.b2.yn.nat.dacc.mean}{18.3}  
\resdef{e1b.phi4-mini-reasoning.b2.yn.nat.dacc.std}{4.8}  
\resdef{e1b.qwen3-4b-thinking-2507.a2.inf.nat.dR.hi}{10.4}  
\resdef{e1b.qwen3-4b-thinking-2507.a2.inf.nat.dR.lo}{3.5}  
\resdef{e1b.qwen3-4b-thinking-2507.a2.inf.nat.dR.mean}{6.8}  
\resdef{e1b.qwen3-4b-thinking-2507.a2.inf.nat.dR.std}{2.7}  
\resdef{e1b.qwen3-4b-thinking-2507.a3-conn.inf.nat.dR.hi}{-38.4}  
\resdef{e1b.qwen3-4b-thinking-2507.a3-conn.inf.nat.dR.lo}{-58.6}  
\resdef{e1b.qwen3-4b-thinking-2507.a3-conn.inf.nat.dR.mean}{-48.5}  
\resdef{e1b.qwen3-4b-thinking-2507.a3-conn.inf.nat.dR.std}{3.0}  
\resdef{e1b.qwen3-4b-thinking-2507.a3-modal.inf.nat.dR.hi}{-27.1}  
\resdef{e1b.qwen3-4b-thinking-2507.a3-modal.inf.nat.dR.lo}{-55.2}  
\resdef{e1b.qwen3-4b-thinking-2507.a3-modal.inf.nat.dR.mean}{-41.7}  
\resdef{e1b.qwen3-4b-thinking-2507.a3-modal.inf.nat.dR.std}{4.8}  
\resdef{e1b.qwen3-4b-thinking-2507.a3-quant.inf.nat.dR.hi}{-43.6}  
\resdef{e1b.qwen3-4b-thinking-2507.a3-quant.inf.nat.dR.lo}{-69.2}  
\resdef{e1b.qwen3-4b-thinking-2507.a3-quant.inf.nat.dR.mean}{-56.4}  
\resdef{e1b.qwen3-4b-thinking-2507.a3-quant.inf.nat.dR.std}{9.7}  
\resdef{e1b.qwen3-4b-thinking-2507.a3.inf.nat.dR.hi}{-28.3}  
\resdef{e1b.qwen3-4b-thinking-2507.a3.inf.nat.dR.lo}{-39.5}  
\resdef{e1b.qwen3-4b-thinking-2507.a3.inf.nat.dR.mean}{-33.9}  
\resdef{e1b.qwen3-4b-thinking-2507.a3.inf.nat.dR.std}{1.0}  
\resdef{e1b.qwen3-4b-thinking-2507.b2.yn.nat.dacc.hi}{11.2}  
\resdef{e1b.qwen3-4b-thinking-2507.b2.yn.nat.dacc.lo}{-0.4}  
\resdef{e1b.qwen3-4b-thinking-2507.b2.yn.nat.dacc.mean}{5.3}  
\resdef{e1b.qwen3-4b-thinking-2507.b2.yn.nat.dacc.std}{3.7}  
\resdef{e1b.qwq-32b.a2.inf.nat.dR.hi}{9.7}  
\resdef{e1b.qwq-32b.a2.inf.nat.dR.lo}{3.1}  
\resdef{e1b.qwq-32b.a2.inf.nat.dR.mean}{6.2}  
\resdef{e1b.qwq-32b.a2.inf.nat.dR.std}{0.4}  
\resdef{e1b.qwq-32b.a3-conn.inf.nat.dR.hi}{13.1}  
\resdef{e1b.qwq-32b.a3-conn.inf.nat.dR.lo}{-6.1}  
\resdef{e1b.qwq-32b.a3-conn.inf.nat.dR.mean}{4.0}  
\resdef{e1b.qwq-32b.a3-conn.inf.nat.dR.std}{7.0}  
\resdef{e1b.qwq-32b.a3-modal.inf.nat.dR.hi}{35.4}  
\resdef{e1b.qwq-32b.a3-modal.inf.nat.dR.lo}{11.5}  
\resdef{e1b.qwq-32b.a3-modal.inf.nat.dR.mean}{22.9}  
\resdef{e1b.qwq-32b.a3-modal.inf.nat.dR.std}{4.8}  
\resdef{e1b.qwq-32b.a3-quant.inf.nat.dR.hi}{-7.7}  
\resdef{e1b.qwq-32b.a3-quant.inf.nat.dR.lo}{-39.7}  
\resdef{e1b.qwq-32b.a3-quant.inf.nat.dR.mean}{-23.1}  
\resdef{e1b.qwq-32b.a3-quant.inf.nat.dR.std}{3.8}  
\resdef{e1b.qwq-32b.a3.inf.nat.dR.hi}{25.0}  
\resdef{e1b.qwq-32b.a3.inf.nat.dR.lo}{12.1}  
\resdef{e1b.qwq-32b.a3.inf.nat.dR.mean}{18.5}  
\resdef{e1b.qwq-32b.a3.inf.nat.dR.std}{1.8}  
\resdef{e1b.qwq-32b.b2.yn.nat.dacc.hi}{18.9}  
\resdef{e1b.qwq-32b.b2.yn.nat.dacc.lo}{3.9}  
\resdef{e1b.qwq-32b.b2.yn.nat.dacc.mean}{11.2}  
\resdef{e1b.qwq-32b.b2.yn.nat.dacc.std}{1.6}  
\resdef{e1b.skywork-or1-7b.a2.inf.nat.dR.hi}{-13.2}  
\resdef{e1b.skywork-or1-7b.a2.inf.nat.dR.lo}{-28.0}  
\resdef{e1b.skywork-or1-7b.a2.inf.nat.dR.mean}{-20.5}  
\resdef{e1b.skywork-or1-7b.a2.inf.nat.dR.std}{3.5}  
\resdef{e1b.skywork-or1-7b.a3-conn.inf.nat.dR.hi}{-20.2}  
\resdef{e1b.skywork-or1-7b.a3-conn.inf.nat.dR.lo}{-49.5}  
\resdef{e1b.skywork-or1-7b.a3-conn.inf.nat.dR.mean}{-34.3}  
\resdef{e1b.skywork-or1-7b.a3-conn.inf.nat.dR.std}{16.7}  
\resdef{e1b.skywork-or1-7b.a3-modal.inf.nat.dR.hi}{16.7}  
\resdef{e1b.skywork-or1-7b.a3-modal.inf.nat.dR.lo}{-15.6}  
\resdef{e1b.skywork-or1-7b.a3-modal.inf.nat.dR.mean}{1.0}  
\resdef{e1b.skywork-or1-7b.a3-modal.inf.nat.dR.std}{9.5}  
\resdef{e1b.skywork-or1-7b.a3-quant.inf.nat.dR.hi}{3.8}  
\resdef{e1b.skywork-or1-7b.a3-quant.inf.nat.dR.lo}{-29.5}  
\resdef{e1b.skywork-or1-7b.a3-quant.inf.nat.dR.mean}{-14.1}  
\resdef{e1b.skywork-or1-7b.a3-quant.inf.nat.dR.std}{5.9}  
\resdef{e1b.skywork-or1-7b.a3.inf.nat.dR.hi}{-3.8}  
\resdef{e1b.skywork-or1-7b.a3.inf.nat.dR.lo}{-17.1}  
\resdef{e1b.skywork-or1-7b.a3.inf.nat.dR.mean}{-10.4}  
\resdef{e1b.skywork-or1-7b.a3.inf.nat.dR.std}{4.9}  
\resdef{e1b.skywork-or1-7b.b2.yn.nat.dacc.hi}{19.6}  
\resdef{e1b.skywork-or1-7b.b2.yn.nat.dacc.lo}{1.8}  
\resdef{e1b.skywork-or1-7b.b2.yn.nat.dacc.mean}{10.5}  
\resdef{e1b.skywork-or1-7b.b2.yn.nat.dacc.std}{6.6}  
\resdef{e1b.smollm3-3b.a2.inf.nat.dR.hi}{6.2}  
\resdef{e1b.smollm3-3b.a2.inf.nat.dR.lo}{-0.4}  
\resdef{e1b.smollm3-3b.a2.inf.nat.dR.mean}{2.9}  
\resdef{e1b.smollm3-3b.a2.inf.nat.dR.std}{1.7}  
\resdef{e1b.smollm3-3b.a3-conn.inf.nat.dR.hi}{29.3}  
\resdef{e1b.smollm3-3b.a3-conn.inf.nat.dR.lo}{-5.1}  
\resdef{e1b.smollm3-3b.a3-conn.inf.nat.dR.mean}{12.1}  
\resdef{e1b.smollm3-3b.a3-conn.inf.nat.dR.std}{3.0}  
\resdef{e1b.smollm3-3b.a3-modal.inf.nat.dR.hi}{31.2}  
\resdef{e1b.smollm3-3b.a3-modal.inf.nat.dR.lo}{2.1}  
\resdef{e1b.smollm3-3b.a3-modal.inf.nat.dR.mean}{16.7}  
\resdef{e1b.smollm3-3b.a3-modal.inf.nat.dR.std}{1.8}  
\resdef{e1b.smollm3-3b.a3-quant.inf.nat.dR.hi}{15.4}  
\resdef{e1b.smollm3-3b.a3-quant.inf.nat.dR.lo}{-28.2}  
\resdef{e1b.smollm3-3b.a3-quant.inf.nat.dR.mean}{-6.4}  
\resdef{e1b.smollm3-3b.a3-quant.inf.nat.dR.std}{8.0}  
\resdef{e1b.smollm3-3b.a3.inf.nat.dR.hi}{20.5}  
\resdef{e1b.smollm3-3b.a3.inf.nat.dR.lo}{6.5}  
\resdef{e1b.smollm3-3b.a3.inf.nat.dR.mean}{13.5}  
\resdef{e1b.smollm3-3b.a3.inf.nat.dR.std}{0.6}  
\resdef{e1b.smollm3-3b.b2.yn.nat.dacc.hi}{6.7}  
\resdef{e1b.smollm3-3b.b2.yn.nat.dacc.lo}{-6.3}  
\resdef{e1b.smollm3-3b.b2.yn.nat.dacc.mean}{0.4}  
\resdef{e1b.smollm3-3b.b2.yn.nat.dacc.std}{4.0}  
\resdef{e1sf.qwen3-14b.a3f-conn.inf.nat.dR.hi}{-12.7}  
\resdef{e1sf.qwen3-14b.a3f-conn.inf.nat.dR.lo}{-19.1}  
\resdef{e1sf.qwen3-14b.a3f-conn.inf.nat.dR.mean}{-15.9}  
\resdef{e1sf.qwen3-14b.a3f-conn.nli.nat.dR.hi}{-1.0}  
\resdef{e1sf.qwen3-14b.a3f-conn.nli.nat.dR.lo}{-4.0}  
\resdef{e1sf.qwen3-14b.a3f-conn.nli.nat.dR.mean}{-2.4}  
\resdef{e1sf.qwen3-14b.a3f-modal.inf.nat.dR.mean}{1.2}  
\resdef{e1sf.qwen3-14b.a3f-quant.inf.nat.dR.hi}{-18.3}  
\resdef{e1sf.qwen3-14b.a3f-quant.inf.nat.dR.lo}{-27.1}  
\resdef{e1sf.qwen3-14b.a3f-quant.inf.nat.dR.mean}{-22.6}  
\resdef{e1sf.qwen3-14b.a3f-quant.nli.nat.dR.hi}{-8.6}  
\resdef{e1sf.qwen3-14b.a3f-quant.nli.nat.dR.lo}{-14.4}  
\resdef{e1sf.qwen3-14b.a3f-quant.nli.nat.dR.mean}{-11.4}  
\resdef{e1sf.qwen3-30b-a3b.a3f-conn.inf.nat.dR.hi}{-36.8}  
\resdef{e1sf.qwen3-30b-a3b.a3f-conn.inf.nat.dR.lo}{-45.1}  
\resdef{e1sf.qwen3-30b-a3b.a3f-conn.inf.nat.dR.mean}{-40.9}  
\resdef{e1sf.qwen3-30b-a3b.a3f-conn.nli.nat.dR.hi}{-9.3}  
\resdef{e1sf.qwen3-30b-a3b.a3f-conn.nli.nat.dR.lo}{-14.3}  
\resdef{e1sf.qwen3-30b-a3b.a3f-conn.nli.nat.dR.mean}{-11.8}  
\resdef{e1sf.qwen3-30b-a3b.a3f-quant.inf.nat.dR.hi}{-31.4}  
\resdef{e1sf.qwen3-30b-a3b.a3f-quant.inf.nat.dR.lo}{-39.8}  
\resdef{e1sf.qwen3-30b-a3b.a3f-quant.inf.nat.dR.mean}{-35.7}  
\resdef{e1sf.qwen3-30b-a3b.a3f-quant.nli.nat.dR.hi}{-10.1}  
\resdef{e1sf.qwen3-30b-a3b.a3f-quant.nli.nat.dR.lo}{-15.7}  
\resdef{e1sf.qwen3-30b-a3b.a3f-quant.nli.nat.dR.mean}{-12.8}  
\resdef{e1sf.qwen3-32b.a3f-conn.inf.nat.dR.hi}{-48.3}  
\resdef{e1sf.qwen3-32b.a3f-conn.inf.nat.dR.lo}{-57.0}  
\resdef{e1sf.qwen3-32b.a3f-conn.inf.nat.dR.mean}{-52.7}  
\resdef{e1sf.qwen3-32b.a3f-conn.nli.nat.dR.hi}{-16.3}  
\resdef{e1sf.qwen3-32b.a3f-conn.nli.nat.dR.lo}{-23.1}  
\resdef{e1sf.qwen3-32b.a3f-conn.nli.nat.dR.mean}{-19.7}  
\resdef{e1sf.qwen3-32b.a3f-modal.inf.nat.dR.mean}{-13.6}  
\resdef{e1sf.qwen3-32b.a3f-quant.inf.nat.dR.hi}{-50.8}  
\resdef{e1sf.qwen3-32b.a3f-quant.inf.nat.dR.lo}{-60.1}  
\resdef{e1sf.qwen3-32b.a3f-quant.inf.nat.dR.mean}{-55.6}  
\resdef{e1sf.qwen3-32b.a3f-quant.nli.nat.dR.hi}{-33.1}  
\resdef{e1sf.qwen3-32b.a3f-quant.nli.nat.dR.lo}{-42.9}  
\resdef{e1sf.qwen3-32b.a3f-quant.nli.nat.dR.mean}{-38.0}  
\resdef{e1sf.qwen3-4b.a3f-conn.inf.nat.dR.hi}{-48.4}  
\resdef{e1sf.qwen3-4b.a3f-conn.inf.nat.dR.lo}{-57.6}  
\resdef{e1sf.qwen3-4b.a3f-conn.inf.nat.dR.mean}{-53.0}  
\resdef{e1sf.qwen3-4b.a3f-conn.nli.nat.dR.hi}{-9.2}  
\resdef{e1sf.qwen3-4b.a3f-conn.nli.nat.dR.lo}{-14.7}  
\resdef{e1sf.qwen3-4b.a3f-conn.nli.nat.dR.mean}{-11.9}  
\resdef{e1sf.qwen3-4b.a3f-quant.inf.nat.dR.hi}{-45.6}  
\resdef{e1sf.qwen3-4b.a3f-quant.inf.nat.dR.lo}{-53.9}  
\resdef{e1sf.qwen3-4b.a3f-quant.inf.nat.dR.mean}{-49.8}  
\resdef{e1sf.qwen3-4b.a3f-quant.nli.nat.dR.hi}{-7.3}  
\resdef{e1sf.qwen3-4b.a3f-quant.nli.nat.dR.lo}{-12.7}  
\resdef{e1sf.qwen3-4b.a3f-quant.nli.nat.dR.mean}{-9.9}  
\resdef{e1sf.qwen3-8b.a3f-conn.inf.nat.dR.hi}{-33.8}  
\resdef{e1sf.qwen3-8b.a3f-conn.inf.nat.dR.lo}{-41.6}  
\resdef{e1sf.qwen3-8b.a3f-conn.inf.nat.dR.mean}{-37.7}  
\resdef{e1sf.qwen3-8b.a3f-conn.nli.nat.dR.hi}{-10.6}  
\resdef{e1sf.qwen3-8b.a3f-conn.nli.nat.dR.lo}{-17.1}  
\resdef{e1sf.qwen3-8b.a3f-conn.nli.nat.dR.mean}{-13.8}  
\resdef{e1sf.qwen3-8b.a3f-quant.inf.nat.dR.hi}{-33.3}  
\resdef{e1sf.qwen3-8b.a3f-quant.inf.nat.dR.lo}{-42.1}  
\resdef{e1sf.qwen3-8b.a3f-quant.inf.nat.dR.mean}{-37.7}  
\resdef{e1sf.qwen3-8b.a3f-quant.inf.nat.dR.n}{300}  
\resdef{e1sf.qwen3-8b.a3f-quant.inf.off.R.mean}{38.8}  
\resdef{e1sf.qwen3-8b.a3f-quant.nli.nat.dR.hi}{-29.9}  
\resdef{e1sf.qwen3-8b.a3f-quant.nli.nat.dR.lo}{-38.2}  
\resdef{e1sf.qwen3-8b.a3f-quant.nli.nat.dR.mean}{-34.0}  
\resdef{e1xt.qwen3-14b.b3.mc.nat.acc.hi}{94.7}  
\resdef{e1xt.qwen3-14b.b3.mc.nat.acc.lo}{88.3}  
\resdef{e1xt.qwen3-14b.b3.mc.nat.acc.mean}{91.6}  
\resdef{e1xt.qwen3-14b.b3.mc.nat.acc.std}{1.4}  
\resdef{e1xt.qwen3-14b.b3.mc.nat.dacc.hi}{10.1}  
\resdef{e1xt.qwen3-14b.b3.mc.nat.dacc.lo}{3.4}  
\resdef{e1xt.qwen3-14b.b3.mc.nat.dacc.mean}{6.7}  
\resdef{e1xt.qwen3-14b.b3.mc.nat.dacc.std}{0.6}  
\resdef{e1xt.qwen3-14b.b3.mc.off.acc.hi}{89.1}  
\resdef{e1xt.qwen3-14b.b3.mc.off.acc.lo}{80.4}  
\resdef{e1xt.qwen3-14b.b3.mc.off.acc.mean}{84.9}  
\resdef{e1xt.qwen3-14b.b3.mc.off.acc.std}{1.1}  
\resdef{e1xt.qwen3-4b.b3.mc.nat.acc.hi}{90.9}  
\resdef{e1xt.qwen3-4b.b3.mc.nat.acc.lo}{83.4}  
\resdef{e1xt.qwen3-4b.b3.mc.nat.acc.mean}{87.3}  
\resdef{e1xt.qwen3-4b.b3.mc.nat.acc.std}{0.0}  
\resdef{e1xt.qwen3-4b.b3.mc.nat.dacc.hi}{14.0}  
\resdef{e1xt.qwen3-4b.b3.mc.nat.dacc.lo}{6.9}  
\resdef{e1xt.qwen3-4b.b3.mc.nat.dacc.mean}{10.5}  
\resdef{e1xt.qwen3-4b.b3.mc.nat.dacc.std}{1.2}  
\resdef{e1xt.qwen3-4b.b3.mc.off.acc.hi}{81.6}  
\resdef{e1xt.qwen3-4b.b3.mc.off.acc.lo}{71.8}  
\resdef{e1xt.qwen3-4b.b3.mc.off.acc.mean}{76.9}  
\resdef{e1xt.qwen3-4b.b3.mc.off.acc.std}{1.2}  
\resdef{e1xt.qwen3-8b.b3.mc.nat.acc.hi}{92.7}  
\resdef{e1xt.qwen3-8b.b3.mc.nat.acc.lo}{85.6}  
\resdef{e1xt.qwen3-8b.b3.mc.nat.acc.mean}{89.3}  
\resdef{e1xt.qwen3-8b.b3.mc.nat.acc.std}{1.0}  
\resdef{e1xt.qwen3-8b.b3.mc.nat.dacc.hi}{14.8}  
\resdef{e1xt.qwen3-8b.b3.mc.nat.dacc.lo}{6.8}  
\resdef{e1xt.qwen3-8b.b3.mc.nat.dacc.mean}{10.7}  
\resdef{e1xt.qwen3-8b.b3.mc.nat.dacc.std}{0.5}  
\resdef{e1xt.qwen3-8b.b3.mc.off.acc.hi}{83.4}  
\resdef{e1xt.qwen3-8b.b3.mc.off.acc.lo}{73.5}  
\resdef{e1xt.qwen3-8b.b3.mc.off.acc.mean}{78.5}  
\resdef{e1xt.qwen3-8b.b3.mc.off.acc.std}{0.6}  
\resdef{e1xt.qwen3-8b.b4.mc.nat.dacc.mean}{10.1}  
\resdef{a6.qwen3-8b.a6-naturally.can.c0512.acc.mean}{81.3}  
\resdef{a6.qwen3-8b.a6-naturally.can.nat.acc.mean}{85.3}  
\resdef{a6.qwen3-8b.a6-naturally.can.off.acc.mean}{76.0}  
\resdef{a6.qwen3-8b.a6-naturally.rec.c0512.acc.mean}{40.0}  
\resdef{a6.qwen3-8b.a6-naturally.rec.nat.acc.mean}{38.7}  
\resdef{a6.qwen3-8b.a6-naturally.rec.off.acc.mean}{36.0}  
\resdef{a6.qwen3-8b.a6-scalar.can.c0512.acc.mean}{85.5}  
\resdef{a6.qwen3-8b.a6-scalar.can.nat.acc.mean}{94.2}  
\resdef{a6.qwen3-8b.a6-scalar.can.off.acc.mean}{46.4}  
\resdef{a6.qwen3-8b.a6-scalar.rec.c0512.acc.mean}{89.9}  
\resdef{a6.qwen3-8b.a6-scalar.rec.nat.acc.mean}{65.2}  
\resdef{a6.qwen3-8b.a6-scalar.rec.off.acc.mean}{100.0}  
\resdef{e1probe.granite-3p3-8b.a3.infflip.nat.dR.hi}{15.2}  
\resdef{e1probe.granite-3p3-8b.a3.infflip.nat.dR.lo}{3.1}  
\resdef{e1probe.granite-3p3-8b.a3.infflip.nat.dR.mean}{9.2}  
\resdef{e1probe.granite-3p3-8b.a3.infprior.nat.dR.mean}{6.5}  
\resdef{e1probe.llama-nemotron-nano-8b.a3.infflip.nat.dR.mean}{17.5}  
\resdef{e1probe.llama-nemotron-nano-8b.a3.infprior.nat.dR.mean}{25.4}  
\resdef{e1probe.magistral-small.a3.infflip.nat.dR.hi}{-46.7}  
\resdef{e1probe.magistral-small.a3.infflip.nat.dR.lo}{-59.0}  
\resdef{e1probe.magistral-small.a3.infflip.nat.dR.mean}{-52.9}  
\resdef{e1probe.magistral-small.a3.infprior.nat.dR.hi}{18.3}  
\resdef{e1probe.magistral-small.a3.infprior.nat.dR.lo}{10.1}  
\resdef{e1probe.magistral-small.a3.infprior.nat.dR.mean}{14.1}  
\resdef{e1probe.qwen3-8b.a3.infflip.nat.dR.mean}{22.7}  
\resdef{e1probe.qwen3p5-9b.a3.infflip.nat.dR.mean}{-31.7}  
\resdef{e1q.qwen3p5-0p8b.a2.inf.nat.dR.hi}{-12.4}  
\resdef{e1q.qwen3p5-0p8b.a2.inf.nat.dR.lo}{-22.5}  
\resdef{e1q.qwen3p5-0p8b.a2.inf.nat.dR.mean}{-17.2}  
\resdef{e1q.qwen3p5-0p8b.a2.inf.nat.dR.std}{1.3}  
\resdef{e1q.qwen3p5-0p8b.a3-conn.inf.nat.dR.hi}{-1.0}  
\resdef{e1q.qwen3p5-0p8b.a3-conn.inf.nat.dR.lo}{-27.3}  
\resdef{e1q.qwen3p5-0p8b.a3-conn.inf.nat.dR.mean}{-14.1}  
\resdef{e1q.qwen3p5-0p8b.a3-conn.inf.nat.dR.std}{6.3}  
\resdef{e1q.qwen3p5-0p8b.a3-modal.inf.nat.dR.hi}{-3.1}  
\resdef{e1q.qwen3p5-0p8b.a3-modal.inf.nat.dR.lo}{-19.8}  
\resdef{e1q.qwen3p5-0p8b.a3-modal.inf.nat.dR.mean}{-10.4}  
\resdef{e1q.qwen3p5-0p8b.a3-modal.inf.nat.dR.std}{9.5}  
\resdef{e1q.qwen3p5-0p8b.a3-quant.inf.nat.dR.hi}{2.6}  
\resdef{e1q.qwen3p5-0p8b.a3-quant.inf.nat.dR.lo}{-26.9}  
\resdef{e1q.qwen3p5-0p8b.a3-quant.inf.nat.dR.mean}{-11.5}  
\resdef{e1q.qwen3p5-0p8b.a3-quant.inf.nat.dR.std}{3.8}  
\resdef{e1q.qwen3p5-0p8b.a3.inf.nat.dR.hi}{-5.5}  
\resdef{e1q.qwen3p5-0p8b.a3.inf.nat.dR.lo}{-13.7}  
\resdef{e1q.qwen3p5-0p8b.a3.inf.nat.dR.mean}{-9.4}  
\resdef{e1q.qwen3p5-0p8b.a3.inf.nat.dR.std}{2.8}  
\resdef{e1q.qwen3p5-0p8b.b2.yn.nat.dacc.hi}{2.1}  
\resdef{e1q.qwen3p5-0p8b.b2.yn.nat.dacc.lo}{-13.3}  
\resdef{e1q.qwen3p5-0p8b.b2.yn.nat.dacc.mean}{-5.6}  
\resdef{e1q.qwen3p5-0p8b.b2.yn.nat.dacc.std}{2.2}  
\resdef{e1q.qwen3p5-2b.a2.inf.nat.dR.hi}{17.4}  
\resdef{e1q.qwen3p5-2b.a2.inf.nat.dR.lo}{9.9}  
\resdef{e1q.qwen3p5-2b.a2.inf.nat.dR.mean}{13.5}  
\resdef{e1q.qwen3p5-2b.a2.inf.nat.dR.std}{2.0}  
\resdef{e1q.qwen3p5-2b.a3-conn.inf.nat.dR.hi}{-16.2}  
\resdef{e1q.qwen3p5-2b.a3-conn.inf.nat.dR.lo}{-47.5}  
\resdef{e1q.qwen3p5-2b.a3-conn.inf.nat.dR.mean}{-32.3}  
\resdef{e1q.qwen3p5-2b.a3-conn.inf.nat.dR.std}{8.7}  
\resdef{e1q.qwen3p5-2b.a3-modal.inf.nat.dR.hi}{-8.3}  
\resdef{e1q.qwen3p5-2b.a3-modal.inf.nat.dR.lo}{-37.5}  
\resdef{e1q.qwen3p5-2b.a3-modal.inf.nat.dR.mean}{-22.9}  
\resdef{e1q.qwen3p5-2b.a3-modal.inf.nat.dR.std}{7.2}  
\resdef{e1q.qwen3p5-2b.a3-quant.inf.nat.dR.hi}{1.3}  
\resdef{e1q.qwen3p5-2b.a3-quant.inf.nat.dR.lo}{-30.8}  
\resdef{e1q.qwen3p5-2b.a3-quant.inf.nat.dR.mean}{-14.1}  
\resdef{e1q.qwen3p5-2b.a3-quant.inf.nat.dR.std}{2.2}  
\resdef{e1q.qwen3p5-2b.a3.inf.nat.dR.hi}{-4.6}  
\resdef{e1q.qwen3p5-2b.a3.inf.nat.dR.lo}{-16.8}  
\resdef{e1q.qwen3p5-2b.a3.inf.nat.dR.mean}{-10.6}  
\resdef{e1q.qwen3p5-2b.a3.inf.nat.dR.std}{3.9}  
\resdef{e1q.qwen3p5-2b.b2.yn.nat.dacc.hi}{4.2}  
\resdef{e1q.qwen3p5-2b.b2.yn.nat.dacc.lo}{-8.1}  
\resdef{e1q.qwen3p5-2b.b2.yn.nat.dacc.mean}{-1.8}  
\resdef{e1q.qwen3p5-2b.b2.yn.nat.dacc.std}{0.6}  
\resdef{e1q.qwen3p5-35b-a3b.a2.inf.nat.dR.hi}{10.8}  
\resdef{e1q.qwen3p5-35b-a3b.a2.inf.nat.dR.lo}{3.5}  
\resdef{e1q.qwen3p5-35b-a3b.a2.inf.nat.dR.mean}{6.8}  
\resdef{e1q.qwen3p5-35b-a3b.a2.inf.nat.dR.std}{1.0}  
\resdef{e1q.qwen3p5-35b-a3b.a3-conn.inf.nat.dR.hi}{5.1}  
\resdef{e1q.qwen3p5-35b-a3b.a3-conn.inf.nat.dR.lo}{-9.1}  
\resdef{e1q.qwen3p5-35b-a3b.a3-conn.inf.nat.dR.mean}{-1.0}  
\resdef{e1q.qwen3p5-35b-a3b.a3-conn.inf.nat.dR.std}{3.5}  
\resdef{e1q.qwen3p5-35b-a3b.a3-modal.inf.nat.dR.hi}{21.9}  
\resdef{e1q.qwen3p5-35b-a3b.a3-modal.inf.nat.dR.lo}{3.1}  
\resdef{e1q.qwen3p5-35b-a3b.a3-modal.inf.nat.dR.mean}{11.5}  
\resdef{e1q.qwen3p5-35b-a3b.a3-modal.inf.nat.dR.std}{3.6}  
\resdef{e1q.qwen3p5-35b-a3b.a3-quant.inf.nat.dR.hi}{5.1}  
\resdef{e1q.qwen3p5-35b-a3b.a3-quant.inf.nat.dR.lo}{-16.7}  
\resdef{e1q.qwen3p5-35b-a3b.a3-quant.inf.nat.dR.mean}{-5.1}  
\resdef{e1q.qwen3p5-35b-a3b.a3-quant.inf.nat.dR.std}{2.2}  
\resdef{e1q.qwen3p5-35b-a3b.a3.inf.nat.dR.hi}{12.7}  
\resdef{e1q.qwen3p5-35b-a3b.a3.inf.nat.dR.lo}{3.2}  
\resdef{e1q.qwen3p5-35b-a3b.a3.inf.nat.dR.mean}{7.9}  
\resdef{e1q.qwen3p5-35b-a3b.a3.inf.nat.dR.std}{1.3}  
\resdef{e1q.qwen3p5-35b-a3b.b2.yn.nat.dacc.hi}{8.4}  
\resdef{e1q.qwen3p5-35b-a3b.b2.yn.nat.dacc.lo}{0.7}  
\resdef{e1q.qwen3p5-35b-a3b.b2.yn.nat.dacc.mean}{4.2}  
\resdef{e1q.qwen3p5-35b-a3b.b2.yn.nat.dacc.std}{1.1}  
\resdef{e1q.qwen3p5-4b.a2.inf.nat.dR.hi}{5.3}  
\resdef{e1q.qwen3p5-4b.a2.inf.nat.dR.lo}{-0.6}  
\resdef{e1q.qwen3p5-4b.a2.inf.nat.dR.mean}{2.3}  
\resdef{e1q.qwen3p5-4b.a2.inf.nat.dR.std}{1.7}  
\resdef{e1q.qwen3p5-4b.a3-conn.inf.nat.dR.hi}{-22.7}  
\resdef{e1q.qwen3p5-4b.a3-conn.inf.nat.dR.lo}{-47.0}  
\resdef{e1q.qwen3p5-4b.a3-conn.inf.nat.dR.mean}{-34.8}  
\resdef{e1q.qwen3p5-4b.a3-conn.inf.nat.dR.std}{7.8}  
\resdef{e1q.qwen3p5-4b.a3-modal.inf.nat.dR.hi}{9.9}  
\resdef{e1q.qwen3p5-4b.a3-modal.inf.nat.dR.lo}{-9.9}  
\resdef{e1q.qwen3p5-4b.a3-modal.inf.nat.dR.mean}{-0.0}  
\resdef{e1q.qwen3p5-4b.a3-modal.inf.nat.dR.std}{6.8}  
\resdef{e1q.qwen3p5-4b.a3-quant.inf.nat.dR.hi}{-21.8}  
\resdef{e1q.qwen3p5-4b.a3-quant.inf.nat.dR.lo}{-56.4}  
\resdef{e1q.qwen3p5-4b.a3-quant.inf.nat.dR.mean}{-39.1}  
\resdef{e1q.qwen3p5-4b.a3-quant.inf.nat.dR.std}{3.8}  
\resdef{e1q.qwen3p5-4b.a3.inf.nat.dR.hi}{2.8}  
\resdef{e1q.qwen3p5-4b.a3.inf.nat.dR.lo}{-9.3}  
\resdef{e1q.qwen3p5-4b.a3.inf.nat.dR.mean}{-3.2}  
\resdef{e1q.qwen3p5-4b.a3.inf.nat.dR.std}{3.2}  
\resdef{e1q.qwen3p5-4b.b2.yn.nat.dacc.hi}{10.4}  
\resdef{e1q.qwen3p5-4b.b2.yn.nat.dacc.lo}{-2.8}  
\resdef{e1q.qwen3p5-4b.b2.yn.nat.dacc.mean}{3.9}  
\resdef{e1q.qwen3p5-4b.b2.yn.nat.dacc.std}{1.4}  
\resdef{e1q.qwen3p5-9b.a2.inf.nat.dR.hi}{7.4}  
\resdef{e1q.qwen3p5-9b.a2.inf.nat.dR.lo}{1.9}  
\resdef{e1q.qwen3p5-9b.a2.inf.nat.dR.mean}{4.4}  
\resdef{e1q.qwen3p5-9b.a2.inf.nat.dR.std}{1.2}  
\resdef{e1q.qwen3p5-9b.a3-conn.inf.nat.dR.hi}{-11.1}  
\resdef{e1q.qwen3p5-9b.a3-conn.inf.nat.dR.lo}{-36.4}  
\resdef{e1q.qwen3p5-9b.a3-conn.inf.nat.dR.mean}{-23.2}  
\resdef{e1q.qwen3p5-9b.a3-conn.inf.nat.dR.std}{1.6}  
\resdef{e1q.qwen3p5-9b.a3-modal.inf.nat.dR.hi}{9.9}  
\resdef{e1q.qwen3p5-9b.a3-modal.inf.nat.dR.lo}{-4.2}  
\resdef{e1q.qwen3p5-9b.a3-modal.inf.nat.dR.mean}{2.6}  
\resdef{e1q.qwen3p5-9b.a3-modal.inf.nat.dR.std}{5.4}  
\resdef{e1q.qwen3p5-9b.a3-quant.inf.nat.dR.hi}{-24.4}  
\resdef{e1q.qwen3p5-9b.a3-quant.inf.nat.dR.lo}{-55.1}  
\resdef{e1q.qwen3p5-9b.a3-quant.inf.nat.dR.mean}{-39.1}  
\resdef{e1q.qwen3p5-9b.a3-quant.inf.nat.dR.std}{3.8}  
\resdef{e1q.qwen3p5-9b.a3.inf.nat.dR.hi}{-2.8}  
\resdef{e1q.qwen3p5-9b.a3.inf.nat.dR.lo}{-12.6}  
\resdef{e1q.qwen3p5-9b.a3.inf.nat.dR.mean}{-7.6}  
\resdef{e1q.qwen3p5-9b.a3.inf.nat.dR.std}{1.4}  
\resdef{e1q.qwen3p5-9b.b2.yn.nat.dacc.hi}{11.8}  
\resdef{e1q.qwen3p5-9b.b2.yn.nat.dacc.lo}{0.9}  
\resdef{e1q.qwen3p5-9b.b2.yn.nat.dacc.mean}{6.0}  
\resdef{e1q.qwen3p5-9b.b2.yn.nat.dacc.std}{2.2}  
\resdef{e1q.qwen3p8-27b-elow.a2.inf.nat.dR.hi}{19.2}  
\resdef{e1q.qwen3p8-27b-elow.a2.inf.nat.dR.lo}{9.9}  
\resdef{e1q.qwen3p8-27b-elow.a2.inf.nat.dR.mean}{14.3}  
\resdef{e1q.qwen3p8-27b-elow.a2.inf.nat.dR.std}{2.0}  
\resdef{e1q.qwen3p8-27b-elow.a3-conn.inf.nat.dR.hi}{-22.2}  
\resdef{e1q.qwen3p8-27b-elow.a3-conn.inf.nat.dR.lo}{-52.5}  
\resdef{e1q.qwen3p8-27b-elow.a3-conn.inf.nat.dR.mean}{-37.4}  
\resdef{e1q.qwen3p8-27b-elow.a3-conn.inf.nat.dR.std}{12.6}  
\resdef{e1q.qwen3p8-27b-elow.a3-modal.inf.nat.dR.hi}{9.4}  
\resdef{e1q.qwen3p8-27b-elow.a3-modal.inf.nat.dR.lo}{-13.5}  
\resdef{e1q.qwen3p8-27b-elow.a3-modal.inf.nat.dR.mean}{-2.1}  
\resdef{e1q.qwen3p8-27b-elow.a3-modal.inf.nat.dR.std}{11.0}  
\resdef{e1q.qwen3p8-27b-elow.a3-quant.inf.nat.dR.hi}{-3.8}  
\resdef{e1q.qwen3p8-27b-elow.a3-quant.inf.nat.dR.lo}{-43.6}  
\resdef{e1q.qwen3p8-27b-elow.a3-quant.inf.nat.dR.mean}{-23.1}  
\resdef{e1q.qwen3p8-27b-elow.a3-quant.inf.nat.dR.std}{3.8}  
\resdef{e1q.qwen3p8-27b-elow.a3.inf.nat.dR.hi}{-3.2}  
\resdef{e1q.qwen3p8-27b-elow.a3.inf.nat.dR.lo}{-14.2}  
\resdef{e1q.qwen3p8-27b-elow.a3.inf.nat.dR.mean}{-8.7}  
\resdef{e1q.qwen3p8-27b-elow.a3.inf.nat.dR.std}{3.9}  
\resdef{e1q.qwen3p8-27b-elow.b2.yn.nat.dacc.hi}{14.4}  
\resdef{e1q.qwen3p8-27b-elow.b2.yn.nat.dacc.lo}{3.9}  
\resdef{e1q.qwen3p8-27b-elow.b2.yn.nat.dacc.mean}{8.8}  
\resdef{e1q.qwen3p8-27b-elow.b2.yn.nat.dacc.std}{1.6}  
\resdef{e1q.qwen3p8-27b-emed.a2.inf.nat.dR.hi}{15.0}  
\resdef{e1q.qwen3p8-27b-emed.a2.inf.nat.dR.lo}{6.6}  
\resdef{e1q.qwen3p8-27b-emed.a2.inf.nat.dR.mean}{10.6}  
\resdef{e1q.qwen3p8-27b-emed.a2.inf.nat.dR.std}{1.8}  
\resdef{e1q.qwen3p8-27b-emed.a3-conn.inf.nat.dR.hi}{-1.0}  
\resdef{e1q.qwen3p8-27b-emed.a3-conn.inf.nat.dR.lo}{-24.2}  
\resdef{e1q.qwen3p8-27b-emed.a3-conn.inf.nat.dR.mean}{-12.1}  
\resdef{e1q.qwen3p8-27b-emed.a3-conn.inf.nat.dR.std}{3.0}  
\resdef{e1q.qwen3p8-27b-emed.a3-modal.inf.nat.dR.hi}{28.1}  
\resdef{e1q.qwen3p8-27b-emed.a3-modal.inf.nat.dR.lo}{5.2}  
\resdef{e1q.qwen3p8-27b-emed.a3-modal.inf.nat.dR.mean}{15.6}  
\resdef{e1q.qwen3p8-27b-emed.a3-modal.inf.nat.dR.std}{6.2}  
\resdef{e1q.qwen3p8-27b-emed.a3-quant.inf.nat.dR.hi}{16.7}  
\resdef{e1q.qwen3p8-27b-emed.a3-quant.inf.nat.dR.lo}{-17.9}  
\resdef{e1q.qwen3p8-27b-emed.a3-quant.inf.nat.dR.mean}{-0.0}  
\resdef{e1q.qwen3p8-27b-emed.a3-quant.inf.nat.dR.std}{0.0}  
\resdef{e1q.qwen3p8-27b-emed.a3.inf.nat.dR.hi}{4.4}  
\resdef{e1q.qwen3p8-27b-emed.a3.inf.nat.dR.lo}{-4.6}  
\resdef{e1q.qwen3p8-27b-emed.a3.inf.nat.dR.mean}{-0.0}  
\resdef{e1q.qwen3p8-27b-emed.a3.inf.nat.dR.std}{0.5}  
\resdef{e1q.qwen3p8-27b-emed.b2.yn.nat.dacc.hi}{15.8}  
\resdef{e1q.qwen3p8-27b-emed.b2.yn.nat.dacc.lo}{2.1}  
\resdef{e1q.qwen3p8-27b-emed.b2.yn.nat.dacc.mean}{8.8}  
\resdef{e1q.qwen3p8-27b-emed.b2.yn.nat.dacc.std}{0.6}  
\resdef{e1q.qwen3p8-27b.a2.inf.nat.dR.hi}{20.3}  
\resdef{e1q.qwen3p8-27b.a2.inf.nat.dR.lo}{11.0}  
\resdef{e1q.qwen3p8-27b.a2.inf.nat.dR.mean}{15.5}  
\resdef{e1q.qwen3p8-27b.a2.inf.nat.dR.std}{3.0}  
\resdef{e1q.qwen3p8-27b.a3-conn.inf.nat.dR.hi}{-10.1}  
\resdef{e1q.qwen3p8-27b.a3-conn.inf.nat.dR.lo}{-35.4}  
\resdef{e1q.qwen3p8-27b.a3-conn.inf.nat.dR.mean}{-22.2}  
\resdef{e1q.qwen3p8-27b.a3-conn.inf.nat.dR.std}{4.6}  
\resdef{e1q.qwen3p8-27b.a3-modal.inf.nat.dR.hi}{-1.0}  
\resdef{e1q.qwen3p8-27b.a3-modal.inf.nat.dR.lo}{-16.7}  
\resdef{e1q.qwen3p8-27b.a3-modal.inf.nat.dR.mean}{-8.3}  
\resdef{e1q.qwen3p8-27b.a3-modal.inf.nat.dR.std}{1.8}  
\resdef{e1q.qwen3p8-27b.a3-quant.inf.nat.dR.hi}{0.0}  
\resdef{e1q.qwen3p8-27b.a3-quant.inf.nat.dR.lo}{-35.9}  
\resdef{e1q.qwen3p8-27b.a3-quant.inf.nat.dR.mean}{-16.7}  
\resdef{e1q.qwen3p8-27b.a3-quant.inf.nat.dR.std}{9.7}  
\resdef{e1q.qwen3p8-27b.a3.inf.nat.dR.hi}{-4.1}  
\resdef{e1q.qwen3p8-27b.a3.inf.nat.dR.lo}{-13.7}  
\resdef{e1q.qwen3p8-27b.a3.inf.nat.dR.mean}{-8.9}  
\resdef{e1q.qwen3p8-27b.a3.inf.nat.dR.std}{3.0}  
\resdef{e1q.qwen3p8-27b.b2.yn.nat.dacc.hi}{17.2}  
\resdef{e1q.qwen3p8-27b.b2.yn.nat.dacc.lo}{4.6}  
\resdef{e1q.qwen3p8-27b.b2.yn.nat.dacc.mean}{10.5}  
\resdef{e1q.qwen3p8-27b.b2.yn.nat.dacc.std}{1.1}  
\resdef{e1ds.qwen3-14b.circa.mc.nat.dacc.hi}{30.0}  
\resdef{e1ds.qwen3-14b.circa.mc.nat.dacc.lo}{18.9}  
\resdef{e1ds.qwen3-14b.circa.mc.nat.dacc.mean}{24.4}  
\resdef{e1ds.qwen3-14b.nope.nli.nat.dacc.hi}{14.9}  
\resdef{e1ds.qwen3-14b.nope.nli.nat.dacc.lo}{4.4}  
\resdef{e1ds.qwen3-14b.nope.nli.nat.dacc.mean}{9.6}  
\resdef{e1ds.qwen3-30b-a3b.circa.mc.nat.dacc.hi}{12.7}  
\resdef{e1ds.qwen3-30b-a3b.circa.mc.nat.dacc.lo}{4.0}  
\resdef{e1ds.qwen3-30b-a3b.circa.mc.nat.dacc.mean}{8.3}  
\resdef{e1ds.qwen3-30b-a3b.nope.nli.nat.dacc.hi}{15.5}  
\resdef{e1ds.qwen3-30b-a3b.nope.nli.nat.dacc.lo}{5.3}  
\resdef{e1ds.qwen3-30b-a3b.nope.nli.nat.dacc.mean}{10.4}  
\resdef{e1ds.qwen3-32b.circa.mc.nat.dacc.hi}{7.1}  
\resdef{e1ds.qwen3-32b.circa.mc.nat.dacc.lo}{-0.9}  
\resdef{e1ds.qwen3-32b.circa.mc.nat.dacc.mean}{3.1}  
\resdef{e1ds.qwen3-32b.nope.nli.nat.dacc.hi}{14.4}  
\resdef{e1ds.qwen3-32b.nope.nli.nat.dacc.lo}{3.8}  
\resdef{e1ds.qwen3-32b.nope.nli.nat.dacc.mean}{9.2}  
\resdef{e1ds.qwen3-4b.circa.mc.nat.dacc.hi}{20.4}  
\resdef{e1ds.qwen3-4b.circa.mc.nat.dacc.lo}{10.4}  
\resdef{e1ds.qwen3-4b.circa.mc.nat.dacc.mean}{15.3}  
\resdef{e1ds.qwen3-4b.nope.nli.nat.dacc.hi}{13.1}  
\resdef{e1ds.qwen3-4b.nope.nli.nat.dacc.lo}{2.3}  
\resdef{e1ds.qwen3-4b.nope.nli.nat.dacc.mean}{7.8}  
\resdef{e1ds.qwen3-8b.circa.mc.nat.dacc.hi}{22.7}  
\resdef{e1ds.qwen3-8b.circa.mc.nat.dacc.lo}{12.4}  
\resdef{e1ds.qwen3-8b.circa.mc.nat.dacc.mean}{17.5}  
\resdef{e1ds.qwen3-8b.nope.nli.nat.dacc.hi}{15.5}  
\resdef{e1ds.qwen3-8b.nope.nli.nat.dacc.lo}{4.5}  
\resdef{e1ds.qwen3-8b.nope.nli.nat.dacc.mean}{9.9}  
\resdef{e1dsx.granite-3p3-8b.circa.mc.nat.dacc.hi}{5.2}  
\resdef{e1dsx.granite-3p3-8b.circa.mc.nat.dacc.lo}{-2.5}  
\resdef{e1dsx.granite-3p3-8b.circa.mc.nat.dacc.mean}{1.3}  
\resdef{e1dsx.granite-3p3-8b.nope.nli.nat.dacc.hi}{9.6}  
\resdef{e1dsx.granite-3p3-8b.nope.nli.nat.dacc.lo}{1.4}  
\resdef{e1dsx.granite-3p3-8b.nope.nli.nat.dacc.mean}{5.6}  
\resdef{e1dsx.llama-nemotron-nano-8b.circa.mc.nat.dacc.hi}{32.9}  
\resdef{e1dsx.llama-nemotron-nano-8b.circa.mc.nat.dacc.lo}{23.1}  
\resdef{e1dsx.llama-nemotron-nano-8b.circa.mc.nat.dacc.mean}{28.1}  
\resdef{e1dsx.llama-nemotron-nano-8b.nope.nli.nat.dacc.hi}{8.5}  
\resdef{e1dsx.llama-nemotron-nano-8b.nope.nli.nat.dacc.lo}{-0.4}  
\resdef{e1dsx.llama-nemotron-nano-8b.nope.nli.nat.dacc.mean}{4.1}  
\resdef{e1dsx.nemotron-nano-9b-v2.circa.mc.nat.dacc.hi}{20.5}  
\resdef{e1dsx.nemotron-nano-9b-v2.circa.mc.nat.dacc.lo}{11.1}  
\resdef{e1dsx.nemotron-nano-9b-v2.circa.mc.nat.dacc.mean}{15.7}  
\resdef{e1dsx.nemotron-nano-9b-v2.nope.nli.nat.dacc.hi}{9.8}  
\resdef{e1dsx.nemotron-nano-9b-v2.nope.nli.nat.dacc.lo}{1.0}  
\resdef{e1dsx.nemotron-nano-9b-v2.nope.nli.nat.dacc.mean}{5.4}  
\resdef{e1dsx.phi4-reasoning.circa.mc.nat.dacc.hi}{17.9}  
\resdef{e1dsx.phi4-reasoning.circa.mc.nat.dacc.lo}{8.9}  
\resdef{e1dsx.phi4-reasoning.circa.mc.nat.dacc.mean}{13.2}  
\resdef{e1dsx.phi4-reasoning.nope.nli.nat.dacc.hi}{22.6}  
\resdef{e1dsx.phi4-reasoning.nope.nli.nat.dacc.lo}{11.1}  
\resdef{e1dsx.phi4-reasoning.nope.nli.nat.dacc.mean}{16.9}  
\resdef{e1dsx.qwen3p5-4b.circa.mc.nat.dacc.hi}{20.4}  
\resdef{e1dsx.qwen3p5-4b.circa.mc.nat.dacc.lo}{12.9}  
\resdef{e1dsx.qwen3p5-4b.circa.mc.nat.dacc.mean}{16.7}  
\resdef{e1dsx.qwen3p5-4b.nope.nli.nat.dacc.hi}{10.4}  
\resdef{e1dsx.qwen3p5-4b.nope.nli.nat.dacc.lo}{1.4}  
\resdef{e1dsx.qwen3p5-4b.nope.nli.nat.dacc.mean}{5.8}  
\resdef{e1dsx.qwen3p5-9b.circa.mc.nat.dacc.hi}{11.6}  
\resdef{e1dsx.qwen3p5-9b.circa.mc.nat.dacc.lo}{4.9}  
\resdef{e1dsx.qwen3p5-9b.circa.mc.nat.dacc.mean}{8.3}  
\resdef{e1dsx.qwen3p5-9b.nope.nli.nat.dacc.hi}{10.3}  
\resdef{e1dsx.qwen3p5-9b.nope.nli.nat.dacc.lo}{3.1}  
\resdef{e1dsx.qwen3p5-9b.nope.nli.nat.dacc.mean}{6.6}  
\resdef{e1dsx.skywork-or1-7b.circa.mc.nat.dacc.hi}{38.0}  
\resdef{e1dsx.skywork-or1-7b.circa.mc.nat.dacc.lo}{27.9}  
\resdef{e1dsx.skywork-or1-7b.circa.mc.nat.dacc.mean}{32.8}  
\resdef{e1dsx.skywork-or1-7b.nope.nli.nat.dacc.hi}{30.6}  
\resdef{e1dsx.skywork-or1-7b.nope.nli.nat.dacc.lo}{17.9}  
\resdef{e1dsx.skywork-or1-7b.nope.nli.nat.dacc.mean}{24.3}  
\resdef{e1dsx.smollm3-3b.circa.mc.nat.dacc.hi}{6.3}  
\resdef{e1dsx.smollm3-3b.circa.mc.nat.dacc.lo}{-3.7}  
\resdef{e1dsx.smollm3-3b.circa.mc.nat.dacc.mean}{1.2}  
\resdef{e1dsx.smollm3-3b.nope.nli.nat.dacc.hi}{10.8}  
\resdef{e1dsx.smollm3-3b.nope.nli.nat.dacc.lo}{1.2}  
\resdef{e1dsx.smollm3-3b.nope.nli.nat.dacc.mean}{6.0}  
\resdef{e1sfx.exaone-deep-7p8b.a3f-conn.inf.nat.dR.hi}{-28.2}  
\resdef{e1sfx.exaone-deep-7p8b.a3f-conn.inf.nat.dR.lo}{-37.6}  
\resdef{e1sfx.exaone-deep-7p8b.a3f-conn.inf.nat.dR.mean}{-33.0}  
\resdef{e1sfx.exaone-deep-7p8b.a3f-conn.nli.nat.dR.hi}{-26.9}  
\resdef{e1sfx.exaone-deep-7p8b.a3f-conn.nli.nat.dR.lo}{-33.2}  
\resdef{e1sfx.exaone-deep-7p8b.a3f-conn.nli.nat.dR.mean}{-30.0}  
\resdef{e1sfx.exaone-deep-7p8b.a3f-modal.inf.nat.dR.mean}{10.2}  
\resdef{e1sfx.exaone-deep-7p8b.a3f-quant.inf.nat.dR.hi}{-28.6}  
\resdef{e1sfx.exaone-deep-7p8b.a3f-quant.inf.nat.dR.lo}{-37.8}  
\resdef{e1sfx.exaone-deep-7p8b.a3f-quant.inf.nat.dR.mean}{-33.2}  
\resdef{e1sfx.exaone-deep-7p8b.a3f-quant.nli.nat.dR.hi}{-24.3}  
\resdef{e1sfx.exaone-deep-7p8b.a3f-quant.nli.nat.dR.lo}{-30.4}  
\resdef{e1sfx.exaone-deep-7p8b.a3f-quant.nli.nat.dR.mean}{-27.2}  
\resdef{e1sfx.granite-3p3-8b.a3f-conn.inf.nat.dR.hi}{22.1}  
\resdef{e1sfx.granite-3p3-8b.a3f-conn.inf.nat.dR.lo}{15.7}  
\resdef{e1sfx.granite-3p3-8b.a3f-conn.inf.nat.dR.mean}{18.9}  
\resdef{e1sfx.granite-3p3-8b.a3f-conn.nli.nat.dR.hi}{-4.4}  
\resdef{e1sfx.granite-3p3-8b.a3f-conn.nli.nat.dR.lo}{-13.2}  
\resdef{e1sfx.granite-3p3-8b.a3f-conn.nli.nat.dR.mean}{-8.8}  
\resdef{e1sfx.granite-3p3-8b.a3f-quant.inf.nat.dR.hi}{15.6}  
\resdef{e1sfx.granite-3p3-8b.a3f-quant.inf.nat.dR.lo}{9.3}  
\resdef{e1sfx.granite-3p3-8b.a3f-quant.inf.nat.dR.mean}{12.6}  
\resdef{e1sfx.granite-3p3-8b.a3f-quant.nli.nat.dR.hi}{-18.3}  
\resdef{e1sfx.granite-3p3-8b.a3f-quant.nli.nat.dR.lo}{-27.2}  
\resdef{e1sfx.granite-3p3-8b.a3f-quant.nli.nat.dR.mean}{-22.8}  
\resdef{e1sfx.llama-nemotron-nano-8b.a3f-conn.inf.nat.dR.hi}{5.4}  
\resdef{e1sfx.llama-nemotron-nano-8b.a3f-conn.inf.nat.dR.lo}{2.9}  
\resdef{e1sfx.llama-nemotron-nano-8b.a3f-conn.inf.nat.dR.mean}{4.1}  
\resdef{e1sfx.llama-nemotron-nano-8b.a3f-conn.nli.nat.dR.hi}{-2.4}  
\resdef{e1sfx.llama-nemotron-nano-8b.a3f-conn.nli.nat.dR.lo}{-6.1}  
\resdef{e1sfx.llama-nemotron-nano-8b.a3f-conn.nli.nat.dR.mean}{-4.2}  
\resdef{e1sfx.llama-nemotron-nano-8b.a3f-quant.inf.nat.dR.hi}{15.1}  
\resdef{e1sfx.llama-nemotron-nano-8b.a3f-quant.inf.nat.dR.lo}{9.9}  
\resdef{e1sfx.llama-nemotron-nano-8b.a3f-quant.inf.nat.dR.mean}{12.4}  
\resdef{e1sfx.llama-nemotron-nano-8b.a3f-quant.nli.nat.dR.hi}{-6.2}  
\resdef{e1sfx.llama-nemotron-nano-8b.a3f-quant.nli.nat.dR.lo}{-11.6}  
\resdef{e1sfx.llama-nemotron-nano-8b.a3f-quant.nli.nat.dR.mean}{-8.7}  
\resdef{e1sfx.mimo-7b-rl.a3f-conn.nli.nat.dR.mean}{-23.6}  
\resdef{e1sfx.mimo-7b-rl.a3f-quant.nli.nat.dR.mean}{-21.2}  
\resdef{e1sfx.mimo-7b-sft.a3f-conn.nli.nat.dR.mean}{-26.7}  
\resdef{e1sfx.mimo-7b-sft.a3f-quant.nli.nat.dR.mean}{-26.1}  
\resdef{e1sfx.nemotron-nano-9b-v2.a3f-conn.inf.nat.dR.hi}{-35.7}  
\resdef{e1sfx.nemotron-nano-9b-v2.a3f-conn.inf.nat.dR.lo}{-45.3}  
\resdef{e1sfx.nemotron-nano-9b-v2.a3f-conn.inf.nat.dR.mean}{-40.4}  
\resdef{e1sfx.nemotron-nano-9b-v2.a3f-conn.nli.nat.dR.hi}{0.0}  
\resdef{e1sfx.nemotron-nano-9b-v2.a3f-conn.nli.nat.dR.lo}{-2.2}  
\resdef{e1sfx.nemotron-nano-9b-v2.a3f-conn.nli.nat.dR.mean}{-1.0}  
\resdef{e1sfx.nemotron-nano-9b-v2.a3f-quant.inf.nat.dR.hi}{-50.7}  
\resdef{e1sfx.nemotron-nano-9b-v2.a3f-quant.inf.nat.dR.lo}{-61.0}  
\resdef{e1sfx.nemotron-nano-9b-v2.a3f-quant.inf.nat.dR.mean}{-56.0}  
\resdef{e1sfx.nemotron-nano-9b-v2.a3f-quant.nli.nat.dR.hi}{1.6}  
\resdef{e1sfx.nemotron-nano-9b-v2.a3f-quant.nli.nat.dR.lo}{-1.4}  
\resdef{e1sfx.nemotron-nano-9b-v2.a3f-quant.nli.nat.dR.mean}{0.0}  
\resdef{e1sfx.phi4-mini-reasoning.a3f-conn.inf.nat.dR.hi}{-11.9}  
\resdef{e1sfx.phi4-mini-reasoning.a3f-conn.inf.nat.dR.lo}{-20.0}  
\resdef{e1sfx.phi4-mini-reasoning.a3f-conn.inf.nat.dR.mean}{-15.9}  
\resdef{e1sfx.phi4-mini-reasoning.a3f-conn.nli.nat.dR.hi}{-22.2}  
\resdef{e1sfx.phi4-mini-reasoning.a3f-conn.nli.nat.dR.lo}{-28.9}  
\resdef{e1sfx.phi4-mini-reasoning.a3f-conn.nli.nat.dR.mean}{-25.6}  
\resdef{e1sfx.phi4-mini-reasoning.a3f-quant.inf.nat.dR.hi}{-17.2}  
\resdef{e1sfx.phi4-mini-reasoning.a3f-quant.inf.nat.dR.lo}{-24.7}  
\resdef{e1sfx.phi4-mini-reasoning.a3f-quant.inf.nat.dR.mean}{-21.0}  
\resdef{e1sfx.phi4-mini-reasoning.a3f-quant.nli.nat.dR.hi}{-14.2}  
\resdef{e1sfx.phi4-mini-reasoning.a3f-quant.nli.nat.dR.lo}{-21.1}  
\resdef{e1sfx.phi4-mini-reasoning.a3f-quant.nli.nat.dR.mean}{-17.7}  
\resdef{e1sfx.phi4-reasoning.a3f-conn.inf.nat.dR.hi}{-37.3}  
\resdef{e1sfx.phi4-reasoning.a3f-conn.inf.nat.dR.lo}{-44.9}  
\resdef{e1sfx.phi4-reasoning.a3f-conn.inf.nat.dR.mean}{-41.1}  
\resdef{e1sfx.phi4-reasoning.a3f-conn.nli.nat.dR.hi}{-21.2}  
\resdef{e1sfx.phi4-reasoning.a3f-conn.nli.nat.dR.lo}{-28.8}  
\resdef{e1sfx.phi4-reasoning.a3f-conn.nli.nat.dR.mean}{-25.0}  
\resdef{e1sfx.phi4-reasoning.a3f-quant.inf.nat.dR.hi}{-30.0}  
\resdef{e1sfx.phi4-reasoning.a3f-quant.inf.nat.dR.lo}{-38.2}  
\resdef{e1sfx.phi4-reasoning.a3f-quant.inf.nat.dR.mean}{-34.2}  
\resdef{e1sfx.phi4-reasoning.a3f-quant.nli.nat.dR.hi}{-23.0}  
\resdef{e1sfx.phi4-reasoning.a3f-quant.nli.nat.dR.lo}{-30.9}  
\resdef{e1sfx.phi4-reasoning.a3f-quant.nli.nat.dR.mean}{-26.9}  
\resdef{e1sfx.qwen3p5-0p8b.a3f-conn.inf.nat.dR.hi}{-10.9}  
\resdef{e1sfx.qwen3p5-0p8b.a3f-conn.inf.nat.dR.lo}{-17.9}  
\resdef{e1sfx.qwen3p5-0p8b.a3f-conn.inf.nat.dR.mean}{-14.3}  
\resdef{e1sfx.qwen3p5-0p8b.a3f-conn.nli.nat.dR.hi}{-4.8}  
\resdef{e1sfx.qwen3p5-0p8b.a3f-conn.nli.nat.dR.lo}{-12.0}  
\resdef{e1sfx.qwen3p5-0p8b.a3f-conn.nli.nat.dR.mean}{-8.3}  
\resdef{e1sfx.qwen3p5-0p8b.a3f-quant.inf.nat.dR.hi}{-2.1}  
\resdef{e1sfx.qwen3p5-0p8b.a3f-quant.inf.nat.dR.lo}{-8.1}  
\resdef{e1sfx.qwen3p5-0p8b.a3f-quant.inf.nat.dR.mean}{-5.0}  
\resdef{e1sfx.qwen3p5-0p8b.a3f-quant.nli.nat.dR.hi}{14.9}  
\resdef{e1sfx.qwen3p5-0p8b.a3f-quant.nli.nat.dR.lo}{6.2}  
\resdef{e1sfx.qwen3p5-0p8b.a3f-quant.nli.nat.dR.mean}{10.7}  
\resdef{e1sfx.qwen3p5-2b.a3f-conn.inf.nat.dR.hi}{-20.9}  
\resdef{e1sfx.qwen3p5-2b.a3f-conn.inf.nat.dR.lo}{-29.9}  
\resdef{e1sfx.qwen3p5-2b.a3f-conn.inf.nat.dR.mean}{-25.4}  
\resdef{e1sfx.qwen3p5-2b.a3f-conn.nli.nat.dR.hi}{-5.9}  
\resdef{e1sfx.qwen3p5-2b.a3f-conn.nli.nat.dR.lo}{-10.2}  
\resdef{e1sfx.qwen3p5-2b.a3f-conn.nli.nat.dR.mean}{-8.0}  
\resdef{e1sfx.qwen3p5-2b.a3f-quant.inf.nat.dR.hi}{-26.8}  
\resdef{e1sfx.qwen3p5-2b.a3f-quant.inf.nat.dR.lo}{-36.3}  
\resdef{e1sfx.qwen3p5-2b.a3f-quant.inf.nat.dR.mean}{-31.6}  
\resdef{e1sfx.qwen3p5-2b.a3f-quant.nli.nat.dR.hi}{-6.4}  
\resdef{e1sfx.qwen3p5-2b.a3f-quant.nli.nat.dR.lo}{-11.1}  
\resdef{e1sfx.qwen3p5-2b.a3f-quant.nli.nat.dR.mean}{-8.7}  
\resdef{e1sfx.qwen3p5-4b.a3f-conn.inf.nat.dR.hi}{-22.0}  
\resdef{e1sfx.qwen3p5-4b.a3f-conn.inf.nat.dR.lo}{-30.6}  
\resdef{e1sfx.qwen3p5-4b.a3f-conn.inf.nat.dR.mean}{-26.2}  
\resdef{e1sfx.qwen3p5-4b.a3f-conn.nli.nat.dR.hi}{-11.3}  
\resdef{e1sfx.qwen3p5-4b.a3f-conn.nli.nat.dR.lo}{-17.0}  
\resdef{e1sfx.qwen3p5-4b.a3f-conn.nli.nat.dR.mean}{-14.1}  
\resdef{e1sfx.qwen3p5-4b.a3f-quant.inf.nat.dR.hi}{-45.3}  
\resdef{e1sfx.qwen3p5-4b.a3f-quant.inf.nat.dR.lo}{-55.4}  
\resdef{e1sfx.qwen3p5-4b.a3f-quant.inf.nat.dR.mean}{-50.3}  
\resdef{e1sfx.qwen3p5-4b.a3f-quant.nli.nat.dR.hi}{-10.6}  
\resdef{e1sfx.qwen3p5-4b.a3f-quant.nli.nat.dR.lo}{-16.4}  
\resdef{e1sfx.qwen3p5-4b.a3f-quant.nli.nat.dR.mean}{-13.4}  
\resdef{e1sfx.qwen3p5-9b.a3f-conn.inf.nat.dR.hi}{-21.1}  
\resdef{e1sfx.qwen3p5-9b.a3f-conn.inf.nat.dR.lo}{-29.7}  
\resdef{e1sfx.qwen3p5-9b.a3f-conn.inf.nat.dR.mean}{-25.3}  
\resdef{e1sfx.qwen3p5-9b.a3f-conn.nli.nat.dR.hi}{-4.8}  
\resdef{e1sfx.qwen3p5-9b.a3f-conn.nli.nat.dR.lo}{-9.0}  
\resdef{e1sfx.qwen3p5-9b.a3f-conn.nli.nat.dR.mean}{-6.8}  
\resdef{e1sfx.qwen3p5-9b.a3f-quant.inf.nat.dR.hi}{-25.7}  
\resdef{e1sfx.qwen3p5-9b.a3f-quant.inf.nat.dR.lo}{-35.1}  
\resdef{e1sfx.qwen3p5-9b.a3f-quant.inf.nat.dR.mean}{-30.4}  
\resdef{e1sfx.qwen3p5-9b.a3f-quant.nli.nat.dR.hi}{-8.1}  
\resdef{e1sfx.qwen3p5-9b.a3f-quant.nli.nat.dR.lo}{-13.7}  
\resdef{e1sfx.qwen3p5-9b.a3f-quant.nli.nat.dR.mean}{-10.8}  
\resdef{e1sfx.qwen3p8-27b-elow.a3f-conn.nli.nat.dR.mean}{-8.6}  
\resdef{e1sfx.qwen3p8-27b-elow.a3f-quant.nli.nat.dR.mean}{-4.0}  
\resdef{e1sfx.qwen3p8-27b-emed.a3f-conn.nli.nat.dR.mean}{-2.3}  
\resdef{e1sfx.qwen3p8-27b-emed.a3f-quant.nli.nat.dR.mean}{-1.6}  
\resdef{e1sfx.qwen3p8-27b.a3f-conn.inf.nat.dR.hi}{-17.7}  
\resdef{e1sfx.qwen3p8-27b.a3f-conn.inf.nat.dR.lo}{-26.1}  
\resdef{e1sfx.qwen3p8-27b.a3f-conn.inf.nat.dR.mean}{-21.7}  
\resdef{e1sfx.qwen3p8-27b.a3f-conn.nli.nat.dR.hi}{-1.6}  
\resdef{e1sfx.qwen3p8-27b.a3f-conn.nli.nat.dR.lo}{-4.7}  
\resdef{e1sfx.qwen3p8-27b.a3f-conn.nli.nat.dR.mean}{-3.0}  
\resdef{e1sfx.qwen3p8-27b.a3f-quant-negn.inf.nat.dR.mean}{-59.7}  
\resdef{e1sfx.qwen3p8-27b.a3f-quant-negp.inf.nat.dR.mean}{56.2}  
\resdef{e1sfx.qwen3p8-27b.a3f-quant.inf.nat.dR.hi}{-7.2}  
\resdef{e1sfx.qwen3p8-27b.a3f-quant.inf.nat.dR.lo}{-20.1}  
\resdef{e1sfx.qwen3p8-27b.a3f-quant.inf.nat.dR.mean}{-13.7}  
\resdef{e1sfx.qwen3p8-27b.a3f-quant.nli.nat.dR.hi}{-0.4}  
\resdef{e1sfx.qwen3p8-27b.a3f-quant.nli.nat.dR.lo}{-2.2}  
\resdef{e1sfx.qwen3p8-27b.a3f-quant.nli.nat.dR.mean}{-1.2}  
\resdef{e1sfx.r1distill-llama-8b.a3f-conn.inf.nat.dR.hi}{-58.4}  
\resdef{e1sfx.r1distill-llama-8b.a3f-conn.inf.nat.dR.lo}{-66.8}  
\resdef{e1sfx.r1distill-llama-8b.a3f-conn.inf.nat.dR.mean}{-62.7}  
\resdef{e1sfx.r1distill-llama-8b.a3f-conn.nli.nat.dR.hi}{-30.2}  
\resdef{e1sfx.r1distill-llama-8b.a3f-conn.nli.nat.dR.lo}{-36.6}  
\resdef{e1sfx.r1distill-llama-8b.a3f-conn.nli.nat.dR.mean}{-33.3}  
\resdef{e1sfx.r1distill-llama-8b.a3f-quant.inf.nat.dR.hi}{-45.8}  
\resdef{e1sfx.r1distill-llama-8b.a3f-quant.inf.nat.dR.lo}{-54.4}  
\resdef{e1sfx.r1distill-llama-8b.a3f-quant.inf.nat.dR.mean}{-50.1}  
\resdef{e1sfx.r1distill-llama-8b.a3f-quant.nli.nat.dR.hi}{-28.1}  
\resdef{e1sfx.r1distill-llama-8b.a3f-quant.nli.nat.dR.lo}{-35.0}  
\resdef{e1sfx.r1distill-llama-8b.a3f-quant.nli.nat.dR.mean}{-31.4}  
\resdef{e1sfx.skywork-or1-7b.a3f-conn.inf.nat.dR.hi}{-25.0}  
\resdef{e1sfx.skywork-or1-7b.a3f-conn.inf.nat.dR.lo}{-35.2}  
\resdef{e1sfx.skywork-or1-7b.a3f-conn.inf.nat.dR.mean}{-30.1}  
\resdef{e1sfx.skywork-or1-7b.a3f-conn.nli.nat.dR.hi}{-28.8}  
\resdef{e1sfx.skywork-or1-7b.a3f-conn.nli.nat.dR.lo}{-37.4}  
\resdef{e1sfx.skywork-or1-7b.a3f-conn.nli.nat.dR.mean}{-33.1}  
\resdef{e1sfx.skywork-or1-7b.a3f-quant.inf.nat.dR.hi}{-14.0}  
\resdef{e1sfx.skywork-or1-7b.a3f-quant.inf.nat.dR.lo}{-23.9}  
\resdef{e1sfx.skywork-or1-7b.a3f-quant.inf.nat.dR.mean}{-19.0}  
\resdef{e1sfx.skywork-or1-7b.a3f-quant.nli.nat.dR.hi}{-24.6}  
\resdef{e1sfx.skywork-or1-7b.a3f-quant.nli.nat.dR.lo}{-32.4}  
\resdef{e1sfx.skywork-or1-7b.a3f-quant.nli.nat.dR.mean}{-28.4}  
\resdef{e1sfx.smollm3-3b.a3f-conn.inf.nat.dR.hi}{8.9}  
\resdef{e1sfx.smollm3-3b.a3f-conn.inf.nat.dR.lo}{-1.2}  
\resdef{e1sfx.smollm3-3b.a3f-conn.inf.nat.dR.mean}{3.9}  
\resdef{e1sfx.smollm3-3b.a3f-conn.nli.nat.dR.hi}{2.1}  
\resdef{e1sfx.smollm3-3b.a3f-conn.nli.nat.dR.lo}{-0.1}  
\resdef{e1sfx.smollm3-3b.a3f-conn.nli.nat.dR.mean}{1.0}  
\resdef{e1sfx.smollm3-3b.a3f-quant.inf.nat.dR.hi}{-2.8}  
\resdef{e1sfx.smollm3-3b.a3f-quant.inf.nat.dR.lo}{-11.6}  
\resdef{e1sfx.smollm3-3b.a3f-quant.inf.nat.dR.mean}{-7.2}  
\resdef{e1sfx.smollm3-3b.a3f-quant.nli.nat.dR.hi}{6.3}  
\resdef{e1sfx.smollm3-3b.a3f-quant.nli.nat.dR.lo}{1.6}  
\resdef{e1sfx.smollm3-3b.a3f-quant.nli.nat.dR.mean}{3.9}  
\resdef{e1gds.gpt-oss-20b.a3-quant.infflip.high.dR.mean}{-11.7}  
\resdef{e1gds.gpt-oss-20b.a3xde.inf.high.dR.mean}{-30.4}  
\resdef{e1gds.gpt-oss-20b.circa.mc.high.dacc.hi}{4.5}  
\resdef{e1gds.gpt-oss-20b.circa.mc.high.dacc.lo}{-1.7}  
\resdef{e1gds.gpt-oss-20b.circa.mc.high.dacc.mean}{1.3}  
\resdef{e1gds.gpt-oss-20b.nope.nli.high.dacc.hi}{6.6}  
\resdef{e1gds.gpt-oss-20b.nope.nli.high.dacc.lo}{0.2}  
\resdef{e1gds.gpt-oss-20b.nope.nli.high.dacc.mean}{3.4}  
\resdef{e1pangu.openpangu-embedded-7b-model.a2.inf.nat.dR.hi}{11.5}  
\resdef{e1pangu.openpangu-embedded-7b-model.a2.inf.nat.dR.lo}{3.1}  
\resdef{e1pangu.openpangu-embedded-7b-model.a2.inf.nat.dR.mean}{7.3}  
\resdef{e1pangu.openpangu-embedded-7b-model.a2.inf.nat.dR.std}{2.4}  
\resdef{e1pangu.openpangu-embedded-7b-model.a3-conn.inf.nat.dR.hi}{3.0}  
\resdef{e1pangu.openpangu-embedded-7b-model.a3-conn.inf.nat.dR.lo}{-14.1}  
\resdef{e1pangu.openpangu-embedded-7b-model.a3-conn.inf.nat.dR.mean}{-4.0}  
\resdef{e1pangu.openpangu-embedded-7b-model.a3-conn.inf.nat.dR.std}{1.8}  
\resdef{e1pangu.openpangu-embedded-7b-model.a3-modal.inf.nat.dR.hi}{17.7}  
\resdef{e1pangu.openpangu-embedded-7b-model.a3-modal.inf.nat.dR.lo}{3.1}  
\resdef{e1pangu.openpangu-embedded-7b-model.a3-modal.inf.nat.dR.mean}{10.4}  
\resdef{e1pangu.openpangu-embedded-7b-model.a3-modal.inf.nat.dR.std}{6.5}  
\resdef{e1pangu.openpangu-embedded-7b-model.a3-quant.inf.nat.dR.hi}{24.4}  
\resdef{e1pangu.openpangu-embedded-7b-model.a3-quant.inf.nat.dR.lo}{5.1}  
\resdef{e1pangu.openpangu-embedded-7b-model.a3-quant.inf.nat.dR.mean}{14.1}  
\resdef{e1pangu.openpangu-embedded-7b-model.a3-quant.inf.nat.dR.std}{11.1}  
\resdef{e1pangu.openpangu-embedded-7b-model.a3.inf.nat.dR.hi}{16.6}  
\resdef{e1pangu.openpangu-embedded-7b-model.a3.inf.nat.dR.lo}{8.0}  
\resdef{e1pangu.openpangu-embedded-7b-model.a3.inf.nat.dR.mean}{12.3}  
\resdef{e1pangu.openpangu-embedded-7b-model.a3.inf.nat.dR.std}{0.9}  
\resdef{e1pangu.openpangu-embedded-7b-model.a3f-conn.inf.nat.dR.hi}{-2.1}  
\resdef{e1pangu.openpangu-embedded-7b-model.a3f-conn.inf.nat.dR.lo}{-7.6}  
\resdef{e1pangu.openpangu-embedded-7b-model.a3f-conn.inf.nat.dR.mean}{-4.8}  
\resdef{e1pangu.openpangu-embedded-7b-model.a3f-conn.nli.nat.dR.hi}{1.1}  
\resdef{e1pangu.openpangu-embedded-7b-model.a3f-conn.nli.nat.dR.lo}{-1.3}  
\resdef{e1pangu.openpangu-embedded-7b-model.a3f-conn.nli.nat.dR.mean}{-0.1}  
\resdef{e1pangu.openpangu-embedded-7b-model.a3f-quant.inf.nat.dR.hi}{9.8}  
\resdef{e1pangu.openpangu-embedded-7b-model.a3f-quant.inf.nat.dR.lo}{3.6}  
\resdef{e1pangu.openpangu-embedded-7b-model.a3f-quant.inf.nat.dR.mean}{6.7}  
\resdef{e1pangu.openpangu-embedded-7b-model.a3f-quant.nli.nat.dR.hi}{1.6}  
\resdef{e1pangu.openpangu-embedded-7b-model.a3f-quant.nli.nat.dR.lo}{-2.9}  
\resdef{e1pangu.openpangu-embedded-7b-model.a3f-quant.nli.nat.dR.mean}{-0.7}  
\resdef{e1pangu.openpangu-embedded-7b-model.b2.yn.nat.dacc.hi}{17.0}  
\resdef{e1pangu.openpangu-embedded-7b-model.b2.yn.nat.dacc.lo}{0.7}  
\resdef{e1pangu.openpangu-embedded-7b-model.b2.yn.nat.dacc.mean}{8.9}  
\resdef{e1pangu.openpangu-embedded-7b-model.b2.yn.nat.dacc.std}{2.2}  
\resdef{e1pangu.openpangu-embedded-7b-v1p1.a2.inf.nat.dR.hi}{-9.5}  
\resdef{e1pangu.openpangu-embedded-7b-v1p1.a2.inf.nat.dR.lo}{-23.0}  
\resdef{e1pangu.openpangu-embedded-7b-v1p1.a2.inf.nat.dR.mean}{-16.1}  
\resdef{e1pangu.openpangu-embedded-7b-v1p1.a2.inf.nat.dR.std}{2.3}  
\resdef{e1pangu.openpangu-embedded-7b-v1p1.a3-conn.inf.nat.dR.hi}{0.0}  
\resdef{e1pangu.openpangu-embedded-7b-v1p1.a3-conn.inf.nat.dR.lo}{-8.1}  
\resdef{e1pangu.openpangu-embedded-7b-v1p1.a3-conn.inf.nat.dR.mean}{-3.0}  
\resdef{e1pangu.openpangu-embedded-7b-v1p1.a3-conn.inf.nat.dR.std}{3.0}  
\resdef{e1pangu.openpangu-embedded-7b-v1p1.a3-modal.inf.nat.dR.hi}{21.9}  
\resdef{e1pangu.openpangu-embedded-7b-v1p1.a3-modal.inf.nat.dR.lo}{1.0}  
\resdef{e1pangu.openpangu-embedded-7b-v1p1.a3-modal.inf.nat.dR.mean}{10.4}  
\resdef{e1pangu.openpangu-embedded-7b-v1p1.a3-modal.inf.nat.dR.std}{1.8}  
\resdef{e1pangu.openpangu-embedded-7b-v1p1.a3-quant.inf.nat.dR.hi}{-0.0}  
\resdef{e1pangu.openpangu-embedded-7b-v1p1.a3-quant.inf.nat.dR.lo}{-26.9}  
\resdef{e1pangu.openpangu-embedded-7b-v1p1.a3-quant.inf.nat.dR.mean}{-12.8}  
\resdef{e1pangu.openpangu-embedded-7b-v1p1.a3-quant.inf.nat.dR.std}{2.2}  
\resdef{e1pangu.openpangu-embedded-7b-v1p1.a3.inf.nat.dR.hi}{12.5}  
\resdef{e1pangu.openpangu-embedded-7b-v1p1.a3.inf.nat.dR.lo}{3.6}  
\resdef{e1pangu.openpangu-embedded-7b-v1p1.a3.inf.nat.dR.mean}{7.9}  
\resdef{e1pangu.openpangu-embedded-7b-v1p1.a3.inf.nat.dR.std}{1.2}  
\resdef{e1pangu.openpangu-embedded-7b-v1p1.a3f-conn.inf.nat.dR.hi}{0.0}  
\resdef{e1pangu.openpangu-embedded-7b-v1p1.a3f-conn.inf.nat.dR.lo}{-3.2}  
\resdef{e1pangu.openpangu-embedded-7b-v1p1.a3f-conn.inf.nat.dR.mean}{-1.6}  
\resdef{e1pangu.openpangu-embedded-7b-v1p1.a3f-conn.nli.nat.dR.hi}{-1.8}  
\resdef{e1pangu.openpangu-embedded-7b-v1p1.a3f-conn.nli.nat.dR.lo}{-4.3}  
\resdef{e1pangu.openpangu-embedded-7b-v1p1.a3f-conn.nli.nat.dR.mean}{-3.0}  
\resdef{e1pangu.openpangu-embedded-7b-v1p1.a3f-quant.inf.nat.dR.hi}{-11.4}  
\resdef{e1pangu.openpangu-embedded-7b-v1p1.a3f-quant.inf.nat.dR.lo}{-19.4}  
\resdef{e1pangu.openpangu-embedded-7b-v1p1.a3f-quant.inf.nat.dR.mean}{-15.4}  
\resdef{e1pangu.openpangu-embedded-7b-v1p1.a3f-quant.nli.nat.dR.hi}{-6.0}  
\resdef{e1pangu.openpangu-embedded-7b-v1p1.a3f-quant.nli.nat.dR.lo}{-11.2}  
\resdef{e1pangu.openpangu-embedded-7b-v1p1.a3f-quant.nli.nat.dR.mean}{-8.4}  
\resdef{e1pangu.openpangu-embedded-7b-v1p1.b2.yn.nat.dacc.hi}{9.8}  
\resdef{e1pangu.openpangu-embedded-7b-v1p1.b2.yn.nat.dacc.lo}{-5.6}  
\resdef{e1pangu.openpangu-embedded-7b-v1p1.b2.yn.nat.dacc.mean}{2.1}  
\resdef{e1pangu.openpangu-embedded-7b-v1p1.b2.yn.nat.dacc.std}{4.2}  
\resdef{derived.fullsets.nli.drop}{21}  
\resdef{derived.fullsets.nli.tested}{24}  
\resdef{derived.glmm.qwen3-14b.coef}{-0.049}  
\resdef{derived.glmm.qwen3-14b.p}{<.001}  
\resdef{derived.glmm.qwen3-32b.coef}{-0.037}  
\resdef{derived.glmm.qwen3-32b.p}{.230}  
\resdef{derived.glmm.qwen3-4b.coef}{-0.067}  
\resdef{derived.glmm.qwen3-4b.p}{<.001}  
\resdef{derived.glmm.qwen3-8b.coef}{-0.063}  
\resdef{derived.glmm.qwen3-8b.hi}{-0.039}  
\resdef{derived.glmm.qwen3-8b.lo}{-0.087}  
\resdef{derived.glmm.qwen3-8b.p}{<.001}  
\resdef{derived.holm.full.survive}{47}  
\resdef{derived.holm.full.tests}{48}  
\resdef{derived.main.counted}{30}  
\resdef{derived.main.formatdep}{1}  
\resdef{derived.main.no_sig_change}{5}  
\resdef{derived.main.rise}{4}  
\resdef{derived.main.tested}{40}  
\resdef{derived.tost.a2.equivalent}{2}  
\resdef{derived.tost.a2.tested}{42}  
\resdef{e10.qwen3-32b.a3-conn.inf.nat.R.hi}{15.2}  
\resdef{e10.qwen3-32b.a3-conn.inf.nat.R.lo}{2.0}  
\resdef{e10.qwen3-32b.a3-conn.inf.nat.R.mean}{8.1}  
\resdef{e10.qwen3-32b.a3-conn.inf.nat.R.std}{3.5}  
\resdef{e10.qwen3-32b.a3-conn.inf.off.R.hi}{67.7}  
\resdef{e10.qwen3-32b.a3-conn.inf.off.R.lo}{43.4}  
\resdef{e10.qwen3-32b.a3-conn.inf.off.R.mean}{55.6}  
\resdef{e10.qwen3-32b.a3-conn.inf.off.R.std}{3.5}  
\resdef{e10.qwen3-32b.a3-conn.inf.sint.R.hi}{35.4}  
\resdef{e10.qwen3-32b.a3-conn.inf.sint.R.lo}{17.2}  
\resdef{e10.qwen3-32b.a3-conn.inf.sint.R.mean}{26.3}  
\resdef{e10.qwen3-32b.a3-conn.inf.sint.R.std}{4.6}  
\resdef{e10.qwen3-32b.a3-conn.inf.slog.R.hi}{14.1}  
\resdef{e10.qwen3-32b.a3-conn.inf.slog.R.lo}{3.0}  
\resdef{e10.qwen3-32b.a3-conn.inf.slog.R.mean}{8.1}  
\resdef{e10.qwen3-32b.a3-conn.inf.slog.R.std}{3.5}  
\resdef{e10.qwen3-32b.a3-conn.inf.sneu.R.hi}{11.1}  
\resdef{e10.qwen3-32b.a3-conn.inf.sneu.R.lo}{2.0}  
\resdef{e10.qwen3-32b.a3-conn.inf.sneu.R.mean}{6.1}  
\resdef{e10.qwen3-32b.a3-conn.inf.sneu.R.std}{6.1}  
\resdef{e10.qwen3-32b.a3-quant.inf.nat.R.hi}{26.9}  
\resdef{e10.qwen3-32b.a3-quant.inf.nat.R.lo}{6.4}  
\resdef{e10.qwen3-32b.a3-quant.inf.nat.R.mean}{15.4}  
\resdef{e10.qwen3-32b.a3-quant.inf.nat.R.std}{3.8}  
\resdef{e10.qwen3-32b.a3-quant.inf.off.R.hi}{80.8}  
\resdef{e10.qwen3-32b.a3-quant.inf.off.R.lo}{48.7}  
\resdef{e10.qwen3-32b.a3-quant.inf.off.R.mean}{65.4}  
\resdef{e10.qwen3-32b.a3-quant.inf.off.R.std}{3.8}  
\resdef{e10.qwen3-32b.a3-quant.inf.sint.R.hi}{43.6}  
\resdef{e10.qwen3-32b.a3-quant.inf.sint.R.lo}{16.7}  
\resdef{e10.qwen3-32b.a3-quant.inf.sint.R.mean}{29.5}  
\resdef{e10.qwen3-32b.a3-quant.inf.sint.R.std}{5.9}  
\resdef{e10.qwen3-32b.a3-quant.inf.slog.R.hi}{16.7}  
\resdef{e10.qwen3-32b.a3-quant.inf.slog.R.lo}{2.6}  
\resdef{e10.qwen3-32b.a3-quant.inf.slog.R.mean}{9.0}  
\resdef{e10.qwen3-32b.a3-quant.inf.slog.R.std}{9.7}  
\resdef{e10.qwen3-32b.a3-quant.inf.sneu.R.hi}{17.9}  
\resdef{e10.qwen3-32b.a3-quant.inf.sneu.R.lo}{3.8}  
\resdef{e10.qwen3-32b.a3-quant.inf.sneu.R.mean}{10.3}  
\resdef{e10.qwen3-32b.a3-quant.inf.sneu.R.std}{5.9}  
\resdef{e10.qwen3-32b.b2.yn.nat.acc.hi}{89.5}  
\resdef{e10.qwen3-32b.b2.yn.nat.acc.lo}{77.5}  
\resdef{e10.qwen3-32b.b2.yn.nat.acc.mean}{83.5}  
\resdef{e10.qwen3-32b.b2.yn.nat.acc.std}{1.6}  
\resdef{e10.qwen3-32b.b2.yn.off.acc.hi}{90.2}  
\resdef{e10.qwen3-32b.b2.yn.off.acc.lo}{76.5}  
\resdef{e10.qwen3-32b.b2.yn.off.acc.mean}{83.9}  
\resdef{e10.qwen3-32b.b2.yn.off.acc.std}{1.2}  
\resdef{e10.qwen3-32b.b2.yn.sint.acc.hi}{90.5}  
\resdef{e10.qwen3-32b.b2.yn.sint.acc.lo}{78.2}  
\resdef{e10.qwen3-32b.b2.yn.sint.acc.mean}{84.6}  
\resdef{e10.qwen3-32b.b2.yn.sint.acc.std}{2.2}  
\resdef{e10.qwen3-32b.b2.yn.slog.acc.hi}{86.7}  
\resdef{e10.qwen3-32b.b2.yn.slog.acc.lo}{73.7}  
\resdef{e10.qwen3-32b.b2.yn.slog.acc.mean}{80.4}  
\resdef{e10.qwen3-32b.b2.yn.slog.acc.std}{2.2}  
\resdef{e10.qwen3-32b.b2.yn.sneu.acc.hi}{90.5}  
\resdef{e10.qwen3-32b.b2.yn.sneu.acc.lo}{78.9}  
\resdef{e10.qwen3-32b.b2.yn.sneu.acc.mean}{84.9}  
\resdef{e10.qwen3-32b.b2.yn.sneu.acc.std}{2.2}  
\resdef{e10.qwen3-8b.a3-conn.inf.nat.R.hi}{3.0}  
\resdef{e10.qwen3-8b.a3-conn.inf.nat.R.lo}{0.0}  
\resdef{e10.qwen3-8b.a3-conn.inf.nat.R.mean}{1.0}  
\resdef{e10.qwen3-8b.a3-conn.inf.nat.R.std}{1.8}  
\resdef{e10.qwen3-8b.a3-conn.inf.off.R.hi}{49.5}  
\resdef{e10.qwen3-8b.a3-conn.inf.off.R.lo}{25.3}  
\resdef{e10.qwen3-8b.a3-conn.inf.off.R.mean}{37.4}  
\resdef{e10.qwen3-8b.a3-conn.inf.off.R.std}{3.5}  
\resdef{e10.qwen3-8b.a3-conn.inf.sint.R.hi}{25.3}  
\resdef{e10.qwen3-8b.a3-conn.inf.sint.R.lo}{6.1}  
\resdef{e10.qwen3-8b.a3-conn.inf.sint.R.mean}{15.2}  
\resdef{e10.qwen3-8b.a3-conn.inf.sint.R.std}{6.1}  
\resdef{e10.qwen3-8b.a3-conn.inf.slog.R.hi}{8.1}  
\resdef{e10.qwen3-8b.a3-conn.inf.slog.R.lo}{1.0}  
\resdef{e10.qwen3-8b.a3-conn.inf.slog.R.mean}{4.0}  
\resdef{e10.qwen3-8b.a3-conn.inf.slog.R.std}{3.5}  
\resdef{e10.qwen3-8b.a3-conn.inf.sneu.R.hi}{3.0}  
\resdef{e10.qwen3-8b.a3-conn.inf.sneu.R.lo}{0.0}  
\resdef{e10.qwen3-8b.a3-conn.inf.sneu.R.mean}{1.0}  
\resdef{e10.qwen3-8b.a3-conn.inf.sneu.R.std}{1.8}  
\resdef{e10.qwen3-8b.a3-quant.inf.nat.R.hi}{0.0}  
\resdef{e10.qwen3-8b.a3-quant.inf.nat.R.lo}{0.0}  
\resdef{e10.qwen3-8b.a3-quant.inf.nat.R.mean}{0.0}  
\resdef{e10.qwen3-8b.a3-quant.inf.nat.R.std}{0.0}  
\resdef{e10.qwen3-8b.a3-quant.inf.off.R.hi}{42.3}  
\resdef{e10.qwen3-8b.a3-quant.inf.off.R.lo}{17.9}  
\resdef{e10.qwen3-8b.a3-quant.inf.off.R.mean}{29.5}  
\resdef{e10.qwen3-8b.a3-quant.inf.off.R.std}{5.9}  
\resdef{e10.qwen3-8b.a3-quant.inf.sint.R.hi}{20.5}  
\resdef{e10.qwen3-8b.a3-quant.inf.sint.R.lo}{6.4}  
\resdef{e10.qwen3-8b.a3-quant.inf.sint.R.mean}{12.8}  
\resdef{e10.qwen3-8b.a3-quant.inf.sint.R.std}{8.0}  
\resdef{e10.qwen3-8b.a3-quant.inf.slog.R.hi}{16.7}  
\resdef{e10.qwen3-8b.a3-quant.inf.slog.R.lo}{2.6}  
\resdef{e10.qwen3-8b.a3-quant.inf.slog.R.mean}{9.0}  
\resdef{e10.qwen3-8b.a3-quant.inf.slog.R.std}{5.9}  
\resdef{e10.qwen3-8b.a3-quant.inf.sneu.R.hi}{6.4}  
\resdef{e10.qwen3-8b.a3-quant.inf.sneu.R.lo}{0.0}  
\resdef{e10.qwen3-8b.a3-quant.inf.sneu.R.mean}{2.6}  
\resdef{e10.qwen3-8b.a3-quant.inf.sneu.R.std}{2.2}  
\resdef{e10.qwen3-8b.b2.yn.nat.acc.hi}{89.8}  
\resdef{e10.qwen3-8b.b2.yn.nat.acc.lo}{76.5}  
\resdef{e10.qwen3-8b.b2.yn.nat.acc.mean}{83.5}  
\resdef{e10.qwen3-8b.b2.yn.nat.acc.std}{1.6}  
\resdef{e10.qwen3-8b.b2.yn.off.acc.hi}{83.9}  
\resdef{e10.qwen3-8b.b2.yn.off.acc.lo}{68.4}  
\resdef{e10.qwen3-8b.b2.yn.off.acc.mean}{76.5}  
\resdef{e10.qwen3-8b.b2.yn.off.acc.std}{2.6}  
\resdef{e10.qwen3-8b.b2.yn.sint.acc.hi}{84.9}  
\resdef{e10.qwen3-8b.b2.yn.sint.acc.lo}{71.2}  
\resdef{e10.qwen3-8b.b2.yn.sint.acc.mean}{78.2}  
\resdef{e10.qwen3-8b.b2.yn.sint.acc.std}{2.2}  
\resdef{e10.qwen3-8b.b2.yn.slog.acc.hi}{85.3}  
\resdef{e10.qwen3-8b.b2.yn.slog.acc.lo}{71.9}  
\resdef{e10.qwen3-8b.b2.yn.slog.acc.mean}{78.9}  
\resdef{e10.qwen3-8b.b2.yn.slog.acc.std}{2.8}  
\resdef{e10.qwen3-8b.b2.yn.sneu.acc.hi}{84.9}  
\resdef{e10.qwen3-8b.b2.yn.sneu.acc.lo}{70.2}  
\resdef{e10.qwen3-8b.b2.yn.sneu.acc.mean}{77.5}  
\resdef{e10.qwen3-8b.b2.yn.sneu.acc.std}{2.2}  
\resdef{e10.qwen3p5-9b.a3-quant.inf.nat.R.mean}{1.3}  
\resdef{e10.qwen3p5-9b.a3-quant.inf.off.R.mean}{39.7}  
\resdef{e10.qwen3p5-9b.a3-quant.inf.sint.R.mean}{44.9}  
\resdef{e10.qwen3p5-9b.a3-quant.inf.slog.R.mean}{5.1}  
\resdef{e10.qwen3p5-9b.a3-quant.inf.sneu.R.mean}{11.5}  
\resdef{e1xl.qwen3-14b.a3xde.inf.nat.dR.mean}{-19.6}  
\resdef{e1xl.qwen3-8b.a3xde.inf.nat.dR.mean}{-37.9}  
\resdef{e1xl.qwen3-8b.a3xen.inf.nat.dR.mean}{-39.2}  
\resdef{e1xl.qwen3-8b.a3xzh.inf.nat.dR.mean}{-17.1}  
\resdef{e1xl.qwen3p5-9b.a3xde.inf.nat.dR.mean}{-27.9}  
\resdef{e1xl.qwen3p5-9b.a3xen.inf.nat.dR.mean}{-16.7}  
\resdef{e1xl.qwen3p5-9b.a3xzh.inf.nat.dR.mean}{-25.8}  
\resdef{m1dev.qwen3-8b.d.control.gates.mean}{0}  
\resdef{m1dev.qwen3-8b.d.control.literal.mean}{86.1}  
\resdef{m1dev.qwen3-8b.d.control.presup.mean}{64.5}  
\resdef{m1dev.qwen3-8b.d.control.score.mean}{73.3}  
\resdef{m1dev.qwen3-8b.d.control.speaker.mean}{60.4}  
\resdef{m1dev.qwen3-8b.d.control.tokfc.mean}{782.4}  
\resdef{m1dev.qwen3-8b.d.nat.gates.mean}{1}  
\resdef{m1dev.qwen3-8b.d.nat.literal.mean}{86.8}  
\resdef{m1dev.qwen3-8b.d.nat.presup.mean}{67.2}  
\resdef{m1dev.qwen3-8b.d.nat.score.mean}{73.2}  
\resdef{m1dev.qwen3-8b.d.nat.speaker.mean}{59.6}  
\resdef{m1dev.qwen3-8b.d.nat.tokfc.mean}{784.9}  
\resdef{m1dev.qwen3-8b.d.sit.gates.mean}{0}  
\resdef{m1dev.qwen3-8b.d.sit.literal.mean}{86.1}  
\resdef{m1dev.qwen3-8b.d.sit.presup.mean}{65.5}  
\resdef{m1dev.qwen3-8b.d.sit.score.mean}{73.2}  
\resdef{m1dev.qwen3-8b.d.sit.speaker.mean}{60.4}  
\resdef{m1dev.qwen3-8b.d.sit.tokfc.mean}{796.9}  
\resdef{m1dev.qwen3-8b.e.dpo.gates.mean}{0}  
\resdef{m1dev.qwen3-8b.e.dpo.literal.mean}{86.7}  
\resdef{m1dev.qwen3-8b.e.dpo.presup.mean}{65.0}  
\resdef{m1dev.qwen3-8b.e.dpo.score.mean}{73.9}  
\resdef{m1dev.qwen3-8b.e.dpo.speaker.mean}{61.1}  
\resdef{m1dev.qwen3-8b.e.dpo.tokfc.mean}{859.3}  
\resdef{m1dev.qwen3-8b.e.dposhuf.gates.mean}{0}  
\resdef{m1dev.qwen3-8b.e.dposhuf.literal.mean}{86.7}  
\resdef{m1dev.qwen3-8b.e.dposhuf.presup.mean}{65.0}  
\resdef{m1dev.qwen3-8b.e.dposhuf.score.mean}{73.5}  
\resdef{m1dev.qwen3-8b.e.dposhuf.speaker.mean}{60.3}  
\resdef{m1dev.qwen3-8b.e.dposhuf.tokfc.mean}{797.8}  
\resdef{m1dev.qwen3-8b.e.nat.score.mean}{73.2}  
\resdef{m1dev.qwen3-8b.e.nat.speaker.mean}{59.6}  
\resdef{m1dev.qwen3-8b.v1.dual.gates.mean}{0}  
\resdef{m1dev.qwen3-8b.v1.dual.literal.mean}{79.4}  
\resdef{m1dev.qwen3-8b.v1.dual.presup.mean}{52.8}  
\resdef{m1dev.qwen3-8b.v1.dual.score.mean}{65.6}  
\resdef{m1dev.qwen3-8b.v1.dual.speaker.mean}{51.8}  
\resdef{m1dev.qwen3-8b.v1.dual.tokfc.mean}{507.8}  
\resdef{m1dev.qwen3-8b.v1.grice.gates.mean}{0}  
\resdef{m1dev.qwen3-8b.v1.grice.literal.mean}{82.5}  
\resdef{m1dev.qwen3-8b.v1.grice.presup.mean}{54.6}  
\resdef{m1dev.qwen3-8b.v1.grice.score.mean}{69.2}  
\resdef{m1dev.qwen3-8b.v1.grice.speaker.mean}{55.9}  
\resdef{m1dev.qwen3-8b.v1.grice.tokfc.mean}{660.2}  
\resdef{m1dev.qwen3-8b.v1.intent.gates.mean}{0}  
\resdef{m1dev.qwen3-8b.v1.intent.literal.mean}{82.6}  
\resdef{m1dev.qwen3-8b.v1.intent.presup.mean}{57.2}  
\resdef{m1dev.qwen3-8b.v1.intent.score.mean}{68.5}  
\resdef{m1dev.qwen3-8b.v1.intent.speaker.mean}{54.3}  
\resdef{m1dev.qwen3-8b.v1.intent.tokfc.mean}{392.6}  
\resdef{m1dev.qwen3-8b.v1.nat.gates.mean}{1}  
\resdef{m1dev.qwen3-8b.v1.nat.literal.mean}{86.9}  
\resdef{m1dev.qwen3-8b.v1.nat.presup.mean}{66.5}  
\resdef{m1dev.qwen3-8b.v1.nat.score.mean}{73.4}  
\resdef{m1dev.qwen3-8b.v1.nat.speaker.mean}{59.8}  
\resdef{m1dev.qwen3-8b.v1.nat.tokfc.mean}{786.9}  
\resdef{m1dev.qwen3-8b.v1.nat512.gates.mean}{0}  
\resdef{m1dev.qwen3-8b.v1.nat512.literal.mean}{83.8}  
\resdef{m1dev.qwen3-8b.v1.nat512.presup.mean}{66.1}  
\resdef{m1dev.qwen3-8b.v1.nat512.score.mean}{72.9}  
\resdef{m1dev.qwen3-8b.v1.nat512.speaker.mean}{62.0}  
\resdef{m1dev.qwen3-8b.v1.nat512.tokfc.mean}{451.1}  
\resdef{m1dev.qwen3-8b.v1.nt.gates.mean}{0}  
\resdef{m1dev.qwen3-8b.v1.nt.literal.mean}{73.5}  
\resdef{m1dev.qwen3-8b.v1.nt.presup.mean}{52.6}  
\resdef{m1dev.qwen3-8b.v1.nt.score.mean}{66.5}  
\resdef{m1dev.qwen3-8b.v1.nt.speaker.mean}{59.5}  
\resdef{m1dev.qwen3-8b.v1.nt.tokfc.mean}{5.4}  
\resdef{m1dev.qwen3-8b.v1.route.gates.mean}{0}  
\resdef{m1dev.qwen3-8b.v1.route.literal.mean}{81.2}  
\resdef{m1dev.qwen3-8b.v1.route.presup.mean}{55.1}  
\resdef{m1dev.qwen3-8b.v1.route.score.mean}{68.0}  
\resdef{m1dev.qwen3-8b.v1.route.speaker.mean}{54.9}  
\resdef{m1dev.qwen3-8b.v1.route.tokfc.mean}{396.1}  
\resdef{m1dev.qwen3-8b.v1.taskaware.gates.mean}{0}  
\resdef{m1dev.qwen3-8b.v1.taskaware.literal.mean}{88.7}  
\resdef{m1dev.qwen3-8b.v1.taskaware.presup.mean}{64.8}  
\resdef{m1dev.qwen3-8b.v1.taskaware.score.mean}{75.5}  
\resdef{m1dev.qwen3-8b.v1.taskaware.speaker.mean}{62.4}  
\resdef{m1dev.qwen3-8b.v1.taskaware.tokfc.mean}{815.5}  
\resdef{m1dev.qwen3-8b.v1.taskcheck.gates.mean}{0}  
\resdef{m1dev.qwen3-8b.v1.taskcheck.literal.mean}{81.7}  
\resdef{m1dev.qwen3-8b.v1.taskcheck.presup.mean}{54.3}  
\resdef{m1dev.qwen3-8b.v1.taskcheck.score.mean}{66.7}  
\resdef{m1dev.qwen3-8b.v1.taskcheck.speaker.mean}{51.8}  
\resdef{m1dev.qwen3-8b.v1.taskcheck.tokfc.mean}{275.1}  
\resdef{m1dev.qwen3-8b.v1.userroute.gates.mean}{0}  
\resdef{m1dev.qwen3-8b.v1.userroute.literal.mean}{86.9}  
\resdef{m1dev.qwen3-8b.v1.userroute.presup.mean}{65.9}  
\resdef{m1dev.qwen3-8b.v1.userroute.score.mean}{73.3}  
\resdef{m1dev.qwen3-8b.v1.userroute.speaker.mean}{59.7}  
\resdef{m1dev.qwen3-8b.v1.userroute.tokfc.mean}{877.6}  
\resdef{m1dev.qwen3-8b.v2.nat.gates.mean}{1}  
\resdef{m1dev.qwen3-8b.v2.nat.literal.mean}{86.8}  
\resdef{m1dev.qwen3-8b.v2.nat.presup.mean}{67.2}  
\resdef{m1dev.qwen3-8b.v2.nat.score.mean}{73.2}  
\resdef{m1dev.qwen3-8b.v2.nat.speaker.mean}{59.6}  
\resdef{m1dev.qwen3-8b.v2.nat.tokfc.mean}{784.9}  
\resdef{m1dev.qwen3-8b.v2.neutral.gates.mean}{0}  
\resdef{m1dev.qwen3-8b.v2.neutral.literal.mean}{86.6}  
\resdef{m1dev.qwen3-8b.v2.neutral.presup.mean}{65.8}  
\resdef{m1dev.qwen3-8b.v2.neutral.score.mean}{72.7}  
\resdef{m1dev.qwen3-8b.v2.neutral.speaker.mean}{58.9}  
\resdef{m1dev.qwen3-8b.v2.neutral.tokfc.mean}{920.8}  
\resdef{m1dev.qwen3-8b.v2.sitcheck.gates.mean}{0}  
\resdef{m1dev.qwen3-8b.v2.sitcheck.literal.mean}{87.3}  
\resdef{m1dev.qwen3-8b.v2.sitcheck.presup.mean}{60.9}  
\resdef{m1dev.qwen3-8b.v2.sitcheck.score.mean}{68.3}  
\resdef{m1dev.qwen3-8b.v2.sitcheck.speaker.mean}{49.3}  
\resdef{m1dev.qwen3-8b.v2.sitcheck.tokfc.mean}{965.7}  
\resdef{m1dev.qwen3-8b.v2.sitcheck.tr.speaker.r2w.mean}{163}  
\resdef{m1dev.qwen3-8b.v2.sitcheck.tr.speaker.w2r.mean}{27}  
\resdef{m1dev.qwen3-8b.v2.taskaware.gates.mean}{0}  
\resdef{m1dev.qwen3-8b.v2.taskaware.literal.mean}{89.1}  
\resdef{m1dev.qwen3-8b.v2.taskaware.presup.mean}{66.5}  
\resdef{m1dev.qwen3-8b.v2.taskaware.score.mean}{75.8}  
\resdef{m1dev.qwen3-8b.v2.taskaware.speaker.mean}{62.5}  
\resdef{m1dev.qwen3-8b.v2.taskaware.tokfc.mean}{811.4}  
\resdef{m1dev.qwen3-8b.v3.control.literal.mean}{86.1}  
\resdef{m1dev.qwen3-8b.v3.control.presup.mean}{64.5}  
\resdef{m1dev.qwen3-8b.v3.control.score.mean}{73.3}  
\resdef{m1dev.qwen3-8b.v3.control.speaker.mean}{60.4}  
\resdef{m1dev.qwen3-8b.v3.nat.score.mean}{73.2}  
\resdef{m1dev.qwen3-8b.v3.nobal.literal.mean}{86.4}  
\resdef{m1dev.qwen3-8b.v3.nobal.presup.mean}{65.2}  
\resdef{m1dev.qwen3-8b.v3.nobal.score.mean}{73.5}  
\resdef{m1dev.qwen3-8b.v3.nobal.speaker.mean}{60.6}  
\resdef{m1dev.qwen3-8b.v3.sit.literal.mean}{86.1}  
\resdef{m1dev.qwen3-8b.v3.sit.presup.mean}{65.5}  
\resdef{m1dev.qwen3-8b.v3.sit.score.mean}{73.2}  
\resdef{m1dev.qwen3-8b.v3.sit.speaker.mean}{60.4}  
\resdef{m1dev.qwen3-8b.x.nat.score.mean}{73.0}  
\resdef{m1dev.qwen3-8b.x.nat.speaker.mean}{59.7}  
\resdef{m1dev.qwen3-8b.x.routed.mean}{283}  
\resdef{m1dev.qwen3-8b.x.x1.gates.mean}{1}  
\resdef{m1dev.qwen3-8b.x.x1.score.mean}{73.0}  
\resdef{m1dev.qwen3-8b.x.x1.speaker.mean}{59.7}  
\resdef{m1dev.qwen3-8b.x.x1w0.score.mean}{73.0}  
\resdef{m1dev.qwen3-8b.x.x1w0.speaker.mean}{59.7}  
\resdef{m1dev.qwen3-8b.x.x2.gates.mean}{1}  
\resdef{m1dev.qwen3-8b.x.x2.score.mean}{73.0}  
\resdef{m1dev.qwen3-8b.x.x2.speaker.mean}{59.7}  
\resdef{m1dev.qwen3-8b.x.x2w0.score.mean}{73.0}  
\resdef{m1dev.qwen3-8b.x.x2w0.speaker.mean}{59.7}  
\resdef{m1dev.r1-0528.d.control.gates.mean}{0}  
\resdef{m1dev.r1-0528.d.control.literal.mean}{85.0}  
\resdef{m1dev.r1-0528.d.control.presup.mean}{64.7}  
\resdef{m1dev.r1-0528.d.control.score.mean}{71.3}  
\resdef{m1dev.r1-0528.d.control.speaker.mean}{57.6}  
\resdef{m1dev.r1-0528.d.control.tokfc.mean}{1400.0}  
\resdef{m1dev.r1-0528.d.nat.gates.mean}{1}  
\resdef{m1dev.r1-0528.d.nat.literal.mean}{84.9}  
\resdef{m1dev.r1-0528.d.nat.presup.mean}{60.4}  
\resdef{m1dev.r1-0528.d.nat.score.mean}{71.0}  
\resdef{m1dev.r1-0528.d.nat.speaker.mean}{57.1}  
\resdef{m1dev.r1-0528.d.nat.tokfc.mean}{1641.0}  
\resdef{m1dev.r1-0528.d.sit.gates.mean}{0}  
\resdef{m1dev.r1-0528.d.sit.literal.mean}{84.8}  
\resdef{m1dev.r1-0528.d.sit.presup.mean}{65.4}  
\resdef{m1dev.r1-0528.d.sit.score.mean}{71.7}  
\resdef{m1dev.r1-0528.d.sit.speaker.mean}{58.5}  
\resdef{m1dev.r1-0528.d.sit.tokfc.mean}{1384.7}  
\resdef{m1dev.r1-0528.e.dpo.gates.mean}{0}  
\resdef{m1dev.r1-0528.e.dpo.literal.mean}{85.2}  
\resdef{m1dev.r1-0528.e.dpo.presup.mean}{64.8}  
\resdef{m1dev.r1-0528.e.dpo.score.mean}{72.1}  
\resdef{m1dev.r1-0528.e.dpo.speaker.mean}{59.1}  
\resdef{m1dev.r1-0528.e.dpo.tokfc.mean}{1687.8}  
\resdef{m1dev.r1-0528.e.dposhuf.gates.mean}{0}  
\resdef{m1dev.r1-0528.e.dposhuf.literal.mean}{84.8}  
\resdef{m1dev.r1-0528.e.dposhuf.presup.mean}{62.9}  
\resdef{m1dev.r1-0528.e.dposhuf.score.mean}{71.0}  
\resdef{m1dev.r1-0528.e.dposhuf.speaker.mean}{57.2}  
\resdef{m1dev.r1-0528.e.dposhuf.tokfc.mean}{1646.7}  
\resdef{m1dev.r1-0528.e.nat.score.mean}{71.0}  
\resdef{m1dev.r1-0528.e.nat.speaker.mean}{57.1}  
\resdef{m1dev.r1-0528.x.nat.score.mean}{71.1}  
\resdef{m1dev.r1-0528.x.nat.speaker.mean}{57.1}  
\resdef{m1dev.r1-0528.x.x1.gates.mean}{1}  
\resdef{m1dev.r1-0528.x.x1.score.mean}{71.1}  
\resdef{m1dev.r1-0528.x.x1.speaker.mean}{57.1}  
\resdef{m1dev.r1-0528.x.x1w0.score.mean}{71.1}  
\resdef{m1dev.r1-0528.x.x1w0.speaker.mean}{57.1}  
\resdef{m1dev.r1-0528.x.x2.gates.mean}{1}  
\resdef{m1dev.r1-0528.x.x2.score.mean}{71.1}  
\resdef{m1dev.r1-0528.x.x2.speaker.mean}{57.1}  
\resdef{m1dev.r1-0528.x.x2w0.score.mean}{71.1}  
\resdef{m1dev.r1-0528.x.x2w0.speaker.mean}{57.1}  
\resdef{e1gsf.gpt-oss-20b.a3f-conn.nli.high.dR.mean}{-0.9}  
\resdef{e1gsf.gpt-oss-20b.a3f-quant.nli.high.dR.mean}{-1.1}  
\resdef{e1r1.r1-0528-qwen3-8b.a2.inf.nat.dR.hi}{-11.3}  
\resdef{e1r1.r1-0528-qwen3-8b.a2.inf.nat.dR.lo}{-23.2}  
\resdef{e1r1.r1-0528-qwen3-8b.a2.inf.nat.dR.mean}{-17.2}  
\resdef{e1r1.r1-0528-qwen3-8b.a2.inf.nat.dR.std}{3.0}  
\resdef{e1r1.r1-0528-qwen3-8b.a3-conn.inf.nat.dR.hi}{-4.0}  
\resdef{e1r1.r1-0528-qwen3-8b.a3-conn.inf.nat.dR.lo}{-17.2}  
\resdef{e1r1.r1-0528-qwen3-8b.a3-conn.inf.nat.dR.mean}{-10.1}  
\resdef{e1r1.r1-0528-qwen3-8b.a3-conn.inf.nat.dR.std}{4.6}  
\resdef{e1r1.r1-0528-qwen3-8b.a3-modal.inf.nat.dR.hi}{0.0}  
\resdef{e1r1.r1-0528-qwen3-8b.a3-modal.inf.nat.dR.lo}{-13.5}  
\resdef{e1r1.r1-0528-qwen3-8b.a3-modal.inf.nat.dR.mean}{-6.2}  
\resdef{e1r1.r1-0528-qwen3-8b.a3-modal.inf.nat.dR.std}{5.4}  
\resdef{e1r1.r1-0528-qwen3-8b.a3-quant.inf.nat.dR.hi}{-9.0}  
\resdef{e1r1.r1-0528-qwen3-8b.a3-quant.inf.nat.dR.lo}{-42.3}  
\resdef{e1r1.r1-0528-qwen3-8b.a3-quant.inf.nat.dR.mean}{-24.4}  
\resdef{e1r1.r1-0528-qwen3-8b.a3-quant.inf.nat.dR.std}{2.2}  
\resdef{e1r1.r1-0528-qwen3-8b.a3.inf.nat.dR.hi}{2.4}  
\resdef{e1r1.r1-0528-qwen3-8b.a3.inf.nat.dR.lo}{-7.2}  
\resdef{e1r1.r1-0528-qwen3-8b.a3.inf.nat.dR.mean}{-2.4}  
\resdef{e1r1.r1-0528-qwen3-8b.a3.inf.nat.dR.std}{2.6}  
\resdef{e1r1.r1-0528-qwen3-8b.a3f-conn.nli.nat.dR.mean}{-6.3}  
\resdef{e1r1.r1-0528-qwen3-8b.a3f-quant.nli.nat.dR.mean}{-17.6}  
\resdef{e1r1.r1-0528-qwen3-8b.b2.yn.nat.dacc.hi}{13.3}  
\resdef{e1r1.r1-0528-qwen3-8b.b2.yn.nat.dacc.lo}{-1.8}  
\resdef{e1r1.r1-0528-qwen3-8b.b2.yn.nat.dacc.mean}{6.0}  
\resdef{e1r1.r1-0528-qwen3-8b.b2.yn.nat.dacc.std}{2.6}  
\resdef{x3test.arms.taskaware.score_diff}{2.3}  
\resdef{x3test.arms.x1.ci_hi}{1.6}  
\resdef{x3test.arms.x1.ci_lo}{1.2}  
\resdef{x3test.arms.x1.score_diff}{1.4}  
\resdef{x3test.minlb}{0.0}  
\resdef{gold.open.kappa-ab.mean}{0.18}  
\resdef{gold.v4.a1ctx-acc.mean}{100.0}  
\resdef{gold.v4.a1ctx-acc.wilsonlo}{86.2}  
\resdef{gold.v4.acc-class-lit.mean}{81.2}  
\resdef{gold.v4.acc-class-prag.mean}{73.1}  
\resdef{gold.v4.acc-human-a.mean}{77.4}  
\resdef{gold.v4.kappa-human-a-bin.mean}{0.59}  
\resdef{gold.v4.kappa-human-a.hi}{0.79}  
\resdef{gold.v4.kappa-human-a.lo}{0.34}  
\resdef{gold.v4.kappa-human-a.mean}{0.57}  
\resdef{gold.v4.kappa-human-b-bin.mean}{0.18}  
\resdef{gold.v4.kappa-human-b.hi}{0.31}  
\resdef{gold.v4.kappa-human-b.lo}{0.02}  
\resdef{gold.v4.kappa-human-b.mean}{0.14}  
\resdef{gold.v4.kappa-human-b3.mean}{0.03}  
\resdef{gold.v4.n-a1ctx}{24}  
\resdef{gold.v4.n-coding}{50}  
\resdef{gold.gpt6.a1ctx-acc.mean}{100.0}  
\resdef{gold.gpt6.kappa-a.hi}{0.80}  
\resdef{gold.gpt6.kappa-a.lo}{0.34}  
\resdef{gold.gpt6.kappa-a.mean}{0.58}  
\resdef{gold.gpt6.kappa-b.hi}{0.29}  
\resdef{gold.gpt6.kappa-b.lo}{-0.09}  
\resdef{gold.gpt6.kappa-b.mean}{0.10}  
\resdef{gold.gpt6.n-a1ctx}{20}  
\resdef{gold.gpt6.n-coding}{40}  
\resdef{gold.judge.kappa-human-gptoss20b.hi}{0.54}  
\resdef{gold.judge.kappa-human-gptoss20b.lo}{0.12}  
\resdef{gold.judge.kappa-human-gptoss20b.mean}{0.33}  

\draftfalse

\title{Entertain, then Cancel: How Thinking Makes \\ Reasoning Models Read  \emph{Some} and \emph{Or} Literally}

\author{
\bfseries Zhizhuo Kou\textsuperscript{1} \quad Jian Yang \textsuperscript{1} \quad
Pengfei Hao\textsuperscript{2} \quad Junyu Luo \textsuperscript{3} \\
\bfseries Jingshu Peng \textsuperscript{1} \quad
Sirui Han\textsuperscript{1}\NoHyper\footnotemark[1]\endNoHyper \quad Yike Guo\textsuperscript{1} \\[2pt]
\normalfont\textsuperscript{1}The Hong Kong University of Science and Technology \\
\normalfont\textsuperscript{2}The Hong Kong University of Science and Technology (Guangzhou) \quad
\textsuperscript{3} Peking University \\[2pt]
\normalfont\small\texttt{zkouaa@connect.ust.hk, siruihan@ust.hk}
}

\begin{document}
\maketitle

\begin{abstract}
When a speaker says ``some students passed'', listeners infer that not all of them did. Reasoning models now let users decide how long a model thinks before it answers, but we do not know what extra thinking does to such everyday inferences. We study this directly: we keep a model's weights fixed, change only how much it thinks, and record which reading it chooses. A second tool cuts the model's own reasoning at successive points and asks for an answer, showing where the reading changes. Thinking makes models read logical words as logic: it cancels the implicatures of \emph{some} (\emph{not all}) and \emph{or} (\emph{not both}) in most tested open reasoning models and in GPT-5, while context-driven inferences such as indirect answers improve. The trace shows why: models first entertain the enriched reading and cancel it later. Answering from that early reading, an exploratory inference-time remedy, recovers these implicatures without losing literal accuracy. Evaluations of reasoning models should measure how they interpret, not only how accurate they are. Code and data: \url{https://github.com/kouzhizhuo/entertain-then-cancel}.

\end{abstract}

\section{Introduction}
\label{sec:intro}

\begin{figure}[t]
  \centering
  \includegraphics[width=\columnwidth]{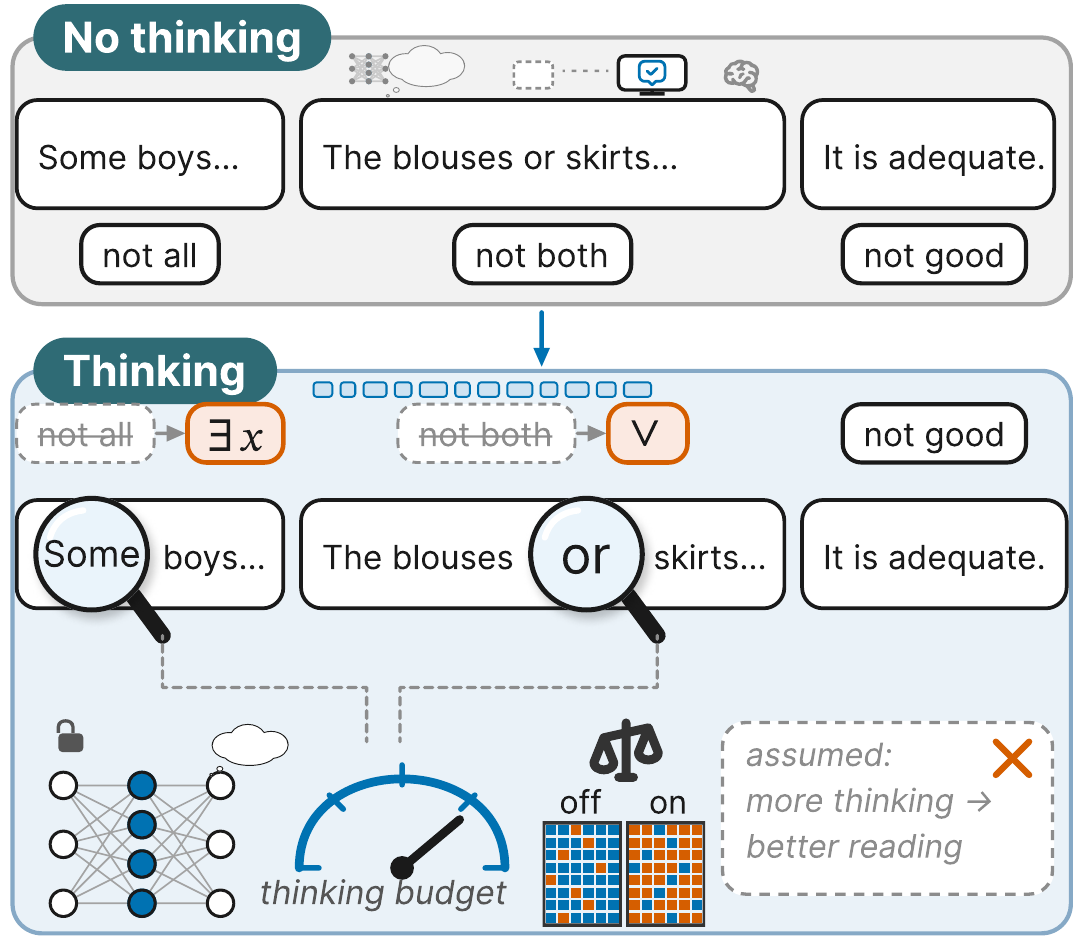}
  \caption{\textbf{Thinking acts as a logic lens on logical words.} \form{some} and \form{or} are read as $\exists x$ and $\vee$, while \form{adequate} keeps \gloss{not good}.}
  \label{fig:teaser}
  \vspace{-12pt}
\end{figure}

A listener hearing ``Some boys play'' usually concludes that not all of them do. Although \form{some} remains logically true even if every boy plays, listeners reason that a speaker knowing they all play would have said \form{all}. This step from what is said to what is meant is \emph{pragmatic enrichment}, and the inference from \form{some} to \gloss{not all} is the textbook case \citep{grice1975logic}. Reasoning models now treat thinking as a setting the user controls: a mode that can be switched on or off, a budget of thinking tokens, or an effort level \citep{yang2025qwen3,openai2025gptoss,deepseekai2025r1}. More thinking usually helps on mathematics and code. Does it also help a model understand what a speaker means? Existing evaluations score whether a model recovers an intended meaning \citep{hu-etal-2023-fine,ruis2023goldilocks,sravanthi-etal-2024-pub}. Recent studies report that more thinking can hurt theory of mind \citep{gong2026think} and that Qwen3 recognizes fewer scalar implicatures in thinking mode \citep{spinoso2026evaluating}. No study yet measures \textbf{what deliberation does to the reading a model chooses} when both readings are allowed.

Human pragmatics offers two competing predictions. On the \emph{effort account}, enrichment costs effort, so more thinking should yield more of it \citep{sperber1986relevance,bott2004some}. On the \emph{default-and-cancel account}, an implicature such as \form{some}\,$\rightarrow$\,\gloss{not all} arises by default and is cancelled when the context or a check of the literal meaning speaks against it \citep{levinson2000presumptive}.

Our idea is simple. Because thinking is a setting, we can hold a model's weights fixed and turn only this dial, as in a dose--response experiment, and record which reading the model chooses (\Cref{fig:teaser}). To see where the reading changes, we cut the model's own reasoning text, its \emph{trace}, after its first $t$ tokens and ask for an answer at each cut. We apply both tools to \res{derived.main.tested} open and seven closed reasoning models, on items that separate logical words (\form{some}, \form{or}, modals) from lexical scales (\form{good}/\form{excellent}), presuppositions and indirect answers. The headline is that thinking makes models read logical words as logic, while inferences that need the context improve; along the trace, models entertain the enriched reading early and cancel it.

Our contributions are as follows: \ding{182} \textbf{Thinking Cancels Implicature:} Extended reasoning cancels scalar implicatures for \form{some} and \form{or} across \res{derived.main.counted}/\res{derived.main.tested} open models and GPT-5, as well as in label-free three-way choices (\res{derived.fullsets.nli.drop}/\res{derived.fullsets.nli.tested} models). \ding{183} \textbf{Diagnostic Protocol:} We introduce a fixed-weight framework combining dose--response profiling with a trace-prefix diagnostic to track interpretation dynamics while controlling for yes-bias. \ding{184} \textbf{Entertain, Then Cancel:} Models typically entertain pragmatic readings in early reasoning tokens before switching to literal readings in completed outputs. \ding{185} \textbf{Early-Trace Extraction:} Extracting answers after the first 128 trace tokens improves accuracy on scalar felicity items across two 8B models without compromising literal performance. \ding{186} \textbf{Released Resources:} We release new Bott--Noveck sentences with expert-checked speaker-knowledge scenarios, expanded IMPPRES test sets, full trace logs, and machine-translated German/Chinese slices.

\section{Background and Related Work}
\label{sec:background}

\paragraph{Two kinds of implicature.}
\citet{grice1975logic} separated \emph{generalized} implicatures, which a word carries unless the context blocks them (\form{some}\,$\rightarrow$\,\gloss{not all}, \form{or}\,$\rightarrow$\,\gloss{not both}), from \emph{particularized} ones, which arise only in a specific context (``I wore gloves'' in reply to ``Did you leave fingerprints?'' means \gloss{no}). Generalized implicatures attach both to logical words and to lexical scales such as \form{good}/\form{excellent}, whose enrichment rates vary widely across scales \citep{vantiel2016scalar}.

\paragraph{Three predictions for a thinking model.}
The \emph{effort account} treats every implicature as an inference that costs processing effort \citep{sperber1986relevance}. Adults who reject ``Some elephants are mammals'' respond more slowly, and they reject it less often under time pressure or memory load \citep{bott2004some,deneys2007logical,marty2013scalar}, although this is contested \citep{grodner2010some,degen2015processing}. If thinking tokens act as processing resources, more thinking should mean more enrichment.
The \emph{default-and-cancel account} holds generalized implicatures as defaults that a listener cancels when the context or a check of the truth conditions speaks against them \citep{levinson2000presumptive}. If a model spends its thinking on checking strict entailment, it should cancel the implicatures of logical words, while particularized implicatures, reachable only by inference, should hold or improve.
A third, developmental view notes that children accept underinformative \form{some} more frequently than adults \citep{noveck2001children}. If enrichment tracks competence, thinking may amplify a model's existing preference, with a direction that changes with capability.
We use \emph{default} for the reading chosen without thinking and \emph{cancel} for a shift towards the literal reading as thinking grows, without committing to a cognitive mechanism. \Cref{app:theory} lists the pre-registered hypotheses and falsifiers.

\begin{figure}[t]
  \centering
  \includegraphics[width=\columnwidth]{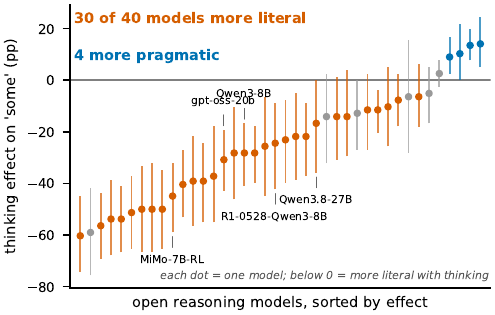}
  \caption{\textbf{Thinking makes \form{some} more literal in most open reasoning models.} Vermilion: counted literal shift; blue: more pragmatic (\Cref{fig:hero-full}).}
  \label{fig:hero}
  \vspace{-12pt}
\end{figure}

\paragraph{Prior evidence in language models.}
Chain-of-thought prompts improve implicature resolution \citep{kim2023pope,ruis2023goldilocks}, yet chain-of-thought was also seen to ``encourage literal interpretations'' \citep{park-etal-2024-pragmatic}. More reasoning lowers theory-of-mind accuracy \citep{gong2026think}, and in ImplicatureX Qwen3's thinking mode recognises fewer scalar implicatures and more cancellations at one budget \citep{spinoso2026evaluating}. These studies measure accuracy at fixed settings. We measure the reading chosen as a dose--response on fixed weights, separate implicature types, and locate the change in the trace and in training (\Cref{app:related}).

\section{Method}
\label{sec:method}

\begin{figure*}[t]
  \centering
  \includegraphics[width=\textwidth]{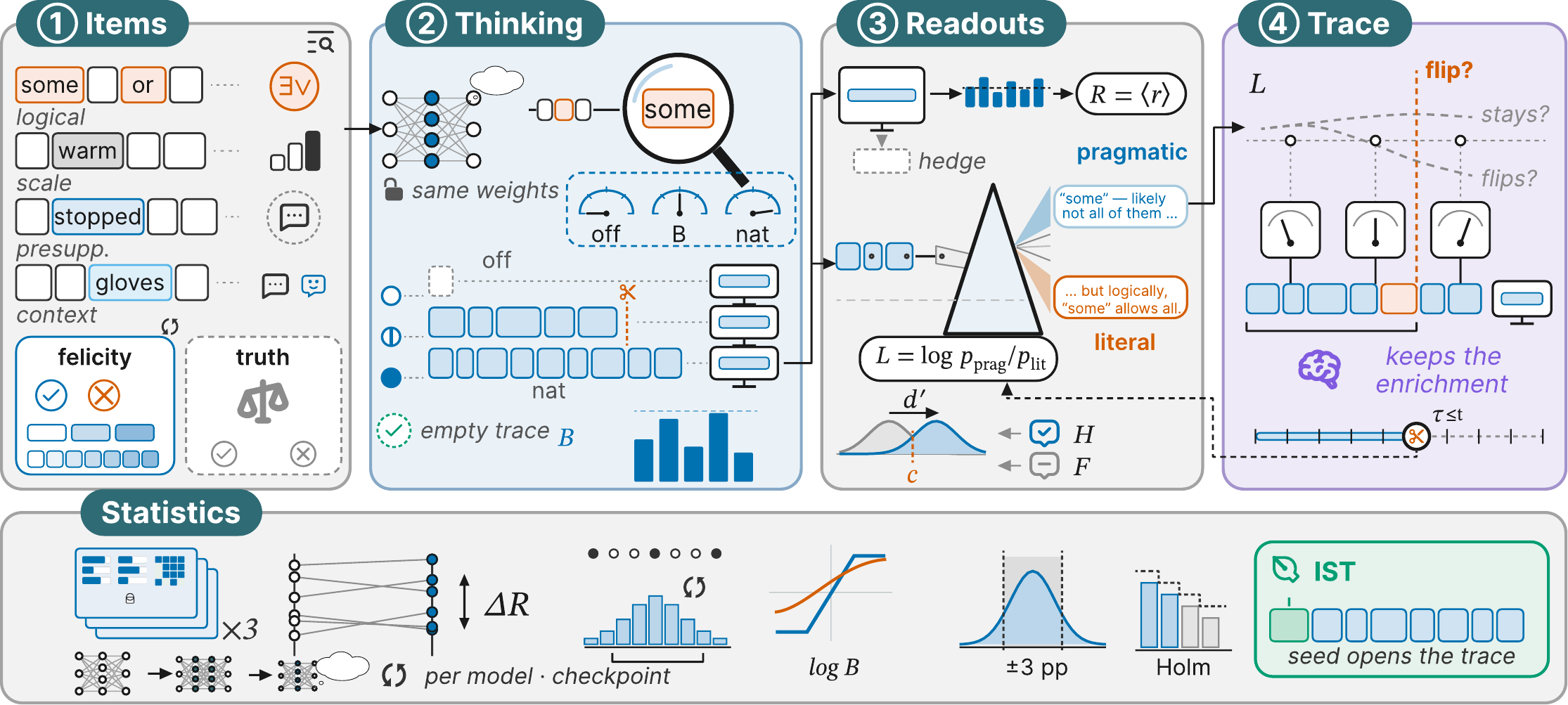}
  \caption{\textbf{A reading probe on fixed weights.} Items with a trigger (\ding{172}) are answered with thinking off, capped or natural (\ding{173}); we read the chosen reading $\rate$ and the log-odds $L$ (\ding{174}), also after successive trace prefixes (\ding{175}), for every model and checkpoint.}
  \label{fig:method}
  \vspace{-12pt}
\end{figure*}

As shows in (\Cref{fig:method}): a dose protocol that changes only how much a model thinks, and a trace diagnostic that shows where its reading changes.

\paragraph{\ding{172} Items and framings.}
Each item is a sentence with a trigger and a question about what the speaker meant. In our running example a speaker says ``Some boys play'', and the model is asked whether it would conclude that the speaker believes ``Not all boys play''; \gloss{yes} is the pragmatic reading and \gloss{no} the literal one.
Logical-word items (\form{some}, \form{or}, modals) come from IMPPRES \citep{jeretic-etal-2020-natural}; lexical-scale items from scalar-diversity sets with human enrichment rates \citep{vantiel2016scalar,gotzner2018scalar}; presupposition items from IMPPRES; and particularized implicatures from LUDWIG \citep{ruis2023goldilocks}, PragMega \citep{hu-etal-2023-fine}, Fig-QA \citep{liu-etal-2022-testing}, PUB \citep{sravanthi-etal-2024-pub} and Circa \citep{louis-etal-2020-id}, with NOPE as a second presupposition set \citep{parrish-etal-2021-nope}.
Primary framings ask what the speaker meant or whether the sentence is an adequate description (\emph{felicity}). A true/false question serves as a logic-mode comparison, as the literal answer is strictly correct in that context. Options are shuffled on every trial. Headline numbers are computed on a held-out half of the items; replications on full released sets are marked as such.

\paragraph{\ding{173} Thinking.}
The same weights answer with thinking off, with thinking stopped after a budget of $B$ tokens, or with natural stopping, subject to an 8{,}192-token safety cap (\NAT). For Qwen3 \citep{yang2025qwen3}, ``off'' is the model's own switch, which differs from the thinking template only by an empty thinking block; for gpt-oss we use its low and high reasoning effort \citep{openai2025gptoss}. Capped budgets append either Qwen3's own stop phrase or only the closing tag, because the stop phrase mentions limited time.

\paragraph{\ding{174} Readouts.}
Each answer ends in a line \tok{Answer: <option>}; anything else counts as a hedge. With $r_{imk}$ the share of committed answers that take the pragmatic reading for item $i$, model $m$ and condition $k$, the \emph{reading chosen} is
\begin{equation}
  \rate_{mk} \;=\; \frac{1}{|I|}\sum_{i\in I} r_{imk},
  \label{eq:rate}
\end{equation}
and the thinking effect is $\Delta\rate=\rate_{\NAT}-\rate_{\text{off}}$, in percentage points. For particularized implicatures and presuppositions $\rate$ is accuracy.
A second readout needs no answer parsing: after the trace $\trace$ we force the answer prefix and compare the two options,
\begin{equation}
  L_{ik} \;=\; \log \frac{p(\text{pragmatic}\mid x_i, \trace_{ik})}{p(\text{literal}\mid x_i, \trace_{ik})}.
  \label{eq:logodds}
\end{equation}
On category sentences we also compute signal-detection sensitivity $d'=z(H)-z(F)$ and criterion $c$ from hits and false alarms on control sentences (\Cref{app:method}).

\paragraph{\ding{175} Trace.}
To see where the reading changes, we cut each natural trace after its first $t$ tokens, $t \in \{0, 32, \ldots, 2048\}$, close it, and read $\rate$ and $L$ from that prefix $\trace_{\leq t}$. The curve $\rate(t)$ describes, on average over items, how the reading endorsed after a prefix changes along the trace; it is an intervention on the trace, not a readout of a hidden interpretation. Per trace, the diagnostic labels an \emph{entertain-then-cancel} trajectory when the answer forced at an early prefix is pragmatic and the final answer is literal, and a \emph{literal-throughout} trajectory when no prefix yields the pragmatic answer. The probe is anchored at both ends: with an empty prefix it reproduces the model's no-thinking reading (\res{e7f.qwen3-8b.a3f-quant.inf.ntforced.R.mean}\% against \res{e1sf.qwen3-8b.a3f-quant.inf.off.R.mean}\% for \form{some} in Qwen3-8B), and every item's intended reading comes from source gold or, for the speaker-knowledge scenarios, from a blind author check that chose the intended reading in all \res{gold.v4.n-a1ctx} checked scenarios.

\paragraph{Models.}
The primary models are Qwen3 from 0.6B to 32B, whose thinking switch leaves everything else fixed; OLMo-3-Think 7B and 32B and gpt-oss 20b and 120b add two families with same-recipe size pairs \citep{olmo2025olmo3,openai2025gptoss}. Ten further open reasoning models and seven closed models widen the panel. To trace the origin of the effect we score checkpoints along base $\rightarrow$ SFT $\rightarrow$ DPO $\rightarrow$ RL lineages \citep{olmo2025olmo2,olmo2025olmo3,lambert2024tulu}. \Cref{app:models,app:method} give model identifiers, prompts and all settings.

\paragraph{Statistics.}
Each condition runs with three decoding seeds at identical sampling settings. Intervals are 95\% cluster-bootstrap confidence intervals over items, paired across conditions; a difference is significant when its interval excludes 0. The pre-registered tests are a logistic model of the pragmatic reading on the log thinking budget, the phenomenon class and their interaction, fitted with item-clustered robust errors in place of the planned random effects; equivalence tests at $\pm3$\,pp; and Holm correction within each model family (\Cref{app:stats}). Several pre-registered predictions did not hold; \Cref{app:prereg} reports every outcome.

\section{Thinking Makes \emph{Some} and \emph{Or} More Literal}
\label{sec:find}

This section shows that thinking makes logical words literal, that the shift survives every analysis choice, where in the trace it happens, and at which training stage the literal default appears.


\subsection{Thinking Cancels the Implicatures of \emph{Some} and \emph{Or}}
\label{sec:find:split}

\begin{table*}[t]
  \centering\small
  \setlength{\tabcolsep}{3.5pt}
  \resizebox{\textwidth}{!}{%
  \begin{tabular}{@{}l ccc cc cc@{}}
    \toprule
    & \multicolumn{3}{c}{\textsc{Logical words}} & \multicolumn{2}{c}{\textsc{Rest of the battery}} & \multicolumn{2}{c}{\textsc{Three-way choice}} \\
    \cmidrule(lr){2-4}\cmidrule(lr){5-6}\cmidrule(l){7-8}
    Model & \form{some} & \form{or} & modals & lexical scales & indirect ans. & \form{some} & \form{or} \\
    \midrule
    \multicolumn{8}{@{}l}{\textsc{Qwen3 (thinking off $\rightarrow$ natural thinking)}} \\
    Qwen3-4B & \cellcolor{cInterv!21}\res{e1.qwen3-4b.a3-quant.inf.nat.dR.mean}{\scriptsize\,\textcolor{gray}{[\res{e1.qwen3-4b.a3-quant.inf.nat.dR.lo}, \res{e1.qwen3-4b.a3-quant.inf.nat.dR.hi}]}} & \cellcolor{cInterv!24}\underline{\res{e1.qwen3-4b.a3-conn.inf.nat.dR.mean}{\scriptsize\,\textcolor{gray}{[\res{e1.qwen3-4b.a3-conn.inf.nat.dR.lo}, \res{e1.qwen3-4b.a3-conn.inf.nat.dR.hi}]}}} & \cellcolor{cInterv!21}\res{e1.qwen3-4b.a3-modal.inf.nat.dR.mean}{\scriptsize\,\textcolor{gray}{[\res{e1.qwen3-4b.a3-modal.inf.nat.dR.lo}, \res{e1.qwen3-4b.a3-modal.inf.nat.dR.hi}]}} & \cellcolor{cInterv!14}\res{e1.qwen3-4b.a2.inf.nat.dR.mean}{\scriptsize\,\textcolor{gray}{[\res{e1.qwen3-4b.a2.inf.nat.dR.lo}, \res{e1.qwen3-4b.a2.inf.nat.dR.hi}]}} & \cellcolor{cInterv!5}\res{e1.qwen3-4b.b2.yn.nat.dacc.mean}{\scriptsize\,\textcolor{gray}{[\res{e1.qwen3-4b.b2.yn.nat.dacc.lo}, \res{e1.qwen3-4b.b2.yn.nat.dacc.hi}]}}$^{\circ}$ & \cellcolor{cInterv!15}\res{e1sf.qwen3-4b.a3f-quant.nli.nat.dR.mean} & \cellcolor{cInterv!16}\res{e1sf.qwen3-4b.a3f-conn.nli.nat.dR.mean}  \\
    Qwen3-8B & \cellcolor{cInterv!20}\res{e1.qwen3-8b.a3-quant.inf.nat.dR.mean}{\scriptsize\,\textcolor{gray}{[\res{e1.qwen3-8b.a3-quant.inf.nat.dR.lo}, \res{e1.qwen3-8b.a3-quant.inf.nat.dR.hi}]}} & \cellcolor{cInterv!21}\res{e1.qwen3-8b.a3-conn.inf.nat.dR.mean}{\scriptsize\,\textcolor{gray}{[\res{e1.qwen3-8b.a3-conn.inf.nat.dR.lo}, \res{e1.qwen3-8b.a3-conn.inf.nat.dR.hi}]}} & \cellcolor{cInterv!16}\res{e1.qwen3-8b.a3-modal.inf.nat.dR.mean}{\scriptsize\,\textcolor{gray}{[\res{e1.qwen3-8b.a3-modal.inf.nat.dR.lo}, \res{e1.qwen3-8b.a3-modal.inf.nat.dR.hi}]}} & \cellcolor{cOurs!12}\res{e1.qwen3-8b.a2.inf.nat.dR.mean}{\scriptsize\,\textcolor{gray}{[\res{e1.qwen3-8b.a2.inf.nat.dR.lo}, \res{e1.qwen3-8b.a2.inf.nat.dR.hi}]}} & \cellcolor{cOurs!13}\res{e1.qwen3-8b.b2.yn.nat.dacc.mean}{\scriptsize\,\textcolor{gray}{[\res{e1.qwen3-8b.b2.yn.nat.dacc.lo}, \res{e1.qwen3-8b.b2.yn.nat.dacc.hi}]}}$^{\circ}$ & \cellcolor{cInterv!21}\underline{\res{e1sf.qwen3-8b.a3f-quant.nli.nat.dR.mean}} & \cellcolor{cInterv!16}\res{e1sf.qwen3-8b.a3f-conn.nli.nat.dR.mean}  \\
    Qwen3-14B & \cellcolor{cInterv!19}\res{e1.qwen3-14b.a3-quant.inf.nat.dR.mean}{\scriptsize\,\textcolor{gray}{[\res{e1.qwen3-14b.a3-quant.inf.nat.dR.lo}, \res{e1.qwen3-14b.a3-quant.inf.nat.dR.hi}]}} & \cellcolor{cInterv!16}\res{e1.qwen3-14b.a3-conn.inf.nat.dR.mean}{\scriptsize\,\textcolor{gray}{[\res{e1.qwen3-14b.a3-conn.inf.nat.dR.lo}, \res{e1.qwen3-14b.a3-conn.inf.nat.dR.hi}]}} & \cellcolor{cOurs!11}\res{e1.qwen3-14b.a3-modal.inf.nat.dR.mean}{\scriptsize\,\textcolor{gray}{[\res{e1.qwen3-14b.a3-modal.inf.nat.dR.lo}, \res{e1.qwen3-14b.a3-modal.inf.nat.dR.hi}]}}$^{\circ}$ & \cellcolor{cOurs!13}\res{e1.qwen3-14b.a2.inf.nat.dR.mean}{\scriptsize\,\textcolor{gray}{[\res{e1.qwen3-14b.a2.inf.nat.dR.lo}, \res{e1.qwen3-14b.a2.inf.nat.dR.hi}]}} & \cellcolor{cOurs!7}\res{e1.qwen3-14b.b2.yn.nat.dacc.mean}{\scriptsize\,\textcolor{gray}{[\res{e1.qwen3-14b.b2.yn.nat.dacc.lo}, \res{e1.qwen3-14b.b2.yn.nat.dacc.hi}]}}$^{\circ}$ & \cellcolor{cInterv!16}\res{e1sf.qwen3-14b.a3f-quant.nli.nat.dR.mean} & \cellcolor{cInterv!9}\res{e1sf.qwen3-14b.a3f-conn.nli.nat.dR.mean}  \\
    Qwen3-30B-A3B & \cellcolor{cInterv!20}\res{e1s.qwen3-30b-a3b.a3-quant.inf.nat.dR.mean}{\scriptsize\,\textcolor{gray}{[\res{e1s.qwen3-30b-a3b.a3-quant.inf.nat.dR.lo}, \res{e1s.qwen3-30b-a3b.a3-quant.inf.nat.dR.hi}]}} & \cellcolor{cInterv!22}\res{e1s.qwen3-30b-a3b.a3-conn.inf.nat.dR.mean}{\scriptsize\,\textcolor{gray}{[\res{e1s.qwen3-30b-a3b.a3-conn.inf.nat.dR.lo}, \res{e1s.qwen3-30b-a3b.a3-conn.inf.nat.dR.hi}]}} & \cellcolor{cInterv!18}\res{e1s.qwen3-30b-a3b.a3-modal.inf.nat.dR.mean}{\scriptsize\,\textcolor{gray}{[\res{e1s.qwen3-30b-a3b.a3-modal.inf.nat.dR.lo}, \res{e1s.qwen3-30b-a3b.a3-modal.inf.nat.dR.hi}]}} & \cellcolor{cOurs!6}\res{e1s.qwen3-30b-a3b.a2.inf.nat.dR.mean}{\scriptsize\,\textcolor{gray}{[\res{e1s.qwen3-30b-a3b.a2.inf.nat.dR.lo}, \res{e1s.qwen3-30b-a3b.a2.inf.nat.dR.hi}]}}$^{\circ}$ & \cellcolor{cOurs!15}\res{e1s.qwen3-30b-a3b.b2.yn.nat.dacc.mean}{\scriptsize\,\textcolor{gray}{[\res{e1s.qwen3-30b-a3b.b2.yn.nat.dacc.lo}, \res{e1s.qwen3-30b-a3b.b2.yn.nat.dacc.hi}]}} & \cellcolor{cInterv!16}\res{e1sf.qwen3-30b-a3b.a3f-quant.nli.nat.dR.mean} & \cellcolor{cInterv!16}\res{e1sf.qwen3-30b-a3b.a3f-conn.nli.nat.dR.mean}  \\
    Qwen3-32B & \cellcolor{cInterv!23}\underline{\res{e1s.qwen3-32b.a3-quant.inf.nat.dR.mean}{\scriptsize\,\textcolor{gray}{[\res{e1s.qwen3-32b.a3-quant.inf.nat.dR.lo}, \res{e1s.qwen3-32b.a3-quant.inf.nat.dR.hi}]}}} & \cellcolor{cInterv!22}\res{e1s.qwen3-32b.a3-conn.inf.nat.dR.mean}{\scriptsize\,\textcolor{gray}{[\res{e1s.qwen3-32b.a3-conn.inf.nat.dR.lo}, \res{e1s.qwen3-32b.a3-conn.inf.nat.dR.hi}]}} & \cellcolor{cInterv!16}\res{e1s.qwen3-32b.a3-modal.inf.nat.dR.mean}{\scriptsize\,\textcolor{gray}{[\res{e1s.qwen3-32b.a3-modal.inf.nat.dR.lo}, \res{e1s.qwen3-32b.a3-modal.inf.nat.dR.hi}]}}$^{\circ}$ & \cellcolor{cOurs!11}\res{e1s.qwen3-32b.a2.inf.nat.dR.mean}{\scriptsize\,\textcolor{gray}{[\res{e1s.qwen3-32b.a2.inf.nat.dR.lo}, \res{e1s.qwen3-32b.a2.inf.nat.dR.hi}]}} & \cellcolor{cOurs!9}\res{e1s.qwen3-32b.b2.yn.nat.dacc.mean}{\scriptsize\,\textcolor{gray}{[\res{e1s.qwen3-32b.b2.yn.nat.dacc.lo}, \res{e1s.qwen3-32b.b2.yn.nat.dacc.hi}]}}$^{\circ}$ & \cellcolor{cInterv!21}\textbf{\res{e1sf.qwen3-32b.a3f-quant.nli.nat.dR.mean}} & \cellcolor{cInterv!18}\res{e1sf.qwen3-32b.a3f-conn.nli.nat.dR.mean}  \\
    \addlinespace[0.5mm]\midrule
    \multicolumn{8}{@{}l}{\textsc{Qwen3.5 and Qwen3.8 (thinking off $\rightarrow$ natural thinking)}} \\
    Qwen3.5-0.8B & \cellcolor{cInterv!16}\res{e1q.qwen3p5-0p8b.a3-quant.inf.nat.dR.mean}{\scriptsize\,\textcolor{gray}{[\res{e1q.qwen3p5-0p8b.a3-quant.inf.nat.dR.lo}, \res{e1q.qwen3p5-0p8b.a3-quant.inf.nat.dR.hi}]}}$^{\circ}$ & \cellcolor{cInterv!17}\res{e1q.qwen3p5-0p8b.a3-conn.inf.nat.dR.mean}{\scriptsize\,\textcolor{gray}{[\res{e1q.qwen3p5-0p8b.a3-conn.inf.nat.dR.lo}, \res{e1q.qwen3p5-0p8b.a3-conn.inf.nat.dR.hi}]}} & \cellcolor{cInterv!15}\res{e1q.qwen3p5-0p8b.a3-modal.inf.nat.dR.mean}{\scriptsize\,\textcolor{gray}{[\res{e1q.qwen3p5-0p8b.a3-modal.inf.nat.dR.lo}, \res{e1q.qwen3p5-0p8b.a3-modal.inf.nat.dR.hi}]}} & \cellcolor{cInterv!18}\res{e1q.qwen3p5-0p8b.a2.inf.nat.dR.mean}{\scriptsize\,\textcolor{gray}{[\res{e1q.qwen3p5-0p8b.a2.inf.nat.dR.lo}, \res{e1q.qwen3p5-0p8b.a2.inf.nat.dR.hi}]}} & \cellcolor{cInterv!12}\res{e1q.qwen3p5-0p8b.b2.yn.nat.dacc.mean}{\scriptsize\,\textcolor{gray}{[\res{e1q.qwen3p5-0p8b.b2.yn.nat.dacc.lo}, \res{e1q.qwen3p5-0p8b.b2.yn.nat.dacc.hi}]}}$^{\circ}$ & \cellcolor{cOurs!15}\res{e1sfx.qwen3p5-0p8b.a3f-quant.nli.nat.dR.mean} & \cellcolor{cInterv!14}\res{e1sfx.qwen3p5-0p8b.a3f-conn.nli.nat.dR.mean}  \\
    Qwen3.5-2B & \cellcolor{cInterv!17}\res{e1q.qwen3p5-2b.a3-quant.inf.nat.dR.mean}{\scriptsize\,\textcolor{gray}{[\res{e1q.qwen3p5-2b.a3-quant.inf.nat.dR.lo}, \res{e1q.qwen3p5-2b.a3-quant.inf.nat.dR.hi}]}}$^{\circ}$ & \cellcolor{cInterv!21}\res{e1q.qwen3p5-2b.a3-conn.inf.nat.dR.mean}{\scriptsize\,\textcolor{gray}{[\res{e1q.qwen3p5-2b.a3-conn.inf.nat.dR.lo}, \res{e1q.qwen3p5-2b.a3-conn.inf.nat.dR.hi}]}} & \cellcolor{cInterv!19}\res{e1q.qwen3p5-2b.a3-modal.inf.nat.dR.mean}{\scriptsize\,\textcolor{gray}{[\res{e1q.qwen3p5-2b.a3-modal.inf.nat.dR.lo}, \res{e1q.qwen3p5-2b.a3-modal.inf.nat.dR.hi}]}} & \cellcolor{cOurs!16}\res{e1q.qwen3p5-2b.a2.inf.nat.dR.mean}{\scriptsize\,\textcolor{gray}{[\res{e1q.qwen3p5-2b.a2.inf.nat.dR.lo}, \res{e1q.qwen3p5-2b.a2.inf.nat.dR.hi}]}} & \cellcolor{cInterv!8}\res{e1q.qwen3p5-2b.b2.yn.nat.dacc.mean}{\scriptsize\,\textcolor{gray}{[\res{e1q.qwen3p5-2b.b2.yn.nat.dacc.lo}, \res{e1q.qwen3p5-2b.b2.yn.nat.dacc.hi}]}}$^{\circ}$ & \cellcolor{cInterv!14}\res{e1sfx.qwen3p5-2b.a3f-quant.nli.nat.dR.mean} & \cellcolor{cInterv!14}\res{e1sfx.qwen3p5-2b.a3f-conn.nli.nat.dR.mean}  \\
    Qwen3.5-4B & \cellcolor{cInterv!21}\res{e1q.qwen3p5-4b.a3-quant.inf.nat.dR.mean}{\scriptsize\,\textcolor{gray}{[\res{e1q.qwen3p5-4b.a3-quant.inf.nat.dR.lo}, \res{e1q.qwen3p5-4b.a3-quant.inf.nat.dR.hi}]}} & \cellcolor{cInterv!21}\res{e1q.qwen3p5-4b.a3-conn.inf.nat.dR.mean}{\scriptsize\,\textcolor{gray}{[\res{e1q.qwen3p5-4b.a3-conn.inf.nat.dR.lo}, \res{e1q.qwen3p5-4b.a3-conn.inf.nat.dR.hi}]}} & \cellcolor{cOurs!3}\res{e1q.qwen3p5-4b.a3-modal.inf.nat.dR.mean}{\scriptsize\,\textcolor{gray}{[\res{e1q.qwen3p5-4b.a3-modal.inf.nat.dR.lo}, \res{e1q.qwen3p5-4b.a3-modal.inf.nat.dR.hi}]}}$^{\circ}$ & \cellcolor{cOurs!9}\res{e1q.qwen3p5-4b.a2.inf.nat.dR.mean}{\scriptsize\,\textcolor{gray}{[\res{e1q.qwen3p5-4b.a2.inf.nat.dR.lo}, \res{e1q.qwen3p5-4b.a2.inf.nat.dR.hi}]}}$^{\circ}$ & \cellcolor{cOurs!11}\res{e1q.qwen3p5-4b.b2.yn.nat.dacc.mean}{\scriptsize\,\textcolor{gray}{[\res{e1q.qwen3p5-4b.b2.yn.nat.dacc.lo}, \res{e1q.qwen3p5-4b.b2.yn.nat.dacc.hi}]}}$^{\circ}$ & \cellcolor{cInterv!16}\res{e1sfx.qwen3p5-4b.a3f-quant.nli.nat.dR.mean} & \cellcolor{cInterv!17}\res{e1sfx.qwen3p5-4b.a3f-conn.nli.nat.dR.mean}  \\
    Qwen3.5-9B & \cellcolor{cInterv!21}\res{e1q.qwen3p5-9b.a3-quant.inf.nat.dR.mean}{\scriptsize\,\textcolor{gray}{[\res{e1q.qwen3p5-9b.a3-quant.inf.nat.dR.lo}, \res{e1q.qwen3p5-9b.a3-quant.inf.nat.dR.hi}]}} & \cellcolor{cInterv!19}\res{e1q.qwen3p5-9b.a3-conn.inf.nat.dR.mean}{\scriptsize\,\textcolor{gray}{[\res{e1q.qwen3p5-9b.a3-conn.inf.nat.dR.lo}, \res{e1q.qwen3p5-9b.a3-conn.inf.nat.dR.hi}]}} & \cellcolor{cOurs!9}\res{e1q.qwen3p5-9b.a3-modal.inf.nat.dR.mean}{\scriptsize\,\textcolor{gray}{[\res{e1q.qwen3p5-9b.a3-modal.inf.nat.dR.lo}, \res{e1q.qwen3p5-9b.a3-modal.inf.nat.dR.hi}]}}$^{\circ}$ & \cellcolor{cOurs!11}\res{e1q.qwen3p5-9b.a2.inf.nat.dR.mean}{\scriptsize\,\textcolor{gray}{[\res{e1q.qwen3p5-9b.a2.inf.nat.dR.lo}, \res{e1q.qwen3p5-9b.a2.inf.nat.dR.hi}]}} & \cellcolor{cOurs!13}\res{e1q.qwen3p5-9b.b2.yn.nat.dacc.mean}{\scriptsize\,\textcolor{gray}{[\res{e1q.qwen3p5-9b.b2.yn.nat.dacc.lo}, \res{e1q.qwen3p5-9b.b2.yn.nat.dacc.hi}]}} & \cellcolor{cInterv!15}\res{e1sfx.qwen3p5-9b.a3f-quant.nli.nat.dR.mean} & \cellcolor{cInterv!13}\res{e1sfx.qwen3p5-9b.a3f-conn.nli.nat.dR.mean}  \\
    Qwen3.8-27B & \cellcolor{cInterv!17}\res{e1q.qwen3p8-27b.a3-quant.inf.nat.dR.mean}{\scriptsize\,\textcolor{gray}{[\res{e1q.qwen3p8-27b.a3-quant.inf.nat.dR.lo}, \res{e1q.qwen3p8-27b.a3-quant.inf.nat.dR.hi}]}}$^{\circ}$ & \cellcolor{cInterv!19}\res{e1q.qwen3p8-27b.a3-conn.inf.nat.dR.mean}{\scriptsize\,\textcolor{gray}{[\res{e1q.qwen3p8-27b.a3-conn.inf.nat.dR.lo}, \res{e1q.qwen3p8-27b.a3-conn.inf.nat.dR.hi}]}} & \cellcolor{cInterv!14}\res{e1q.qwen3p8-27b.a3-modal.inf.nat.dR.mean}{\scriptsize\,\textcolor{gray}{[\res{e1q.qwen3p8-27b.a3-modal.inf.nat.dR.lo}, \res{e1q.qwen3p8-27b.a3-modal.inf.nat.dR.hi}]}} & \cellcolor{cOurs!17}\res{e1q.qwen3p8-27b.a2.inf.nat.dR.mean}{\scriptsize\,\textcolor{gray}{[\res{e1q.qwen3p8-27b.a2.inf.nat.dR.lo}, \res{e1q.qwen3p8-27b.a2.inf.nat.dR.hi}]}} & \cellcolor{cOurs!15}\res{e1q.qwen3p8-27b.b2.yn.nat.dacc.mean}{\scriptsize\,\textcolor{gray}{[\res{e1q.qwen3p8-27b.b2.yn.nat.dacc.lo}, \res{e1q.qwen3p8-27b.b2.yn.nat.dacc.hi}]}} & \cellcolor{cInterv!7}\res{e1sfx.qwen3p8-27b.a3f-quant.nli.nat.dR.mean} & \cellcolor{cInterv!10}\res{e1sfx.qwen3p8-27b.a3f-conn.nli.nat.dR.mean}  \\
    Qwen3.8-27B (low effort) & \cellcolor{cInterv!19}\res{e1q.qwen3p8-27b-elow.a3-quant.inf.nat.dR.mean}{\scriptsize\,\textcolor{gray}{[\res{e1q.qwen3p8-27b-elow.a3-quant.inf.nat.dR.lo}, \res{e1q.qwen3p8-27b-elow.a3-quant.inf.nat.dR.hi}]}} & \cellcolor{cInterv!21}\res{e1q.qwen3p8-27b-elow.a3-conn.inf.nat.dR.mean}{\scriptsize\,\textcolor{gray}{[\res{e1q.qwen3p8-27b-elow.a3-conn.inf.nat.dR.lo}, \res{e1q.qwen3p8-27b-elow.a3-conn.inf.nat.dR.hi}]}} & \cellcolor{cInterv!9}\res{e1q.qwen3p8-27b-elow.a3-modal.inf.nat.dR.mean}{\scriptsize\,\textcolor{gray}{[\res{e1q.qwen3p8-27b-elow.a3-modal.inf.nat.dR.lo}, \res{e1q.qwen3p8-27b-elow.a3-modal.inf.nat.dR.hi}]}}$^{\circ}$ & \cellcolor{cOurs!17}\res{e1q.qwen3p8-27b-elow.a2.inf.nat.dR.mean}{\scriptsize\,\textcolor{gray}{[\res{e1q.qwen3p8-27b-elow.a2.inf.nat.dR.lo}, \res{e1q.qwen3p8-27b-elow.a2.inf.nat.dR.hi}]}} & \cellcolor{cOurs!14}\res{e1q.qwen3p8-27b-elow.b2.yn.nat.dacc.mean}{\scriptsize\,\textcolor{gray}{[\res{e1q.qwen3p8-27b-elow.b2.yn.nat.dacc.lo}, \res{e1q.qwen3p8-27b-elow.b2.yn.nat.dacc.hi}]}} & \cellcolor{cInterv!11}\res{e1sfx.qwen3p8-27b-elow.a3f-quant.nli.nat.dR.mean} & \cellcolor{cInterv!14}\res{e1sfx.qwen3p8-27b-elow.a3f-conn.nli.nat.dR.mean}  \\
    Qwen3.8-27B (medium effort) & \cellcolor{cOurs!3}\res{e1q.qwen3p8-27b-emed.a3-quant.inf.nat.dR.mean}{\scriptsize\,\textcolor{gray}{[\res{e1q.qwen3p8-27b-emed.a3-quant.inf.nat.dR.lo}, \res{e1q.qwen3p8-27b-emed.a3-quant.inf.nat.dR.hi}]}}$^{\circ}$ & \cellcolor{cInterv!16}\res{e1q.qwen3p8-27b-emed.a3-conn.inf.nat.dR.mean}{\scriptsize\,\textcolor{gray}{[\res{e1q.qwen3p8-27b-emed.a3-conn.inf.nat.dR.lo}, \res{e1q.qwen3p8-27b-emed.a3-conn.inf.nat.dR.hi}]}} & \cellcolor{cOurs!17}\res{e1q.qwen3p8-27b-emed.a3-modal.inf.nat.dR.mean}{\scriptsize\,\textcolor{gray}{[\res{e1q.qwen3p8-27b-emed.a3-modal.inf.nat.dR.lo}, \res{e1q.qwen3p8-27b-emed.a3-modal.inf.nat.dR.hi}]}} & \cellcolor{cOurs!15}\res{e1q.qwen3p8-27b-emed.a2.inf.nat.dR.mean}{\scriptsize\,\textcolor{gray}{[\res{e1q.qwen3p8-27b-emed.a2.inf.nat.dR.lo}, \res{e1q.qwen3p8-27b-emed.a2.inf.nat.dR.hi}]}} & \cellcolor{cOurs!14}\res{e1q.qwen3p8-27b-emed.b2.yn.nat.dacc.mean}{\scriptsize\,\textcolor{gray}{[\res{e1q.qwen3p8-27b-emed.b2.yn.nat.dacc.lo}, \res{e1q.qwen3p8-27b-emed.b2.yn.nat.dacc.hi}]}} & \cellcolor{cInterv!8}\res{e1sfx.qwen3p8-27b-emed.a3f-quant.nli.nat.dR.mean} & \cellcolor{cInterv!9}\res{e1sfx.qwen3p8-27b-emed.a3f-conn.nli.nat.dR.mean}  \\
    \addlinespace[0.5mm]\midrule
    \multicolumn{8}{@{}l}{\textsc{Other open families}} \\
    MiMo-7B-SFT & \cellcolor{cInterv!24}\textbf{\res{e1b.mimo-7b-sft.a3-quant.inf.nat.dR.mean}{\scriptsize\,\textcolor{gray}{[\res{e1b.mimo-7b-sft.a3-quant.inf.nat.dR.lo}, \res{e1b.mimo-7b-sft.a3-quant.inf.nat.dR.hi}]}}} & \cellcolor{cInterv!22}\res{e1b.mimo-7b-sft.a3-conn.inf.nat.dR.mean}{\scriptsize\,\textcolor{gray}{[\res{e1b.mimo-7b-sft.a3-conn.inf.nat.dR.lo}, \res{e1b.mimo-7b-sft.a3-conn.inf.nat.dR.hi}]}} & \cellcolor{cInterv!19}\res{e1b.mimo-7b-sft.a3-modal.inf.nat.dR.mean}{\scriptsize\,\textcolor{gray}{[\res{e1b.mimo-7b-sft.a3-modal.inf.nat.dR.lo}, \res{e1b.mimo-7b-sft.a3-modal.inf.nat.dR.hi}]}} & \cellcolor{cInterv!19}\res{e1b.mimo-7b-sft.a2.inf.nat.dR.mean}{\scriptsize\,\textcolor{gray}{[\res{e1b.mimo-7b-sft.a2.inf.nat.dR.lo}, \res{e1b.mimo-7b-sft.a2.inf.nat.dR.hi}]}} & \cellcolor{cOurs!18}\res{e1b.mimo-7b-sft.b2.yn.nat.dacc.mean}{\scriptsize\,\textcolor{gray}{[\res{e1b.mimo-7b-sft.b2.yn.nat.dacc.lo}, \res{e1b.mimo-7b-sft.b2.yn.nat.dacc.hi}]}} & \cellcolor{cInterv!20}\res{e1sfx.mimo-7b-sft.a3f-quant.nli.nat.dR.mean} & \cellcolor{cInterv!20}\underline{\res{e1sfx.mimo-7b-sft.a3f-conn.nli.nat.dR.mean}}  \\
    MiMo-7B-RL & \cellcolor{cInterv!22}\res{e1b.mimo-7b-rl.a3-quant.inf.nat.dR.mean}{\scriptsize\,\textcolor{gray}{[\res{e1b.mimo-7b-rl.a3-quant.inf.nat.dR.lo}, \res{e1b.mimo-7b-rl.a3-quant.inf.nat.dR.hi}]}} & \cellcolor{cInterv!22}\res{e1b.mimo-7b-rl.a3-conn.inf.nat.dR.mean}{\scriptsize\,\textcolor{gray}{[\res{e1b.mimo-7b-rl.a3-conn.inf.nat.dR.lo}, \res{e1b.mimo-7b-rl.a3-conn.inf.nat.dR.hi}]}} & \cellcolor{cInterv!6}\res{e1b.mimo-7b-rl.a3-modal.inf.nat.dR.mean}{\scriptsize\,\textcolor{gray}{[\res{e1b.mimo-7b-rl.a3-modal.inf.nat.dR.lo}, \res{e1b.mimo-7b-rl.a3-modal.inf.nat.dR.hi}]}}$^{\circ}$ & \cellcolor{cInterv!14}\res{e1b.mimo-7b-rl.a2.inf.nat.dR.mean}{\scriptsize\,\textcolor{gray}{[\res{e1b.mimo-7b-rl.a2.inf.nat.dR.lo}, \res{e1b.mimo-7b-rl.a2.inf.nat.dR.hi}]}} & \cellcolor{cOurs!20}\textbf{\res{e1b.mimo-7b-rl.b2.yn.nat.dacc.mean}{\scriptsize\,\textcolor{gray}{[\res{e1b.mimo-7b-rl.b2.yn.nat.dacc.lo}, \res{e1b.mimo-7b-rl.b2.yn.nat.dacc.hi}]}}} & \cellcolor{cInterv!19}\res{e1sfx.mimo-7b-rl.a3f-quant.nli.nat.dR.mean} & \cellcolor{cInterv!19}\res{e1sfx.mimo-7b-rl.a3f-conn.nli.nat.dR.mean}  \\
    openPangu-Embedded-7B-V1.1$^\S$ & \cellcolor{cInterv!16}\res{e1pangu.openpangu-embedded-7b-v1p1.a3-quant.inf.nat.dR.mean}{\scriptsize\,\textcolor{gray}{[\res{e1pangu.openpangu-embedded-7b-v1p1.a3-quant.inf.nat.dR.lo}, \res{e1pangu.openpangu-embedded-7b-v1p1.a3-quant.inf.nat.dR.hi}]}}$^{\circ}$ & \cellcolor{cInterv!10}\res{e1pangu.openpangu-embedded-7b-v1p1.a3-conn.inf.nat.dR.mean}{\scriptsize\,\textcolor{gray}{[\res{e1pangu.openpangu-embedded-7b-v1p1.a3-conn.inf.nat.dR.lo}, \res{e1pangu.openpangu-embedded-7b-v1p1.a3-conn.inf.nat.dR.hi}]}}$^{\circ}$ & \cellcolor{cOurs!15}\res{e1pangu.openpangu-embedded-7b-v1p1.a3-modal.inf.nat.dR.mean}{\scriptsize\,\textcolor{gray}{[\res{e1pangu.openpangu-embedded-7b-v1p1.a3-modal.inf.nat.dR.lo}, \res{e1pangu.openpangu-embedded-7b-v1p1.a3-modal.inf.nat.dR.hi}]}} & \cellcolor{cInterv!17}\res{e1pangu.openpangu-embedded-7b-v1p1.a2.inf.nat.dR.mean}{\scriptsize\,\textcolor{gray}{[\res{e1pangu.openpangu-embedded-7b-v1p1.a2.inf.nat.dR.lo}, \res{e1pangu.openpangu-embedded-7b-v1p1.a2.inf.nat.dR.hi}]}} & \cellcolor{cOurs!9}\res{e1pangu.openpangu-embedded-7b-v1p1.b2.yn.nat.dacc.mean}{\scriptsize\,\textcolor{gray}{[\res{e1pangu.openpangu-embedded-7b-v1p1.b2.yn.nat.dacc.lo}, \res{e1pangu.openpangu-embedded-7b-v1p1.b2.yn.nat.dacc.hi}]}}$^{\circ}$ & \cellcolor{cInterv!14}\res{e1pangu.openpangu-embedded-7b-v1p1.a3f-quant.nli.nat.dR.mean} & \cellcolor{cInterv!10}\res{e1pangu.openpangu-embedded-7b-v1p1.a3f-conn.nli.nat.dR.mean}  \\
    openPangu-Embedded-7B (v1.0)$^\S$ & \cellcolor{cOurs!17}\res{e1pangu.openpangu-embedded-7b-model.a3-quant.inf.nat.dR.mean}{\scriptsize\,\textcolor{gray}{[\res{e1pangu.openpangu-embedded-7b-model.a3-quant.inf.nat.dR.lo}, \res{e1pangu.openpangu-embedded-7b-model.a3-quant.inf.nat.dR.hi}]}} & \cellcolor{cInterv!11}\res{e1pangu.openpangu-embedded-7b-model.a3-conn.inf.nat.dR.mean}{\scriptsize\,\textcolor{gray}{[\res{e1pangu.openpangu-embedded-7b-model.a3-conn.inf.nat.dR.lo}, \res{e1pangu.openpangu-embedded-7b-model.a3-conn.inf.nat.dR.hi}]}}$^{\circ}$ & \cellcolor{cOurs!15}\res{e1pangu.openpangu-embedded-7b-model.a3-modal.inf.nat.dR.mean}{\scriptsize\,\textcolor{gray}{[\res{e1pangu.openpangu-embedded-7b-model.a3-modal.inf.nat.dR.lo}, \res{e1pangu.openpangu-embedded-7b-model.a3-modal.inf.nat.dR.hi}]}} & \cellcolor{cOurs!14}\res{e1pangu.openpangu-embedded-7b-model.a2.inf.nat.dR.mean}{\scriptsize\,\textcolor{gray}{[\res{e1pangu.openpangu-embedded-7b-model.a2.inf.nat.dR.lo}, \res{e1pangu.openpangu-embedded-7b-model.a2.inf.nat.dR.hi}]}} & \cellcolor{cOurs!14}\res{e1pangu.openpangu-embedded-7b-model.b2.yn.nat.dacc.mean}{\scriptsize\,\textcolor{gray}{[\res{e1pangu.openpangu-embedded-7b-model.b2.yn.nat.dacc.lo}, \res{e1pangu.openpangu-embedded-7b-model.b2.yn.nat.dacc.hi}]}} & \cellcolor{cInterv!6}\res{e1pangu.openpangu-embedded-7b-model.a3f-quant.nli.nat.dR.mean} & \cellcolor{cInterv!3}\res{e1pangu.openpangu-embedded-7b-model.a3f-conn.nli.nat.dR.mean}  \\
    gpt-oss-20b (high$-$low effort) & \cellcolor{cInterv!20}\res{e3.gpt-oss-20b.a3-quant.inf.high.dR.mean}{\scriptsize\,\textcolor{gray}{[\res{e3.gpt-oss-20b.a3-quant.inf.high.dR.lo}, \res{e3.gpt-oss-20b.a3-quant.inf.high.dR.hi}]}} & \cellcolor{cInterv!19}\res{e3.gpt-oss-20b.a3-conn.inf.high.dR.mean}{\scriptsize\,\textcolor{gray}{[\res{e3.gpt-oss-20b.a3-conn.inf.high.dR.lo}, \res{e3.gpt-oss-20b.a3-conn.inf.high.dR.hi}]}} & \cellcolor{cInterv!18}\res{e3.gpt-oss-20b.a3-modal.inf.high.dR.mean}{\scriptsize\,\textcolor{gray}{[\res{e3.gpt-oss-20b.a3-modal.inf.high.dR.lo}, \res{e3.gpt-oss-20b.a3-modal.inf.high.dR.hi}]}} & \cellcolor{cInterv!4}\res{e3.gpt-oss-20b.a2.inf.high.dR.mean}{\scriptsize\,\textcolor{gray}{[\res{e3.gpt-oss-20b.a2.inf.high.dR.lo}, \res{e3.gpt-oss-20b.a2.inf.high.dR.hi}]}}$^{\circ}$ & \cellcolor{cOurs!10}\res{e3.gpt-oss-20b.b2.yn.high.dacc.mean}{\scriptsize\,\textcolor{gray}{[\res{e3.gpt-oss-20b.b2.yn.high.dacc.lo}, \res{e3.gpt-oss-20b.b2.yn.high.dacc.hi}]}}$^{\circ}$ & \cellcolor{cInterv!7}\res{e1gsf.gpt-oss-20b.a3f-quant.nli.high.dR.mean} & \cellcolor{cInterv!6}\res{e1gsf.gpt-oss-20b.a3f-conn.nli.high.dR.mean}  \\
    \addlinespace[0.5mm]\midrule
    \multicolumn{8}{@{}l}{\textsc{DeepSeek (R1 distills always think: forced-empty block $\rightarrow$ natural; V4: thinking off $\rightarrow$ on)}} \\
    DeepSeek-R1-0528-Qwen3-8B & \cellcolor{cInterv!19}\res{e1r1.r1-0528-qwen3-8b.a3-quant.inf.nat.dR.mean}{\scriptsize\,\textcolor{gray}{[\res{e1r1.r1-0528-qwen3-8b.a3-quant.inf.nat.dR.lo}, \res{e1r1.r1-0528-qwen3-8b.a3-quant.inf.nat.dR.hi}]}} & \cellcolor{cInterv!15}\res{e1r1.r1-0528-qwen3-8b.a3-conn.inf.nat.dR.mean}{\scriptsize\,\textcolor{gray}{[\res{e1r1.r1-0528-qwen3-8b.a3-conn.inf.nat.dR.lo}, \res{e1r1.r1-0528-qwen3-8b.a3-conn.inf.nat.dR.hi}]}} & \cellcolor{cInterv!13}\res{e1r1.r1-0528-qwen3-8b.a3-modal.inf.nat.dR.mean}{\scriptsize\,\textcolor{gray}{[\res{e1r1.r1-0528-qwen3-8b.a3-modal.inf.nat.dR.lo}, \res{e1r1.r1-0528-qwen3-8b.a3-modal.inf.nat.dR.hi}]}}$^{\circ}$ & \cellcolor{cInterv!18}\res{e1r1.r1-0528-qwen3-8b.a2.inf.nat.dR.mean}{\scriptsize\,\textcolor{gray}{[\res{e1r1.r1-0528-qwen3-8b.a2.inf.nat.dR.lo}, \res{e1r1.r1-0528-qwen3-8b.a2.inf.nat.dR.hi}]}} & \cellcolor{cOurs!13}\res{e1r1.r1-0528-qwen3-8b.b2.yn.nat.dacc.mean}{\scriptsize\,\textcolor{gray}{[\res{e1r1.r1-0528-qwen3-8b.b2.yn.nat.dacc.lo}, \res{e1r1.r1-0528-qwen3-8b.b2.yn.nat.dacc.hi}]}}$^{\circ}$ & \cellcolor{cInterv!18}\res{e1r1.r1-0528-qwen3-8b.a3f-quant.nli.nat.dR.mean} & \cellcolor{cInterv!13}\res{e1r1.r1-0528-qwen3-8b.a3f-conn.nli.nat.dR.mean}  \\
    R1-Distill-Llama-8B & \cellcolor{cInterv!23}\res{e1p.r1distill-llama-8b.a3-quant.inf.nat.dR.mean}{\scriptsize\,\textcolor{gray}{[\res{e1p.r1distill-llama-8b.a3-quant.inf.nat.dR.lo}, \res{e1p.r1distill-llama-8b.a3-quant.inf.nat.dR.hi}]}} & \cellcolor{cInterv!23}\res{e1p.r1distill-llama-8b.a3-conn.inf.nat.dR.mean}{\scriptsize\,\textcolor{gray}{[\res{e1p.r1distill-llama-8b.a3-conn.inf.nat.dR.lo}, \res{e1p.r1distill-llama-8b.a3-conn.inf.nat.dR.hi}]}} & \cellcolor{cInterv!21}\res{e1p.r1distill-llama-8b.a3-modal.inf.nat.dR.mean}{\scriptsize\,\textcolor{gray}{[\res{e1p.r1distill-llama-8b.a3-modal.inf.nat.dR.lo}, \res{e1p.r1distill-llama-8b.a3-modal.inf.nat.dR.hi}]}} & \cellcolor{cInterv!21}\res{e1p.r1distill-llama-8b.a2.inf.nat.dR.mean}{\scriptsize\,\textcolor{gray}{[\res{e1p.r1distill-llama-8b.a2.inf.nat.dR.lo}, \res{e1p.r1distill-llama-8b.a2.inf.nat.dR.hi}]}} & \cellcolor{cOurs!20}\underline{\res{e1p.r1distill-llama-8b.b2.yn.nat.dacc.mean}{\scriptsize\,\textcolor{gray}{[\res{e1p.r1distill-llama-8b.b2.yn.nat.dacc.lo}, \res{e1p.r1distill-llama-8b.b2.yn.nat.dacc.hi}]}}} & \cellcolor{cInterv!20}\res{e1sfx.r1distill-llama-8b.a3f-quant.nli.nat.dR.mean} & \cellcolor{cInterv!21}\textbf{\res{e1sfx.r1distill-llama-8b.a3f-conn.nli.nat.dR.mean}}  \\
    \addlinespace[0.5mm]\midrule
    \multicolumn{8}{@{}l}{\textsc{Closed models (highest$-$lowest dose)}} \\
    GPT-5 & \cellcolor{cInterv!18}\res{e1f.gpt-5.a3-quant.inf.eff-high.dR.mean}{\scriptsize\,\textcolor{gray}{[\res{e1f.gpt-5.a3-quant.inf.eff-high.dR.lo}, \res{e1f.gpt-5.a3-quant.inf.eff-high.dR.hi}]}} & \cellcolor{cInterv!18}\res{e1f.gpt-5.a3-conn.inf.eff-high.dR.mean}{\scriptsize\,\textcolor{gray}{[\res{e1f.gpt-5.a3-conn.inf.eff-high.dR.lo}, \res{e1f.gpt-5.a3-conn.inf.eff-high.dR.hi}]}} & \cellcolor{cInterv!16}\res{e1f.gpt-5.a3-modal.inf.eff-high.dR.mean}{\scriptsize\,\textcolor{gray}{[\res{e1f.gpt-5.a3-modal.inf.eff-high.dR.lo}, \res{e1f.gpt-5.a3-modal.inf.eff-high.dR.hi}]}} & \cellcolor{cOurs!14}\res{e1f.gpt-5.a2.inf.eff-high.dR.mean}{\scriptsize\,\textcolor{gray}{[\res{e1f.gpt-5.a2.inf.eff-high.dR.lo}, \res{e1f.gpt-5.a2.inf.eff-high.dR.hi}]}} & \cellcolor{cOurs!13}\res{e1f.gpt-5.b2.yn.eff-high.dacc.mean}{\scriptsize\,\textcolor{gray}{[\res{e1f.gpt-5.b2.yn.eff-high.dacc.lo}, \res{e1f.gpt-5.b2.yn.eff-high.dacc.hi}]}}$^{\circ}$ & \cellcolor{cInterv!13}\res{e1f.gpt-5.a3-quant.nli3.eff-high.dRx.mean} & \cellcolor{cInterv!16}\res{e1f.gpt-5.a3-conn.nli3.eff-high.dRx.mean}  \\
    Claude Sonnet 4.6$^\dagger$ & \cellcolor{cOurs!19}\res{e1f.claude-sonnet-4p6.a3-quant.inf.bud16k.dR.mean}{\scriptsize\,\textcolor{gray}{[\res{e1f.claude-sonnet-4p6.a3-quant.inf.bud16k.dR.lo}, \res{e1f.claude-sonnet-4p6.a3-quant.inf.bud16k.dR.hi}]}} & \cellcolor{cOurs!15}\res{e1f.claude-sonnet-4p6.a3-conn.inf.bud16k.dR.mean}{\scriptsize\,\textcolor{gray}{[\res{e1f.claude-sonnet-4p6.a3-conn.inf.bud16k.dR.lo}, \res{e1f.claude-sonnet-4p6.a3-conn.inf.bud16k.dR.hi}]}}$^{\circ}$ & \cellcolor{cOurs!20}\res{e1f.claude-sonnet-4p6.a3-modal.inf.bud16k.dR.mean}{\scriptsize\,\textcolor{gray}{[\res{e1f.claude-sonnet-4p6.a3-modal.inf.bud16k.dR.lo}, \res{e1f.claude-sonnet-4p6.a3-modal.inf.bud16k.dR.hi}]}} & \cellcolor{cOurs!18}\res{e1f.claude-sonnet-4p6.a2.inf.bud16k.dR.mean}{\scriptsize\,\textcolor{gray}{[\res{e1f.claude-sonnet-4p6.a2.inf.bud16k.dR.lo}, \res{e1f.claude-sonnet-4p6.a2.inf.bud16k.dR.hi}]}} & \cellcolor{cOurs!9}\res{e1f.claude-sonnet-4p6.b2.yn.bud16k.dacc.mean}{\scriptsize\,\textcolor{gray}{[\res{e1f.claude-sonnet-4p6.b2.yn.bud16k.dacc.lo}, \res{e1f.claude-sonnet-4p6.b2.yn.bud16k.dacc.hi}]}}$^{\circ}$ & \cellcolor{cInterv!12}\res{e1f.claude-sonnet-4p6.a3-quant.nli3.bud16k.dRx.mean} & \cellcolor{cInterv!17}\res{e1f.claude-sonnet-4p6.a3-conn.nli3.bud16k.dRx.mean}  \\
    \bottomrule
  \end{tabular}}
  \par\smallskip{\footnotesize Baseline: thinking off, or a forced-empty block where no switch exists; $^\dagger$Reverses in the three-way choice. $^\S$Vendor system prompt; fast mode via ` /no\_think'. Three-way: no answer polarity, means only (\Cref{app:frontier}). All models: \Cref{tab:main,tab:frontier}.\par}
  \caption{\textbf{Thinking reduces the pragmatic reading of logical words across families.} Change with thinking (pp) [95\% CI]; \textbf{bold}/\underline{underlined}: largest/second absolute change per column; $^{\\circ}$: CI includes 0.}
  \vspace{-12pt}
  \label{tab:results}
\end{table*}

Thinking reduces the pragmatic reading of \form{some}, \form{or}, or both in most models, whereas other inference types behave differently. \Cref{tab:results} reports the major open families (see also \Cref{fig:major}), while \Cref{fig:hero} and \Cref{fig:hero-heatmap} detail the item families. Each effect compares natural thinking with thinking off, or with a forced-empty thinking block for models without a switch; for gpt-oss and closed models it compares the highest with the lowest dose.
Natural thinking changes the pragmatic reading of \form{some} in Qwen3-8B by \res{e1.qwen3-8b.a3-quant.inf.nat.dR.mean}\,pp (95\% CI \res{e1.qwen3-8b.a3-quant.inf.nat.dR.lo} to \res{e1.qwen3-8b.a3-quant.inf.nat.dR.hi}) and that of \form{or} by \res{e1.qwen3-8b.a3-conn.inf.nat.dR.mean}\,pp; in Qwen3-32B the changes are \res{e1s.qwen3-32b.a3-quant.inf.nat.dR.mean} and \res{e1s.qwen3-32b.a3-conn.inf.nat.dR.mean}\,pp. High reasoning effort does the same in gpt-oss-20b (\res{e3.gpt-oss-20b.a3-quant.inf.high.dR.mean} and \res{e3.gpt-oss-20b.a3-conn.inf.high.dR.mean}\,pp against low effort), and in OLMo-3-32B-Think the \form{some} reading changes by \res{e1s.olmo3-32b-think.a3-quant.inf.nat.dR.mean}\,pp. It holds in DeepSeek's distilled thinkers too, which always think: against a forced-empty thinking block, natural thinking changes the \form{some} reading of R1-Distill-Llama-8B by \res{e1p.r1distill-llama-8b.a3-quant.inf.nat.dR.mean}\,pp and of R1-Distill-Llama-70B by \res{e1p.r1distill-llama-70b.a3-quant.inf.nat.dR.mean}\,pp, and the newest distilled model, DeepSeek-R1-0528-Qwen3-8B, by \res{e1r1.r1-0528-qwen3-8b.a3-quant.inf.nat.dR.mean}\,pp. 

Across \res{derived.main.tested} open reasoning models, the shift on \form{some} or \form{or} is significant (95\% CI excluding 0, uncorrected) and survives the pre-declared format and polarity checks in \res{derived.main.counted}; \res{derived.main.rise} models move the other way, \res{derived.main.formatdep} shifts only in the Yes/No format, and \res{derived.main.no_sig_change} show no significant change (\Cref{tab:main}). 

The count applies a counting rule declared before the full-roster runs; the outcome of the pre-registered aggregate test is reported separately in \Cref{app:prereg}. Models outside the count are discussed in \Cref{app:outside}.
On the held-out half of the full IMPPRES sets (\res{e1sf.qwen3-8b.a3f-quant.inf.nat.dR.n} items per word), the changes for Qwen3-8B are \res{e1sf.qwen3-8b.a3f-quant.inf.nat.dR.mean}\,pp for \form{some} and \res{e1sf.qwen3-8b.a3f-conn.inf.nat.dR.mean}\,pp for \form{or}, which replicates the core result.
Among closed models, high effort changes GPT-5's reading of \form{some} by \res{e1f.gpt-5.a3-quant.inf.eff-high.dR.mean}\,pp and of \form{or} by \res{e1f.gpt-5.a3-conn.inf.eff-high.dR.mean}\,pp against minimal effort, and the shift holds in a second answer format (\Cref{app:frontier}). DeepSeek-V4 goes the other way: switching thinking on raises the pragmatic reading of \form{some} by \res{e1f.deepseek-v4-pro.a3-quant.inf.think.dR.mean}\,pp in V4-Pro and \res{e1f.deepseek-v4-flash.a3-quant.inf.think.dR.mean}\,pp in V4-Flash, and that of lexical scales by \res{e1f.deepseek-v4-pro.a2.inf.think.dR.mean}\,pp in V4-Pro (Yes/No format).

Modals the third logical-word trigger shift less consistently: on the full held-out sets, the change is \res{e1sf.qwen3-32b.a3f-modal.inf.nat.dR.mean}\,pp in Qwen3-32B but \res{e1sf.qwen3-14b.a3f-modal.inf.nat.dR.mean}\,pp in Qwen3-14B, and EXAONE-Deep and OLMo-3-32B-Think read modals \emph{more} pragmatically with thinking (\res{e1sfx.exaone-deep-7p8b.a3f-modal.inf.nat.dR.mean} and \res{e1s.olmo3-32b-think.a3-modal.inf.nat.dR.mean}\,pp). The robust effect is on \form{some} and \form{or}.

On the full held-out sets, run for the models listed in \Cref{tab:fullsets}, \res{derived.holm.full.survive} of \res{derived.holm.full.tests} \form{some} and \form{or} shifts survive Holm correction within their model family. Thinking also separates the two classes: the interaction between the log thinking budget and the class of phenomenon is negative in Qwen3-4B, 8B and 14B (8B: \res{derived.glmm.qwen3-8b.coef}, 95\% CI \res{derived.glmm.qwen3-8b.lo} to \res{derived.glmm.qwen3-8b.hi}, $p$ \res{derived.glmm.qwen3-8b.p}), and has the same sign, though not significant, at 32B.

The rest of the battery diverges. From 8B up, Qwen3 interprets lexical scales slightly \emph{more} pragmatically with thinking (\res{e1.qwen3-8b.a2.inf.nat.dR.mean}\,pp at 8B, \res{e1s.qwen3-32b.a2.inf.nat.dR.mean}\,pp at 32B). Presupposition accuracy rises by \res{e1.qwen3-8b.a4.nli.nat.dacc.mean}\,pp in Qwen3-8B. Context-driven implicatures improve: accuracy on figurative language (Fig-QA) rises by \res{e1xt.qwen3-8b.b3.mc.nat.dacc.mean}\,pp and on six PUB tasks by \res{e1xt.qwen3-8b.b4.mc.nat.dacc.mean}\,pp, and both gains are significant at every Qwen3 size from 4B to 32B; on two further sets, accuracy rises on indirect answers in Circa \citep{louis-etal-2020-id} by \res{e1ds.qwen3-8b.circa.mc.nat.dacc.mean}\,pp and on naturally occurring presuppositions in NOPE \citep{parrish-etal-2021-nope} by \res{e1ds.qwen3-8b.nope.nli.nat.dacc.mean}\,pp in Qwen3-8B, and no model among the 14 we tested on these two sets gets significantly worse (\Cref{tab:breadth}); on LUDWIG indirect answers, accuracy changes by \res{e1.qwen3-8b.b2.yn.nat.dacc.mean}\,pp. Below 8B, Qwen3 loses lexical scales as well (\res{e1.qwen3-4b.a2.inf.nat.dR.mean}\,pp at 4B), so the lexical effect changes sign between Qwen3-4B and Qwen3-8B. Equivalence within $\pm3$\,pp on lexical scales holds in only \res{derived.tost.a2.equivalent} of \res{derived.tost.a2.tested} models. Thinking reduces enrichment of \form{some}, \form{or} or both in \res{derived.main.counted} of \res{derived.main.tested} open models; in Qwen3 from 8B up it increases lexical enrichment and improves context-driven inference.

\begin{figure*}[t]
  \centering
  \IfFileExists{figures/fig3_trace.pdf}{\includegraphics[width=\textwidth]{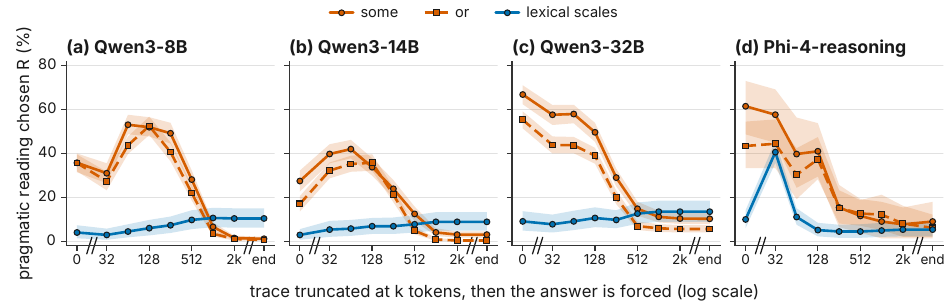}}{\figplaceholder{\textwidth}{4.4cm}}
  \caption{\textbf{The enriched reading is endorsed early in the trace and abandoned later.} Pragmatic reading $\rate$ after the first $t$ tokens of the model's own trace, with 95\% bands; panel (d) and the lexical-scale lines come from an earlier, smaller run of the probe.}
  \label{fig:trace}
\end{figure*}

\subsection{Sensitivity Analysis: The Shift Survives Stopping, Baseline, Format and Item Choices}
\label{sec:find:control}

\begin{figure}[t]
  \centering
  \includegraphics[width=\columnwidth]{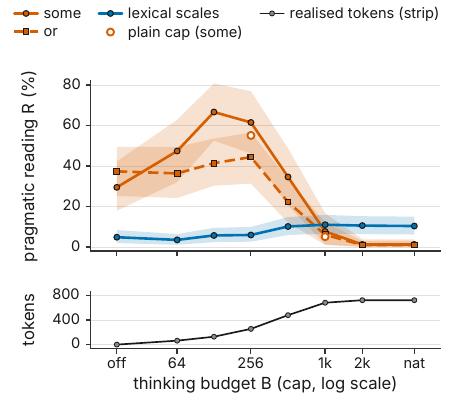}
  \caption{\textbf{Pragmatic reading rate for Qwen3-8B vs.\ thinking budget.} Capped with stop phrase (hollow: plain cap). Lower strip: realized tokens. Bands: 95\% cluster-bootstrap CIs.}
  \label{fig:budget}
  \vspace{-12pt}
\end{figure}

Full-length thinking lowers the pragmatic reading of \form{some} and \form{or} under every analysis choice (\Cref{tab:sens}): with a forced-empty block as the baseline, under either stopping procedure, on the held-out full sets and in a three-way choice without Yes/No labels, in both Qwen3-8B and Qwen3-32B. Along the budget (\Cref{fig:budget}) the reading of \form{some} first rises, by \res{e1.qwen3-8b.a3-quant.inf.c0128.dR.mean}\,pp at 128 tokens, and falls below its no-thinking level only beyond 512 tokens (\res{e1.qwen3-8b.a3-quant.inf.c1024.dR.mean}\,pp at 1024): mirroring the trace result (\Cref{sec:find:trace}).
\begin{table}[ht]
  \centering\small
  \setlength{\tabcolsep}{4pt}
  \begin{tabular}{@{}lrr@{}}
    \toprule
    Analysis choice & \form{some} & \form{or} \\
    \midrule
    \multicolumn{3}{@{}l}{\textsc{Qwen3-8B}} \\
    Reference: core, Yes/No, off $\rightarrow$ natural & \cellcolor{cInterv!20}\res{e1.qwen3-8b.a3-quant.inf.nat.dR.mean} & \cellcolor{cInterv!21}\res{e1.qwen3-8b.a3-conn.inf.nat.dR.mean}  \\
    Baseline: forced-empty thinking block & \res{e1.qwen3-8b.a3-quant.inf.nat.dRb0.mean} & \res{e1.qwen3-8b.a3-conn.inf.nat.dRb0.mean}  \\
    Budget 128, stop phrase & \cellcolor{cOurs!21}\res{e1.qwen3-8b.a3-quant.inf.c0128.dR.mean} & \cellcolor{cOurs!11}\res{e1.qwen3-8b.a3-conn.inf.c0128.dR.mean}  \\
    Budget 512, stop phrase & \cellcolor{cOurs!12}\res{e1.qwen3-8b.a3-quant.inf.c0512.dR.mean} & \cellcolor{cInterv!17}\res{e1.qwen3-8b.a3-conn.inf.c0512.dR.mean}  \\
    Budget 1024, stop phrase & \cellcolor{cInterv!19}\res{e1.qwen3-8b.a3-quant.inf.c1024.dR.mean} & \cellcolor{cInterv!20}\res{e1.qwen3-8b.a3-conn.inf.c1024.dR.mean}  \\
    Budget 1024, plain cap & \cellcolor{cInterv!19}\res{e1.qwen3-8b.a3-quant.inf.p1024.dR.mean} & \cellcolor{cInterv!21}\res{e1.qwen3-8b.a3-conn.inf.p1024.dR.mean}  \\
    Items: held-out full set & \cellcolor{cInterv!21}\res{e1sf.qwen3-8b.a3f-quant.inf.nat.dR.mean} & \cellcolor{cInterv!21}\res{e1sf.qwen3-8b.a3f-conn.inf.nat.dR.mean}  \\
    Format: three-way choice & \cellcolor{cInterv!21}\res{e1sf.qwen3-8b.a3f-quant.nli.nat.dR.mean} & \cellcolor{cInterv!16}\res{e1sf.qwen3-8b.a3f-conn.nli.nat.dR.mean}  \\
    \addlinespace[0.5mm]\midrule
    \multicolumn{3}{@{}l}{\textsc{Qwen3-32B}} \\
    Reference: core, Yes/No, off $\rightarrow$ natural & \cellcolor{cInterv!23}\res{e1s.qwen3-32b.a3-quant.inf.nat.dR.mean} & \cellcolor{cInterv!22}\res{e1s.qwen3-32b.a3-conn.inf.nat.dR.mean}  \\
    Items: held-out full set & \cellcolor{cInterv!23}\res{e1sf.qwen3-32b.a3f-quant.inf.nat.dR.mean} & \cellcolor{cInterv!23}\res{e1sf.qwen3-32b.a3f-conn.inf.nat.dR.mean}  \\
    Format: three-way choice & \cellcolor{cInterv!21}\res{e1sf.qwen3-32b.a3f-quant.nli.nat.dR.mean} & \cellcolor{cInterv!18}\res{e1sf.qwen3-32b.a3f-conn.nli.nat.dR.mean}  \\
    \bottomrule
  \end{tabular}
  \caption{\textbf{Full thinking lowers the reading under every analysis choice.} Change in the pragmatic reading with thinking (pp); short budgets raise it before it falls.}
  \label{tab:sens}
\end{table}

Four further checks, on how thinking stops, the baseline, answer bias and speaker knowledge, are in \Cref{app:checks}. The shift persists under both stopping procedures and in a three-way choice without Yes/No labels. The three-way prompt asks whether the implicature follows from the sentence, so it removes answer polarity but also moves the question from speaker belief to inference; we use it as a polarity check rather than a framing-matched replication (prompts in \Cref{app:prompts}).

\begin{table}[t]
  \centering
  \small
  \setlength{\tabcolsep}{3pt}
  \resizebox{\columnwidth}{!}{%
  \begin{tabular}{@{}l cccc@{}}
    \toprule
    & \multicolumn{4}{c}{\textbf{Default reading (\%)}} \\
    \cmidrule(l){2-5}
    \textbf{Stage} & \form{some} & \form{or} & lexical & LUDWIG \\
    \midrule
    \multicolumn{5}{@{}l}{\textit{OLMo-3-Think (native thinking; cost = NAT $-$ forced-empty)}} \\
    base & \cellcolor{cCmpA!12}\cellpm{e2}{olmo3-7b-base.a3-quant.inf.lp4.R} & \cellcolor{cCmpA!12}\cellpm{e2}{olmo3-7b-base.a3-conn.inf.lp4.R} & \cellcolor{cCmpA!7}\cellpm{e2}{olmo3-7b-base.a2.inf.lp4.R} & \cellcolor{cCmpA!11}\cellpm{e2}{olmo3-7b-base.b2.yn.lp4.acc} \\
    Think-SFT & \cellcolor{cCmpA!7}\cellpm{e2}{olmo3-think-sft.a3-quant.inf.lp4.R} & \cellcolor{cCmpA!9}\cellpm{e2}{olmo3-think-sft.a3-conn.inf.lp4.R} & \cellcolor{cCmpA!3}\cellpm{e2}{olmo3-think-sft.a2.inf.lp4.R} & \cellcolor{cCmpA!12}\cellpm{e2}{olmo3-think-sft.b2.yn.lp4.acc} \\
    Think-DPO & \cellcolor{cCmpA!9}\cellpm{e2}{olmo3-think-dpo.a3-quant.inf.lp4.R} & \cellcolor{cCmpA!8}\cellpm{e2}{olmo3-think-dpo.a3-conn.inf.lp4.R} & \cellcolor{cCmpA!3}\cellpm{e2}{olmo3-think-dpo.a2.inf.lp4.R} & \cellcolor{cCmpA!12}\cellpm{e2}{olmo3-think-dpo.b2.yn.lp4.acc} \\
    RL step 100 & \cellcolor{cCmpA!8}\cellpm{e2}{olmo3-think-s0100.a3-quant.inf.lp4.R} & \cellcolor{cCmpA!9}\cellpm{e2}{olmo3-think-s0100.a3-conn.inf.lp4.R} & \cellcolor{cCmpA!3}\cellpm{e2}{olmo3-think-s0100.a2.inf.lp4.R} & \cellcolor{cCmpA!12}\cellpm{e2}{olmo3-think-s0100.b2.yn.lp4.acc} \\
    RL step 500 & \cellcolor{cCmpA!7}\cellpm{e2}{olmo3-think-s0500.a3-quant.inf.lp4.R} & \cellcolor{cCmpA!9}\cellpm{e2}{olmo3-think-s0500.a3-conn.inf.lp4.R} & \cellcolor{cCmpA!3}\cellpm{e2}{olmo3-think-s0500.a2.inf.lp4.R} & \cellcolor{cCmpA!12}\cellpm{e2}{olmo3-think-s0500.b2.yn.lp4.acc} \\
    RL step 1375 (final) & \cellcolor{cCmpA!6}\cellpm{e2}{olmo3-think-s1375.a3-quant.inf.lp4.R} & \cellcolor{cCmpA!8}\cellpm{e2}{olmo3-think-s1375.a3-conn.inf.lp4.R} & \cellcolor{cCmpA!3}\cellpm{e2}{olmo3-think-s1375.a2.inf.lp4.R} & \cellcolor{cCmpA!12}\cellpm{e2}{olmo3-think-s1375.b2.yn.lp4.acc} \\
    \midrule
    \multicolumn{5}{@{}l}{\textit{OLMo-3 RL-Zero from the base (no thinking SFT)}} \\
    Math, step 100 & \cellcolor{cCmpA!12}\cellpm{e2}{olmo3-rlzero-math-s0100.a3-quant.inf.lp4.R} & \cellcolor{cCmpA!12}\cellpm{e2}{olmo3-rlzero-math-s0100.a3-conn.inf.lp4.R} & \cellcolor{cCmpA!7}\cellpm{e2}{olmo3-rlzero-math-s0100.a2.inf.lp4.R} & \cellcolor{cCmpA!11}\cellpm{e2}{olmo3-rlzero-math-s0100.b2.yn.lp4.acc} \\
    Math, final & \cellcolor{cCmpA!12}\cellpm{e2}{olmo3-rlzero-math-final.a3-quant.inf.lp4.R} & \cellcolor{cCmpA!12}\cellpm{e2}{olmo3-rlzero-math-final.a3-conn.inf.lp4.R} & \cellcolor{cCmpA!7}\cellpm{e2}{olmo3-rlzero-math-final.a2.inf.lp4.R} & \cellcolor{cCmpA!11}\cellpm{e2}{olmo3-rlzero-math-final.b2.yn.lp4.acc} \\
    IF, final & \cellcolor{cCmpA!14}\cellpm{e2}{olmo3-rlzero-if-final.a3-quant.inf.lp4.R} & \cellcolor{cCmpA!13}\cellpm{e2}{olmo3-rlzero-if-final.a3-conn.inf.lp4.R} & \cellcolor{cCmpA!7}\cellpm{e2}{olmo3-rlzero-if-final.a2.inf.lp4.R} & \cellcolor{cCmpA!11}\cellpm{e2}{olmo3-rlzero-if-final.b2.yn.lp4.acc} \\
    General, final & \cellcolor{cCmpA!12}\cellpm{e2}{olmo3-rlzero-general-final.a3-quant.inf.lp4.R} & \cellcolor{cCmpA!12}\cellpm{e2}{olmo3-rlzero-general-final.a3-conn.inf.lp4.R} & \cellcolor{cCmpA!8}\cellpm{e2}{olmo3-rlzero-general-final.a2.inf.lp4.R} & \cellcolor{cCmpA!11}\cellpm{e2}{olmo3-rlzero-general-final.b2.yn.lp4.acc} \\
    Mix, final & \cellcolor{cCmpA!14}\cellpm{e2}{olmo3-rlzero-mix-final.a3-quant.inf.lp4.R} & \cellcolor{cCmpA!13}\cellpm{e2}{olmo3-rlzero-mix-final.a3-conn.inf.lp4.R} & \cellcolor{cCmpA!9}\cellpm{e2}{olmo3-rlzero-mix-final.a2.inf.lp4.R} & \cellcolor{cCmpA!11}\cellpm{e2}{olmo3-rlzero-mix-final.b2.yn.lp4.acc} \\
    \midrule
    \multicolumn{5}{@{}l}{\textit{OLMo-2-7B (cost = prompted CoT $-$ none)}} \\
    base & \cellcolor{cCmpA!8}\cellpm{e2}{olmo2-7b-base.a3-quant.inf.lp4.R} & \cellcolor{cCmpA!6}\cellpm{e2}{olmo2-7b-base.a3-conn.inf.lp4.R} & \cellcolor{cCmpA!5}\cellpm{e2}{olmo2-7b-base.a2.inf.lp4.R} & \cellcolor{cCmpA!11}\cellpm{e2}{olmo2-7b-base.b2.yn.lp4.acc} \\
    SFT & \cellcolor{cCmpA!7}\cellpm{e2}{olmo2-7b-sft.a3-quant.inf.lp4.R} & \cellcolor{cCmpA!6}\cellpm{e2}{olmo2-7b-sft.a3-conn.inf.lp4.R} & \cellcolor{cCmpA!6}\cellpm{e2}{olmo2-7b-sft.a2.inf.lp4.R} & \cellcolor{cCmpA!12}\cellpm{e2}{olmo2-7b-sft.b2.yn.lp4.acc} \\
    DPO & \cellcolor{cCmpA!7}\cellpm{e2}{olmo2-7b-dpo.a3-quant.inf.lp4.R} & \cellcolor{cCmpA!4}\cellpm{e2}{olmo2-7b-dpo.a3-conn.inf.lp4.R} & \cellcolor{cCmpA!6}\cellpm{e2}{olmo2-7b-dpo.a2.inf.lp4.R} & \cellcolor{cCmpA!11}\cellpm{e2}{olmo2-7b-dpo.b2.yn.lp4.acc} \\
    \rlvr (Instruct) & \cellcolor{cCmpA!5}\cellpm{e2}{olmo2-7b-instruct.a3-quant.inf.lp4.R} & \cellcolor{cCmpA!4}\cellpm{e2}{olmo2-7b-instruct.a3-conn.inf.lp4.R} & \cellcolor{cCmpA!6}\cellpm{e2}{olmo2-7b-instruct.a2.inf.lp4.R} & \cellcolor{cCmpA!11}\cellpm{e2}{olmo2-7b-instruct.b2.yn.lp4.acc} \\
    \midrule
    \multicolumn{5}{@{}l}{\textit{Tülu-3-8B}} \\
    SFT & \cellcolor{cCmpA!8}\cellpm{e2}{tulu3-8b-sft.a3-quant.inf.lp4.R} & \cellcolor{cCmpA!3}\cellpm{e2}{tulu3-8b-sft.a3-conn.inf.lp4.R} & \cellcolor{cCmpA!3}\cellpm{e2}{tulu3-8b-sft.a2.inf.lp4.R} & \cellcolor{cCmpA!12}\cellpm{e2}{tulu3-8b-sft.b2.yn.lp4.acc} \\
    DPO & \cellcolor{cCmpA!4}\cellpm{e2}{tulu3-8b-dpo.a3-quant.inf.lp4.R} & \cellcolor{cCmpA!2}\cellpm{e2}{tulu3-8b-dpo.a3-conn.inf.lp4.R} & \cellcolor{cCmpA!3}\cellpm{e2}{tulu3-8b-dpo.a2.inf.lp4.R} & \cellcolor{cCmpA!12}\cellpm{e2}{tulu3-8b-dpo.b2.yn.lp4.acc} \\
    \rlvr & \cellcolor{cCmpA!4}\cellpm{e2}{tulu3-8b-rlvr.a3-quant.inf.lp4.R} & \cellcolor{cCmpA!2}\cellpm{e2}{tulu3-8b-rlvr.a3-conn.inf.lp4.R} & \cellcolor{cCmpA!3}\cellpm{e2}{tulu3-8b-rlvr.a2.inf.lp4.R} & \cellcolor{cCmpA!12}\cellpm{e2}{tulu3-8b-rlvr.b2.yn.lp4.acc} \\
    \rlvr (3.1) & \cellcolor{cCmpA!4}\cellpm{e2}{tulu31-8b.a3-quant.inf.lp4.R} & \cellcolor{cCmpA!2}\cellpm{e2}{tulu31-8b.a3-conn.inf.lp4.R} & \cellcolor{cCmpA!3}\cellpm{e2}{tulu31-8b.a2.inf.lp4.R} & \cellcolor{cCmpA!12}\cellpm{e2}{tulu31-8b.b2.yn.lp4.acc} \\
    \midrule
    \multicolumn{5}{@{}l}{\textit{Qwen2.5-7B: distillation and zero-RL}} \\
    base & \cellcolor{cCmpA!7}\cellpm{e2}{qwen25-7b.a3-quant.inf.lp4.R} & \cellcolor{cCmpA!3}\cellpm{e2}{qwen25-7b.a3-conn.inf.lp4.R} & \cellcolor{cCmpA!6}\cellpm{e2}{qwen25-7b.a2.inf.lp4.R} & \cellcolor{cCmpA!11}\cellpm{e2}{qwen25-7b.b2.yn.lp4.acc} \\
    R1-Distill (SFT on traces) & \cellcolor{cCmpA!13}\cellpm{e2}{r1distill-qwen-7b.a3-quant.inf.lp4.R} & \cellcolor{cCmpA!11}\cellpm{e2}{r1distill-qwen-7b.a3-conn.inf.lp4.R} & \cellcolor{cCmpA!8}\cellpm{e2}{r1distill-qwen-7b.a2.inf.lp4.R} & \cellcolor{cCmpA!12}\cellpm{e2}{r1distill-qwen-7b.b2.yn.lp4.acc} \\
    SimpleRL-Zoo (zero-RL) & \cellcolor{cCmpA!8}\cellpm{e2}{simplerl-zoo-7b.a3-quant.inf.lp4.R} & \cellcolor{cCmpA!3}\cellpm{e2}{simplerl-zoo-7b.a3-conn.inf.lp4.R} & \cellcolor{cCmpA!6}\cellpm{e2}{simplerl-zoo-7b.a2.inf.lp4.R} & \cellcolor{cCmpA!11}\cellpm{e2}{simplerl-zoo-7b.b2.yn.lp4.acc} \\
    Open-Reasoner-Zero & \cellcolor{cCmpA!7}\cellpm{e2}{orz-7b.a3-quant.inf.lp4.R} & \cellcolor{cCmpA!3}\cellpm{e2}{orz-7b.a3-conn.inf.lp4.R} & \cellcolor{cCmpA!3}\cellpm{e2}{orz-7b.a2.inf.lp4.R} & \cellcolor{cCmpA!11}\cellpm{e2}{orz-7b.b2.yn.lp4.acc} \\
    \bottomrule
  \end{tabular}}
  \caption{\textbf{The literal default appears with thinking SFT in OLMo-3 and DPO in Tülu-3.} Default reading (\%) under a fixed 4-shot prompt, mean $\pm$ s.d..}
  \label{tab:stages}
  \vspace{-12pt}
\end{table}

\subsection{Early in the Trace the Enriched Reading Is Endorsed; at the End It Is Rejected}
\label{sec:find:trace}

Forcing an answer after each prefix of the natural trace gives intervention-conditioned answer trajectories (\Cref{fig:trace}; \res{e7f.qwen3-8b.a3f-quant.inf.p0128.R.n} held-out \form{some} and \res{e7f.qwen3-8b.a3f-conn.inf.p0128.R.n} \form{or} items per model, three seeds). With an empty trace, the reading matches the no-thinking baseline (\res{e7f.qwen3-8b.a3f-quant.inf.ntforced.R.mean}\% against \res{e1sf.qwen3-8b.a3f-quant.inf.off.R.mean}\% for \form{some} in Qwen3-8B).
In Qwen3-8B the pragmatic reading of \form{some} first \emph{rises}, from \res{e7f.qwen3-8b.a3f-quant.inf.ntforced.R.mean}\% with an empty trace to \res{e7f.qwen3-8b.a3f-quant.inf.p0128.R.mean}\% after 128 tokens (+\res{e7f.qwen3-8b.a3f-quant.inf.p0128.dR.mean}\,pp, CI \res{e7f.qwen3-8b.a3f-quant.inf.p0128.dR.lo} to \res{e7f.qwen3-8b.a3f-quant.inf.p0128.dR.hi}), and then falls to \res{e7f.qwen3-8b.a3f-quant.inf.nat.R.mean}\% by the end of the trace; \form{or} follows the same arc (+\res{e7f.qwen3-8b.a3f-conn.inf.p0128.dR.mean}\,pp after 128 tokens). Qwen3-14B shows it on \form{or} (+\res{e7f.qwen3-14b.a3f-conn.inf.p0128.dR.mean}\,pp after 128 tokens, \res{e7f.qwen3-14b.a3f-conn.inf.nat.R.mean}\% at the end).
Independent generations stopped at a budget of 128 tokens show the same rise (\res{e1.qwen3-8b.a3-quant.inf.c0128.R.mean}\% for \form{some} in Qwen3-8B). Qwen3-32B shows no early rise, only the later fall; most of it comes after the first 128 tokens (\res{e7f.qwen3-32b.a3f-quant.inf.p0128.dR.mean}\,pp at 128 tokens against \res{e7f.qwen3-32b.a3f-quant.inf.nat.dR.mean}\,pp at the end).
Lexical scales take a different path: their pragmatic reading rises slowly through the trace (\res{e7.qwen3-8b.a2.inf.nat.dR.mean}\,pp in Qwen3-8B).

Within single traces, in \res{e7f.qwen3-8b.a3f-quant.inf.traj.a.share}\% of Qwen3-8B \form{some} traces, the answer forced at an early prefix is pragmatic while the final answer is literal, whereas \res{e7f.qwen3-8b.a3f-quant.inf.traj.b.share}\% are literal throughout. 

Across Qwen3 models and words, this pragmatic-to-literal trajectory ranges from \res{e7f.qwen3-14b.a3f-conn.inf.traj.a.share}\% to \res{e7f.qwen3-8b.a3f-conn.inf.traj.a.share}\%. These rates count a pragmatic answer at any of up to eight sampled prefixes.
Beyond Qwen3, Phi-4-reasoning follows the Qwen3-32B profile: no early rise, a fall from \res{e7.phi4-reasoning.a3-quant.inf.ntforced.R.mean}\% to \res{e7.phi4-reasoning.a3-quant.inf.nat.R.mean}\% on \form{some}, and pragmatic-then-literal trajectories in \res{e7.phi4-reasoning.a3-quant.inf.traj.a.share}\% of traces (\Cref{app:results}).
On the newest generation, Qwen3.8-27B cancels \form{or} along its trace, from \res{e7.qwen3p8-27b.a3-conn.inf.ntforced.R.mean}\% with an empty trace to \res{e7.qwen3p8-27b.a3-conn.inf.nat.R.mean}\% at the end (pragmatic-then-literal in \res{e7.qwen3p8-27b.a3-conn.inf.traj.a.share}\% of traces), while its \form{some} reading falls less (\res{e7.qwen3p8-27b.a3-quant.inf.ntforced.R.mean}\% to \res{e7.qwen3p8-27b.a3-quant.inf.nat.R.mean}\%; \Cref{app:results}). Qwen3.5 shows the full arc, including the early rise: Qwen3.5-9B reads \form{some} pragmatically in \res{e7.qwen3p5-9b.a3-quant.inf.ntforced.R.mean}\% of items with an empty trace, \res{e7.qwen3p5-9b.a3-quant.inf.p0032.R.mean}\% after 32 tokens and \res{e7.qwen3p5-9b.a3-quant.inf.nat.R.mean}\% at the end (\Cref{fig:trace-newest}), although for Qwen3.5 many truncated traces yield no valid answer, so this arc rests on the traces that do (\Cref{app:results}).
Among further families, EXAONE-Deep and Nemotron-Nano-9B hold the enriched reading over the first hundred tokens and then drop it (\form{some}: \res{e7.exaone-deep-7p8b.a3-quant.inf.p0128.R.mean}\% after 128 tokens, \res{e7.exaone-deep-7p8b.a3-quant.inf.nat.R.mean}\% at the end; \res{e7.nemotron-nano-9b-v2.a3-quant.inf.p0128.R.mean}\% and \res{e7.nemotron-nano-9b-v2.a3-quant.inf.nat.R.mean}\%), Phi-4-mini-reasoning shows a weaker arc, and Skywork-OR1 drops within the first 32 tokens and then stays flat (\Cref{app:results}).
We call this pattern \emph{entertain, then cancel}: in almost every model we probe the enriched reading is endorsed early in the trace and lower by its end, in most models far below the no-thinking level.

For Qwen3.5-9B, \res{e7f.qwen3p5-9b.a3f-quant.inf.traj_n.dropped_invalid} of its \form{some} traces give an invalid answer at some prefix and are dropped (\Cref{app:trace}), so its trajectory rates describe the valid subset; the trajectory claim rests on the models with near-complete traces.
The enriched reading of a logical word is endorsed early in the trace, in smaller models even more than without thinking, and abandoned later.

\subsection{Post-Training Moves the Default Reading}
\label{sec:origin:stages}

If thinking cancels a default, training for reasoning might also shift the default itself. \Cref{tab:stages} reports the scores across every stage of three lineages without deliberation (using answer log-probabilities), where we measure the within-stage thinking effect wherever applicable.
Post-training for reasoning moves the default reading of \form{some}, at a stage and in a direction that depend on the recipe.

In OLMo-3, the pragmatic default falls from \res{e2.olmo3-7b-base.a3-quant.inf.lp4.R.mean}\% in the base model to \res{e2.olmo3-think-sft.a3-quant.inf.lp4.R.mean}\% after supervised fine-tuning on thinking traces, whereas reinforcement learning on math from the same base (RL-Zero) leaves it at \res{e2.olmo3-rlzero-math-final.a3-quant.inf.lp4.R.mean}\%.
In Tülu-3 the change comes with preference tuning (DPO), from \res{e2.tulu3-8b-sft.a3-quant.inf.lp4.R.mean}\% to \res{e2.tulu3-8b-dpo.a3-quant.inf.lp4.R.mean}\%, and the following reinforcement learning from verifiable rewards (\rlvr) leaves it there.
As the default turns literal, the additional thinking effect shrinks: within OLMo-3-Think the cost of thinking on \form{some} shrinks from \res{e2t.olmo3-think-dpo.a3-quant.inf.nat.dR.mean}\,pp at the preference-tuning stage to \res{e2t.olmo3-think-s1375.a3-quant.inf.nat.dR.mean}\,pp after the final RL step.
Distillation from reasoning traces does not have this effect in every lineage: R1-Distill-Qwen-7B reads \form{some} pragmatically in \res{e2.r1distill-qwen-7b.a3-quant.inf.lp4.R.mean}\% of items, against \res{e2.qwen25-7b.a3-quant.inf.lp4.R.mean}\% for its base model.

The literal default of logical words emerges during post-training, at a recipe-dependent stage (thinking SFT in OLMo-3, preference tuning in Tülu-3), and not in the RL-Zero and distillation comparisons we tested.

\section{Selective Thinking for Scalar Items (Exploratory)}
\label{sec:x3}

If thinking first entertains an implicature and then cancels it, the reading to keep is already in the trace. For questions that ask what a speaker of a \form{some} or \form{or} sentence believes, we take the answer the model gives after the first 128 tokens of its own trace instead of the answer after the completed trace; every other item keeps the natural-thinking answer. Routing reads only the prompt (the question template and the speaker's sentence), never the dataset or label. On routed items the answer comes entirely from the prefix (weight $w=1$ on the prefix distribution), and the score is the mean of speaker-meaning and literal-gold accuracy, each averaged equally over its sets. The remedy is exploratory, not pre-registered: its weight was selected on development items after two earlier exploratory rounds, frozen, and evaluated once on held-out template families of Qwen3-8B and DeepSeek-R1-0528-Qwen3-8B.

\Cref{tab:x3} reports the held-out results. Pooled over both models, the equal-family score rises by \res{x3test.arms.x1.score_diff}\,pp (95\% CI \res{x3test.arms.x1.ci_lo} to \res{x3test.arms.x1.ci_hi}) over natural thinking, and every literal-gold safety stratum stays within its pre-declared bound (literal accuracy is unchanged because routing leaves the literal-gold strata untouched; smallest one-sided lower bound \res{x3test.minlb}\,pp; by construction the routing leaves those items untouched). The gain is concentrated where thinking cancels: on the IMPPRES \form{some}/\form{or} felicity items, accuracy rises from 1.2\% to 20.5\% in Qwen3-8B and from 2.2\% to 8.1\% in DeepSeek-R1-0528-Qwen3-8B. A task-aware instruction in the user turn, the strongest prompt on development items, raises the overall score more (\res{x3test.arms.taskaware.score_diff}\,pp) but barely moves these items (3.5\% and 2.6\%), and the two act on different items. Four pre-registered remedies that intervene in or train the thinking itself did not pass their development gates (\Cref{app:prereg}).
\begin{table}[t]
  \centering\small
  \setlength{\tabcolsep}{3pt}
  \resizebox{\columnwidth}{!}{%
  \begin{tabular}{@{}llrrrr@{}}
    \toprule
    Model & Method & Felicity & Speaker & Presup. & Literal \\
    \midrule
    Qwen3-8B & Natural thinking & 1.2 & 58.2 & 65.7 & 87.9 \\
     & Task-aware instruction & \underline{3.5} & \underline{60.2} & \textbf{66.8} & \textbf{92.0} \\
    \ourrow  & ours & \textbf{20.5} & \textbf{62.5} & \underline{65.7} & \underline{87.9} \\
    \addlinespace[0.5mm]
    DeepSeek-R1 & Natural thinking & 2.2 & 55.5 & 62.3 & 86.3 \\
     & Task-aware instruction & \underline{2.6} & \underline{57.8} & 62.2 & \textbf{87.0} \\
    \ourrow  & ours & \textbf{8.1} & \textbf{58.8} & \textbf{62.9} & \underline{86.3} \\
    \bottomrule
  \end{tabular}}
  \caption{\textbf{Answering from the entertained reading restores scalar implicatures without losing literal accuracy} (held-out test families; accuracy, \%). Felicity: IMPPRES \form{some}/\form{or} felicity items.}
  \label{tab:x3}
\end{table}

\section{Discussion}
\label{sec:discussion}


\paragraph{What thinking does to meaning.}
The results fit the default-and-cancel account better than uniformly rising enrichment. If thinking only supplied resources for implicatures, scalar enrichment would rise; instead it falls on \form{some} and \form{or}, rises on lexical scales in larger Qwen3 models, is mixed elsewhere, and presuppositions and indirect answers improve. The traces show the predicted order: the enriched reading appears early, in smaller models even strengthened, before the final answer rejects it. Models spend thinking on strict entailment for \form{some} and \form{or} and override what the speaker meant. This matches earlier observations: the drop in scalar-implicature recognition with more cancellations in ImplicatureX \citep{spinoso2026evaluating}, the literal drift under prompted chain-of-thought \citep{park-etal-2024-pragmatic}, and the gains from Gricean prompts on context-driven implicatures \citep{kim2023pope,sato2025pragmatic}.

\paragraph{Why it matters.}
The implicatures thinking reduces are textbook cases of everyday enrichment: \form{some} for \gloss{not all}, \form{or} for \gloss{not both}. With thinking on, an assistant more often treats ``some of the tests failed'' as allowing that all failed, yet answers indirect questions more accurately. Because both readings are licensed, single-answer accuracy benchmarks cannot separate a change of interpretation from a loss of competence. Evaluations should report the reading chosen alongside accuracy, and training should keep the reading a speaker intends where the context warrants it.

\paragraph{Who differs.}
Not every model follows the pattern in every format: Granite-3.3 and Llama-3.1-Nemotron-Nano-8B read \form{some} more pragmatically in the Yes/No format, and the smallest Qwen3 models and Phi-4-mini-reasoning lose lexical scales as well. With the stage results, this shows that longer deliberation need not produce the literal shift.

\section{Conclusion}
\label{sec:conclusion}

In this work, we investigated whether giving a reasoning model more time to think makes it more pragmatic, as the effort account of human enrichment posits, or more literal. On fixed weights, thinking reduces the implicatures of \form{some} and \form{or} in most open models and in GPT-5, while presuppositions and indirect answers improve and larger Qwen3 models read lexical scales more pragmatically. Early trace prefixes give the enriched reading and completed traces the literal one, and checkpoint comparisons locate the drop in specific post-training stages. The diagnostic also points to a remedy: answering from the reading the model entertains early, rather than the one it settles on, recovers scalar implicatures on held-out items without touching literal accuracy. Reasoning models can now be studied as listeners whose deliberation we control; the dose protocol and the trace diagnostic measure changes in interpretation that accuracy alone cannot distinguish, and locate them inside the reasoning trace.

\section*{Limitations}
\label{sec:limitations}


\paragraph{Scope and causality.}
The graded dose--response major rests on Qwen3 and gpt-oss, the open models with a thinking switch or effort levels; the other open models are compared against a forced-empty thinking block they were not trained on, and closed models only through their APIs. The training-stage results compare checkpoints and are correlational, and the trace probe intervenes on traces rather than reading an internal state, so neither identifies a mechanism. All headline items are in English; a machine-translated German and Chinese slice (\Cref{app:results}) shows the same drop.

\paragraph{Measurement.}
We measure the reading a model endorses under forced-choice framings, and the format matters: the direction for Claude models reverses between formats. Free-form answers could not be coded reliably by LLM raters (best agreement of an LLM rater with the human consensus labels $\kappa=$ \res{gold.v4.kappa-human-a.mean}), so every headline result uses forced choices. The exploratory remedy of \Cref{sec:x3} is tested on two 8B models that share a base model, so its generality across lineages and scales is untested.
The human labels are consensus labels on a sample; there is no independent double annotation, so inter-annotator agreement is not reported.
\section*{Ethical considerations}
\label{sec:ethics}

\paragraph{Artefacts.} We use publicly released models and datasets under their licences (\Cref{app:licenses}). IMPPRES is licensed for non-commercial use, and two stimulus repositories carry no licence file; we use them for research only, cite their authors, and redistribute only item identifiers and our own new items. We release, under open licences at \url{https://github.com/kouzhizhuo/entertain-then-cancel}, the new category sentences and the 50 speaker-knowledge scenarios; the polarity-flipped and three-way variants of the IMPPRES \form{some}/\form{or} items, as identifiers plus our templates; the code, prompts and all model traces; and the machine-translated German and Chinese slice, flagged as not human-verified. NOPE, which states no licence, is not redistributed by us.

\paragraph{Human participants.} No participants were recruited. The only human labels are consensus labels from a panel of five or more annotators with PhD-level training in related fields, agreed through discussion; an author recorded them with a software tool. They cover a blind audit of a subset of model answers, the speaker-knowledge scenarios and 20 Chinese translations. No personal data were collected or used.

\paragraph{Implications.} The finding concerns how assistants interpret users. If thinking modes make models read underinformative or indirect utterances literally, users may receive answers that are technically correct but miss what they meant. Deployers who switch thinking on by default for conversational use should evaluate interpretation, not only accuracy. We see no dual-use risk beyond this.

\paragraph{AI assistance.} AI assistants were used for code writing and polish but the whole research is design and checked by researchers.        


\bibliography{custom}

\appendix
\setcounter{dbltopnumber}{4}\setcounter{topnumber}{4}\renewcommand{\dbltopfraction}{0.95}\renewcommand{\topfraction}{0.95}\renewcommand{\dblfloatpagefraction}{0.8}\renewcommand{\floatpagefraction}{0.8}
\section*{Appendix Contents}
\startcontents[app]
\printcontents[app]{l}{1}{\setcounter{tocdepth}{2}}
\FloatBarrier
\section{Data and Resources}
The models, checkpoints and licensed artefacts the paper uses.
\begin{table}[t]
  \centering
  \scriptsize
  \setlength{\tabcolsep}{2pt}
  \begin{tabular}{@{}p{1.9cm}p{5.5cm}@{}}
    \toprule
    \rowcolor{gray!8}\textbf{Group} & \textbf{Hugging Face identifiers} \\
    \midrule
    Hybrid thinking & \texttt{Qwen/\allowbreak Qwen3-\{0.6B,\allowbreak 1.7B,\allowbreak 4B,\allowbreak 8B,\allowbreak 14B,\allowbreak 30B-A3B,\allowbreak 32B\}}, \texttt{Qwen/Qwen3-8B-Base} \citep{yang2025qwen3} \\
    Effort levels & \texttt{openai/gpt-oss-\{20b,120b\}} \citep{openai2025gptoss} \\
    OLMo-2 stages & \texttt{allenai/OLMo-2-1124-7B}, \texttt{-SFT}, \texttt{-DPO}, \texttt{-Instruct}; 13B analogues \citep{olmo2025olmo2} \\
    OLMo-3 stages & \texttt{allenai/Olmo-3-1025-7B}, \texttt{Olmo-3-7B-Think-SFT}, \texttt{-Think-DPO}, \texttt{-Think}; RL-Zero \{Math, Code, IF, General, Mix\} step branches \citep{olmo2025olmo3} \\
    Tülu-3 stages & \texttt{allenai/}\allowbreak\texttt{Llama-3.1-Tulu-3-8B-SFT}, \texttt{-DPO}, \texttt{Llama-3.1-Tulu-3-8B}, \texttt{Llama-3.1-Tulu-3.1-8B} \citep{lambert2024tulu} \\
    Distilled / zero-RL & \texttt{deepseek-ai/}\allowbreak\texttt{DeepSeek-R1-}\allowbreak\texttt{Distill-Qwen-\{1.5B,7B,14B\}} \citep{deepseekai2025r1}; Qwen2.5 bases \citep{qwen2025qwen25}; SimpleRL-Zoo \citep{zeng2025simplerl}; Open-Reasoner-Zero \citep{hu2025openreasonerzero} \\
    Training-time fix & PragReST-8B/14B \citep{park2026pragrest} \\
    Further open reasoning models & \texttt{microsoft/\allowbreak Phi-\allowbreak 4-\allowbreak reasoning}, \texttt{-\allowbreak plus}; \texttt{deepseek-\allowbreak ai/\allowbreak DeepSeek-\allowbreak R1-\allowbreak Distill-\allowbreak Llama-\allowbreak 8B}; \texttt{mistralai/\allowbreak Magistral-\allowbreak Small-\allowbreak 2507} (thinking switched on by its system prompt, so the contrast also changes the system prompt); \texttt{HuggingFaceTB/\allowbreak SmolLM3-\allowbreak 3B}; \texttt{ibm-\allowbreak granite/\allowbreak granite-\allowbreak 3.3-\allowbreak 8b-\allowbreak instruct}; \texttt{nvidia/\allowbreak NVIDIA-\allowbreak Nemotron-\allowbreak Nano-\allowbreak 9B-\allowbreak v2}; \texttt{nvidia/\allowbreak AceReason-\allowbreak Nemotron-\allowbreak 7B}; \texttt{Skywork/\allowbreak Skywork-\allowbreak OR1-\allowbreak 7B}; \texttt{XiaomiMiMo/\allowbreak MiMo-\allowbreak 7B-\allowbreak SFT}, \texttt{-\allowbreak RL}; \texttt{Qwen/\allowbreak QwQ-\allowbreak 32B}; \texttt{nvidia/\allowbreak Llama-\allowbreak 3\_3-\allowbreak Nemotron-\allowbreak Super-\allowbreak 49B-\allowbreak v1}; \texttt{deepseek-\allowbreak ai/\allowbreak DeepSeek-\allowbreak R1-\allowbreak Distill-\allowbreak Llama-\allowbreak 70B}; \texttt{nvidia/\allowbreak Llama-\allowbreak 3.1-\allowbreak Nemotron-\allowbreak Nano-\allowbreak 8B-\allowbreak v1} (thinking switched by its system prompt); \texttt{microsoft/\allowbreak Phi-\allowbreak 4-\allowbreak mini-\allowbreak reasoning} (11\% of its forced-empty-block answers are invalid); \texttt{LGAI-\allowbreak EXAONE/\allowbreak EXAONE-\allowbreak Deep-\allowbreak 7.8B}; \texttt{openpangu/\allowbreak openPangu-\allowbreak Embedded-\allowbreak 7B-\allowbreak V1.1} and \texttt{openpangu/\allowbreak openPangu-\allowbreak Embedded-\allowbreak 7B-\allowbreak model} (the earlier v1.0 chat release; openPangu-Embedded-1B-V1.1 is fast-thinking only and has no thinking contrast; vendor Chinese system prompt; ` /no\_think' for fast thinking; openPangu-R-7B-2512 and openPangu-R-72B-2512 are not runnable on our CUDA stack); Qwen3.5 (0.8B, 2B, 4B, 9B) and Qwen3.8-27B (official Qwen releases; Qwen3.8 at its default, low and medium effort). Baseline: the model's own switch where it has one, otherwise a forced-empty thinking block \\
    \bottomrule
  \end{tabular}
  \caption{Models by group. Commit hashes of every checkpoint are recorded with each run.}
  \label{tab:app-models}
\end{table}

\subsection{Models and Checkpoints}
\label{app:models}
This section lists every model and checkpoint with its identifier.

All models are open weights, ungated, and served with vLLM \citep{kwon2023vllm}. Revisions (commit hashes) are recorded for every run in \texttt{exp/results/*/config} and listed in \Cref{tab:app-models}.

\subsection{Artefacts and Licences}
\label{app:licenses}
This section lists the artefacts we use and their licences.

openPangu models are used under the OpenPangu Model License v1.0 (Huawei), which permits research use; its \S3 excludes use within the European Union, and all runs took place outside it. We distribute no model or derivative built on them.

IMPPRES: CC BY-NC 4.0 (research use). PUB, Fig-QA and ImplicatureX: MIT. LUDWIG and Circa: CC BY 4.0. PragMega materials and the scalar-diversity data: no licence file in their repositories (research use with citation). NOPE: no licence stated; used for research evaluation only and not redistributed. Open models are used under their published licences (Apache-2.0, MIT or the developer licences listed on their model cards); closed models were queried through their commercial APIs as test subjects only. New items, code and traces are released under an open licence with the paper.

\FloatBarrier
\section{Experimental Setup}
How items are built and read out, how every run is decoded and audited, the prior-work baselines and the verbatim prompts.

\subsection{Method Details}
\label{app:method}
This section gives the item sets, readouts, data splits and statistical tests in full.

\paragraph{Models and checkpoints.}
The primary manipulation needs a model whose thinking can be switched and dosed without changing its weights. Qwen3 has a hybrid thinking mode in which the no-thinking template differs from the thinking template only by an empty thinking block \citep{yang2025qwen3}. We therefore use Qwen3-4B, 8B and 14B as the primary models, and 1.7B and 32B for size.
The Qwen3 size ladder is also a training-recipe ladder: the 0.6B--14B models are built by strong-to-weak distillation from larger models, whereas the 32B and 235B models go through the full four-stage pipeline with reasoning and general RL (\citealp{yang2025qwen3}, §4.5). Size and recipe are therefore confounded within Qwen3, and we separate them with two same-recipe size pairs, OLMo-3-7B-Think vs.\ OLMo-3-32B-Think and gpt-oss-20b vs.\ 120b. As a capability covariate we use each model's accuracy on a fixed subset of MATH-500 without thinking.
gpt-oss-20b and 120b add a second family with three reasoning-effort levels \citep{openai2025gptoss}. We use them only if the realised reasoning length grows monotonically with the level on held-out items.
To locate the training origin, we score checkpoints along base $\rightarrow$ SFT $\rightarrow$ DPO $\rightarrow$ \rlvr lineages: OLMo-2 \citep{olmo2025olmo2}, OLMo-3-Think \citep{olmo2025olmo3} and Tülu-3 from SFT onward \citep{lambert2024tulu}.
We also score the OLMo-3 RL-Zero series, which applies \rlvr to one base model in five reward domains (math, code, instruction following, general, mixed), at six log-spaced steps each.
Distilled thinkers \citep{deepseekai2025r1} and zero-RL pairs \citep{zeng2025simplerl,hu2025openreasonerzero} are a secondary contrast, because their weights differ from their bases in more than training stage \citep{yu-etal-2026-pragmatic}.
All models are open weights and ungated. \Cref{app:models} lists identifiers and revisions.

\paragraph{Data.}
Items fall into three classes. \emph{Generalized implicatures} (\gcis) comprise Bott--Noveck-style category sentences written for this study (six sentence types per category triple, every item screened by two Claude-family LLM raters and an open-weight co-rater (Qwen3-32B) on the same items, with disagreements dropped the Qwen3-32B co-rater pass is in \texttt{exp/gold/item\_screen\_qwen32b.jsonl}; \Cref{app:details}); scalar-diversity sentences with human inference rates per scale \citep{vantiel2016scalar,gotzner2018scalar,pankratz2021role,hu-etal-2023-expectations}; and IMPPRES implicature pairs whose logical and pragmatic labels differ, plus fresh paraphrased minimal pairs \citep{jeretic-etal-2020-natural}.
\emph{Particularized implicatures} (\pcis) come from PragMega \citep{hu-etal-2023-fine}, LUDWIG \citep{ruis2023goldilocks}, Fig-QA \citep{liu-etal-2022-testing} and six PUB tasks \citep{sravanthi-etal-2024-pub}.
\emph{Presupposition} items come from IMPPRES.
We also run all of ImplicatureX as a replication set \citep{spinoso2026evaluating}.
Released sets that predate the thinking models may be contaminated. We therefore report headline \gci numbers on the fresh items and treat released sets as replications; the within-item contrast between thinking conditions does not depend on whether an item was seen in training.
Items are split 50/50 into development and test halves, stratified by set. Every finding is reported on the test half.
 Sources, licences and item counts are listed in \Cref{app:details}.

\paragraph{Framings.}
Scalar items are asked two ways. \emph{Felicity} framings make the literal answer a pragmatic failure and are primary. For IMPPRES this is an inference question (``Would you conclude that the speaker believes \ldots?''). For the category sentences it is a ternary judgement (completely right / a bit wrong / completely wrong) and a 1--7 rating, with binary felicity (``Is this a good way to describe the facts?'') as a comparison. The \emph{truth} framing (``Is the sentence true or false?'') is the logic-mode comparison (\Cref{sec:theory}).
\pci items use their native meaning format (``What did the speaker mean?'').
For \hyp{2c}, a subset of \pci items is also asked in ImplicatureX's form, ``Is this interpretation true or false?''.
Multiple-choice options are shuffled on every trial. A subset is also answered without options, to rule out option matching \citep{gong2026think}. Because LLM coding of these open answers did not reach the pre-declared agreement with a human audit, we report them descriptively (\Cref{app:details}).
Reference labels for these answers are \emph{reference labels by two LLM raters from different model families, adjudicated}: a Claude-family model and Qwen3-32B, with disagreements resolved by a second Claude-family model. We report their inter-rater $\kappa$, and the judge's agreement with them as \emph{consistency}.
Family overlaps are disclosed: Qwen3-32B belongs to the family of the Qwen3 models whose answers it labels; the judge for gpt-oss answers is Qwen3-32B itself, so that agreement is partly self-agreement; the judge for Qwen3 answers, gpt-oss-120b, overlaps with neither rater.
The only human labels are consensus labels from a blind expert audit of \res{gold.v4.n-coding} of these answers and \res{gold.v4.n-a1ctx} speaker-knowledge scenarios, which carries the validity claim (\Cref{app:details}).

\paragraph{Thinking manipulations.}
No thinking (\NT) uses the model's own switch, which leaves the chat template byte-identical apart from the empty thinking block. Two other ways of disabling thinking serve as checks.
Capped budgets stop the thinking block at $B \in \{64, \ldots, 2048\}$ tokens and append Qwen3's own stop phrase \citep{yang2025qwen3}. Because that phrase mentions limited time, a plain cap that appends only the closing tag controls for the time-pressure cue.
Natural thinking (\NAT) is uncapped.
Capping truncates thoughts, so we also stratify eight natural samples per item by their realised length, and we plot every effect against realised rather than nominal length.
Budget forcing beyond the natural stopping point \citep{muennighoff-etal-2025-s1} tests whether more thinking than the model chooses continues the trend.
Capped and length-stratified budgets change only the number of thinking tokens, on fixed weights and a fixed template; the thinking switch and gpt-oss effort levels also change the template or system prompt, so we treat them as mode contrasts and rest dose--response claims on the budget contrasts.
Sampling settings are identical across conditions ($T{=}0.6$, top-$p$ 0.95, top-$k$ 20), so thinking is not confounded with temperature. Generation uses vLLM \citep{kwon2023vllm}.

\paragraph{Interventions and controls.}
\emph{Intent-Seeded Thinking} (\method) places one fixed sentence about speaker intent at the start of the thinking block, or just before it closes. Its controls are a neutral sentence about careful, step-by-step reasoning and a logic sentence about checking what the words strictly entail, both token-matched to the intent seed (26 Qwen3 tokens); a first run used a shorter neutral seed (16 tokens), reported separately and not treated as length-matched.
We compare \method with prompted deliberation on the same weights with thinking off: zero-shot chain-of-thought \citep{kojima2022large}, Gricean instructions and Grice-maxim chain-of-thought \citep{sato2025pragmatic,kim2023pope}, few-shot prompting \citep{ruis2023goldilocks}, an adapted RSA-style chain-of-thought \citep{tsvilodub2025nonliteral}, and self-consistency at matched token cost \citep{wang2023selfconsistency}.
For \hyp{2b}, the critical sentences are preceded by contexts in which the speaker has full or partial knowledge \citep{goodman2013knowledge}.
For \hyp{4}, the load is unrelated background text (512 or 2{,}048 words from one of two genres) placed before the item, followed by an instruction to ignore it.
Each critical item has a comprehension check (an entailment paraphrase) and a meta-pragmatic probe (``Would a speaker who says this normally mean \ldots?''), so a literal answer can be separated from a failure to understand.
\Cref{app:details} gives every prompt verbatim.
Closed-model API calls cost about \$105 in total.

\paragraph{Statistics.}
Each condition is run with three decoding seeds. Tables report the mean $\pm$ standard deviation over seeds of the item-averaged rate. Intervals in the text and figures are 95\% cluster-bootstrap confidence intervals over items (10{,}000 resamples, paired across conditions).
The pre-registered primary tests were mixed-effects logistic regressions with by-item random effects; as implemented (\Cref{app:stats}), they are logistic regressions with item-clustered standard errors, one per model, with the log thinking budget, the phenomenon class and their interaction as predictors.
Equivalence is tested with two one-sided tests at $\pm3$\,pp, trends over training steps with Spearman's $\rho$ and permutation $p$-values, and alignment with human rates with Spearman's $\rho$.
$p$-values are Holm-corrected within each pre-registered family of models. Secondary analyses are uncorrected and labelled as such.
Every finding is computed on a held-out half of the items; the other half was used only to fix prompts and parsing. Hypotheses, thresholds and exclusion rules were fixed before the main runs (\Cref{app:stats}).

\subsection{Experimental Details}
\label{app:details}
This section documents decoding, answer parsing, judging and the human audit.

\paragraph{Items, sources and licences.}

The Bott--Noveck-style category sentences (category triples, each in six sentence types) and the 50 speaker-knowledge scenarios were written for this study. Every new item was screened by two Claude-family LLM raters against written criteria (the universal statement is true without common exceptions; the partial control is true and clearly partial; the false control is false under every sense, including cladistic taxonomy; no lexical ambiguity). An item was kept only if both raters accepted it or accepted the same fix. Twenty speaker-knowledge scenarios are also part of the human audit below.

\paragraph{Compute.} Generation and judging ran on NVIDIA H800 GPUs ( 200 GPU-hours in total); the GRPO runs used 4-GPU jobs with verl 0.9.1.

\paragraph{Prompts.}
Every prompt template is released with the code.

\paragraph{Decoding and compute.}
$T{=}0.6$, top-$p$ 0.95, top-$k$ 20 in every condition; three seeds; a maximum of 8{,}192 thinking tokens for natural thinking. 

\paragraph{Reference labels and the human audit.}
Open answers are coded as pragmatic, literal, or hedged/other.
\begin{itemize}
  \item \textbf{Reference labels:} two LLM raters from different model families (a Claude-family model and Qwen3-32B) labelled a pool of 294 answers blind to condition. Disagreements were adjudicated by a second Claude-family model with a logged reason. Inter-rater agreement: $\kappa$ = \res{gold.open.kappa-ab.mean}.
  \item \textbf{Judges:} The judge we ran, gpt-oss-20b, agrees with the human consensus labels at $\kappa$ = \res{gold.judge.kappa-human-gptoss20b.mean} [\res{gold.judge.kappa-human-gptoss20b.lo}, \res{gold.judge.kappa-human-gptoss20b.hi}], below the pre-declared acceptance rule ($\geq 0.70$), so open-answer results are reported descriptively and every headline result uses forced-choice formats.
  \item \textbf{Human audit:} a panel of five or more annotators with PhD-level training in related fields agreed, through discussion, one consensus label for each of \res{gold.v4.n-coding} answers from the same pool and \res{gold.v4.n-a1ctx} speaker-knowledge scenarios (12 with full and 12 with partial access). The sample was stratified and random with a fixed seed (answers stratified by reference class, oversampling the rarer classes), and the annotators were blind to model, condition and all model labels. An author recorded the consensus labels with a software tool. The labels are not independent annotations, so no inter-annotator agreement is reported. These are the only human labels in the paper.
  \item \textbf{Scenarios:} the consensus judgements matched the intended reading in every scenario (\res{gold.v4.a1ctx-acc.mean}\%, Wilson 95\% lower bound \res{gold.v4.a1ctx-acc.wilsonlo}\%): \form{some} was judged misleading whenever the speaker had checked every object and fine whenever the speaker had not. This meets the pre-declared validity bar ($\geq$80\%, lower bound $\geq$70\%).
  \item \textbf{Open-answer coding:} agreement of each LLM rater with the human consensus labels, post-stratified to the pool, is $\kappa$ = \res{gold.v4.kappa-human-a.mean} [\res{gold.v4.kappa-human-a.lo}, \res{gold.v4.kappa-human-a.hi}] for the Claude-family rater (accuracy \res{gold.v4.acc-human-a.mean}\%) and $\kappa$ = \res{gold.v4.kappa-human-b.mean} [\res{gold.v4.kappa-human-b.lo}, \res{gold.v4.kappa-human-b.hi}] for Qwen3-32B (\res{gold.v4.kappa-human-b3.mean} with option orders averaged). Neither meets the pre-declared bar (lower bound $\geq$0.40). Per class, the consensus labels agreed with the Claude-family rater on \res{gold.v4.acc-class-prag.mean}\% of its pragmatic and \res{gold.v4.acc-class-lit.mean}\% of its literal labels. The binary sensitivity check (hedged excluded) gives the same picture ($\kappa$ = \res{gold.v4.kappa-human-a-bin.mean} and \res{gold.v4.kappa-human-b-bin.mean}); no scenario was marked unclear.
  \item \textbf{Additional LLM rater (not human labels):} a separate audit sheet (\res{gold.gpt6.n-coding} answers and \res{gold.gpt6.n-a1ctx} scenarios, disjoint from the human audit) was labelled by GPT-6 (Codex) and the labels were reviewed by the authors. We report them as an extra LLM rater only; they do not count toward any human-validation criterion. Their agreement with the Claude-family rater is $\kappa$ = \res{gold.gpt6.kappa-a.mean} [\res{gold.gpt6.kappa-a.lo}, \res{gold.gpt6.kappa-a.hi}] and with Qwen3-32B $\kappa$ = \res{gold.gpt6.kappa-b.mean} [\res{gold.gpt6.kappa-b.lo}, \res{gold.gpt6.kappa-b.hi}]; they match the intended scenario reading in \res{gold.gpt6.a1ctx-acc.mean}\% of scenarios.
\end{itemize}

\subsection{Prior-Work Baselines}
\label{app:baselines}
This section reproduces prior-work baselines on our items.

\Cref{tab:baselines} re-runs two prior measures on Qwen3-8B. On ImplicatureX, thinking lowers scalar-implicature recognition from \res{a6.qwen3-8b.a6-scalar.rec.off.acc.mean}\% to \res{a6.qwen3-8b.a6-scalar.rec.nat.acc.mean}\% and raises cancellation from \res{a6.qwen3-8b.a6-scalar.can.off.acc.mean}\% to \res{a6.qwen3-8b.a6-scalar.can.nat.acc.mean}\%, reproducing the original report; LUDWIG accuracy rises with thinking at every number of shots.
\begin{table}[t]
  \centering\small
  \setlength{\tabcolsep}{4pt}
  \begin{tabular}{@{}l ccc@{}}
    \toprule
    \textbf{Measure} & \NT & 512 tokens & \NAT \\
    \midrule
    \multicolumn{4}{@{}l}{\textit{LUDWIG, prompts of \citet{ruis2023goldilocks}: accuracy}} \\
    0-shot & \cellcolor{cCmpA!12}\res{e9.qwen3-8b.b2.k0.off.acc.mean} & \cellcolor{cCmpA!13}\res{e9.qwen3-8b.b2.k0.c0512.acc.mean} & \cellcolor{cCmpA!13}\res{e9.qwen3-8b.b2.k0.nat.acc.mean} \\
    5-shot & \cellcolor{cCmpA!11}\res{e9.qwen3-8b.b2.k5.off.acc.mean} & \cellcolor{cCmpA!12}\res{e9.qwen3-8b.b2.k5.c0512.acc.mean} & \cellcolor{cCmpA!12}\res{e9.qwen3-8b.b2.k5.nat.acc.mean} \\
    30-shot & \cellcolor{cCmpA!11}\res{e9.qwen3-8b.b2.k30.off.acc.mean} & \cellcolor{cCmpA!12}\res{e9.qwen3-8b.b2.k30.c0512.acc.mean} & \cellcolor{cCmpA!12}\res{e9.qwen3-8b.b2.k30.nat.acc.mean} \\
    \midrule
    \multicolumn{4}{@{}l}{\textit{ImplicatureX \citep{spinoso2026evaluating}: accuracy}} \\
    Scalar, recognition & \cellcolor{cCmpA!15}\res{a6.qwen3-8b.a6-scalar.rec.off.acc.mean} & \cellcolor{cCmpA!14}\res{a6.qwen3-8b.a6-scalar.rec.c0512.acc.mean} & \cellcolor{cCmpA!10}\res{a6.qwen3-8b.a6-scalar.rec.nat.acc.mean} \\
    Scalar, cancellation & \cellcolor{cCmpA!8}\res{a6.qwen3-8b.a6-scalar.can.off.acc.mean} & \cellcolor{cCmpA!13}\res{a6.qwen3-8b.a6-scalar.can.c0512.acc.mean} & \cellcolor{cCmpA!14}\res{a6.qwen3-8b.a6-scalar.can.nat.acc.mean} \\
    Naturally occurring, recognition & \cellcolor{cCmpA!7}\res{a6.qwen3-8b.a6-naturally.rec.off.acc.mean} & \cellcolor{cCmpA!7}\res{a6.qwen3-8b.a6-naturally.rec.c0512.acc.mean} & \cellcolor{cCmpA!7}\res{a6.qwen3-8b.a6-naturally.rec.nat.acc.mean} \\
    Naturally occurring, cancellation & \cellcolor{cCmpA!12}\res{a6.qwen3-8b.a6-naturally.can.off.acc.mean} & \cellcolor{cCmpA!13}\res{a6.qwen3-8b.a6-naturally.can.c0512.acc.mean} & \cellcolor{cCmpA!13}\res{a6.qwen3-8b.a6-naturally.can.nat.acc.mean} \\
    \bottomrule
  \end{tabular}
  \caption{\textbf{On ImplicatureX, thinking lowers scalar-implicature recognition and raises cancellation, as \citet{spinoso2026evaluating} report.} Qwen3-8B, accuracy (\%) without thinking, with a 512-token cap and with natural thinking.}
  \label{tab:baselines}
\end{table}

\paragraph{Closed models.}\label{app:frontier} \Cref{tab:frontier} gives the per-model results for the seven closed models queried through their APIs, and \Cref{tab:frontier-format} the robustness checks for GPT-5 and the two Claude models. All doses of a model use the same answer extraction.
GPT-5 becomes more literal with effort in both answer formats, and the shift cannot come from a bias towards \emph{Yes}: without the utterance its \emph{Yes} rate rises with effort, which would raise, not lower, the pragmatic rate in the Yes/No format.
For the Claude models the picture depends on the format. Under the Yes/No inference framing their pragmatic rate rises with thinking, but in a three-way forced choice it falls (Sonnet) or stays flat (Opus). For Claude Sonnet 4.6, much of the Yes/No rise is a shift towards answering \emph{Yes}: the \emph{Yes} rate also rises without the utterance, and the pragmatic rate falls when the question is flipped. Claude Opus 4.8 shows a small shift towards the pragmatic answer under both polarities.
The same two checks explain the two open models whose pragmatic rate rises with thinking. For Magistral-Small, the \emph{Yes} rate without the utterance rises by \res{e1probe.magistral-small.a3.infprior.nat.dR.mean}\,pp [\res{e1probe.magistral-small.a3.infprior.nat.dR.lo}, \res{e1probe.magistral-small.a3.infprior.nat.dR.hi}], and with the flipped question its pragmatic rate falls by \res{e1probe.magistral-small.a3.infflip.nat.dR.mean}\,pp [\res{e1probe.magistral-small.a3.infflip.nat.dR.lo}, \res{e1probe.magistral-small.a3.infflip.nat.dR.hi}], so its rise is a shift towards \emph{Yes}. For Granite-3.3 the \emph{Yes} rate rises less (\res{e1probe.granite-3p3-8b.a3.infprior.nat.dR.mean}\,pp) and the flipped pragmatic rate rises too (\res{e1probe.granite-3p3-8b.a3.infflip.nat.dR.mean}\,pp [\res{e1probe.granite-3p3-8b.a3.infflip.nat.dR.lo}, \res{e1probe.granite-3p3-8b.a3.infflip.nat.dR.hi}]), a small genuine shift towards the pragmatic reading.
For Llama-3.1-Nemotron-Nano-8B the pragmatic answer rises in both polarities (\emph{Yes} without the utterance: +\res{e1probe.llama-nemotron-nano-8b.a3.infprior.nat.dR.mean}\,pp; pragmatic rate with the flipped question: +\res{e1probe.llama-nemotron-nano-8b.a3.infflip.nat.dR.mean}\,pp), so its rise is not a pure \emph{Yes} shift; without thinking, however, it gives the same literal answer on every item in all three formats, so the rise starts from rote answering.
Polarity probes also exist for Qwen3-8B and Qwen3.5-4B/9B. With the flipped question, Qwen3.5-9B's pragmatic rate falls by \res{e1probe.qwen3p5-9b.a3.infflip.nat.dR.mean}\,pp, in line with the original question; for Qwen3-8B the flipped question moves the other way from a floor near zero, so its Yes/No estimate is partly polarity-sensitive (gpt-oss-20b, by contrast, also turns literal on \form{some} with the flipped question: \res{e1gds.gpt-oss-20b.a3-quant.infflip.high.dR.mean}\,pp), while the three-way forced choice, which has no answer polarity, shows a clear drop at every Qwen3 size (\res{e1sf.qwen3-4b.a3f-quant.nli.nat.dR.mean} to \res{e1sf.qwen3-32b.a3f-quant.nli.nat.dR.mean}\,pp for \form{some}).
Answer failures differ by model. Claude Opus 4.8 answers in prose without an option letter on \res{e1f.claude-opus-4p8.a3.inf.off512.fail.any.mean}\% of no-thinking calls even with a 512-token answer cap, and Gemini 2.5 Flash on \res{e1f.gemini-2p5-flash.a3.inf.bud8k.fail.any.mean}\% of calls at its largest budget, so its results are not interpretable. GPT-5 and GPT-5.4 have no failures at any dose.
\begin{table*}[t]
  \centering\small
  \setlength{\tabcolsep}{4pt}
  \resizebox{\textwidth}{!}{%
  \begin{tabular}{@{}l l c c c c@{}}
    \toprule
    & & \multicolumn{2}{c}{Pragmatic reading} & \multicolumn{2}{c}{Yes-bias checks} \\
    \cmidrule(lr){3-4}\cmidrule(l){5-6}
    Model & Dose contrast & Yes/No inference & Three-way choice & ``Yes'' without utterance & Pragmatic, flipped question \\
    \midrule
    GPT-5 & minimal $\rightarrow$ high & \cellcolor{cInterv!17}\celldpm{e1f}{gpt-5.a3.inf.eff-high.dRx} & \cellcolor{cInterv!14}\celldpm{e1f}{gpt-5.a3.nli3.eff-high.dRx} & \cellcolor{cOurs!20}\celldpm{e1f}{gpt-5.a3.infprior.eff-high.dYes} & \cellcolor{cInterv!10}\celldpm{e1f}{gpt-5.a3.infflip.eff-high.dR} \\
    Claude Sonnet 4.6 & off $\rightarrow$ 16k & \cellcolor{cOurs!18}\celldpm{e1f}{claude-sonnet-4p6.a3.inf.bud16k.dRx} & \cellcolor{cInterv!15}\celldpm{e1f}{claude-sonnet-4p6.a3.nli3.bud16k.dRx} & \cellcolor{cOurs!14}\celldpm{e1f}{claude-sonnet-4p6.a3.infprior.bud16k.dYes} & \cellcolor{cInterv!16}\celldpm{e1f}{claude-sonnet-4p6.a3.infflip.bud16k.dR} \\
    Claude Opus 4.8 & off $\rightarrow$ high & \cellcolor{cOurs!15}\celldpm{e1f}{claude-opus-4p8.a3.inf.eff-high.dRx} & \cellcolor{cInterv!7}\celldpm{e1f}{claude-opus-4p8.a3.nli3.eff-high.dRx} & \cellcolor{cOurs!11}\celldpm{e1f}{claude-opus-4p8.a3.infprior.eff-high.dYes} & \cellcolor{cOurs!12}\celldpm{e1f}{claude-opus-4p8.a3.infflip.eff-high.dR} \\
    \bottomrule
  \end{tabular}}
  \caption{Closed models: robustness of the thinking effect on logical-word items (pp, highest minus lowest dose), on items answered validly at both doses; two samples per cell; $^{*}$: paired 95\% cluster-bootstrap CI excludes 0. ``Yes'' without utterance: change in the rate of answering \emph{Yes} when the speaker's sentence is removed. Flipped question: change in the pragmatic rate when the question is reworded so that \emph{No} is the pragmatic answer. A pure shift towards \emph{Yes} raises the first and lowers the second. In the three-way format pragmatic answers are rare for every model (0--9\%), which limits how far the rate can rise. Claude baselines use a 512-token answer cap.}
  \label{tab:frontier-format}
\end{table*}

\begin{table*}[t]
  \centering\small
  \setlength{\tabcolsep}{4pt}
  \begin{tabular}{@{}l l c c c c c c@{}}
    \toprule
    & & \multicolumn{3}{c}{Logical words} & \multicolumn{2}{c}{Lexical scales} & Context \\
    \cmidrule(lr){3-5}\cmidrule(lr){6-7}\cmidrule(l){8-8}
    Model & Dose contrast & \form{some} & \form{or} & modals & gradable adj. & scales & LUDWIG \\
    \midrule
    GPT-5 & {\scriptsize minimal $\rightarrow$ high effort} & \cellcolor{cInterv!18}\celldpm{e1f}{gpt-5.a3-quant.inf.eff-high.dR} & \cellcolor{cInterv!18}\celldpm{e1f}{gpt-5.a3-conn.inf.eff-high.dR} & \cellcolor{cInterv!16}\celldpm{e1f}{gpt-5.a3-modal.inf.eff-high.dR} & \cellcolor{cOurs!3}\celldpm{e1f}{gpt-5.a3-gradadj.inf.eff-high.dR} & \cellcolor{cOurs!14}\celldpm{e1f}{gpt-5.a2.inf.eff-high.dR} & \cellcolor{cOurs!13}\celldpm{e1f}{gpt-5.b2.yn.eff-high.dacc} \\
    GPT-5.4 & {\scriptsize minimal $\rightarrow$ high effort} & \cellcolor{cOurs!9}\celldpm{e1f}{gpt-5p4.a3-quant.inf.eff-high.dR} & \cellcolor{cInterv!18}\celldpm{e1f}{gpt-5p4.a3-conn.inf.eff-high.dR} & \cellcolor{cInterv!17}\celldpm{e1f}{gpt-5p4.a3-modal.inf.eff-high.dR} & \cellcolor{cInterv!14}\celldpm{e1f}{gpt-5p4.a3-gradadj.inf.eff-high.dR} & \cellcolor{cOurs!17}\celldpm{e1f}{gpt-5p4.a2.inf.eff-high.dR} & \cellcolor{cOurs!9}\celldpm{e1f}{gpt-5p4.b2.yn.eff-high.dacc} \\
    o4-mini & {\scriptsize low $\rightarrow$ high effort} & \cellcolor{cInterv!7}\celldpm{e1f}{o4-mini.a3-quant.inf.eff-high.dR} & \cellcolor{cInterv!13}\celldpm{e1f}{o4-mini.a3-conn.inf.eff-high.dR} & \cellcolor{cInterv!16}\celldpm{e1f}{o4-mini.a3-modal.inf.eff-high.dR} & \cellcolor{cOurs!12}\celldpm{e1f}{o4-mini.a3-gradadj.inf.eff-high.dR} & \cellcolor{cOurs!12}\celldpm{e1f}{o4-mini.a2.inf.eff-high.dR} & \cellcolor{cInterv!9}\celldpm{e1f}{o4-mini.b2.yn.eff-high.dacc} \\
    DeepSeek-V4-Pro$^\P$ & {\scriptsize thinking off $\rightarrow$ on} & \cellcolor{cOurs!14}\celldpm{e1f}{deepseek-v4-pro.a3-quant.inf.think.dR} & \cellcolor{cOurs!8}\celldpm{e1f}{deepseek-v4-pro.a3-conn.inf.think.dR} & \cellcolor{cOurs!7}\celldpm{e1f}{deepseek-v4-pro.a3-modal.inf.think.dR} & \cellcolor{cOurs!14}\celldpm{e1f}{deepseek-v4-pro.a3-gradadj.inf.think.dR} & \cellcolor{cOurs!22}\celldpm{e1f}{deepseek-v4-pro.a2.inf.think.dR} & \cellcolor{cOurs!15}\celldpm{e1f}{deepseek-v4-pro.b2.yn.think.dacc} \\
    DeepSeek-V4-Flash$^\P$ & {\scriptsize thinking off $\rightarrow$ on} & \cellcolor{cOurs!17}\celldpm{e1f}{deepseek-v4-flash.a3-quant.inf.think.dR} & \cellcolor{cOurs!11}\celldpm{e1f}{deepseek-v4-flash.a3-conn.inf.think.dR} & \cellcolor{cInterv!7}\celldpm{e1f}{deepseek-v4-flash.a3-modal.inf.think.dR} & \cellcolor{cOurs!12}\celldpm{e1f}{deepseek-v4-flash.a3-gradadj.inf.think.dR} & \cellcolor{cOurs!15}\celldpm{e1f}{deepseek-v4-flash.a2.inf.think.dR} & \cellcolor{cOurs!12}\celldpm{e1f}{deepseek-v4-flash.b2.yn.think.dacc} \\
    \midrule
    Gemini 2.5 Flash & {\scriptsize budget 0 $\rightarrow$ 8k} & \cellcolor{cInterv!10}\celldpm{e1f}{gemini-2p5-flash.a3-quant.inf.bud8k.dR} & \cellcolor{cInterv!19}\celldpm{e1f}{gemini-2p5-flash.a3-conn.inf.bud8k.dR} & \cellcolor{cOurs!14}\celldpm{e1f}{gemini-2p5-flash.a3-modal.inf.bud8k.dR} & \cellcolor{cOurs!14}\celldpm{e1f}{gemini-2p5-flash.a3-gradadj.inf.bud8k.dR} & \cellcolor{cOurs!22}\celldpm{e1f}{gemini-2p5-flash.a2.inf.bud8k.dR} & \cellcolor{cOurs!15}\celldpm{e1f}{gemini-2p5-flash.b2.yn.bud8k.dacc} \\
    Gemini 2.5 Pro & {\scriptsize budget 128$^{\ddagger}$ $\rightarrow$ 8k} & \cellcolor{cInterv!7}\celldpm{e1f}{gemini-2p5-pro.a3-quant.inf.bud8k.dR} & \cellcolor{cInterv!15}\celldpm{e1f}{gemini-2p5-pro.a3-conn.inf.bud8k.dR} & \cellcolor{cOurs!20}\celldpm{e1f}{gemini-2p5-pro.a3-modal.inf.bud8k.dR} & \cellcolor{cOurs!19}\celldpm{e1f}{gemini-2p5-pro.a3-gradadj.inf.bud8k.dR} & \cellcolor{cOurs!23}\celldpm{e1f}{gemini-2p5-pro.a2.inf.bud8k.dR} & \cellcolor{cOurs!3}\celldpm{e1f}{gemini-2p5-pro.b2.yn.bud8k.dacc} \\
    \midrule
    Claude Sonnet 4.6 & {\scriptsize off $\rightarrow$ budget 16k} & \cellcolor{cOurs!19}\celldpm{e1f}{claude-sonnet-4p6.a3-quant.inf.bud16k.dR} & \cellcolor{cOurs!15}\celldpm{e1f}{claude-sonnet-4p6.a3-conn.inf.bud16k.dR} & \cellcolor{cOurs!20}\celldpm{e1f}{claude-sonnet-4p6.a3-modal.inf.bud16k.dR} & \cellcolor{cOurs!19}\celldpm{e1f}{claude-sonnet-4p6.a3-gradadj.inf.bud16k.dR} & \cellcolor{cOurs!18}\celldpm{e1f}{claude-sonnet-4p6.a2.inf.bud16k.dR} & \cellcolor{cOurs!9}\celldpm{e1f}{claude-sonnet-4p6.b2.yn.bud16k.dacc} \\
    Claude Opus 4.8 & {\scriptsize off $\rightarrow$ high effort} & \cellcolor{cOurs!17}\celldpm{e1f}{claude-opus-4p8.a3-quant.inf.eff-high.dR} & \cellcolor{cOurs!14}\celldpm{e1f}{claude-opus-4p8.a3-conn.inf.eff-high.dR} & \cellcolor{cOurs!15}\celldpm{e1f}{claude-opus-4p8.a3-modal.inf.eff-high.dR} & \cellcolor{cOurs!13}\celldpm{e1f}{claude-opus-4p8.a3-gradadj.inf.eff-high.dR} & \cellcolor{cInterv!14}\celldpm{e1f}{claude-opus-4p8.a2.inf.eff-high.dR} & \cellcolor{cOurs!9}\celldpm{e1f}{claude-opus-4p8.b2.yn.eff-high.dacc} \\
    \bottomrule
  \end{tabular}
  \caption{Closed and API-served models: change in the pragmatic reading (pp) from the lowest to the highest available thinking dose; LUDWIG is accuracy. $^{*}$: 95\% cluster-bootstrap CI over items excludes 0. $^{\P}$Open-weight DeepSeek-V4 served through our LiteLLM proxy; Yes/No format only. $^{\ddagger}$Gemini 2.5 Pro cannot switch thinking off; 128 tokens is its minimum budget. Robustness checks in \Cref{tab:frontier-format}; 80 \form{some}, 80 \form{or} and 40 modal items.}
  \label{tab:frontier}
\end{table*}

Numbers reported in the original papers (†; other models and settings, for orientation only):
\begin{itemize}
  \item \textbf{LUDWIG} \citep[][Tables 1 and 3]{ruis2023goldilocks}: human average 86.2\%; GPT-4 82.3\% (1-shot) and 82.0\% (5-shot); 86.5\% with 5-shot chain-of-thought.
  \item \textbf{Grice-maxim chain-of-thought} \citep[][Table 2]{kim2023pope}: text-davinci-003 on LUDWIG, 0.63 zero-shot, 0.71 few-shot, 0.87 with chain-of-thought.
  \item \textbf{ImplicatureX} \citep[][Table 6]{spinoso2026evaluating}: Qwen3-8B scalar-implicature recognition 1.000 without thinking and 0.804 with a 512-token budget; naturally occurring conversational implicatures 0.420 and 0.380.
  \item \textbf{Option removal} \citep[][§4, Fig.~6a]{gong2026think}: on HiToM, Qwen3-8B in reasoning mode goes from 0.481 to 0.629 without options; the non-reasoning model goes from 0.557 to 0.510.
\end{itemize}
Prompts, shots and decoding for each re-run follow the original papers except for the model; deviations are listed with the prompts in \Cref{app:details}.

\subsection{Verbatim Prompts}
\label{app:prompts}
Each block shows one fully rendered prompt per format or arm (user turn and, where the arm adds one, the assistant prefix), for one example item.
\paragraph{Yes no inference.} Yes/No inference format (felicity framing), forced-choice letters; the arm tail is appended after the user turn.
\begin{footnotesize}
\begin{verbatim}
A speaker says: "Some teenagers criticize Guy."
Second statement: "Not all teenagers criticize
  Guy."

Would you conclude that the speaker believes the
  second statement is true? Answer Yes or No.

Options:
A. Yes
B. No

Respond with only the letter of the best answer.
\end{verbatim}
\end{footnotesize}
\paragraph{Three way choice.} polarity-free three-way format (NLI): the second statement follows / may be true or false / must be false
\begin{footnotesize}
\begin{verbatim}
First sentence: Some teenagers criticize Guy.
Second sentence: Not all teenagers criticize Guy.

Assume the first sentence is true. Does the second
  sentence follow?

Options:
A. Yes, the second sentence must be true if the
  first is true.
B. Not necessarily: the second sentence may be
  true or false.
C. No, the second sentence must be false if the
  first is true.

Respond with only the letter of the best answer.
\end{verbatim}
\end{footnotesize}
\paragraph{Flipped question.} polarity-flipped felicity question (yes-bias probe): literal answer Yes, pragmatic answer No
\begin{footnotesize}
\begin{verbatim}
A speaker says: "Some guys were cooking."
Second statement: "Not all guys were cooking."

After the speaker's statement, is it still left
  open whether the second statement is true?
  Answer Yes or No.

Options:
A. Yes
B. No

Respond with only the letter of the best answer.
\end{verbatim}
\end{footnotesize}
\paragraph{Circa multiple choice.} multiple-choice format with the dataset's own options
\begin{footnotesize}
\begin{verbatim}
Context: Y has just told X that he/she is
  considering switching his/her job.
X asks: "Do you want a new challenge?"
Y answers: "I've been so bored t work."

What does Y's answer mean?

Options:
A. Yes
B. No
C. Neither yes nor no (in the middle)

Respond with only the letter of the best answer.
\end{verbatim}
\end{footnotesize}
\paragraph{Pub multiple choice.} multiple-choice format with the dataset's own options
\begin{footnotesize}
\begin{verbatim}
Context: Y has just travelled from a different
  city to meet X.
X: Do you want to catch a Christmas movie later?
Y: I hope it is funny.

Choose the best option.

Options:
A. Yes
B. No
C. Yes, subject to some conditions
D. In the middle, neither yes nor no
E. Other

Respond with only the letter of the best answer.
\end{verbatim}
\end{footnotesize}
\paragraph{Budget stop phrase.} think\_b<N> (non-plain): the thought is cut at N tokens and this stop phrase is appended before the close tag; shown for N=0 (empty thought)
\begin{footnotesize}
\begin{verbatim}
<|im_start|>user
Sentence: Some elephants are mammals.

Is this sentence a good way to describe the facts?
  Answer Yes or No.

Options:
A. No
B. Yes

Respond with only the letter of the best
  answer.<|im_end|>
<|im_start|>assistant
<think>


Considering the limited time by the user, I have
  to give the solution based on the thinking
  directly now.
</think>


--> 'B. Yes'
\end{verbatim}
\end{footnotesize}
\paragraph{Trace prefix forced close.} think\_b<N>\_plain: the first N thinking tokens of the SAME trace, then the native close tag (no stop phrase)
\begin{footnotesize}
\begin{verbatim}
<|im_start|>user
Sentence: Some elephants are mammals.

Is this sentence a good way to describe the facts?
  Answer Yes or No.

Options:
A. No
B. Yes

Respond with only the letter of the best
  answer.<|im_end|>
<|im_start|>assistant
<think>
Okay, let's see. The question is whether the
  sentence "Some elephants are mammals" is a good
  way to describe the facts. The options are A
  (No) or B (Yes). I need to answer with just the
  letter.

First, I should recall what elephants are
  classified as. Elephants are a type of mammal.
  The scientific classification of elephants is
  under the class Mammalia. So, all elephants are
  mammals. The sentence says "some elephants are
  mammals," which is technically correct because
  it's a subset. But wait, if all elephants are
  mammals, then saying "some" is still true,
  right? Because "some" means at least one. But is
  it a good way to describe the facts? The
  original statement might be considered accurate,
  but maybe it's not the most precise. However,
  the question is about whether it's a good way to
  describe the facts. Since all elephants are
  mammals, the statement is true, but it's not the
  most comprehensive. However, the question is
  asking if it's a good way, not if it's the best.
  So, if the facts are that all elephants are
  mammals, then "some" is still correct.
  Therefore, the answer would be Yes. But wait,
  maybe the answer is No</think>


--> 'B. Yes'
\end{verbatim}
\end{footnotesize}
\paragraph{Nt off.} thinking switched off: empty thought block in the chat template
\begin{footnotesize}
\begin{verbatim}
<|im_start|>user
Sentence: Some elephants are mammals.

Is this sentence a good way to describe the facts?
  Answer Yes or No.

Options:
A. No
B. Yes

Respond with only the letter of the best
  answer.<|im_end|>
<|im_start|>assistant
<think>

</think>


--> 'B. Yes'
\end{verbatim}
\end{footnotesize}
\paragraph{Forced empty nt.} forced-empty thinking (think\_b0\_plain): the native close tag directly after the opener
\begin{footnotesize}
\begin{verbatim}
<|im_start|>user
A speaker says: "Some teenagers criticize Guy."
Second statement: "Not all teenagers criticize
  Guy."

Would you conclude that the speaker believes the
  second statement is true? Answer Yes or No.

Options:
A. No
B. Yes

Respond with only the letter of the best
  answer.<|im_end|>
<|im_start|>assistant
<think>
</think>


--> 'B'
\end{verbatim}
\end{footnotesize}
\paragraph{Nat.} 
\begin{footnotesize}
\begin{verbatim}
<|im_start|>user
A speaker says: "All teenagers criticize Guy."
Second statement: "Some teenagers criticize Guy."

Would you conclude that the speaker believes the
  second statement is false? Answer Yes or No.

Options:
A. No
B. Yes

Respond with only the letter of the best
  answer.<|im_end|>
<|im_start|>assistant
<think>
Okay, let's try to figure this out. So, the first
  statement is "All teenagers criticize Guy," and
  the second is "Some teenagers criticize Guy."
  The question is whether the speaker believes the
  second statement is false. The options are A
  (No) or B (Yes).

Hmm. Let me think. If someone says "All teenagers
  criticize Guy," does that mean they think "Some
  teenagers criticize Guy" is false? Well, in
  logic, "All" implies that there is at least one.
  So if all teenagers do something, then there
  must be some teenagers doing it. So if the first
  statement is true, then the second statement
  should also be true. Because "some" is a weaker
  statement. So if all teenagers do it, then
  certainly some do it. Therefore, if the speaker
  says "All teenagers criticize Guy," they must
  believe that "Some teenagers criticize Guy" is
  true, not false. Therefore, the speaker does not
  believe the second statement is false. So the
  answer would be No, which is option A. Wait, but
  wait. What if the speaker is lying? Wait, the
  question is about what the speaker believes. If
  the speaker says "All teenagers criticize Guy,"
  then their belief is that all do. But does that
  mean they think that "some" is false? No,
  because "some" is a subset of "all." So if all
  are, then some are. Therefore, the speaker would
  believe that the second statement is true, not
  false. Therefore, the answer is No, so option A.
  But wait, the options are A. No and B. Yes. So
  the question is whether the speaker believes the
  second statement is false. If the speaker
  believes the first statement (all), then they
  would believe the second (some) is true, so they
  don't believe the second is false. So the answer
  is No, so option A.
</think>


--> 'A. No'
\end{verbatim}
\end{footnotesize}

\FloatBarrier
\section{Additional Results}
Full per-model results and further analyses behind the main text.

\subsection{Additional Results}
\label{app:results}

\paragraph{Models outside the count.}\label{app:outside} Of those that move the other way, Magistral-Small mostly shifts towards answering \emph{Yes}: with the question flipped so that \emph{No} is pragmatic, its pragmatic rate falls by \res{e1probe.magistral-small.a3.infflip.nat.dR.mean}\,pp. Granite-3.3 and Llama-3.1-Nemotron-Nano-8B move towards the pragmatic reading only in the Yes/No format: in a three-way choice both turn literal (\form{some}: \res{e1sfx.granite-3p3-8b.a3f-quant.nli.nat.dR.mean} and \res{e1sfx.llama-nemotron-nano-8b.a3f-quant.nli.nat.dR.mean}\,pp; \Cref{tab:fullsets}). The rise of openPangu-Embedded-7B-model is, like Magistral's, a shift towards answering \emph{Yes}: with the question flipped its pragmatic rate falls, and in the three-way choice it is flat. In the three-way format, which has no Yes/No labels, \res{derived.fullsets.nli.drop} of the \res{derived.fullsets.nli.tested} models with a full-set run shift towards the literal reading on \form{some} or \form{or}, and only two rise on one word: SmolLM3-3B from its floor (\form{some}: +\res{e1sfx.smollm3-3b.a3f-quant.nli.nat.dR.mean}\,pp) and Qwen3.5-0.8B on \form{some} (+\res{e1sfx.qwen3p5-0p8b.a3f-quant.nli.nat.dR.mean}\,pp), whose \form{or} reading falls (\res{e1sfx.qwen3p5-0p8b.a3f-conn.nli.nat.dR.mean}\,pp). openPangu-Embedded-7B-V1.1, null on the core items, shifts on its full held-out set (\form{some}: \res{e1pangu.openpangu-embedded-7b-v1p1.a3f-quant.inf.nat.dR.mean}\,pp) and in the three-way choice (\res{e1pangu.openpangu-embedded-7b-v1p1.a3f-quant.nli.nat.dR.mean}\,pp).
\begin{table*}[t]
  \centering\small
  \setlength{\tabcolsep}{4pt}
  \begin{tabular}{@{}l cc cc@{}}
    \toprule
    & \multicolumn{2}{c}{Yes/No inference} & \multicolumn{2}{c}{Three-way choice} \\
    \cmidrule(lr){2-3}\cmidrule(l){4-5}
    Model & \form{some} & \form{or} & \form{some} & \form{or} \\
    \midrule
    Qwen3-4B & \cellcolor{cInterv!23}\res{e1sf.qwen3-4b.a3f-quant.inf.nat.dR.mean}{\scriptsize\,[\res{e1sf.qwen3-4b.a3f-quant.inf.nat.dR.lo}, \res{e1sf.qwen3-4b.a3f-quant.inf.nat.dR.hi}]} & \cellcolor{cInterv!23}\res{e1sf.qwen3-4b.a3f-conn.inf.nat.dR.mean}{\scriptsize\,[\res{e1sf.qwen3-4b.a3f-conn.inf.nat.dR.lo}, \res{e1sf.qwen3-4b.a3f-conn.inf.nat.dR.hi}]} & \cellcolor{cInterv!15}\res{e1sf.qwen3-4b.a3f-quant.nli.nat.dR.mean}{\scriptsize\,[\res{e1sf.qwen3-4b.a3f-quant.nli.nat.dR.lo}, \res{e1sf.qwen3-4b.a3f-quant.nli.nat.dR.hi}]} & \cellcolor{cInterv!16}\res{e1sf.qwen3-4b.a3f-conn.nli.nat.dR.mean}{\scriptsize\,[\res{e1sf.qwen3-4b.a3f-conn.nli.nat.dR.lo}, \res{e1sf.qwen3-4b.a3f-conn.nli.nat.dR.hi}]} \\
    Qwen3-8B & \cellcolor{cInterv!21}\res{e1sf.qwen3-8b.a3f-quant.inf.nat.dR.mean}{\scriptsize\,[\res{e1sf.qwen3-8b.a3f-quant.inf.nat.dR.lo}, \res{e1sf.qwen3-8b.a3f-quant.inf.nat.dR.hi}]} & \cellcolor{cInterv!21}\res{e1sf.qwen3-8b.a3f-conn.inf.nat.dR.mean}{\scriptsize\,[\res{e1sf.qwen3-8b.a3f-conn.inf.nat.dR.lo}, \res{e1sf.qwen3-8b.a3f-conn.inf.nat.dR.hi}]} & \cellcolor{cInterv!21}\res{e1sf.qwen3-8b.a3f-quant.nli.nat.dR.mean}{\scriptsize\,[\res{e1sf.qwen3-8b.a3f-quant.nli.nat.dR.lo}, \res{e1sf.qwen3-8b.a3f-quant.nli.nat.dR.hi}]} & \cellcolor{cInterv!16}\res{e1sf.qwen3-8b.a3f-conn.nli.nat.dR.mean}{\scriptsize\,[\res{e1sf.qwen3-8b.a3f-conn.nli.nat.dR.lo}, \res{e1sf.qwen3-8b.a3f-conn.nli.nat.dR.hi}]} \\
    Qwen3-14B & \cellcolor{cInterv!19}\res{e1sf.qwen3-14b.a3f-quant.inf.nat.dR.mean}{\scriptsize\,[\res{e1sf.qwen3-14b.a3f-quant.inf.nat.dR.lo}, \res{e1sf.qwen3-14b.a3f-quant.inf.nat.dR.hi}]} & \cellcolor{cInterv!17}\res{e1sf.qwen3-14b.a3f-conn.inf.nat.dR.mean}{\scriptsize\,[\res{e1sf.qwen3-14b.a3f-conn.inf.nat.dR.lo}, \res{e1sf.qwen3-14b.a3f-conn.inf.nat.dR.hi}]} & \cellcolor{cInterv!16}\res{e1sf.qwen3-14b.a3f-quant.nli.nat.dR.mean}{\scriptsize\,[\res{e1sf.qwen3-14b.a3f-quant.nli.nat.dR.lo}, \res{e1sf.qwen3-14b.a3f-quant.nli.nat.dR.hi}]} & \cellcolor{cInterv!9}\res{e1sf.qwen3-14b.a3f-conn.nli.nat.dR.mean}{\scriptsize\,[\res{e1sf.qwen3-14b.a3f-conn.nli.nat.dR.lo}, \res{e1sf.qwen3-14b.a3f-conn.nli.nat.dR.hi}]} \\
    Qwen3-30B-A3B & \cellcolor{cInterv!21}\res{e1sf.qwen3-30b-a3b.a3f-quant.inf.nat.dR.mean}{\scriptsize\,[\res{e1sf.qwen3-30b-a3b.a3f-quant.inf.nat.dR.lo}, \res{e1sf.qwen3-30b-a3b.a3f-quant.inf.nat.dR.hi}]} & \cellcolor{cInterv!22}\res{e1sf.qwen3-30b-a3b.a3f-conn.inf.nat.dR.mean}{\scriptsize\,[\res{e1sf.qwen3-30b-a3b.a3f-conn.inf.nat.dR.lo}, \res{e1sf.qwen3-30b-a3b.a3f-conn.inf.nat.dR.hi}]} & \cellcolor{cInterv!16}\res{e1sf.qwen3-30b-a3b.a3f-quant.nli.nat.dR.mean}{\scriptsize\,[\res{e1sf.qwen3-30b-a3b.a3f-quant.nli.nat.dR.lo}, \res{e1sf.qwen3-30b-a3b.a3f-quant.nli.nat.dR.hi}]} & \cellcolor{cInterv!16}\res{e1sf.qwen3-30b-a3b.a3f-conn.nli.nat.dR.mean}{\scriptsize\,[\res{e1sf.qwen3-30b-a3b.a3f-conn.nli.nat.dR.lo}, \res{e1sf.qwen3-30b-a3b.a3f-conn.nli.nat.dR.hi}]} \\
    Qwen3-32B & \cellcolor{cInterv!23}\res{e1sf.qwen3-32b.a3f-quant.inf.nat.dR.mean}{\scriptsize\,[\res{e1sf.qwen3-32b.a3f-quant.inf.nat.dR.lo}, \res{e1sf.qwen3-32b.a3f-quant.inf.nat.dR.hi}]} & \cellcolor{cInterv!23}\res{e1sf.qwen3-32b.a3f-conn.inf.nat.dR.mean}{\scriptsize\,[\res{e1sf.qwen3-32b.a3f-conn.inf.nat.dR.lo}, \res{e1sf.qwen3-32b.a3f-conn.inf.nat.dR.hi}]} & \cellcolor{cInterv!21}\res{e1sf.qwen3-32b.a3f-quant.nli.nat.dR.mean}{\scriptsize\,[\res{e1sf.qwen3-32b.a3f-quant.nli.nat.dR.lo}, \res{e1sf.qwen3-32b.a3f-quant.nli.nat.dR.hi}]} & \cellcolor{cInterv!18}\res{e1sf.qwen3-32b.a3f-conn.nli.nat.dR.mean}{\scriptsize\,[\res{e1sf.qwen3-32b.a3f-conn.nli.nat.dR.lo}, \res{e1sf.qwen3-32b.a3f-conn.nli.nat.dR.hi}]} \\
    \midrule
    Qwen3.5-0.8B & \cellcolor{cInterv!12}\res{e1sfx.qwen3p5-0p8b.a3f-quant.inf.nat.dR.mean}{\scriptsize\,[\res{e1sfx.qwen3p5-0p8b.a3f-quant.inf.nat.dR.lo}, \res{e1sfx.qwen3p5-0p8b.a3f-quant.inf.nat.dR.hi}]} & \cellcolor{cInterv!17}\res{e1sfx.qwen3p5-0p8b.a3f-conn.inf.nat.dR.mean}{\scriptsize\,[\res{e1sfx.qwen3p5-0p8b.a3f-conn.inf.nat.dR.lo}, \res{e1sfx.qwen3p5-0p8b.a3f-conn.inf.nat.dR.hi}]} & \cellcolor{cOurs!15}\res{e1sfx.qwen3p5-0p8b.a3f-quant.nli.nat.dR.mean}{\scriptsize\,[\res{e1sfx.qwen3p5-0p8b.a3f-quant.nli.nat.dR.lo}, \res{e1sfx.qwen3p5-0p8b.a3f-quant.nli.nat.dR.hi}]} & \cellcolor{cInterv!14}\res{e1sfx.qwen3p5-0p8b.a3f-conn.nli.nat.dR.mean}{\scriptsize\,[\res{e1sfx.qwen3p5-0p8b.a3f-conn.nli.nat.dR.lo}, \res{e1sfx.qwen3p5-0p8b.a3f-conn.nli.nat.dR.hi}]} \\
    Qwen3.5-2B & \cellcolor{cInterv!20}\res{e1sfx.qwen3p5-2b.a3f-quant.inf.nat.dR.mean}{\scriptsize\,[\res{e1sfx.qwen3p5-2b.a3f-quant.inf.nat.dR.lo}, \res{e1sfx.qwen3p5-2b.a3f-quant.inf.nat.dR.hi}]} & \cellcolor{cInterv!19}\res{e1sfx.qwen3p5-2b.a3f-conn.inf.nat.dR.mean}{\scriptsize\,[\res{e1sfx.qwen3p5-2b.a3f-conn.inf.nat.dR.lo}, \res{e1sfx.qwen3p5-2b.a3f-conn.inf.nat.dR.hi}]} & \cellcolor{cInterv!14}\res{e1sfx.qwen3p5-2b.a3f-quant.nli.nat.dR.mean}{\scriptsize\,[\res{e1sfx.qwen3p5-2b.a3f-quant.nli.nat.dR.lo}, \res{e1sfx.qwen3p5-2b.a3f-quant.nli.nat.dR.hi}]} & \cellcolor{cInterv!14}\res{e1sfx.qwen3p5-2b.a3f-conn.nli.nat.dR.mean}{\scriptsize\,[\res{e1sfx.qwen3p5-2b.a3f-conn.nli.nat.dR.lo}, \res{e1sfx.qwen3p5-2b.a3f-conn.nli.nat.dR.hi}]} \\
    Qwen3.5-4B & \cellcolor{cInterv!23}\res{e1sfx.qwen3p5-4b.a3f-quant.inf.nat.dR.mean}{\scriptsize\,[\res{e1sfx.qwen3p5-4b.a3f-quant.inf.nat.dR.lo}, \res{e1sfx.qwen3p5-4b.a3f-quant.inf.nat.dR.hi}]} & \cellcolor{cInterv!20}\res{e1sfx.qwen3p5-4b.a3f-conn.inf.nat.dR.mean}{\scriptsize\,[\res{e1sfx.qwen3p5-4b.a3f-conn.inf.nat.dR.lo}, \res{e1sfx.qwen3p5-4b.a3f-conn.inf.nat.dR.hi}]} & \cellcolor{cInterv!16}\res{e1sfx.qwen3p5-4b.a3f-quant.nli.nat.dR.mean}{\scriptsize\,[\res{e1sfx.qwen3p5-4b.a3f-quant.nli.nat.dR.lo}, \res{e1sfx.qwen3p5-4b.a3f-quant.nli.nat.dR.hi}]} & \cellcolor{cInterv!17}\res{e1sfx.qwen3p5-4b.a3f-conn.nli.nat.dR.mean}{\scriptsize\,[\res{e1sfx.qwen3p5-4b.a3f-conn.nli.nat.dR.lo}, \res{e1sfx.qwen3p5-4b.a3f-conn.nli.nat.dR.hi}]} \\
    Qwen3.5-9B & \cellcolor{cInterv!20}\res{e1sfx.qwen3p5-9b.a3f-quant.inf.nat.dR.mean}{\scriptsize\,[\res{e1sfx.qwen3p5-9b.a3f-quant.inf.nat.dR.lo}, \res{e1sfx.qwen3p5-9b.a3f-quant.inf.nat.dR.hi}]} & \cellcolor{cInterv!19}\res{e1sfx.qwen3p5-9b.a3f-conn.inf.nat.dR.mean}{\scriptsize\,[\res{e1sfx.qwen3p5-9b.a3f-conn.inf.nat.dR.lo}, \res{e1sfx.qwen3p5-9b.a3f-conn.inf.nat.dR.hi}]} & \cellcolor{cInterv!15}\res{e1sfx.qwen3p5-9b.a3f-quant.nli.nat.dR.mean}{\scriptsize\,[\res{e1sfx.qwen3p5-9b.a3f-quant.nli.nat.dR.lo}, \res{e1sfx.qwen3p5-9b.a3f-quant.nli.nat.dR.hi}]} & \cellcolor{cInterv!13}\res{e1sfx.qwen3p5-9b.a3f-conn.nli.nat.dR.mean}{\scriptsize\,[\res{e1sfx.qwen3p5-9b.a3f-conn.nli.nat.dR.lo}, \res{e1sfx.qwen3p5-9b.a3f-conn.nli.nat.dR.hi}]} \\
    Qwen3.8-27B & \cellcolor{cInterv!16}\res{e1sfx.qwen3p8-27b.a3f-quant.inf.nat.dR.mean}{\scriptsize\,[\res{e1sfx.qwen3p8-27b.a3f-quant.inf.nat.dR.lo}, \res{e1sfx.qwen3p8-27b.a3f-quant.inf.nat.dR.hi}]} & \cellcolor{cInterv!19}\res{e1sfx.qwen3p8-27b.a3f-conn.inf.nat.dR.mean}{\scriptsize\,[\res{e1sfx.qwen3p8-27b.a3f-conn.inf.nat.dR.lo}, \res{e1sfx.qwen3p8-27b.a3f-conn.inf.nat.dR.hi}]} & \cellcolor{cInterv!7}\res{e1sfx.qwen3p8-27b.a3f-quant.nli.nat.dR.mean}{\scriptsize\,[\res{e1sfx.qwen3p8-27b.a3f-quant.nli.nat.dR.lo}, \res{e1sfx.qwen3p8-27b.a3f-quant.nli.nat.dR.hi}]} & \cellcolor{cInterv!10}\res{e1sfx.qwen3p8-27b.a3f-conn.nli.nat.dR.mean}{\scriptsize\,[\res{e1sfx.qwen3p8-27b.a3f-conn.nli.nat.dR.lo}, \res{e1sfx.qwen3p8-27b.a3f-conn.nli.nat.dR.hi}]} \\
    \midrule
    Phi-4-reasoning & \cellcolor{cInterv!21}\res{e1sfx.phi4-reasoning.a3f-quant.inf.nat.dR.mean}{\scriptsize\,[\res{e1sfx.phi4-reasoning.a3f-quant.inf.nat.dR.lo}, \res{e1sfx.phi4-reasoning.a3f-quant.inf.nat.dR.hi}]} & \cellcolor{cInterv!22}\res{e1sfx.phi4-reasoning.a3f-conn.inf.nat.dR.mean}{\scriptsize\,[\res{e1sfx.phi4-reasoning.a3f-conn.inf.nat.dR.lo}, \res{e1sfx.phi4-reasoning.a3f-conn.inf.nat.dR.hi}]} & \cellcolor{cInterv!20}\res{e1sfx.phi4-reasoning.a3f-quant.nli.nat.dR.mean}{\scriptsize\,[\res{e1sfx.phi4-reasoning.a3f-quant.nli.nat.dR.lo}, \res{e1sfx.phi4-reasoning.a3f-quant.nli.nat.dR.hi}]} & \cellcolor{cInterv!19}\res{e1sfx.phi4-reasoning.a3f-conn.nli.nat.dR.mean}{\scriptsize\,[\res{e1sfx.phi4-reasoning.a3f-conn.nli.nat.dR.lo}, \res{e1sfx.phi4-reasoning.a3f-conn.nli.nat.dR.hi}]} \\
    Phi-4-mini-reasoning & \cellcolor{cInterv!18}\res{e1sfx.phi4-mini-reasoning.a3f-quant.inf.nat.dR.mean}{\scriptsize\,[\res{e1sfx.phi4-mini-reasoning.a3f-quant.inf.nat.dR.lo}, \res{e1sfx.phi4-mini-reasoning.a3f-quant.inf.nat.dR.hi}]} & \cellcolor{cInterv!17}\res{e1sfx.phi4-mini-reasoning.a3f-conn.inf.nat.dR.mean}{\scriptsize\,[\res{e1sfx.phi4-mini-reasoning.a3f-conn.inf.nat.dR.lo}, \res{e1sfx.phi4-mini-reasoning.a3f-conn.inf.nat.dR.hi}]} & \cellcolor{cInterv!18}\res{e1sfx.phi4-mini-reasoning.a3f-quant.nli.nat.dR.mean}{\scriptsize\,[\res{e1sfx.phi4-mini-reasoning.a3f-quant.nli.nat.dR.lo}, \res{e1sfx.phi4-mini-reasoning.a3f-quant.nli.nat.dR.hi}]} & \cellcolor{cInterv!19}\res{e1sfx.phi4-mini-reasoning.a3f-conn.nli.nat.dR.mean}{\scriptsize\,[\res{e1sfx.phi4-mini-reasoning.a3f-conn.nli.nat.dR.lo}, \res{e1sfx.phi4-mini-reasoning.a3f-conn.nli.nat.dR.hi}]} \\
    R1-Distill-Llama-8B & \cellcolor{cInterv!23}\res{e1sfx.r1distill-llama-8b.a3f-quant.inf.nat.dR.mean}{\scriptsize\,[\res{e1sfx.r1distill-llama-8b.a3f-quant.inf.nat.dR.lo}, \res{e1sfx.r1distill-llama-8b.a3f-quant.inf.nat.dR.hi}]} & \cellcolor{cInterv!24}\res{e1sfx.r1distill-llama-8b.a3f-conn.inf.nat.dR.mean}{\scriptsize\,[\res{e1sfx.r1distill-llama-8b.a3f-conn.inf.nat.dR.lo}, \res{e1sfx.r1distill-llama-8b.a3f-conn.inf.nat.dR.hi}]} & \cellcolor{cInterv!20}\res{e1sfx.r1distill-llama-8b.a3f-quant.nli.nat.dR.mean}{\scriptsize\,[\res{e1sfx.r1distill-llama-8b.a3f-quant.nli.nat.dR.lo}, \res{e1sfx.r1distill-llama-8b.a3f-quant.nli.nat.dR.hi}]} & \cellcolor{cInterv!21}\res{e1sfx.r1distill-llama-8b.a3f-conn.nli.nat.dR.mean}{\scriptsize\,[\res{e1sfx.r1distill-llama-8b.a3f-conn.nli.nat.dR.lo}, \res{e1sfx.r1distill-llama-8b.a3f-conn.nli.nat.dR.hi}]} \\
    Skywork-OR1-7B & \cellcolor{cInterv!18}\res{e1sfx.skywork-or1-7b.a3f-quant.inf.nat.dR.mean}{\scriptsize\,[\res{e1sfx.skywork-or1-7b.a3f-quant.inf.nat.dR.lo}, \res{e1sfx.skywork-or1-7b.a3f-quant.inf.nat.dR.hi}]} & \cellcolor{cInterv!20}\res{e1sfx.skywork-or1-7b.a3f-conn.inf.nat.dR.mean}{\scriptsize\,[\res{e1sfx.skywork-or1-7b.a3f-conn.inf.nat.dR.lo}, \res{e1sfx.skywork-or1-7b.a3f-conn.inf.nat.dR.hi}]} & \cellcolor{cInterv!20}\res{e1sfx.skywork-or1-7b.a3f-quant.nli.nat.dR.mean}{\scriptsize\,[\res{e1sfx.skywork-or1-7b.a3f-quant.nli.nat.dR.lo}, \res{e1sfx.skywork-or1-7b.a3f-quant.nli.nat.dR.hi}]} & \cellcolor{cInterv!21}\res{e1sfx.skywork-or1-7b.a3f-conn.nli.nat.dR.mean}{\scriptsize\,[\res{e1sfx.skywork-or1-7b.a3f-conn.nli.nat.dR.lo}, \res{e1sfx.skywork-or1-7b.a3f-conn.nli.nat.dR.hi}]} \\
    EXAONE-Deep-7.8B & \cellcolor{cInterv!21}\res{e1sfx.exaone-deep-7p8b.a3f-quant.inf.nat.dR.mean}{\scriptsize\,[\res{e1sfx.exaone-deep-7p8b.a3f-quant.inf.nat.dR.lo}, \res{e1sfx.exaone-deep-7p8b.a3f-quant.inf.nat.dR.hi}]} & \cellcolor{cInterv!21}\res{e1sfx.exaone-deep-7p8b.a3f-conn.inf.nat.dR.mean}{\scriptsize\,[\res{e1sfx.exaone-deep-7p8b.a3f-conn.inf.nat.dR.lo}, \res{e1sfx.exaone-deep-7p8b.a3f-conn.inf.nat.dR.hi}]} & \cellcolor{cInterv!20}\res{e1sfx.exaone-deep-7p8b.a3f-quant.nli.nat.dR.mean}{\scriptsize\,[\res{e1sfx.exaone-deep-7p8b.a3f-quant.nli.nat.dR.lo}, \res{e1sfx.exaone-deep-7p8b.a3f-quant.nli.nat.dR.hi}]} & \cellcolor{cInterv!20}\res{e1sfx.exaone-deep-7p8b.a3f-conn.nli.nat.dR.mean}{\scriptsize\,[\res{e1sfx.exaone-deep-7p8b.a3f-conn.nli.nat.dR.lo}, \res{e1sfx.exaone-deep-7p8b.a3f-conn.nli.nat.dR.hi}]} \\
    Nemotron-Nano-9B-v2 & \cellcolor{cInterv!23}\res{e1sfx.nemotron-nano-9b-v2.a3f-quant.inf.nat.dR.mean}{\scriptsize\,[\res{e1sfx.nemotron-nano-9b-v2.a3f-quant.inf.nat.dR.lo}, \res{e1sfx.nemotron-nano-9b-v2.a3f-quant.inf.nat.dR.hi}]} & \cellcolor{cInterv!22}\res{e1sfx.nemotron-nano-9b-v2.a3f-conn.inf.nat.dR.mean}{\scriptsize\,[\res{e1sfx.nemotron-nano-9b-v2.a3f-conn.inf.nat.dR.lo}, \res{e1sfx.nemotron-nano-9b-v2.a3f-conn.inf.nat.dR.hi}]} & \cellcolor{cOurs!3}\textcolor{gray}{\res{e1sfx.nemotron-nano-9b-v2.a3f-quant.nli.nat.dR.mean}{\scriptsize\,[\res{e1sfx.nemotron-nano-9b-v2.a3f-quant.nli.nat.dR.lo}, \res{e1sfx.nemotron-nano-9b-v2.a3f-quant.nli.nat.dR.hi}]}} & \cellcolor{cInterv!6}\textcolor{gray}{\res{e1sfx.nemotron-nano-9b-v2.a3f-conn.nli.nat.dR.mean}{\scriptsize\,[\res{e1sfx.nemotron-nano-9b-v2.a3f-conn.nli.nat.dR.lo}, \res{e1sfx.nemotron-nano-9b-v2.a3f-conn.nli.nat.dR.hi}]}} \\
    Granite-3.3-8B & \cellcolor{cOurs!16}\res{e1sfx.granite-3p3-8b.a3f-quant.inf.nat.dR.mean}{\scriptsize\,[\res{e1sfx.granite-3p3-8b.a3f-quant.inf.nat.dR.lo}, \res{e1sfx.granite-3p3-8b.a3f-quant.inf.nat.dR.hi}]} & \cellcolor{cOurs!18}\res{e1sfx.granite-3p3-8b.a3f-conn.inf.nat.dR.mean}{\scriptsize\,[\res{e1sfx.granite-3p3-8b.a3f-conn.inf.nat.dR.lo}, \res{e1sfx.granite-3p3-8b.a3f-conn.inf.nat.dR.hi}]} & \cellcolor{cInterv!19}\res{e1sfx.granite-3p3-8b.a3f-quant.nli.nat.dR.mean}{\scriptsize\,[\res{e1sfx.granite-3p3-8b.a3f-quant.nli.nat.dR.lo}, \res{e1sfx.granite-3p3-8b.a3f-quant.nli.nat.dR.hi}]} & \cellcolor{cInterv!14}\res{e1sfx.granite-3p3-8b.a3f-conn.nli.nat.dR.mean}{\scriptsize\,[\res{e1sfx.granite-3p3-8b.a3f-conn.nli.nat.dR.lo}, \res{e1sfx.granite-3p3-8b.a3f-conn.nli.nat.dR.hi}]} \\
    Llama-3.1-Nemotron-Nano-8B & \cellcolor{cOurs!16}\res{e1sfx.llama-nemotron-nano-8b.a3f-quant.inf.nat.dR.mean}{\scriptsize\,[\res{e1sfx.llama-nemotron-nano-8b.a3f-quant.inf.nat.dR.lo}, \res{e1sfx.llama-nemotron-nano-8b.a3f-quant.inf.nat.dR.hi}]} & \cellcolor{cOurs!11}\res{e1sfx.llama-nemotron-nano-8b.a3f-conn.inf.nat.dR.mean}{\scriptsize\,[\res{e1sfx.llama-nemotron-nano-8b.a3f-conn.inf.nat.dR.lo}, \res{e1sfx.llama-nemotron-nano-8b.a3f-conn.inf.nat.dR.hi}]} & \cellcolor{cInterv!14}\res{e1sfx.llama-nemotron-nano-8b.a3f-quant.nli.nat.dR.mean}{\scriptsize\,[\res{e1sfx.llama-nemotron-nano-8b.a3f-quant.nli.nat.dR.lo}, \res{e1sfx.llama-nemotron-nano-8b.a3f-quant.nli.nat.dR.hi}]} & \cellcolor{cInterv!11}\res{e1sfx.llama-nemotron-nano-8b.a3f-conn.nli.nat.dR.mean}{\scriptsize\,[\res{e1sfx.llama-nemotron-nano-8b.a3f-conn.nli.nat.dR.lo}, \res{e1sfx.llama-nemotron-nano-8b.a3f-conn.nli.nat.dR.hi}]} \\
    SmolLM3-3B & \cellcolor{cInterv!14}\res{e1sfx.smollm3-3b.a3f-quant.inf.nat.dR.mean}{\scriptsize\,[\res{e1sfx.smollm3-3b.a3f-quant.inf.nat.dR.lo}, \res{e1sfx.smollm3-3b.a3f-quant.inf.nat.dR.hi}]} & \cellcolor{cOurs!11}\textcolor{gray}{\res{e1sfx.smollm3-3b.a3f-conn.inf.nat.dR.mean}{\scriptsize\,[\res{e1sfx.smollm3-3b.a3f-conn.inf.nat.dR.lo}, \res{e1sfx.smollm3-3b.a3f-conn.inf.nat.dR.hi}]}} & \cellcolor{cOurs!11}\res{e1sfx.smollm3-3b.a3f-quant.nli.nat.dR.mean}{\scriptsize\,[\res{e1sfx.smollm3-3b.a3f-quant.nli.nat.dR.lo}, \res{e1sfx.smollm3-3b.a3f-quant.nli.nat.dR.hi}]} & \cellcolor{cOurs!6}\textcolor{gray}{\res{e1sfx.smollm3-3b.a3f-conn.nli.nat.dR.mean}{\scriptsize\,[\res{e1sfx.smollm3-3b.a3f-conn.nli.nat.dR.lo}, \res{e1sfx.smollm3-3b.a3f-conn.nli.nat.dR.hi}]}} \\
    openPangu-Embedded-7B-V1.1 & \cellcolor{cInterv!17}\res{e1pangu.openpangu-embedded-7b-v1p1.a3f-quant.inf.nat.dR.mean}{\scriptsize\,[\res{e1pangu.openpangu-embedded-7b-v1p1.a3f-quant.inf.nat.dR.lo}, \res{e1pangu.openpangu-embedded-7b-v1p1.a3f-quant.inf.nat.dR.hi}]} & \cellcolor{cInterv!8}\textcolor{gray}{\res{e1pangu.openpangu-embedded-7b-v1p1.a3f-conn.inf.nat.dR.mean}{\scriptsize\,[\res{e1pangu.openpangu-embedded-7b-v1p1.a3f-conn.inf.nat.dR.lo}, \res{e1pangu.openpangu-embedded-7b-v1p1.a3f-conn.inf.nat.dR.hi}]}} & \cellcolor{cInterv!14}\res{e1pangu.openpangu-embedded-7b-v1p1.a3f-quant.nli.nat.dR.mean}{\scriptsize\,[\res{e1pangu.openpangu-embedded-7b-v1p1.a3f-quant.nli.nat.dR.lo}, \res{e1pangu.openpangu-embedded-7b-v1p1.a3f-quant.nli.nat.dR.hi}]} & \cellcolor{cInterv!10}\res{e1pangu.openpangu-embedded-7b-v1p1.a3f-conn.nli.nat.dR.mean}{\scriptsize\,[\res{e1pangu.openpangu-embedded-7b-v1p1.a3f-conn.nli.nat.dR.lo}, \res{e1pangu.openpangu-embedded-7b-v1p1.a3f-conn.nli.nat.dR.hi}]} \\
    openPangu-Embedded-7B-model & \cellcolor{cOurs!13}\res{e1pangu.openpangu-embedded-7b-model.a3f-quant.inf.nat.dR.mean}{\scriptsize\,[\res{e1pangu.openpangu-embedded-7b-model.a3f-quant.inf.nat.dR.lo}, \res{e1pangu.openpangu-embedded-7b-model.a3f-quant.inf.nat.dR.hi}]} & \cellcolor{cInterv!12}\res{e1pangu.openpangu-embedded-7b-model.a3f-conn.inf.nat.dR.mean}{\scriptsize\,[\res{e1pangu.openpangu-embedded-7b-model.a3f-conn.inf.nat.dR.lo}, \res{e1pangu.openpangu-embedded-7b-model.a3f-conn.inf.nat.dR.hi}]} & \cellcolor{cInterv!6}\textcolor{gray}{\res{e1pangu.openpangu-embedded-7b-model.a3f-quant.nli.nat.dR.mean}{\scriptsize\,[\res{e1pangu.openpangu-embedded-7b-model.a3f-quant.nli.nat.dR.lo}, \res{e1pangu.openpangu-embedded-7b-model.a3f-quant.nli.nat.dR.hi}]}} & \cellcolor{cInterv!3}\textcolor{gray}{\res{e1pangu.openpangu-embedded-7b-model.a3f-conn.nli.nat.dR.mean}{\scriptsize\,[\res{e1pangu.openpangu-embedded-7b-model.a3f-conn.nli.nat.dR.lo}, \res{e1pangu.openpangu-embedded-7b-model.a3f-conn.nli.nat.dR.hi}]}} \\
    \bottomrule
  \end{tabular}
  \caption{\textbf{On the full held-out sets, \form{some} and \form{or} turn literal with thinking in both formats in most models.} Change with thinking (pp) [95\% CI], 300 items per trigger; grey: CI includes 0. Some models start near the floor in the three-way format.}
  \label{tab:fullsets}
\end{table*}

\paragraph{Four robustness checks.}\label{app:checks} \emph{How thinking stops.} Qwen3's stop phrase mentions limited time. Stopping the trace with only the closing tag gives the same shift at 1{,}024 tokens (\res{e1.qwen3-8b.a3-quant.inf.p1024.dR.mean} against \res{e1.qwen3-8b.a3-quant.inf.c1024.dR.mean}\,pp with the stop phrase).
\emph{The baseline.} Against a forced-empty thinking block instead of the model's switch, natural thinking changes the \form{some} reading of Qwen3-8B by \res{e1.qwen3-8b.a3-quant.inf.nat.dRb0.mean}\,pp.
\emph{Answer bias.} A shift towards answering \emph{No} could imitate a literal reading. For GPT-5 the opposite holds: without the speaker's sentence, its rate of answering \emph{Yes} rises with effort, which would push the pragmatic rate up, not down (\Cref{tab:frontier-format}). For Qwen3-8B answer polarity does matter: with the question reworded so that \emph{No} is the pragmatic answer, thinking raises the pragmatic rate by \res{e1probe.qwen3-8b.a3.infflip.nat.dR.mean}\,pp from a floor near zero. In a three-way forced choice, which has no Yes/No labels, thinking still lowers the pragmatic reading at every Qwen3 size, in Qwen3-8B by \res{e1sf.qwen3-8b.a3f-quant.nli.nat.dR.mean}\,pp for \form{some} and \res{e1sf.qwen3-8b.a3f-conn.nli.nat.dR.mean}\,pp for \form{or} (\Cref{app:frontier}).
\emph{Speaker knowledge.} A pragmatic listener enriches more when the speaker has checked every case \citep{goodman2013knowledge}; the scenarios we built for this were screened blind to condition by the annotator panel, whose consensus judged \form{some} misleading in every full-knowledge scenario and fine in every partial-knowledge one (\res{gold.v4.n-a1ctx} scenarios). Most models make little use of this knowledge: with thinking, the gap between full and partial knowledge is \res{a1ctx.qwen3-32b.a1ctx-fel.fel.nat.ksi.mean}\,pp in Qwen3-32B. The one model that uses it without much deliberation, gpt-oss-120b (\res{a1ctx.gpt-oss-120b.a1ctx-fel.fel.low.ksi.mean}\,pp at low effort), uses it less at high effort (\res{a1ctx.gpt-oss-120b.a1ctx-fel.fel.high.ksi.mean}\,pp).

This section gives the full per-model results and additional analyses behind the main text.

\begin{figure*}[tp]
  \centering
  \IfFileExists{figures/figA_trace_newfam.pdf}{\includegraphics[width=\textwidth]{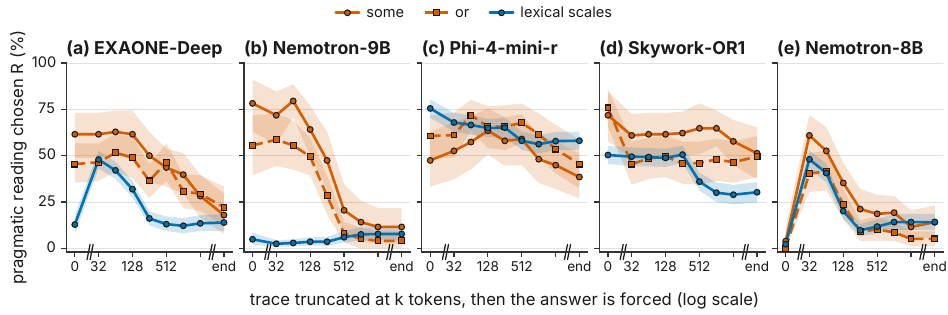}}{\figplaceholder{\textwidth}{4cm}}
  \caption{Trace probe on further families (EXAONE-Deep-7.8B, Nemotron-Nano-9B-v2, Phi-4-mini-reasoning, Skywork-OR1-7B, Llama-3.1-Nemotron-Nano-8B): pragmatic reading $\rate$ after the first $t$ tokens of the model's own trace, with 95\% bands over items. The $y$-axis runs from 0 to 100 (0 to 80 in \Cref{fig:trace}). In panel (e) the value at $t=0$ is the model's collapse to one answer under a forced-empty thinking block.}
  \label{fig:trace-newfam}
\end{figure*}

\begin{figure*}[tp]
  \centering
  \includegraphics[width=\textwidth]{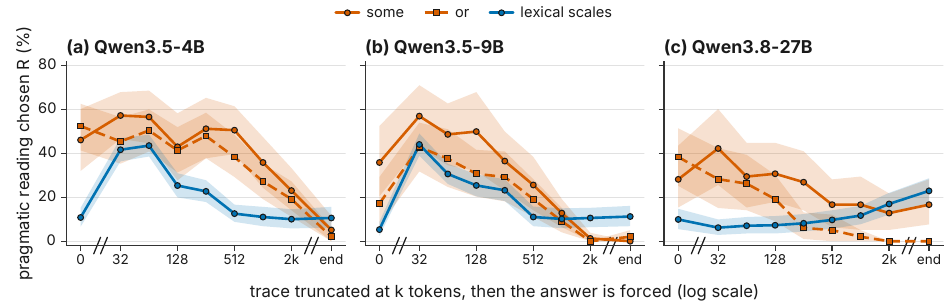}
  \caption{Trace probe on the newest generation (Qwen3.5-4B, Qwen3.5-9B, Qwen3.8-27B): pragmatic reading $\rate$ after the first $t$ tokens of the model's own trace, with 95\% bands over items. The early lexical-scale jump in the Qwen3.5 panels is systematic and confined to sampled answers (see text).}
  \label{fig:trace-newest}
\end{figure*}

\paragraph{Trace probe beyond Qwen3.}\label{app:trace} \Cref{fig:trace-newest} shows the trace probe on the newest generation. In \Cref{fig:trace}d, the jump of Phi-4-reasoning's lexical-scale reading after 32 tokens (\res{e7.phi4-reasoning.a2.inf.p0032.R.mean}\%) appears only in the sampled answers; the answer-position log-odds $L$ does not move (\res{e7.phi4-reasoning.a2.inf.ntforced.L.mean} with an empty trace, \res{e7.phi4-reasoning.a2.inf.p0032.L.mean} after 32 tokens). The same early jump of the lexical-scale line appears in both Qwen3.5 models, over the first 32--64 tokens and with tight intervals (\res{e7.qwen3p5-4b.a2.inf.p0032.R.mean}\% and \res{e7.qwen3p5-4b.a2.inf.p0064.R.mean}\% in Qwen3.5-4B, \res{e7.qwen3p5-9b.a2.inf.p0032.R.mean}\% in Qwen3.5-9B), so it is a systematic early-prefix effect, not noise. It is confined to the sampled answers: the answer-position log-odds stays negative, that is literal-favouring, and moves little (Qwen3.5-9B: \res{e7.qwen3p5-9b.a2.inf.ntforced.L.mean} with an empty trace, \res{e7.qwen3p5-9b.a2.inf.p0032.L.mean} after 32 tokens). The line returns to the no-thinking level by the end. For the further families (\Cref{fig:trace-newfam}), Skywork-OR1 is the exception: its pragmatic reading drops within the first 32 tokens (\form{or}: \res{e7.skywork-or1-7b.a3-conn.inf.ntforced.R.mean}\% to \res{e7.skywork-or1-7b.a3-conn.inf.p0032.R.mean}\%) and then stays flat, so there is no early phase of endorsement. Llama-3.1-Nemotron-Nano-8B starts from a rote floor (\res{e7.llama-nemotron-nano-8b.a3-quant.inf.ntforced.R.mean}\% on \form{some} with an empty trace), jumps after 32 tokens and falls back by the end. Trace-level shares count a trace only if every prefix gives a valid answer. For Qwen3.5 and Phi-4-mini-reasoning, plain truncation often yields invalid answers: in the full-item run for Qwen3.5-9B on \form{some}, \res{e7f.qwen3p5-9b.a3f-quant.inf.traj_n.dropped_invalid} traces are dropped and \res{e7f.qwen3p5-9b.a3f-quant.inf.traj_n.traces} survive, so their curves and shares describe the surviving traces and are indicative only. R1-Distill-Llama-8B is not shown because 39--54\% of its answers after short truncations are invalid.

\paragraph{Further context-driven sets.} \Cref{tab:breadth} gives the Circa and NOPE results summarised in \Cref{sec:find:split}.
\begin{table}[t]
  \centering\small
  \setlength{\tabcolsep}{4pt}
  \resizebox{\columnwidth}{!}{%
  \begin{tabular}{@{}l cc@{}}
    \toprule
    Model & Circa & NOPE \\
    \midrule
    Qwen3-4B & \cellcolor{cOurs!17}\res{e1ds.qwen3-4b.circa.mc.nat.dacc.mean}{\scriptsize\,[\res{e1ds.qwen3-4b.circa.mc.nat.dacc.lo}, \res{e1ds.qwen3-4b.circa.mc.nat.dacc.hi}]} & \cellcolor{cOurs!14}\res{e1ds.qwen3-4b.nope.nli.nat.dacc.mean}{\scriptsize\,[\res{e1ds.qwen3-4b.nope.nli.nat.dacc.lo}, \res{e1ds.qwen3-4b.nope.nli.nat.dacc.hi}]} \\
    Qwen3-8B & \cellcolor{cOurs!18}\res{e1ds.qwen3-8b.circa.mc.nat.dacc.mean}{\scriptsize\,[\res{e1ds.qwen3-8b.circa.mc.nat.dacc.lo}, \res{e1ds.qwen3-8b.circa.mc.nat.dacc.hi}]} & \cellcolor{cOurs!15}\res{e1ds.qwen3-8b.nope.nli.nat.dacc.mean}{\scriptsize\,[\res{e1ds.qwen3-8b.nope.nli.nat.dacc.lo}, \res{e1ds.qwen3-8b.nope.nli.nat.dacc.hi}]} \\
    Qwen3-14B & \cellcolor{cOurs!19}\res{e1ds.qwen3-14b.circa.mc.nat.dacc.mean}{\scriptsize\,[\res{e1ds.qwen3-14b.circa.mc.nat.dacc.lo}, \res{e1ds.qwen3-14b.circa.mc.nat.dacc.hi}]} & \cellcolor{cOurs!15}\res{e1ds.qwen3-14b.nope.nli.nat.dacc.mean}{\scriptsize\,[\res{e1ds.qwen3-14b.nope.nli.nat.dacc.lo}, \res{e1ds.qwen3-14b.nope.nli.nat.dacc.hi}]} \\
    Qwen3-30B-A3B & \cellcolor{cOurs!14}\res{e1ds.qwen3-30b-a3b.circa.mc.nat.dacc.mean}{\scriptsize\,[\res{e1ds.qwen3-30b-a3b.circa.mc.nat.dacc.lo}, \res{e1ds.qwen3-30b-a3b.circa.mc.nat.dacc.hi}]} & \cellcolor{cOurs!15}\res{e1ds.qwen3-30b-a3b.nope.nli.nat.dacc.mean}{\scriptsize\,[\res{e1ds.qwen3-30b-a3b.nope.nli.nat.dacc.lo}, \res{e1ds.qwen3-30b-a3b.nope.nli.nat.dacc.hi}]} \\
    Qwen3-32B & \cellcolor{cOurs!10}\textcolor{gray}{\res{e1ds.qwen3-32b.circa.mc.nat.dacc.mean}{\scriptsize\,[\res{e1ds.qwen3-32b.circa.mc.nat.dacc.lo}, \res{e1ds.qwen3-32b.circa.mc.nat.dacc.hi}]}} & \cellcolor{cOurs!15}\res{e1ds.qwen3-32b.nope.nli.nat.dacc.mean}{\scriptsize\,[\res{e1ds.qwen3-32b.nope.nli.nat.dacc.lo}, \res{e1ds.qwen3-32b.nope.nli.nat.dacc.hi}]} \\
    \midrule
    Qwen3.5-4B & \cellcolor{cOurs!17}\res{e1dsx.qwen3p5-4b.circa.mc.nat.dacc.mean}{\scriptsize\,[\res{e1dsx.qwen3p5-4b.circa.mc.nat.dacc.lo}, \res{e1dsx.qwen3p5-4b.circa.mc.nat.dacc.hi}]} & \cellcolor{cOurs!13}\res{e1dsx.qwen3p5-4b.nope.nli.nat.dacc.mean}{\scriptsize\,[\res{e1dsx.qwen3p5-4b.nope.nli.nat.dacc.lo}, \res{e1dsx.qwen3p5-4b.nope.nli.nat.dacc.hi}]} \\
    Qwen3.5-9B & \cellcolor{cOurs!14}\res{e1dsx.qwen3p5-9b.circa.mc.nat.dacc.mean}{\scriptsize\,[\res{e1dsx.qwen3p5-9b.circa.mc.nat.dacc.lo}, \res{e1dsx.qwen3p5-9b.circa.mc.nat.dacc.hi}]} & \cellcolor{cOurs!13}\res{e1dsx.qwen3p5-9b.nope.nli.nat.dacc.mean}{\scriptsize\,[\res{e1dsx.qwen3p5-9b.nope.nli.nat.dacc.lo}, \res{e1dsx.qwen3p5-9b.nope.nli.nat.dacc.hi}]} \\
    Phi-4-reasoning & \cellcolor{cOurs!16}\res{e1dsx.phi4-reasoning.circa.mc.nat.dacc.mean}{\scriptsize\,[\res{e1dsx.phi4-reasoning.circa.mc.nat.dacc.lo}, \res{e1dsx.phi4-reasoning.circa.mc.nat.dacc.hi}]} & \cellcolor{cOurs!17}\res{e1dsx.phi4-reasoning.nope.nli.nat.dacc.mean}{\scriptsize\,[\res{e1dsx.phi4-reasoning.nope.nli.nat.dacc.lo}, \res{e1dsx.phi4-reasoning.nope.nli.nat.dacc.hi}]} \\
    Nemotron-Nano-9B-v2 & \cellcolor{cOurs!17}\res{e1dsx.nemotron-nano-9b-v2.circa.mc.nat.dacc.mean}{\scriptsize\,[\res{e1dsx.nemotron-nano-9b-v2.circa.mc.nat.dacc.lo}, \res{e1dsx.nemotron-nano-9b-v2.circa.mc.nat.dacc.hi}]} & \cellcolor{cOurs!12}\res{e1dsx.nemotron-nano-9b-v2.nope.nli.nat.dacc.mean}{\scriptsize\,[\res{e1dsx.nemotron-nano-9b-v2.nope.nli.nat.dacc.lo}, \res{e1dsx.nemotron-nano-9b-v2.nope.nli.nat.dacc.hi}]} \\
    Llama-3.1-Nemotron-Nano-8B & \cellcolor{cOurs!20}\res{e1dsx.llama-nemotron-nano-8b.circa.mc.nat.dacc.mean}{\scriptsize\,[\res{e1dsx.llama-nemotron-nano-8b.circa.mc.nat.dacc.lo}, \res{e1dsx.llama-nemotron-nano-8b.circa.mc.nat.dacc.hi}]} & \cellcolor{cOurs!11}\textcolor{gray}{\res{e1dsx.llama-nemotron-nano-8b.nope.nli.nat.dacc.mean}{\scriptsize\,[\res{e1dsx.llama-nemotron-nano-8b.nope.nli.nat.dacc.lo}, \res{e1dsx.llama-nemotron-nano-8b.nope.nli.nat.dacc.hi}]}} \\
    Skywork-OR1-7B & \cellcolor{cOurs!21}\res{e1dsx.skywork-or1-7b.circa.mc.nat.dacc.mean}{\scriptsize\,[\res{e1dsx.skywork-or1-7b.circa.mc.nat.dacc.lo}, \res{e1dsx.skywork-or1-7b.circa.mc.nat.dacc.hi}]} & \cellcolor{cOurs!19}\res{e1dsx.skywork-or1-7b.nope.nli.nat.dacc.mean}{\scriptsize\,[\res{e1dsx.skywork-or1-7b.nope.nli.nat.dacc.lo}, \res{e1dsx.skywork-or1-7b.nope.nli.nat.dacc.hi}]} \\
    Granite-3.3-8B & \cellcolor{cOurs!7}\textcolor{gray}{\res{e1dsx.granite-3p3-8b.circa.mc.nat.dacc.mean}{\scriptsize\,[\res{e1dsx.granite-3p3-8b.circa.mc.nat.dacc.lo}, \res{e1dsx.granite-3p3-8b.circa.mc.nat.dacc.hi}]}} & \cellcolor{cOurs!12}\res{e1dsx.granite-3p3-8b.nope.nli.nat.dacc.mean}{\scriptsize\,[\res{e1dsx.granite-3p3-8b.nope.nli.nat.dacc.lo}, \res{e1dsx.granite-3p3-8b.nope.nli.nat.dacc.hi}]} \\
    SmolLM3-3B & \cellcolor{cOurs!7}\textcolor{gray}{\res{e1dsx.smollm3-3b.circa.mc.nat.dacc.mean}{\scriptsize\,[\res{e1dsx.smollm3-3b.circa.mc.nat.dacc.lo}, \res{e1dsx.smollm3-3b.circa.mc.nat.dacc.hi}]}} & \cellcolor{cOurs!13}\res{e1dsx.smollm3-3b.nope.nli.nat.dacc.mean}{\scriptsize\,[\res{e1dsx.smollm3-3b.nope.nli.nat.dacc.lo}, \res{e1dsx.smollm3-3b.nope.nli.nat.dacc.hi}]} \\
    gpt-oss-20b (high$-$low) & \cellcolor{cOurs!7}\textcolor{gray}{\res{e1gds.gpt-oss-20b.circa.mc.high.dacc.mean}{\scriptsize\,[\res{e1gds.gpt-oss-20b.circa.mc.high.dacc.lo}, \res{e1gds.gpt-oss-20b.circa.mc.high.dacc.hi}]}} & \cellcolor{cOurs!10}\res{e1gds.gpt-oss-20b.nope.nli.high.dacc.mean}{\scriptsize\,[\res{e1gds.gpt-oss-20b.nope.nli.high.dacc.lo}, \res{e1gds.gpt-oss-20b.nope.nli.high.dacc.hi}]} \\
    \bottomrule
  \end{tabular}}
  \caption{\textbf{Thinking raises or keeps accuracy on two further context-driven sets in every model.} Change with thinking (pp) [95\% CI]; grey: CI includes 0; Circa \citep{louis-etal-2020-id}, NOPE \citep{parrish-etal-2021-nope}.}
  \label{tab:breadth}
\end{table}

\paragraph{German and Chinese.} The 160 IMPPRES \form{some}/\form{or} test pairs were machine-translated into German and Chinese by Qwen3-32B, a model from the same family as most subjects; Chinese pairs left untranslated (4 of 160) were dropped. A blind consensus check of 20 Chinese items by the annotator panel found 3 with meaning or trigger errors, including a negated-connective flip, so the Chinese results are not human-verified. With these caveats, thinking lowers the pragmatic reading in both languages as in English: in Qwen3-8B by \res{e1xl.qwen3-8b.a3xde.inf.nat.dR.mean}\,pp in German, \res{e1xl.qwen3-8b.a3xzh.inf.nat.dR.mean}\,pp in Chinese and \res{e1xl.qwen3-8b.a3xen.inf.nat.dR.mean}\,pp on the English sources; in Qwen3-14B by \res{e1xl.qwen3-14b.a3xde.inf.nat.dR.mean}\,pp in German; and in Qwen3.5-9B by \res{e1xl.qwen3p5-9b.a3xde.inf.nat.dR.mean}\,pp in German and \res{e1xl.qwen3p5-9b.a3xzh.inf.nat.dR.mean}\,pp in Chinese (\res{e1xl.qwen3p5-9b.a3xen.inf.nat.dR.mean}\,pp on the English sources); raising gpt-oss-20b from low to high effort changes it by \res{e1gds.gpt-oss-20b.a3xde.inf.high.dR.mean}\,pp in German.

\paragraph{Relation types in the newest generation.} IMPPRES pairs differ in how the two sentences relate (for example, a negated first sentence with a positive second one). In Qwen3.8-27B thinking lowers the pragmatic reading of \form{some} for most relation types (\res{e1sfx.qwen3p8-27b.a3f-quant-negn.inf.nat.dR.mean}\,pp when both are negated) but raises it for a negated first sentence with a positive second one (+\res{e1sfx.qwen3p8-27b.a3f-quant-negp.inf.nat.dR.mean}\,pp), so its pooled change hides opposite effects.

\paragraph{Full held-out sets in both formats.} \Cref{tab:fullsets} gives the effect on the full held-out IMPPRES \form{some} and \form{or} sets for every model with a full-set run.

\begin{figure*}[tp]
  \centering
  \IfFileExists{figures/fig_hero_heatmap.pdf}{\includegraphics[width=\textwidth]{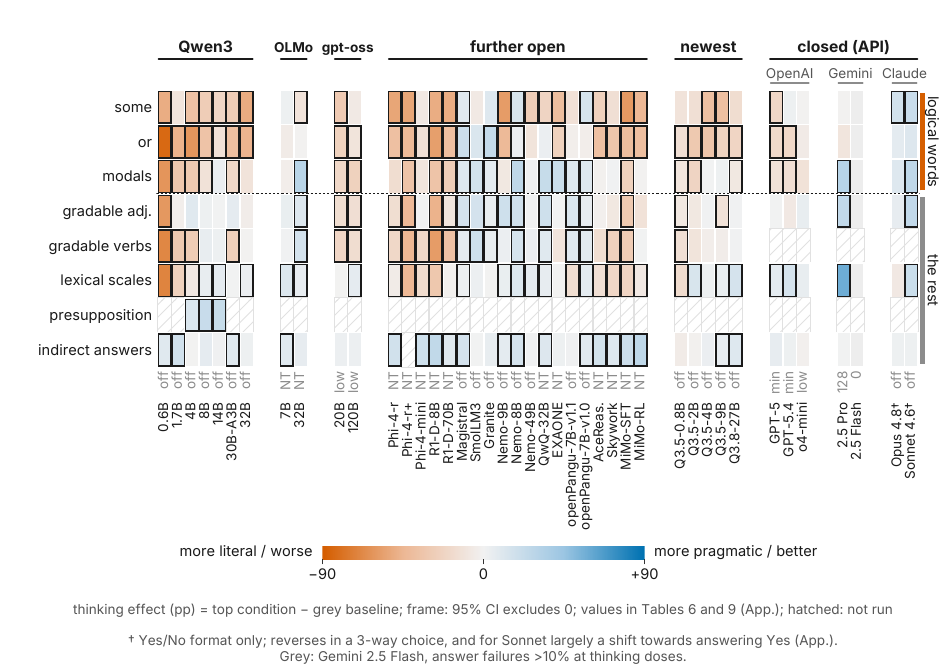}}{\figplaceholder{\textwidth}{5.2cm}}
  \caption{\textbf{Thinking effect for every item family and model.} Change in pragmatic reading with thinking (pp); vermilion: more literal; blue: more pragmatic; framed: 95\% CI excludes 0; hatched: not run. Values in \Cref{tab:main,tab:frontier}.}
  \label{fig:hero-heatmap}
\end{figure*}

\paragraph{Every item family.} \Cref{fig:hero-heatmap} shows the thinking effect for every item family and model behind \Cref{fig:hero}.

\paragraph{Preliminary seed experiments.}\label{sec:origin:ist}

If the literal reading comes from what thinking is spent on, the start of the trace should matter. \emph{Intent-Seeded Thinking} (\method) opens the thinking block with one fixed sentence about what the speaker intends to convey; two control sentences of the same length (26 tokens) ask for careful step-by-step reasoning or for checking what the words strictly entail (\Cref{app:method}).
The intent seed yields the most enrichment of the three seeds (\Cref{tab:ist}). In Qwen3-8B it raises the pragmatic reading of \form{some} from \res{e10.qwen3-8b.a3-quant.inf.nat.R.mean}\% under natural thinking to \res{e10.qwen3-8b.a3-quant.inf.sint.R.mean}\% (\res{e10.qwen3-8b.a3-quant.inf.off.R.mean}\% without thinking), against \res{e10.qwen3-8b.a3-quant.inf.sneu.R.mean}\% with the neutral seed; in Qwen3-32B it raises it from \res{e10.qwen3-32b.a3-quant.inf.nat.R.mean}\% to \res{e10.qwen3-32b.a3-quant.inf.sint.R.mean}\%, while both controls leave it at or below the natural level there. Accuracy on indirect answers moves from \res{e10.qwen3-8b.b2.yn.nat.acc.mean}\% to \res{e10.qwen3-8b.b2.yn.sint.acc.mean}\%.
On the newest generation the seed lifts the reading above its no-thinking level: Qwen3.5-9B reads \form{some} pragmatically in \res{e10.qwen3p5-9b.a3-quant.inf.sint.R.mean}\% of items with the seed, against \res{e10.qwen3p5-9b.a3-quant.inf.nat.R.mean}\% under natural thinking and \res{e10.qwen3p5-9b.a3-quant.inf.off.R.mean}\% without thinking, while the neutral and logic seeds reach \res{e10.qwen3p5-9b.a3-quant.inf.sneu.R.mean}\% and \res{e10.qwen3p5-9b.a3-quant.inf.slog.R.mean}\%.
Prompted chain-of-thought with thinking off moves the reading less than native thinking does, so producing a chain of thought is not enough to cause the full shift.
\begin{table*}[t]
  \centering\small
  \setlength{\tabcolsep}{3pt}
  \begin{tabular}{@{}l ccc ccc@{}}
    \toprule
    & \multicolumn{3}{c}{\textbf{Qwen3-8B}} & \multicolumn{3}{c}{\textbf{Qwen3-32B}} \\
    \cmidrule(lr){2-4}\cmidrule(l){5-7}
    \textbf{Method} & \form{some}\hup & \form{or}\hup & LUDWIG\hup & \form{some}\hup & \form{or}\hup & LUDWIG\hup \\
    \midrule
    No thinking (\NT) & \cellcolor{cCmpA!6}\cellpm{e10}{qwen3-8b.a3-quant.inf.off.R} & \cellcolor{cCmpA!7}\cellpm{e10}{qwen3-8b.a3-conn.inf.off.R} & \cellcolor{cCmpA!12}\cellpm{e10}{qwen3-8b.b2.yn.off.acc} & \cellcolor{cCmpA!11}\cellpm{e10}{qwen3-32b.a3-quant.inf.off.R} & \cellcolor{cCmpA!9}\cellpm{e10}{qwen3-32b.a3-conn.inf.off.R} & \cellcolor{cCmpA!13}\cellpm{e10}{qwen3-32b.b2.yn.off.acc} \\
    Natural thinking (\NAT) & \cellcolor{cCmpA!2}\cellpm{e10}{qwen3-8b.a3-quant.inf.nat.R} & \cellcolor{cCmpA!2}\cellpm{e10}{qwen3-8b.a3-conn.inf.nat.R} & \cellcolor{cCmpA!13}\textbf{\cellpm{e10}{qwen3-8b.b2.yn.nat.acc}} & \cellcolor{cCmpA!4}\cellpm{e10}{qwen3-32b.a3-quant.inf.nat.R} & \cellcolor{cCmpA!3}\cellpm{e10}{qwen3-32b.a3-conn.inf.nat.R} & \cellcolor{cCmpA!13}\cellpm{e10}{qwen3-32b.b2.yn.nat.acc} \\
    Prompted CoT, thinking off$^{a}$ & \cellcolor{cCmpA!3}\underline{\cellpm{e8}{qwen3-8b.a3-quant.inf.cot.R}} & \cellcolor{cCmpA!3}\underline{\cellpm{e8}{qwen3-8b.a3-conn.inf.cot.R}} & \cellcolor{cCmpA!12}\cellpm{e8}{qwen3-8b.b2.yn.cot.acc} & \cellcolor{cCmpA!8}\textbf{\cellpm{e8}{qwen3-32b.a3-quant.inf.cot.R}} & \cellcolor{cCmpA!8}\textbf{\cellpm{e8}{qwen3-32b.a3-conn.inf.cot.R}} & \cellcolor{cCmpA!12}\cellpm{e8}{qwen3-32b.b2.yn.cot.acc} \\
    \midrule
    \multicolumn{7}{@{}l}{\textit{Seed sentence opening the thinking trace (26 tokens each)}} \\
    Neutral seed & \cellcolor{cCmpA!2}\cellpm{e10}{qwen3-8b.a3-quant.inf.sneu.R} & \cellcolor{cCmpA!2}\cellpm{e10}{qwen3-8b.a3-conn.inf.sneu.R} & \cellcolor{cCmpA!12}\cellpm{e10}{qwen3-8b.b2.yn.sneu.acc} & \cellcolor{cCmpA!3}\cellpm{e10}{qwen3-32b.a3-quant.inf.sneu.R} & \cellcolor{cCmpA!3}\cellpm{e10}{qwen3-32b.a3-conn.inf.sneu.R} & \cellcolor{cCmpA!13}\textbf{\cellpm{e10}{qwen3-32b.b2.yn.sneu.acc}} \\
    Logic seed & \cellcolor{cCmpA!3}\cellpm{e10}{qwen3-8b.a3-quant.inf.slog.R} & \cellcolor{cCmpA!3}\cellpm{e10}{qwen3-8b.a3-conn.inf.slog.R} & \cellcolor{cCmpA!12}\underline{\cellpm{e10}{qwen3-8b.b2.yn.slog.acc}} & \cellcolor{cCmpA!3}\cellpm{e10}{qwen3-32b.a3-quant.inf.slog.R} & \cellcolor{cCmpA!3}\cellpm{e10}{qwen3-32b.a3-conn.inf.slog.R} & \cellcolor{cCmpA!12}\cellpm{e10}{qwen3-32b.b2.yn.slog.acc} \\
    Intent seed (\method) & \cellcolor{cCmpA!4}\textbf{\cellpm{e10}{qwen3-8b.a3-quant.inf.sint.R}} & \cellcolor{cCmpA!4}\textbf{\cellpm{e10}{qwen3-8b.a3-conn.inf.sint.R}} & \cellcolor{cCmpA!12}\cellpm{e10}{qwen3-8b.b2.yn.sint.acc} & \cellcolor{cCmpA!6}\underline{\cellpm{e10}{qwen3-32b.a3-quant.inf.sint.R}} & \cellcolor{cCmpA!5}\underline{\cellpm{e10}{qwen3-32b.a3-conn.inf.sint.R}} & \cellcolor{cCmpA!13}\underline{\cellpm{e10}{qwen3-32b.b2.yn.sint.acc}} \\
    \midrule
    \multicolumn{7}{@{}l}{\textit{Training-time fix, same backbone \citep{park2026pragrest}}} \\
    PragReST-8B, \NAT & \cellcolor{cCmpA!4}\cellpm{e1s}{pragrest-8b.a3-quant.inf.nat.R} & \cellcolor{cCmpA!5}\cellpm{e1s}{pragrest-8b.a3-conn.inf.nat.R} & \cellcolor{cCmpA!13}\cellpm{e1s}{pragrest-8b.b2.yn.nat.acc} & \na & \na & \na \\
    \bottomrule
  \end{tabular}
  \par\smallskip{\footnotesize Seeds run with natural thinking; $^{a}$from a separate run with the same settings. PragReST-8B is trained against literal readings \citep{park2026pragrest}.\par}
  \caption{\textbf{The intent seed restores more enrichment than length-matched control seeds.} Pragmatic reading and LUDWIG accuracy (\%); mean $\pm$ s.d.\ over seeds; \textbf{bold}/\underline{underlined}: best/second among thinking rows.}
  \label{tab:ist}
\end{table*}

\noindent\textbf{Takeaway: what the trace starts with steers the reading chosen; a sentence about speaker intent recovers more of the lost enrichment than length-matched control sentences.}

\paragraph{Per-family results and size trends.}
\begin{figure}[t]
  \centering
  \includegraphics[width=\columnwidth]{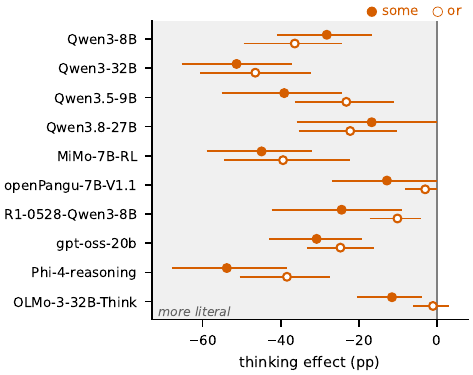}
  \caption{\textbf{Thinking makes \form{some} and \form{or} more literal in every major model family.} Change with thinking (pp, 95\% CI). Size trends: \Cref{fig:size}.}
  \label{fig:major}
\end{figure}
\begin{figure*}[tp]
  \centering
  \includegraphics[width=\textwidth]{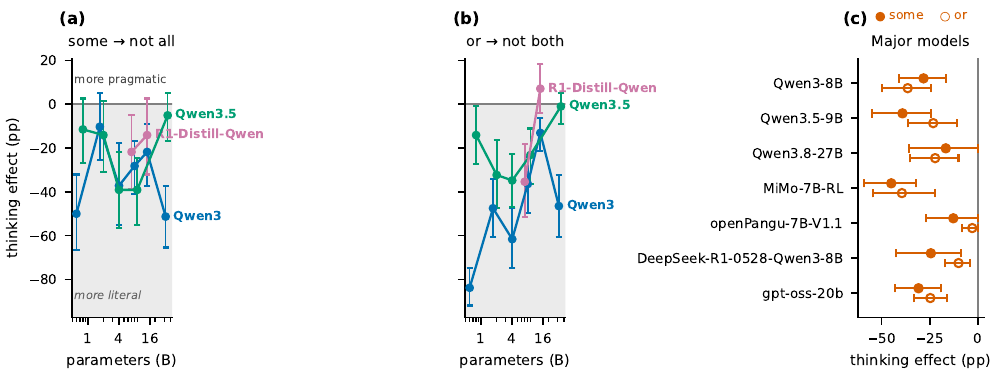}
  \caption{\textbf{The literal shift holds across sizes and families.} Thinking effect on \form{some} and \form{or} (pp, 95\% CI) by model size for Qwen3, Qwen3.5 and R1-Distill-Qwen, with Qwen3.8, MiMo, openPangu and DeepSeek-R1-0528 as single points.}
  \label{fig:size}
\end{figure*}

\paragraph{Thinking effects per model.} \Cref{tab:main} gives the values and intervals behind \Cref{fig:hero}.
\begin{table*}[t]
  \centering
  \scriptsize\renewcommand{\arraystretch}{0.95}
  \setlength{\tabcolsep}{2.4pt}
  \begin{tabular}{@{}l ccccc c@{}}
    \toprule
    & \multicolumn{5}{c}{\textbf{Generalized implicatures: $\Delta\rate$ (pp)}\hup} & \textbf{$\Delta$Acc. (pp)}\hup \\
    \cmidrule(lr){2-6}\cmidrule(l){7-7}
    \textbf{Model} & \form{some} & \form{or} & modals & lexical scales & IMPPRES all & LUDWIG \\
    \midrule
    \multicolumn{7}{@{}l}{\textit{Qwen3, distilled from larger models (0.6B--30B-A3B) or fully RL-trained (32B); \NAT $-$ \NT}} \\
    Qwen3-0.6B$^{\dagger}$ & \cellcolor{cInterv!23}\celldpm{e1s}{qwen3-0p6b.a3-quant.inf.nat.dR} & \cellcolor{cInterv!25}\celldpm{e1s}{qwen3-0p6b.a3-conn.inf.nat.dR} & \cellcolor{cInterv!24}\celldpm{e1s}{qwen3-0p6b.a3-modal.inf.nat.dR} & \cellcolor{cInterv!24}\celldpm{e1s}{qwen3-0p6b.a2.inf.nat.dR} & \cellcolor{cInterv!24}\celldpm{e1s}{qwen3-0p6b.a3.inf.nat.dR} & \cellcolor{cOurs!15}\celldpm{e1s}{qwen3-0p6b.b2.yn.nat.dacc} \\
    Qwen3-1.7B & \cellcolor{cInterv!15}\celldpm{e1s}{qwen3-1p7b.a3-quant.inf.nat.dR} & \cellcolor{cInterv!22}\celldpm{e1s}{qwen3-1p7b.a3-conn.inf.nat.dR} & \cellcolor{cInterv!21}\celldpm{e1s}{qwen3-1p7b.a3-modal.inf.nat.dR} & \cellcolor{cInterv!19}\celldpm{e1s}{qwen3-1p7b.a2.inf.nat.dR} & \cellcolor{cInterv!19}\celldpm{e1s}{qwen3-1p7b.a3.inf.nat.dR} & \cellcolor{cOurs!18}\celldpm{e1s}{qwen3-1p7b.b2.yn.nat.dacc} \\
    Qwen3-4B & \cellcolor{cInterv!21}\celldpm{e1}{qwen3-4b.a3-quant.inf.nat.dR} & \cellcolor{cInterv!24}\celldpm{e1}{qwen3-4b.a3-conn.inf.nat.dR} & \cellcolor{cInterv!21}\celldpm{e1}{qwen3-4b.a3-modal.inf.nat.dR} & \cellcolor{cInterv!14}\celldpm{e1}{qwen3-4b.a2.inf.nat.dR} & \cellcolor{cInterv!20}\celldpm{e1}{qwen3-4b.a3.inf.nat.dR} & \cellcolor{cInterv!5}\celldpm{e1}{qwen3-4b.b2.yn.nat.dacc} \\
    Qwen3-8B & \cellcolor{cInterv!20}\celldpm{e1}{qwen3-8b.a3-quant.inf.nat.dR} & \cellcolor{cInterv!21}\celldpm{e1}{qwen3-8b.a3-conn.inf.nat.dR} & \cellcolor{cInterv!16}\celldpm{e1}{qwen3-8b.a3-modal.inf.nat.dR} & \cellcolor{cOurs!12}\celldpm{e1}{qwen3-8b.a2.inf.nat.dR} & \cellcolor{cInterv!15}\celldpm{e1}{qwen3-8b.a3.inf.nat.dR} & \cellcolor{cOurs!13}\celldpm{e1}{qwen3-8b.b2.yn.nat.dacc} \\
    Qwen3-14B & \cellcolor{cInterv!19}\celldpm{e1}{qwen3-14b.a3-quant.inf.nat.dR} & \cellcolor{cInterv!16}\celldpm{e1}{qwen3-14b.a3-conn.inf.nat.dR} & \cellcolor{cOurs!11}\celldpm{e1}{qwen3-14b.a3-modal.inf.nat.dR} & \cellcolor{cOurs!13}\celldpm{e1}{qwen3-14b.a2.inf.nat.dR} & \cellcolor{cOurs!5}\celldpm{e1}{qwen3-14b.a3.inf.nat.dR} & \cellcolor{cOurs!7}\celldpm{e1}{qwen3-14b.b2.yn.nat.dacc} \\
    Qwen3-30B-A3B & \cellcolor{cInterv!20}\celldpm{e1s}{qwen3-30b-a3b.a3-quant.inf.nat.dR} & \cellcolor{cInterv!22}\celldpm{e1s}{qwen3-30b-a3b.a3-conn.inf.nat.dR} & \cellcolor{cInterv!18}\celldpm{e1s}{qwen3-30b-a3b.a3-modal.inf.nat.dR} & \cellcolor{cOurs!6}\celldpm{e1s}{qwen3-30b-a3b.a2.inf.nat.dR} & \cellcolor{cInterv!17}\celldpm{e1s}{qwen3-30b-a3b.a3.inf.nat.dR} & \cellcolor{cOurs!15}\celldpm{e1s}{qwen3-30b-a3b.b2.yn.nat.dacc} \\
    \cmidrule(l){1-7}
    Qwen3-32B & \cellcolor{cInterv!23}\celldpm{e1s}{qwen3-32b.a3-quant.inf.nat.dR} & \cellcolor{cInterv!22}\celldpm{e1s}{qwen3-32b.a3-conn.inf.nat.dR} & \cellcolor{cInterv!16}\celldpm{e1s}{qwen3-32b.a3-modal.inf.nat.dR} & \cellcolor{cOurs!11}\celldpm{e1s}{qwen3-32b.a2.inf.nat.dR} & \cellcolor{cInterv!18}\celldpm{e1s}{qwen3-32b.a3.inf.nat.dR} & \cellcolor{cOurs!9}\celldpm{e1s}{qwen3-32b.b2.yn.nat.dacc} \\
    \midrule
    \multicolumn{7}{@{}l}{\textit{OLMo-3-Think, same recipe at two sizes; \NAT $-$ forced-empty thinking}} \\
    OLMo-3-7B-Think & \cellcolor{cOurs!9}\celldpm{e1s}{olmo3-7b-think.a3-quant.inf.nat.dR} & \cellcolor{cInterv!12}\celldpm{e1s}{olmo3-7b-think.a3-conn.inf.nat.dR} & \cellcolor{cInterv!11}\celldpm{e1s}{olmo3-7b-think.a3-modal.inf.nat.dR} & \cellcolor{cOurs!16}\celldpm{e1s}{olmo3-7b-think.a2.inf.nat.dR} & \cellcolor{cOurs!9}\celldpm{e1s}{olmo3-7b-think.a3.inf.nat.dR} & \cellcolor{cOurs!15}\celldpm{e1s}{olmo3-7b-think.b2.yn.nat.dacc} \\
    OLMo-3-32B-Think & \cellcolor{cInterv!16}\celldpm{e1s}{olmo3-32b-think.a3-quant.inf.nat.dR} & \cellcolor{cInterv!6}\celldpm{e1s}{olmo3-32b-think.a3-conn.inf.nat.dR} & \cellcolor{cOurs!20}\celldpm{e1s}{olmo3-32b-think.a3-modal.inf.nat.dR} & \cellcolor{cOurs!13}\celldpm{e1s}{olmo3-32b-think.a2.inf.nat.dR} & \cellcolor{cOurs!18}\celldpm{e1s}{olmo3-32b-think.a3.inf.nat.dR} & \cellcolor{cOurs!13}\celldpm{e1s}{olmo3-32b-think.b2.yn.nat.dacc} \\
    \midrule
    \multicolumn{7}{@{}l}{\textit{gpt-oss, reasoning effort high $-$ low}} \\
    gpt-oss-20b & \cellcolor{cInterv!20}\celldpm{e3}{gpt-oss-20b.a3-quant.inf.high.dR} & \cellcolor{cInterv!19}\celldpm{e3}{gpt-oss-20b.a3-conn.inf.high.dR} & \cellcolor{cInterv!18}\celldpm{e3}{gpt-oss-20b.a3-modal.inf.high.dR} & \cellcolor{cInterv!4}\celldpm{e3}{gpt-oss-20b.a2.inf.high.dR} & \cellcolor{cInterv!17}\celldpm{e3}{gpt-oss-20b.a3.inf.high.dR} & \cellcolor{cOurs!10}\celldpm{e3}{gpt-oss-20b.b2.yn.high.dacc} \\
    gpt-oss-120b & \cellcolor{cInterv!13}\celldpm{e3}{gpt-oss-120b.a3-quant.inf.high.dR} & \cellcolor{cInterv!15}\celldpm{e3}{gpt-oss-120b.a3-conn.inf.high.dR} & \cellcolor{cInterv!19}\celldpm{e3}{gpt-oss-120b.a3-modal.inf.high.dR} & \cellcolor{cOurs!13}\celldpm{e3}{gpt-oss-120b.a2.inf.high.dR} & \cellcolor{cInterv!16}\celldpm{e3}{gpt-oss-120b.a3.inf.high.dR} & \cellcolor{cOurs!10}\celldpm{e3}{gpt-oss-120b.b2.yn.high.dacc} \\
    \midrule
    \multicolumn{7}{@{}l}{\textit{Popular models: natural thinking $-$ forced-empty thinking}} \\
    Phi-4-reasoning & \cellcolor{cInterv!23}\celldpm{e1p}{phi4-reasoning.a3-quant.inf.nat.dR} & \cellcolor{cInterv!21}\celldpm{e1p}{phi4-reasoning.a3-conn.inf.nat.dR} & \cellcolor{cInterv!17}\celldpm{e1p}{phi4-reasoning.a3-modal.inf.nat.dR} & \cellcolor{cInterv!12}\celldpm{e1p}{phi4-reasoning.a2.inf.nat.dR} & \cellcolor{cInterv!17}\celldpm{e1p}{phi4-reasoning.a3.inf.nat.dR} & \cellcolor{cOurs!18}\celldpm{e1p}{phi4-reasoning.b2.yn.nat.dacc} \\
    Magistral-Small & \cellcolor{cOurs!15}\celldpm{e1p}{magistral-small.a3-quant.inf.nat.dR} & \cellcolor{cOurs!18}\celldpm{e1p}{magistral-small.a3-conn.inf.nat.dR} & \cellcolor{cOurs!15}\celldpm{e1p}{magistral-small.a3-modal.inf.nat.dR} & \cellcolor{cOurs!10}\celldpm{e1p}{magistral-small.a2.inf.nat.dR} & \cellcolor{cOurs!15}\celldpm{e1p}{magistral-small.a3.inf.nat.dR} & \cellcolor{cOurs!17}\celldpm{e1p}{magistral-small.b2.yn.nat.dacc} \\
    R1-Distill-Llama-8B & \cellcolor{cInterv!23}\celldpm{e1p}{r1distill-llama-8b.a3-quant.inf.nat.dR} & \cellcolor{cInterv!23}\celldpm{e1p}{r1distill-llama-8b.a3-conn.inf.nat.dR} & \cellcolor{cInterv!21}\celldpm{e1p}{r1distill-llama-8b.a3-modal.inf.nat.dR} & \cellcolor{cInterv!21}\celldpm{e1p}{r1distill-llama-8b.a2.inf.nat.dR} & \cellcolor{cInterv!21}\celldpm{e1p}{r1distill-llama-8b.a3.inf.nat.dR} & \cellcolor{cOurs!20}\celldpm{e1p}{r1distill-llama-8b.b2.yn.nat.dacc} \\
    R1-Distill-Llama-70B & \cellcolor{cInterv!23}\celldpm{e1p}{r1distill-llama-70b.a3-quant.inf.nat.dR} & \cellcolor{cInterv!22}\celldpm{e1p}{r1distill-llama-70b.a3-conn.inf.nat.dR} & \cellcolor{cInterv!20}\celldpm{e1p}{r1distill-llama-70b.a3-modal.inf.nat.dR} & \cellcolor{cInterv!8}\celldpm{e1p}{r1distill-llama-70b.a2.inf.nat.dR} & \cellcolor{cInterv!21}\celldpm{e1p}{r1distill-llama-70b.a3.inf.nat.dR} & \cellcolor{cOurs!15}\celldpm{e1p}{r1distill-llama-70b.b2.yn.nat.dacc} \\
    \multicolumn{7}{@{}l}{\textit{Further open reasoning models: natural thinking $-$ thinking off (forced-empty where no switch exists)}} \\
    SmolLM3-3B & \cellcolor{cInterv!13}\celldpm{e1b}{smollm3-3b.a3-quant.inf.nat.dR} & \cellcolor{cOurs!16}\celldpm{e1b}{smollm3-3b.a3-conn.inf.nat.dR} & \cellcolor{cOurs!17}\celldpm{e1b}{smollm3-3b.a3-modal.inf.nat.dR} & \cellcolor{cOurs!10}\celldpm{e1b}{smollm3-3b.a2.inf.nat.dR} & \cellcolor{cOurs!16}\celldpm{e1b}{smollm3-3b.a3.inf.nat.dR} & \cellcolor{cOurs!5}\celldpm{e1b}{smollm3-3b.b2.yn.nat.dacc} \\
    Granite-3.3-8B & \cellcolor{cOurs!15}\celldpm{e1b}{granite-3p3-8b.a3-quant.inf.nat.dR} & \cellcolor{cOurs!19}\celldpm{e1b}{granite-3p3-8b.a3-conn.inf.nat.dR} & \cellcolor{cOurs!14}\celldpm{e1b}{granite-3p3-8b.a3-modal.inf.nat.dR} & \cellcolor{cInterv!5}\celldpm{e1b}{granite-3p3-8b.a2.inf.nat.dR} & \cellcolor{cOurs!16}\celldpm{e1b}{granite-3p3-8b.a3.inf.nat.dR} & \cellcolor{cOurs!12}\celldpm{e1b}{granite-3p3-8b.b2.yn.nat.dacc} \\
    Nemotron-Nano-9B-v2 & \cellcolor{cInterv!23}\celldpm{e1b}{nemotron-nano-9b-v2.a3-quant.inf.nat.dR} & \cellcolor{cInterv!22}\celldpm{e1b}{nemotron-nano-9b-v2.a3-conn.inf.nat.dR} & \cellcolor{cInterv!14}\celldpm{e1b}{nemotron-nano-9b-v2.a3-modal.inf.nat.dR} & \cellcolor{cOurs!12}\celldpm{e1b}{nemotron-nano-9b-v2.a2.inf.nat.dR} & \cellcolor{cInterv!18}\celldpm{e1b}{nemotron-nano-9b-v2.a3.inf.nat.dR} & \cellcolor{cOurs!14}\celldpm{e1b}{nemotron-nano-9b-v2.b2.yn.nat.dacc} \\
    AceReason-Nemotron-7B & \cellcolor{cInterv!19}\celldpm{e1b}{acereason-7b.a3-quant.inf.nat.dR} & \cellcolor{cInterv!21}\celldpm{e1b}{acereason-7b.a3-conn.inf.nat.dR} & \cellcolor{cInterv!9}\celldpm{e1b}{acereason-7b.a3-modal.inf.nat.dR} & \cellcolor{cInterv!17}\celldpm{e1b}{acereason-7b.a2.inf.nat.dR} & \cellcolor{cInterv!16}\celldpm{e1b}{acereason-7b.a3.inf.nat.dR} & \cellcolor{cOurs!17}\celldpm{e1b}{acereason-7b.b2.yn.nat.dacc} \\
    Skywork-OR1-7B & \cellcolor{cInterv!17}\celldpm{e1b}{skywork-or1-7b.a3-quant.inf.nat.dR} & \cellcolor{cInterv!21}\celldpm{e1b}{skywork-or1-7b.a3-conn.inf.nat.dR} & \cellcolor{cOurs!6}\celldpm{e1b}{skywork-or1-7b.a3-modal.inf.nat.dR} & \cellcolor{cInterv!18}\celldpm{e1b}{skywork-or1-7b.a2.inf.nat.dR} & \cellcolor{cInterv!15}\celldpm{e1b}{skywork-or1-7b.a3.inf.nat.dR} & \cellcolor{cOurs!15}\celldpm{e1b}{skywork-or1-7b.b2.yn.nat.dacc} \\
    MiMo-7B-SFT & \cellcolor{cInterv!24}\celldpm{e1b}{mimo-7b-sft.a3-quant.inf.nat.dR} & \cellcolor{cInterv!22}\celldpm{e1b}{mimo-7b-sft.a3-conn.inf.nat.dR} & \cellcolor{cInterv!19}\celldpm{e1b}{mimo-7b-sft.a3-modal.inf.nat.dR} & \cellcolor{cInterv!19}\celldpm{e1b}{mimo-7b-sft.a2.inf.nat.dR} & \cellcolor{cInterv!19}\celldpm{e1b}{mimo-7b-sft.a3.inf.nat.dR} & \cellcolor{cOurs!18}\celldpm{e1b}{mimo-7b-sft.b2.yn.nat.dacc} \\
    MiMo-7B-RL & \cellcolor{cInterv!22}\celldpm{e1b}{mimo-7b-rl.a3-quant.inf.nat.dR} & \cellcolor{cInterv!22}\celldpm{e1b}{mimo-7b-rl.a3-conn.inf.nat.dR} & \cellcolor{cInterv!6}\celldpm{e1b}{mimo-7b-rl.a3-modal.inf.nat.dR} & \cellcolor{cInterv!14}\celldpm{e1b}{mimo-7b-rl.a2.inf.nat.dR} & \cellcolor{cInterv!15}\celldpm{e1b}{mimo-7b-rl.a3.inf.nat.dR} & \cellcolor{cOurs!20}\celldpm{e1b}{mimo-7b-rl.b2.yn.nat.dacc} \\
    QwQ-32B & \cellcolor{cInterv!19}\celldpm{e1b}{qwq-32b.a3-quant.inf.nat.dR} & \cellcolor{cOurs!11}\celldpm{e1b}{qwq-32b.a3-conn.inf.nat.dR} & \cellcolor{cOurs!19}\celldpm{e1b}{qwq-32b.a3-modal.inf.nat.dR} & \cellcolor{cOurs!13}\celldpm{e1b}{qwq-32b.a2.inf.nat.dR} & \cellcolor{cOurs!18}\celldpm{e1b}{qwq-32b.a3.inf.nat.dR} & \cellcolor{cOurs!16}\celldpm{e1b}{qwq-32b.b2.yn.nat.dacc} \\
    Llama-3.3-Nemotron-Super-49B & \cellcolor{cInterv!20}\celldpm{e1b}{llama-nemotron-super-49b.a3-quant.inf.nat.dR} & \cellcolor{cInterv!15}\celldpm{e1b}{llama-nemotron-super-49b.a3-conn.inf.nat.dR} & \cellcolor{cOurs!9}\celldpm{e1b}{llama-nemotron-super-49b.a3-modal.inf.nat.dR} & \cellcolor{cOurs!13}\celldpm{e1b}{llama-nemotron-super-49b.a2.inf.nat.dR} & \cellcolor{cInterv!11}\celldpm{e1b}{llama-nemotron-super-49b.a3.inf.nat.dR} & \cellcolor{cOurs!7}\celldpm{e1b}{llama-nemotron-super-49b.b2.yn.nat.dacc} \\
    Llama-3.1-Nemotron-Nano-8B & \cellcolor{cOurs!16}\celldpm{e1b}{llama-nemotron-nano-8b.a3-quant.inf.nat.dR} & \cellcolor{cOurs!11}\celldpm{e1b}{llama-nemotron-nano-8b.a3-conn.inf.nat.dR} & \cellcolor{cOurs!20}\celldpm{e1b}{llama-nemotron-nano-8b.a3-modal.inf.nat.dR} & \cellcolor{cOurs!16}\celldpm{e1b}{llama-nemotron-nano-8b.a2.inf.nat.dR} & \cellcolor{cOurs!17}\celldpm{e1b}{llama-nemotron-nano-8b.a3.inf.nat.dR} & \cellcolor{cOurs!17}\celldpm{e1b}{llama-nemotron-nano-8b.b2.yn.nat.dacc} \\
    Phi-4-mini-reasoning & \cellcolor{cInterv!14}\celldpm{e1b}{phi4-mini-reasoning.a3-quant.inf.nat.dR} & \cellcolor{cInterv!16}\celldpm{e1b}{phi4-mini-reasoning.a3-conn.inf.nat.dR} & \cellcolor{cInterv!15}\celldpm{e1b}{phi4-mini-reasoning.a3-modal.inf.nat.dR} & \cellcolor{cInterv!18}\celldpm{e1b}{phi4-mini-reasoning.a2.inf.nat.dR} & \cellcolor{cInterv!16}\celldpm{e1b}{phi4-mini-reasoning.a3.inf.nat.dR} & \cellcolor{cOurs!18}\celldpm{e1b}{phi4-mini-reasoning.b2.yn.nat.dacc} \\
    EXAONE-Deep-7.8B & \cellcolor{cInterv!22}\celldpm{e1b}{exaone-deep-7p8b.a3-quant.inf.nat.dR} & \cellcolor{cInterv!19}\celldpm{e1b}{exaone-deep-7p8b.a3-conn.inf.nat.dR} & \cellcolor{cOurs!18}\celldpm{e1b}{exaone-deep-7p8b.a3-modal.inf.nat.dR} & \cellcolor{cInterv!5}\celldpm{e1b}{exaone-deep-7p8b.a2.inf.nat.dR} & \cellcolor{cInterv!13}\celldpm{e1b}{exaone-deep-7p8b.a3.inf.nat.dR} & \cellcolor{cOurs!10}\celldpm{e1b}{exaone-deep-7p8b.b2.yn.nat.dacc} \\
    openPangu-Embedded-7B-V1.1$^{\S}$ & \cellcolor{cInterv!16}\celldpm{e1pangu}{openpangu-embedded-7b-v1p1.a3-quant.inf.nat.dR} & \cellcolor{cInterv!10}\celldpm{e1pangu}{openpangu-embedded-7b-v1p1.a3-conn.inf.nat.dR} & \cellcolor{cOurs!15}\celldpm{e1pangu}{openpangu-embedded-7b-v1p1.a3-modal.inf.nat.dR} & \cellcolor{cInterv!17}\celldpm{e1pangu}{openpangu-embedded-7b-v1p1.a2.inf.nat.dR} & \cellcolor{cOurs!14}\celldpm{e1pangu}{openpangu-embedded-7b-v1p1.a3.inf.nat.dR} & \cellcolor{cOurs!9}\celldpm{e1pangu}{openpangu-embedded-7b-v1p1.b2.yn.nat.dacc} \\
    openPangu-Embedded-7B-model$^{\S}$ & \cellcolor{cOurs!17}\celldpm{e1pangu}{openpangu-embedded-7b-model.a3-quant.inf.nat.dR} & \cellcolor{cInterv!11}\celldpm{e1pangu}{openpangu-embedded-7b-model.a3-conn.inf.nat.dR} & \cellcolor{cOurs!15}\celldpm{e1pangu}{openpangu-embedded-7b-model.a3-modal.inf.nat.dR} & \cellcolor{cOurs!14}\celldpm{e1pangu}{openpangu-embedded-7b-model.a2.inf.nat.dR} & \cellcolor{cOurs!16}\celldpm{e1pangu}{openpangu-embedded-7b-model.a3.inf.nat.dR} & \cellcolor{cOurs!14}\celldpm{e1pangu}{openpangu-embedded-7b-model.b2.yn.nat.dacc} \\
    \multicolumn{7}{@{}l}{\textit{Newest generation: natural thinking $-$ thinking off}} \\
    Qwen3.5-0.8B & \cellcolor{cInterv!16}\celldpm{e1q}{qwen3p5-0p8b.a3-quant.inf.nat.dR} & \cellcolor{cInterv!17}\celldpm{e1q}{qwen3p5-0p8b.a3-conn.inf.nat.dR} & \cellcolor{cInterv!15}\celldpm{e1q}{qwen3p5-0p8b.a3-modal.inf.nat.dR} & \cellcolor{cInterv!18}\celldpm{e1q}{qwen3p5-0p8b.a2.inf.nat.dR} & \cellcolor{cInterv!15}\celldpm{e1q}{qwen3p5-0p8b.a3.inf.nat.dR} & \cellcolor{cInterv!12}\celldpm{e1q}{qwen3p5-0p8b.b2.yn.nat.dacc} \\
    Qwen3.5-2B & \cellcolor{cInterv!17}\celldpm{e1q}{qwen3p5-2b.a3-quant.inf.nat.dR} & \cellcolor{cInterv!21}\celldpm{e1q}{qwen3p5-2b.a3-conn.inf.nat.dR} & \cellcolor{cInterv!19}\celldpm{e1q}{qwen3p5-2b.a3-modal.inf.nat.dR} & \cellcolor{cOurs!16}\celldpm{e1q}{qwen3p5-2b.a2.inf.nat.dR} & \cellcolor{cInterv!15}\celldpm{e1q}{qwen3p5-2b.a3.inf.nat.dR} & \cellcolor{cInterv!8}\celldpm{e1q}{qwen3p5-2b.b2.yn.nat.dacc} \\
    Qwen3.5-4B & \cellcolor{cInterv!21}\celldpm{e1q}{qwen3p5-4b.a3-quant.inf.nat.dR} & \cellcolor{cInterv!21}\celldpm{e1q}{qwen3p5-4b.a3-conn.inf.nat.dR} & \cellcolor{cOurs!3}\celldpm{e1q}{qwen3p5-4b.a3-modal.inf.nat.dR} & \cellcolor{cOurs!9}\celldpm{e1q}{qwen3p5-4b.a2.inf.nat.dR} & \cellcolor{cInterv!10}\celldpm{e1q}{qwen3p5-4b.a3.inf.nat.dR} & \cellcolor{cOurs!11}\celldpm{e1q}{qwen3p5-4b.b2.yn.nat.dacc} \\
    Qwen3.5-9B & \cellcolor{cInterv!21}\celldpm{e1q}{qwen3p5-9b.a3-quant.inf.nat.dR} & \cellcolor{cInterv!19}\celldpm{e1q}{qwen3p5-9b.a3-conn.inf.nat.dR} & \cellcolor{cOurs!9}\celldpm{e1q}{qwen3p5-9b.a3-modal.inf.nat.dR} & \cellcolor{cOurs!11}\celldpm{e1q}{qwen3p5-9b.a2.inf.nat.dR} & \cellcolor{cInterv!14}\celldpm{e1q}{qwen3p5-9b.a3.inf.nat.dR} & \cellcolor{cOurs!13}\celldpm{e1q}{qwen3p5-9b.b2.yn.nat.dacc} \\
    Qwen3.8-27B (default effort) & \cellcolor{cInterv!17}\celldpm{e1q}{qwen3p8-27b.a3-quant.inf.nat.dR} & \cellcolor{cInterv!19}\celldpm{e1q}{qwen3p8-27b.a3-conn.inf.nat.dR} & \cellcolor{cInterv!14}\celldpm{e1q}{qwen3p8-27b.a3-modal.inf.nat.dR} & \cellcolor{cOurs!17}\celldpm{e1q}{qwen3p8-27b.a2.inf.nat.dR} & \cellcolor{cInterv!14}\celldpm{e1q}{qwen3p8-27b.a3.inf.nat.dR} & \cellcolor{cOurs!15}\celldpm{e1q}{qwen3p8-27b.b2.yn.nat.dacc} \\
    Qwen3.8-27B (low effort) & \cellcolor{cInterv!19}\celldpm{e1q}{qwen3p8-27b-elow.a3-quant.inf.nat.dR} & \cellcolor{cInterv!21}\celldpm{e1q}{qwen3p8-27b-elow.a3-conn.inf.nat.dR} & \cellcolor{cInterv!9}\celldpm{e1q}{qwen3p8-27b-elow.a3-modal.inf.nat.dR} & \cellcolor{cOurs!17}\celldpm{e1q}{qwen3p8-27b-elow.a2.inf.nat.dR} & \cellcolor{cInterv!14}\celldpm{e1q}{qwen3p8-27b-elow.a3.inf.nat.dR} & \cellcolor{cOurs!14}\celldpm{e1q}{qwen3p8-27b-elow.b2.yn.nat.dacc} \\
    Qwen3.8-27B (medium effort) & \cellcolor{cOurs!3}\celldpm{e1q}{qwen3p8-27b-emed.a3-quant.inf.nat.dR} & \cellcolor{cInterv!16}\celldpm{e1q}{qwen3p8-27b-emed.a3-conn.inf.nat.dR} & \cellcolor{cOurs!17}\celldpm{e1q}{qwen3p8-27b-emed.a3-modal.inf.nat.dR} & \cellcolor{cOurs!15}\celldpm{e1q}{qwen3p8-27b-emed.a2.inf.nat.dR} & \cellcolor{cOurs!3}\celldpm{e1q}{qwen3p8-27b-emed.a3.inf.nat.dR} & \cellcolor{cOurs!14}\celldpm{e1q}{qwen3p8-27b-emed.b2.yn.nat.dacc} \\
    \midrule
    \multicolumn{7}{@{}l}{\textit{Further distilled and updated models: natural thinking $-$ no or forced-empty thinking}} \\
    Qwen3-4B-Thinking-2507 & \cellcolor{cInterv!23}\celldpm{e1b}{qwen3-4b-thinking-2507.a3-quant.inf.nat.dR} & \cellcolor{cInterv!23}\celldpm{e1b}{qwen3-4b-thinking-2507.a3-conn.inf.nat.dR} & \cellcolor{cInterv!22}\celldpm{e1b}{qwen3-4b-thinking-2507.a3-modal.inf.nat.dR} & \cellcolor{cOurs!13}\celldpm{e1b}{qwen3-4b-thinking-2507.a2.inf.nat.dR} & \cellcolor{cInterv!21}\celldpm{e1b}{qwen3-4b-thinking-2507.a3.inf.nat.dR} & \cellcolor{cOurs!12}\celldpm{e1b}{qwen3-4b-thinking-2507.b2.yn.nat.dacc} \\
    Qwen3.5-35B-A3B & \cellcolor{cInterv!12}\celldpm{e1q}{qwen3p5-35b-a3b.a3-quant.inf.nat.dR} & \cellcolor{cInterv!6}\celldpm{e1q}{qwen3p5-35b-a3b.a3-conn.inf.nat.dR} & \cellcolor{cOurs!16}\celldpm{e1q}{qwen3p5-35b-a3b.a3-modal.inf.nat.dR} & \cellcolor{cOurs!13}\celldpm{e1q}{qwen3p5-35b-a3b.a2.inf.nat.dR} & \cellcolor{cOurs!14}\celldpm{e1q}{qwen3p5-35b-a3b.a3.inf.nat.dR} & \cellcolor{cOurs!11}\celldpm{e1q}{qwen3p5-35b-a3b.b2.yn.nat.dacc} \\
    R1-Distill-Qwen-7B & \cellcolor{cInterv!19}\celldpm{e1s}{r1distill-7b.a3-quant.inf.nat.dR} & \cellcolor{cInterv!21}\celldpm{e1s}{r1distill-7b.a3-conn.inf.nat.dR} & \cellcolor{cInterv!14}\celldpm{e1s}{r1distill-7b.a3-modal.inf.nat.dR} & \cellcolor{cInterv!16}\celldpm{e1s}{r1distill-7b.a2.inf.nat.dR} & \cellcolor{cInterv!16}\celldpm{e1s}{r1distill-7b.a3.inf.nat.dR} & \cellcolor{cOurs!13}\celldpm{e1s}{r1distill-7b.b2.yn.nat.dacc} \\
    R1-Distill-Qwen-14B & \cellcolor{cInterv!17}\celldpm{e1s}{r1distill-14b.a3-quant.inf.nat.dR} & \cellcolor{cOurs!13}\celldpm{e1s}{r1distill-14b.a3-conn.inf.nat.dR} & \cellcolor{cOurs!21}\celldpm{e1s}{r1distill-14b.a3-modal.inf.nat.dR} & \cellcolor{cOurs!11}\celldpm{e1s}{r1distill-14b.a2.inf.nat.dR} & \cellcolor{cOurs!17}\celldpm{e1s}{r1distill-14b.a3.inf.nat.dR} & \cellcolor{cOurs!14}\celldpm{e1s}{r1distill-14b.b2.yn.nat.dacc} \\
    DeepSeek-R1-0528-Qwen3-8B & \cellcolor{cInterv!19}\celldpm{e1r1}{r1-0528-qwen3-8b.a3-quant.inf.nat.dR} & \cellcolor{cInterv!15}\celldpm{e1r1}{r1-0528-qwen3-8b.a3-conn.inf.nat.dR} & \cellcolor{cInterv!13}\celldpm{e1r1}{r1-0528-qwen3-8b.a3-modal.inf.nat.dR} & \cellcolor{cInterv!18}\celldpm{e1r1}{r1-0528-qwen3-8b.a2.inf.nat.dR} & \cellcolor{cInterv!9}\celldpm{e1r1}{r1-0528-qwen3-8b.a3.inf.nat.dR} & \cellcolor{cOurs!13}\celldpm{e1r1}{r1-0528-qwen3-8b.b2.yn.nat.dacc} \\
    \bottomrule
  \end{tabular}
  \par\smallskip{\footnotesize $^{\S}$Run with the vendor's mandatory Chinese system prompt (without it the model degenerates); thinking switched with ` /no\_think' (fast) vs.\ the default (slow).\par}
  \caption{\textbf{Full results: the literal shift on \form{some} and \form{or} holds in \protect\res{derived.main.counted} of \protect\res{derived.main.tested} open reasoning models.} Change in the reading chosen from no thinking (\NT) to natural thinking (\NAT) on the same weights, by item family, on the held-out split; for OLMo-3-Think the baseline is a forced-empty thinking block (the model has no switch), and for gpt-oss the contrast is high minus low reasoning effort. Generalized families use the inference framing: IMPPRES quantifier, connective and modal triggers \citep{jeretic-etal-2020-natural} and scalar-diversity lexical scales \citep{vantiel2016scalar,gotzner2018scalar,pankratz2021role}. LUDWIG: indirect answers \citep{ruis2023goldilocks}. Cells: mean $\pm$ standard deviation over 3 decoding seeds; blue = more pragmatic or more accurate, vermilion = more literal; \sig{*}: the 95\% cluster-bootstrap confidence interval over items excludes 0 (uncorrected; the mixed-effects tests are in \Cref{app:results}). $^{\dagger}$Without thinking, Qwen3-0.6B accepts 86--100\% of inferences (a yes-bias), so it is outside the pre-registered 10--90\% range and is excluded from the dose--response analysis. \Cref{tab:phen} lists the absolute rates.}
  \label{tab:main}
\end{table*}

\paragraph{Training stages.} The pre-registered within-stage thinking effect at each training stage was also run; its outcome is summarised under H3 in \Cref{tab:prereg}.
\begin{table*}[tp]
  \centering
  \scriptsize
  \setlength{\tabcolsep}{4pt}
  \begin{tabularx}{\textwidth}{@{}l >{\raggedright\arraybackslash}X >{\raggedright\arraybackslash}X l@{}}
    \toprule
    \textbf{Hyp.} & \textbf{Prediction} & \textbf{Outcome (evidence)} & \textbf{Status} \\
    \midrule
    \hyp{1} effort & Pragmatic reading on generalized implicatures rises with thinking & Falls for \form{some} or \form{or} (CI excluding 0) in \res{derived.main.counted} of \res{derived.main.tested} open models, among them EXAONE-Deep-7.8B, Phi-4-mini-reasoning, every newest-generation model (Qwen3.5 0.8B--9B, Qwen3.8-27B), QwQ-32B, Nemotron-Super-49B, R1-Distill-Llama-70B, every Qwen3 size from 1.7B, gpt-oss-20b, OLMo-3-32B-Think, Phi-4-reasoning(-plus), R1-Distill-Llama-8B, MiMo, Nemotron-Nano, AceReason and Skywork-OR1; rises in Magistral-Small (a shift towards answering \emph{Yes}, \Cref{app:frontier}) Granite-3.3, Llama-3.1-Nemotron-Nano-8B (Yes/No format only; literal in the three-way choice) and openPangu-Embedded-7B-model (Yes/No format only; flat in the three-way choice); no significant change in gpt-oss-120b, OLMo-3-7B-Think, SmolLM3 and openPangu-Embedded-7B-V1.1 (\Cref{tab:main}); GPT-5 falls in two formats (\Cref{app:frontier}) & not supported \\
    \hyp{1} default-and-cancel & Falls under the felicity framings, $\geq$5\,pp in $\geq$2 of 4B/8B/14B & \cellcolor{cOurs!5}Holds for \form{some}/\form{or} per family; the pre-registered IMPPRES aggregate at 14B is flat (\res{e1.qwen3-14b.a3.inf.nat.dR.mean}\,pp), so the aggregate gate criterion failed & partly supported \\
    \hyp{1b} capability crossover & Negative at small sizes, positive at $\geq$32B on $\geq$2 families, reproduced in a same-recipe pair & Qwen3-32B negative on \form{some}/\form{or}; OLMo-3-Think 7B null and 32B positive on IMPPRES but negative on \form{some}; no family flips sign within a pair on $\geq$2 families & not supported \\
    \hyp{2} split & \pcis improve while \gcis fall; $\geq$3 \pci sets improve or stay flat & \cellcolor{cOurs!15}Accuracy rises on Fig-QA and PUB at every Qwen3 size from 4B to 32B (8B: +\res{e1xt.qwen3-8b.b3.mc.nat.dacc.mean} and +\res{e1xt.qwen3-8b.b4.mc.nat.dacc.mean}\,pp), on Circa and NOPE in 14 models with no decrease (\Cref{tab:breadth}), and on presupposition and LUDWIG in Qwen3-8B; the budget $\times$ class interaction is significant in Qwen3-4B, 8B and 14B (\Cref{app:stats}) & supported \\
    \hyp{2b} speaker knowledge & Default-and-cancel: thinking lowers $\rate$ in both knowledge conditions; effort/capability: thinking widens the gap & \cellcolor{cCmpA!13}Knowledge sensitivity $\approx$0 in Qwen3 and OLMo; in gpt-oss-120b it shrinks with effort (\res{a1ctx.gpt-oss-120b.a1ctx-fel.fel.low.ksi.mean} $\rightarrow$ \res{a1ctx.gpt-oss-120b.a1ctx-fel.fel.high.ksi.mean}\,pp) & default-and-cancel supported (one model) \\
    \hyp{2c} framing $\times$ class & Truth framing denies implicatures of both classes; meaning framing lets \pcis improve & the interpretation-truth arm was not run & not tested \\
    \bottomrule
  \end{tabularx}
  \caption{Outcome of every pre-registered prediction (held-out split). ``Not tested'' marks predictions whose arm was not run.}
  \label{tab:prereg}
\end{table*}

\paragraph{Dose--response over the thinking budget.} \Cref{fig:budget} shows the reading chosen by Qwen3-8B as the thinking budget grows.

\paragraph{Figure items.} \Cref{fig:teaser} shows fragments of IMPPRES \texttt{quantifiers.132} and \texttt{connectives.1092} and of scalar-diversity \texttt{vt16.0}; its readings are glosses, not model output. \Cref{fig:method} is schematic.

\section{Pre-registration and Analysis Details}
The accounts and hypotheses, the outcome of every pre-registered prediction and the statistical procedures.

\begin{table*}[h]
  \centering
  \small
  \setlength{\tabcolsep}{2.6pt}
  \begin{tabular}{@{}l ccc ccc ccc@{}}
    \toprule
    & \multicolumn{3}{c}{\textbf{Qwen3-4B}} & \multicolumn{3}{c}{\textbf{Qwen3-8B}} & \multicolumn{3}{c}{\textbf{Qwen3-14B}} \\
    \cmidrule(lr){2-4}\cmidrule(lr){5-7}\cmidrule(l){8-10}
    \textbf{Phenomenon} & \NT & \NAT & $\Delta$ & \NT & \NAT & $\Delta$ \\
    \midrule
    \multicolumn{10}{@{}l}{\textit{Generalized implicatures: pragmatic reading $\rate$ (\%)\hup}} \\
    Category sentences, ternary & \cellcolor{cCmpA!3}\cellpm{e1}{qwen3-4b.a1.ter.off.R} & \cellcolor{cCmpA!2}\cellpm{e1}{qwen3-4b.a1.ter.nat.R} & \cellcolor{cInterv!9}\celld{e1}{qwen3-4b.a1.ter.nat.dR} & \cellcolor{cCmpA!2}\cellpm{e1}{qwen3-8b.a1.ter.off.R} & \cellcolor{cCmpA!2}\cellpm{e1}{qwen3-8b.a1.ter.nat.R} & \cellcolor{cInterv!9}\celld{e1}{qwen3-8b.a1.ter.nat.dR} & \cellcolor{cCmpA!2}\cellpm{e1}{qwen3-14b.a1.ter.off.R} & \cellcolor{cCmpA!2}\cellpm{e1}{qwen3-14b.a1.ter.nat.R} & \cellcolor{cInterv!9}\celld{e1}{qwen3-14b.a1.ter.nat.dR} \\
    Scalar diversity & \cellcolor{cCmpA!4}\cellpm{e1}{qwen3-4b.a2.inf.off.R} & \cellcolor{cCmpA!3}\cellpm{e1}{qwen3-4b.a2.inf.nat.R} & \cellcolor{cInterv!14}\celld{e1}{qwen3-4b.a2.inf.nat.dR} & \cellcolor{cCmpA!3}\cellpm{e1}{qwen3-8b.a2.inf.off.R} & \cellcolor{cCmpA!3}\cellpm{e1}{qwen3-8b.a2.inf.nat.R} & \cellcolor{cOurs!12}\celld{e1}{qwen3-8b.a2.inf.nat.dR} & \cellcolor{cCmpA!2}\cellpm{e1}{qwen3-14b.a2.inf.off.R} & \cellcolor{cCmpA!3}\cellpm{e1}{qwen3-14b.a2.inf.nat.R} & \cellcolor{cOurs!13}\celld{e1}{qwen3-14b.a2.inf.nat.dR} \\
    IMPPRES scalar & \cellcolor{cCmpA!8}\cellpm{e1}{qwen3-4b.a3.inf.off.R} & \cellcolor{cCmpA!5}\cellpm{e1}{qwen3-4b.a3.inf.nat.R} & \cellcolor{cInterv!20}\celld{e1}{qwen3-4b.a3.inf.nat.dR} & \cellcolor{cCmpA!5}\cellpm{e1}{qwen3-8b.a3.inf.off.R} & \cellcolor{cCmpA!4}\cellpm{e1}{qwen3-8b.a3.inf.nat.R} & \cellcolor{cInterv!15}\celld{e1}{qwen3-8b.a3.inf.nat.dR} & \cellcolor{cCmpA!4}\cellpm{e1}{qwen3-14b.a3.inf.off.R} & \cellcolor{cCmpA!4}\cellpm{e1}{qwen3-14b.a3.inf.nat.R} & \cellcolor{cOurs!5}\celld{e1}{qwen3-14b.a3.inf.nat.dR} \\
    \midrule
    \multicolumn{10}{@{}l}{\textit{Presupposition: projection (\%)\hup}} \\
    IMPPRES presupposition & \cellcolor{cCmpA!10}\cellpm{e1}{qwen3-4b.a4.nli.off.acc} & \cellcolor{cCmpA!11}\cellpm{e1}{qwen3-4b.a4.nli.nat.acc} & \cellcolor{cOurs!16}\celld{e1}{qwen3-4b.a4.nli.nat.dacc} & \cellcolor{cCmpA!8}\cellpm{e1}{qwen3-8b.a4.nli.off.acc} & \cellcolor{cCmpA!11}\cellpm{e1}{qwen3-8b.a4.nli.nat.acc} & \cellcolor{cOurs!19}\celld{e1}{qwen3-8b.a4.nli.nat.dacc} & \cellcolor{cCmpA!9}\cellpm{e1}{qwen3-14b.a4.nli.off.acc} & \cellcolor{cCmpA!12}\cellpm{e1}{qwen3-14b.a4.nli.nat.acc} & \cellcolor{cOurs!19}\celld{e1}{qwen3-14b.a4.nli.nat.dacc} \\
    \midrule
    \multicolumn{10}{@{}l}{\textit{Particularized implicatures: accuracy (\%)\hup}} \\
    LUDWIG indirect answers & \cellcolor{cCmpA!12}\cellpm{e1}{qwen3-4b.b2.yn.off.acc} & \cellcolor{cCmpA!12}\cellpm{e1}{qwen3-4b.b2.yn.nat.acc} & \cellcolor{cInterv!5}\celld{e1}{qwen3-4b.b2.yn.nat.dacc} & \cellcolor{cCmpA!11}\cellpm{e1}{qwen3-8b.b2.yn.off.acc} & \cellcolor{cCmpA!12}\cellpm{e1}{qwen3-8b.b2.yn.nat.acc} & \cellcolor{cOurs!13}\celld{e1}{qwen3-8b.b2.yn.nat.dacc} & \cellcolor{cCmpA!12}\cellpm{e1}{qwen3-14b.b2.yn.off.acc} & \cellcolor{cCmpA!12}\cellpm{e1}{qwen3-14b.b2.yn.nat.acc} & \cellcolor{cOurs!7}\celld{e1}{qwen3-14b.b2.yn.nat.dacc} \\
    Fig-QA metaphors & \cellcolor{cCmpA!12}\cellpm{e1xt}{qwen3-4b.b3.mc.off.acc} & \cellcolor{cCmpA!13}\cellpm{e1xt}{qwen3-4b.b3.mc.nat.acc} & \cellcolor{cOurs!15}\celld{e1xt}{qwen3-4b.b3.mc.nat.dacc} & \cellcolor{cCmpA!12}\cellpm{e1xt}{qwen3-8b.b3.mc.off.acc} & \cellcolor{cCmpA!14}\cellpm{e1xt}{qwen3-8b.b3.mc.nat.acc} & \cellcolor{cOurs!15}\celld{e1xt}{qwen3-8b.b3.mc.nat.dacc} & \cellcolor{cCmpA!13}\cellpm{e1xt}{qwen3-14b.b3.mc.off.acc} & \cellcolor{cCmpA!14}\cellpm{e1xt}{qwen3-14b.b3.mc.nat.acc} & \cellcolor{cOurs!13}\celld{e1xt}{qwen3-14b.b3.mc.nat.dacc} \\
    \bottomrule
  \end{tabular}
  \caption{\textbf{Per-phenomenon results for Qwen3-4B, 8B and 14B.} Reading chosen without (\NT) and with natural thinking (\NAT), mean $\pm$ s.d.\ over three seeds, held-out split; $\Delta$ in pp; --: not run in this form.}
  \label{tab:phen}
\end{table*}

\begin{table*}[tp]
  \centering
  \footnotesize
  \setlength{\tabcolsep}{4pt}
  \begin{tabularx}{\textwidth}{@{}l >{\raggedright\arraybackslash}X >{\raggedright\arraybackslash}X l@{}}
    \toprule
    \textbf{Hyp.} & \textbf{Prediction} & \textbf{Outcome (evidence)} & \textbf{Status} \\
    \midrule
    Scenario validity & Humans find \form{some} misleading only under full access ($\geq$80\%) & \cellcolor{cCmpA!15}\res{gold.v4.a1ctx-acc.mean}\% of \res{gold.v4.n-a1ctx} (blind consensus labels) & met \\
    Open-answer coding & LLM reference rater agrees with the human audit (lower bound of $\kappa \geq 0.40$) & \cellcolor{cCmpA!2}best rater $\kappa$ = \res{gold.v4.kappa-human-a.mean} [\res{gold.v4.kappa-human-a.lo}, \res{gold.v4.kappa-human-a.hi}]; open answers reported descriptively & not met \\
    Bott--Noveck items & Category sentences move with thinking & Near the literal ceiling in every format and condition for Qwen3 4B--32B; only binary felicity at 14B moves & floor set \\
    \bottomrule
  \end{tabularx}
  \caption*{\Cref{tab:prereg}, continued (part 3 of 3).}
\end{table*}

\begin{table*}[tp]
  \centering
  \footnotesize
  \setlength{\tabcolsep}{4pt}
  \begin{tabularx}{\textwidth}{@{}l >{\raggedright\arraybackslash}X >{\raggedright\arraybackslash}X l@{}}
    \toprule
    \textbf{Hyp.} & \textbf{Prediction} & \textbf{Outcome (evidence)} & \textbf{Status} \\
    \midrule
    \hyp{3} training origin & The literal shift appears at \rlvr, not SFT; grows with RL steps & Tülu-3: at DPO, before \rlvr; OLMo-3: at thinking SFT, RL-Zero flat, and the thinking cost shrinks over RL steps (\Cref{tab:stages}) & mixed \\
    \hyp{4} load & A load lowers $\rate$ without thinking & \cellcolor{cCmpA!5}With 2{,}048 words of unrelated preceding text and thinking off, $\rate$ on logical words moves from \res{e4.qwen3-8b.a3.L0.off.R.mean} to \res{e4.qwen3-8b.a3.Lx2048.off.R.mean}\% in Qwen3-8B but from \res{e4.qwen3-14b.a3.L0.off.R.mean} to \res{e4.qwen3-14b.a3.Lx2048.off.R.mean}\% in Qwen3-14B; no consistent direction & not supported \\
    \hyp{5} human ordering & $\rho \geq 0.3$ with human rates per scale & not run & not tested \\
    \hyp{6} restorable & \method restores $\geq$50\% of the shift; a length-matched neutral seed does not & \cellcolor{cCmpA!4}With 26-token seeds, the intent seed recovers \res{e10.qwen3-8b.a3-quant.inf.sint.R.mean}\% of a \res{e10.qwen3-8b.a3-quant.inf.off.R.mean}\% no-thinking reading on \form{some} in Qwen3-8B (\NAT: \res{e10.qwen3-8b.a3-quant.inf.nat.R.mean}\%); neutral and logic seeds recover little or none (\Cref{tab:ist}) & partly met: specific to intent, below 50\% \\
    \hyp{7} locus & The literal preference appears inside the trace, after a marker & \cellcolor{cOurs!19}Truncating the model's own trace: the pragmatic reading of \form{some} first rises (Qwen3-8B, +\res{e7.qwen3-8b.a3-quant.inf.p0128.dR.mean}\,pp after 128 tokens) and falls below the no-thinking level later in the trace (\Cref{fig:trace}); literalness markers such as \form{technically} precede the change in only \res{e7f.qwen3-8b.a3f-quant.inf.marker.pre.mean}\% of changing Qwen3-8B traces, and occur as often in traces that do not change (\res{e7f.qwen3-8b.a3f-quant.inf.marker.ctrl_marker.mean}\%) & crossing supported; marker order not supported \\
    \bottomrule
  \end{tabularx}
  \caption*{\Cref{tab:prereg}, continued (part 2 of 3).}
\end{table*}

\begin{figure*}[tp]
  \centering
  \includegraphics[width=\textwidth]{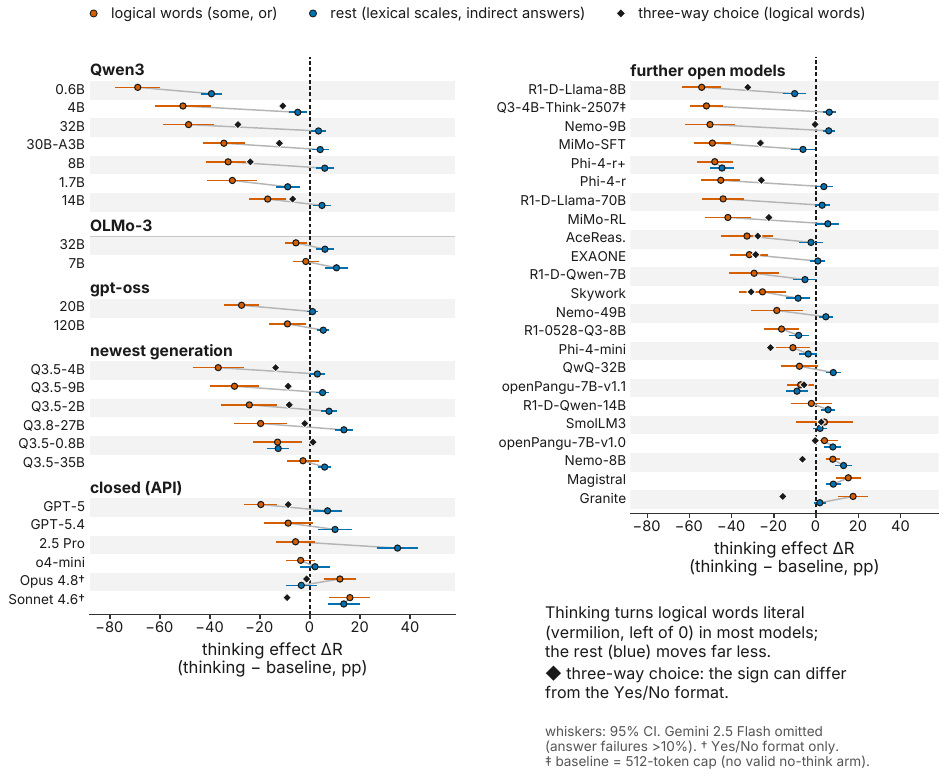}
  \caption{\textbf{Thinking reduces enrichment of \form{some}, \form{or} or both in most tested models; other inference types move differently.} Change in pragmatic reading with thinking (pp, 95\% CI): vermilion, mean of \form{some} and \form{or}; blue, mean of lexical scales and indirect answers; $\blacklozenge$, three-way choice. Every family: \Cref{fig:hero-heatmap}. Q3.5-35B: Qwen3.5-35B-A3B.}
  \label{fig:hero-full}
\end{figure*}

\subsection{Accounts, Hypotheses and Operationalisation}
\label{sec:theory}\label{app:theory}
This section states the theoretical accounts, the pre-registered hypotheses and how each is operationalised.


\paragraph{Generalized and particularized implicatures.}
\citet{grice1975logic} distinguished \emph{generalized conversational implicatures}, which an expression carries unless the context blocks them (\form{some} conveys \gloss{not all}), from \emph{particularized conversational implicatures}, which arise only from a specific context (``I wore gloves'' in answer to ``Did you leave fingerprints?'' conveys \gloss{no}). We call them \gcis and \pcis.
The two major post-Gricean theories disagree about \gcis. On the \emph{default} view \citep{levinson2000presumptive}, a \gci is generated automatically and cancelled only when the context contradicts it. On the relevance-theoretic view \citep{sperber1986relevance}, every implicature is an inference drawn when it pays off in relevance, and drawing it costs processing effort. Both treat \pcis as context-driven inferences.

\paragraph{Adults pay for enrichment; children often do not make it.}
Adults who reject an underinformative sentence such as ``Some elephants are mammals'' take longer than adults who accept it, and a short response deadline yields fewer rejections \citep{bott2004some}. A concurrent memory load does the same \citep{deneys2007logical,dieussaert2011some,marty2013scalar}.
The finding is contested. Scalar inferences depend on context \citep{breheny2006generalised,degen2015investigating} and can be immediate when the alternatives are salient \citep{grodner2010some,degen2015processing}, although other studies find a delay \citep{huang2009online}.
Development adds a second axis. Children accept underinformative \form{some} far more often than adults do \citep{noveck2001children,papafragou2003scalar}. Yet the same children give such sentences the middle value on a three-point scale, so they notice the violation and tolerate it \citep{katsos2011pragmatic}. On this view, enrichment grows with the listener's resources and competence, and the ``logician'' is a developmental stage, not a default.
For \pcis the human record differs again. Conventional indirect requests are understood without first computing their literal meaning \citep{gibbs1983literal}, and salient figurative meanings can be accessed directly \citep{giora1997understanding}.

\paragraph{Test-time thinking as a dose of deliberation.}
A reasoning model with a thinking toggle and a budget exposes what no human experiment can: the same weights answering the same item with more or less deliberation.
We use \emph{default} operationally, as the reading a model chooses at zero deliberation, and \emph{cancel} for a change of that reading towards the literal one as deliberation is added; neither term presupposes a cognitive mechanism.
We consider three profiles.
\begin{itemize}
  \item On the \emph{effort account}, enrichment is effortful, and more thinking yields more pragmatic readings on every phenomenon we test.
  \item On the \emph{default-and-cancel account}, the model holds \gcis as defaults and spends deliberation on truth-conditional verification, a habit that training with verifiable rewards (\rlvr) plausibly rewards. Thinking then \emph{cancels} \gcis while it improves \pcis, whose pragmatic reading can only be reached by inference.
  \item On the \emph{capability account}, deliberation amplifies whichever competence dominates the trace, so the sign of the effect on \gcis depends on the model's capability, much as enrichment depends on age in children.
\end{itemize}
The theory question is therefore:
\begin{center}
\emph{Do reasoning models pay the cost of enrichment, or the cost of cancelling it, and does that depend on capability?}
\end{center}

\begin{claimbox}
Test-time thinking is a dose of deliberation on fixed weights. If pragmatic enrichment in language models is effortful, more thinking yields more enrichment on every phenomenon; if training with verifiable rewards installs a default-and-cancel profile, thinking cancels generalized implicatures while it improves particularized ones; if deliberation amplifies existing competence, the sign on generalized implicatures reverses with capability.
\end{claimbox}

\paragraph{Operationalisation.}
The dependent variable is the \emph{reading chosen}: $\rate$ is the proportion of committed answers that take the pragmatic reading. For \pcis the pragmatic reading is the correct answer, so $\rate$ is accuracy.
The dose is the \emph{thinking budget} $B$, from no thinking (\NT) through capped budgets to natural stopping under an 8{,}192-token safety cap (\NAT). Its log, $x=\log_2(B+1)$, enters the statistical models.
Under a true/false instruction, the literal answer to ``Some elephants are mammals'' is \emph{correct}. A model that reads the question as a logic problem would therefore drift literal with thinking for reasons that have nothing to do with enrichment; we call this \emph{logic-mode compliance}. The primary readouts are judgements of felicity rather than truth, and the true/false framing serves as the logic-mode comparison.
Binary judgements also hide pragmatic tolerance. Category sentences are therefore judged on a ternary scale (completely right / a bit wrong / completely wrong; both lower values count as pragmatic) and on a 1--7 scale \citep{katsos2011pragmatic,degen2015investigating}, with binary felicity and truth judgements as comparisons.
Binary judgements of category sentences stay near the literal ceiling (\res{e1.qwen3-8b.a1.fel.off.lit.mean}\% literal for Qwen3-8B without thinking), whereas a majority of adults reject such sentences (54--100\% across tasks; \citealp{deneys2007logical,marty2013scalar,katsos2011pragmatic}).
\Cref{tab:operationalization} maps each prediction to its measurement.

\begin{table}[t]
  \centering
  \small
  \setlength{\tabcolsep}{3pt}
  \renewcommand{\arraystretch}{1.1}
  \begin{tabularx}{\linewidth}{@{}>{\raggedright\arraybackslash}p{2.0cm} >{\raggedright\arraybackslash}X c@{}}
    \toprule
    \rowcolor{gray!8}\textbf{Prediction} & \textbf{Measurement on LMs} & \textbf{Hyp.} \\
    \midrule
    Effort ordering & $\rate$ on \gcis vs.\ $x=\log_2(B+1)$ under felicity framings (inference; ternary and 1--7 for category sentences); true/false as the logic-mode contrast & \Cref{hyp:direction} \\
    \gci/\pci split & budget $\times$ class interaction on $\rate$ (\pci: accuracy, literal-error share) & \Cref{hyp:split} \\
    Speaker knowledge & $\Delta\rate$ (full $-$ partial knowledge) at \NT vs.\ \NAT & H2b \\
    Framing $\times$ class & \pci accuracy under meaning vs.\ interpretation-truth framing, \NT vs.\ \NAT & H2c \\
    Training origin & default $\rate$ by stage (log-prob scoring); cost of thinking by stage; $\rho$(RL step, literal rate) & \Cref{hyp:origin} \\
    Load & $\rate$ under a memory load, crossed with thinking & \Cref{hyp:load} \\
    Human ordering & Spearman $\rho$ with human rates per scale and item & \Cref{hyp:human} \\
    Restorability & \method vs.\ token-matched neutral and logic seeds; \pci accuracy & \Cref{hyp:restore} \\
    Locus & answer-position log-odds over trace prefixes; marker onset & \Cref{hyp:locus} \\
    \bottomrule
  \end{tabularx}
  \caption{\textbf{From theory to measurement.} Each prediction that separates the effort and default-and-cancel accounts, the quantity we measure, and the hypothesis it tests. $\rate$: proportion of committed answers that take the pragmatic reading; $B$: thinking budget in tokens.}
  \label{tab:operationalization}
\end{table}

\paragraph{Hypotheses.}
Hypotheses, thresholds, falsifiers and the rule that decides between the three profiles were fixed before the main runs (\Cref{app:stats}).

\begin{hypothesis}[Direction on \gcis]\label{hyp:direction}
Under the felicity framings, the effort account predicts that $\rate$ on \gcis rises with $x$; the default-and-cancel account predicts that it falls. A direction counts if $|\Delta\rate|$ between \NT and \NAT is at least 5\,pp with non-overlapping 95\% confidence intervals, with the same sign under capped and length-stratified budgets and under both readouts. Default-and-cancel predicts the shift under both framings, whereas logic-mode compliance predicts it under true/false only. \emph{Falsified} for both accounts if $\Delta\rate$ is equivalent to zero within $\pm3$\,pp at every size.
(\hyp{1b}, capability) The capability account predicts that $\Delta\rate$ is negative for the smallest models and positive for at least one model of 32B or more, that its sign is ordered by a capability measure rather than by training recipe, and that the flip reproduces within a same-recipe size pair.
\end{hypothesis}

\begin{hypothesis}[Split]\label{hyp:split}
Thinking raises accuracy and lowers the literal-error share on \pcis while it lowers $\rate$ on \gcis, a budget $\times$ class interaction. The claim is about the class average, with a random effect for item set. It holds if the interaction is significant and at least three \pci sets improve or stay flat under natural thinking; a drop confined to a capped budget counts as a truncation cost. \emph{Falsified} if \gcis and \pcis move in the same direction, or if \pci accuracy drops under natural thinking on at least half of the sets.
Two corollaries follow.
(\hyp{2b}) A pragmatic listener enriches more when the speaker has full knowledge \citep{goodman2013knowledge}. Default-and-cancel predicts that thinking lowers $\rate$ in both knowledge conditions; the effort and capability accounts predict that thinking widens the gap between them.
(\hyp{2c}) Asking whether an implicated interpretation is \emph{true} should make thinking deny it for \gcis and \pcis alike, as in the one prior same-weights contrast (\citealp{spinoso2026evaluating}, Tables~6--7). Asking what the speaker \emph{meant} should instead let \pcis improve. \emph{Falsified} if \pci accuracy under the meaning framing also drops, or if the truth-framing drop is absent for \pcis.
\end{hypothesis}

\begin{hypothesis}[Training origin]\label{hyp:origin}
The literal shift is installed by \rlvr. We test two estimands. \emph{H3-default}: the default reading, scored without deliberation, turns literal at the \rlvr stage and, in zero-RL runs, with RL steps. \emph{H3-cost}: the within-stage cost of thinking appears or grows at the \rlvr stage. \emph{Falsified} if the change appears at SFT, if at most two of five RL-Zero series trend, or if distilled thinkers show the same cost as RL-trained ones.
\end{hypothesis}

\begin{hypothesis}[Load]\label{hyp:load}
A memory load lowers $\rate$ on \gcis without thinking, as in humans. The load $\times$ thinking interaction shows whether thinking acts as a resource or as the canceller. \emph{Falsified} if load moves $\rate$ by less than 3\,pp at both thinking levels.
\end{hypothesis}

\begin{hypothesis}[Human ordering]\label{hyp:human}
At the smallest budget, literal rates per scale correlate with human rates (Spearman $\rho \geq 0.3$, with a CI excluding 0).
\end{hypothesis}

\begin{hypothesis}[Restorable]\label{hyp:restore}
One fixed sentence about speaker intent, placed inside the thinking channel (\emph{Intent-Seeded Thinking}, \method), restores at least half of a negative \gci shift, with \pci accuracy non-inferior within 1\,pp. \emph{Falsified} if a token-matched neutral sentence restores as much.
\end{hypothesis}

\begin{hypothesis}[Locus]\label{hyp:locus}
Read at successive prefixes of a trace, the answer-position preference starts near its no-thinking value and crosses to the literal reading inside the trace, after literalness markers such as \form{technically} and \form{strictly speaking}.
\end{hypothesis}

\subsection{Pre-registered Predictions and Outcomes}
\label{app:prereg}
This section reports the outcome of every pre-registered prediction.

Hypotheses, thresholds and falsifiers were fixed before the main runs (\Cref{sec:theory}). \Cref{tab:prereg} lists the outcome of each prediction on the held-out split. Several did not hold; we report them in full.

\paragraph{Pre-registered remedies.} We also pre-registered four remedies (a sentence at the start of thinking, a check before thinking ends, and fine-tuning and preference optimisation on the model's own correct traces); all four were run, and none passed its pre-declared development gates (pre-registration file \texttt{METHOD\_DESIGN.md} at \url{https://anonymous.4open.science/r/entertain-then-cancel/}).

\subsection{Statistical Procedures}
\label{app:stats}
This section lists the run settings and statistical procedures.

\Cref{tab:manifest} lists, for every experiment block, the number of models, items per model, decoding seeds, temperature and thinking cap; NAT generations stop at the model's close-of-thinking delimiter or at the cap, and capped traces are forced closed.
\begin{table}[t]
  \centering\small
  \begin{tabular}{@{}lrrrrr@{}}
    \toprule
    \rowcolor{gray!8}Block & Models & Items & Seeds & $T$ & Cap \\
    \midrule
    \texttt{E1} & 3 & 3{,}877 & 3 & 0.6 & 8192 \\
    \texttt{E1S} & 13 & 2{,}052 & 3 & 0.6 & 8192 \\
    \texttt{E1SF} & 5 & 3{,}600 & 3 & 0.6 & 8192 \\
    \texttt{E1SFX} & 17 & 3{,}600 & 3 & 0.6 & 8192 \\
    \texttt{E1Q} & 6 & 2{,}052 & 3 & 0.6 & 8192 \\
    \texttt{E1P} & 7 & 2{,}052 & 3 & 0.6 & 8192 \\
    \texttt{E1B} & 14 & 2{,}052 & 3 & 0.6 & 8192 \\
    \texttt{E1PANGU} & 3 & 5{,}972 & 3 & 0.6 & 8192 \\
    \texttt{E1R1} & 1 & 2{,}052 & 3 & 0.6 & 8192 \\
    \texttt{E3} & 2 & 2{,}052 & 3 & API & -- \\
    \texttt{E1GSF} & 1 & 3{,}600 & 3 & API & -- \\
    \texttt{E1PROBE} & 6 & 320 & 3 & 0.6 & 8192 \\
    \texttt{E1DS} & 17 & 1{,}092 & 3 & 0.6 & 8192 \\
    \texttt{E1XL} & 8 & 476 & 3 & 0.6 & 8192 \\
    \texttt{E10} & 7 & 1{,}492 & 3 & 0.6 & 8192 \\
    \texttt{E7} & 15 & 687 & 3 & 0.6 & 8192 \\
    \texttt{E7F} & 11 & 1{,}800 & 3 & 0.6 & 8192 \\
    \texttt{E2} & 44 & 2{,}052 & 3 & API & -- \\
    \texttt{E2T} & 8 & 1{,}492 & 3 & 0.6 & 8192 \\
    \texttt{E2C} & 3 & 1{,}492 & 3 & 0.6 & 8192 \\
    \texttt{E6} & 4 & 1{,}067 & 3 & 0.6 & 8192 \\
    \texttt{E8} & 6 & 1{,}492 & 3 & 0.6 & 8192 \\
    \texttt{A1CTX} & 6 & 750 & 3 & 0.6 & 8192 \\
    \texttt{A6} & 5 & 813 & 3 & 0.6 & 8192 \\
    \texttt{E9} & 6 & 600 & 3 & 0.6 & 8192 \\
    \texttt{M1DEV} & 3 & 3{,}450 & 1 & 0.6 & 8192 \\
    \bottomrule
  \end{tabular}
  \caption{Run settings per experiment block: models, items per model, decoding seeds, temperature and thinking cap (tokens).}
  \label{tab:manifest}
\end{table}

Hypotheses, thresholds, falsifiers and exclusion rules were fixed before the main runs and are recorded with dates in the project log (anonymised at \url{https://anonymous.4open.science/r/entertain-then-cancel/}).
\begin{itemize}
  \item \textbf{Budget $\times$ class model.} For each Qwen3 model we fit a logistic model of the pragmatic reading on $\log_2(B+1)$ of the realised thinking tokens, the class (generalized vs.\ particularized or presupposition) and their interaction. The plan specified by-item random intercepts and slopes; we fit the same fixed effects with item-clustered robust errors instead (no mixed-model solver was available on our stack), a population-averaged estimate. The interaction is \res{derived.glmm.qwen3-4b.coef}, \res{derived.glmm.qwen3-8b.coef}, \res{derived.glmm.qwen3-14b.coef} and \res{derived.glmm.qwen3-32b.coef} for 4B, 8B, 14B and 32B ($p$ \res{derived.glmm.qwen3-4b.p}, \res{derived.glmm.qwen3-8b.p}, \res{derived.glmm.qwen3-14b.p}, \res{derived.glmm.qwen3-32b.p}).
  \item \textbf{Confidence intervals.} Cluster bootstrap over items, stratified by set, with 10{,}000 resamples. Differences between conditions are computed inside each resample, so they are paired.
  \item \textbf{Equivalence.} Two one-sided tests at $\pm3$\,pp on the change in lexical scales and LUDWIG. Equivalence holds for lexical scales in \res{derived.tost.a2.equivalent} of \res{derived.tost.a2.tested} models; LUDWIG, with 95 items, is too small to establish equivalence at this margin.
  \item \textbf{Multiple comparisons.} Holm correction of the paired-bootstrap $p$-values for the \form{some} and \form{or} shifts within each model family. On the full held-out sets \res{derived.holm.full.survive} of \res{derived.holm.full.tests} shifts survive; on the core sets of 26--33 items, fewer do, as expected from their size. Secondary analyses are uncorrected and labelled as such.
  \item \textbf{Exclusions.} Hedged answers are excluded from $\rate$ and reported, and the extraction-free readout $L$ keeps them. 
  \item \textbf{Power.} Item counts per cell are given with every result; with 26--33 core items per word, the item-level intervals are wide, which is why the headline numbers are replicated on the full held-out sets of 300 items per word.
\end{itemize}

\section{Related Work}
The extended discussion of related work.

\subsection{Related Work}
\label{sec:related}\label{app:related}
This section reviews related work on implicature, reasoning models and test-time compute.


\paragraph{Test-time reasoning and social understanding.}
Several studies find that reasoning does not transfer to social cognition.
Reasoning variants of Qwen3 and DeepSeek do no better than their non-reasoning counterparts on theory-of-mind benchmarks, and o3's accuracy on higher-order belief questions falls as reasoning effort rises \citep{gong2026think}. Reasoning models also trail older models on theory-of-mind variants of cognitive-science experiments \citep{lupu2025decrypto}, and the gains they do show are better described as robustness to prompt variation than as new ability \citep{dehaan2026reasoning,dehaan2026evaluating}.
For presupposition, models with stronger reasoning give less human-like projection judgements \citep{azin-etal-2026-presupposition}, and reasoning adds only 2--11 points on queries with false presuppositions \citep{sathyanathan-etal-2026-evaluating}.
On implicature the record is mixed.
The closest observation is incidental: in ImplicatureX, a benchmark of implicature recognition and cancellation, Qwen3 in thinking mode at a 512-token budget recognises fewer scalar implicatures than the same weights without thinking, and recognises cancellations more often \citep{spinoso2026evaluating}.
Chain-of-thought (CoT) prompts built on Grice's maxims, on relevance theory or on the Rational Speech Act model improve implicature resolution and number-word interpretation \citep{kim2023pope,sato2025pragmatic,tsvilodub2025nonliteral}; CoT lifts GPT-4 to human average on binary implicatures \citep{ruis2023goldilocks}; and training on counterfactual reasoning traces helps further \citep{park2026pragrest,sravanthi-etal-2025-understand}. Yet CoT was also observed to ``encourage literal interpretations'' in Korean \citep{park-etal-2024-pragmatic}.
Unlike ImplicatureX, which observes the drop at a single budget on 46 items without intervals and does not treat it as a thinking effect, and unlike \citet{gong2026think}, who study the same manipulations on theory-of-mind accuracy, we treat the thinking budget as a controlled dose, measure the reading chosen under felicity and truth framings, show that the effect dissociates by implicature type, trace it to an entertain-then-cancel trajectory inside the trace and to specific post-training stages (thinking SFT in OLMo-3, preference tuning in Tülu-3), and test four pre-registered remedies under matched controls.

\paragraph{Pragmatics in language models.}
Benchmarks cover implicature, indirect speech, irony, metaphor, humour, deceit and presupposition \citep{hu-etal-2023-fine,ruis2023goldilocks,sravanthi-etal-2024-pub,jeretic-etal-2020-natural,liu-etal-2022-testing}; see \citet{ma-etal-2025-pragmatics} for a survey.
When models err on these items, their errors favour the literal reading over distractors \citep{hu-etal-2023-fine}.
Work on scalar implicature links model expectations over alternatives to human inference rates \citep{schuster-etal-2020-harnessing,hu-etal-2023-expectations}, probes scalar diversity \citep{lin-etal-2024-probing}, and steers graded interpretations through activations \citep{cho2026continuous}. Thinking and instruct models have also been compared on quantifier production, across models \citep{montero2025quantification}.
Pragmatic competence rises from base to SFT and preference-tuned checkpoints \citep{yu-etal-2026-pragmatic}. This is why comparisons across training stages confound deliberation with post-training.
Unlike these works, we treat the inference budget as the independent variable, hold the weights fixed for the primary contrast, and extend the stage analysis from supervised fine-tuning through preference tuning to \rlvr.

\paragraph{Overthinking and side effects of reasoning training.}
Longer reasoning can lower accuracy on easy problems, on tasks with distractors and on deduction \citep{chen2024overthinking,gema2025inverse,ghosal2025thinking,lee2026lost,sui2025stop}. Explicit reasoning also degrades instruction following \citep{li2025thinking}, and CoT hurts where deliberation hurts humans \citep{liu2024mind}, while its benefits concentrate on math and symbolic tasks \citep{sprague2025cot}.
Reinforcement fine-tuning has side effects of its own: it lowers refusals on unanswerable questions \citep{song2025hallucination}, can raise hallucination \citep{yao2025hallucination}, and may sharpen what the base model already does rather than extend it \citep{yue2025does}.
Budget forcing \citep{muennighoff-etal-2025-s1}, hybrid thinking modes \citep{yang2025qwen3} and effort levels \citep{openai2025gptoss} make the amount of reasoning controllable; recent work also asks how reasoning changes social behaviours such as sycophancy \citep{feng-etal-2026-good}.
These works define harm as lower accuracy on tasks with one correct answer.
Our outcome is a choice between two readings that are both licensed by the sentence. The direction of the shift, not its size, is what distinguishes the two theories.

\paragraph{The processing cost of enrichment in humans.}
Human evidence that scalar enrichment is effortful comes from response times, deadlines and dual-task load \citep{bott2004some,deneys2007logical,dieussaert2011some,marty2013scalar,tomlinson2013possibly}, and from development \citep{noveck2001children}. Counter-evidence shows that the cost depends on context and on the availability of alternatives \citep{breheny2006generalised,grodner2010some,degen2015processing}.
Enrichment rates also vary widely across scales \citep{vantiel2016scalar,gotzner2018scalar,pankratz2021role}, and probabilistic models treat enrichment as inference about a cooperative speaker \citep{frank2012predicting,goodman2016pragmatic,goodman2013knowledge}.
We import the manipulation from this literature (a controlled change in deliberation), not its conclusion. \Cref{app:theory} derives what each account predicts for a model.
\FloatBarrier

\end{document}